\documentclass{article}

\RequirePackage{etex}

\usepackage[margin=1in]{geometry}
\usepackage[english]{babel}
\usepackage[utf8]{inputenc}
\usepackage{subcaption}
\usepackage{amsmath}
\usepackage{amssymb}
\usepackage{amsfonts}
\usepackage{amsthm}
\usepackage{mathrsfs}
\usepackage{dsfont}
\usepackage[all]{xy}
\usepackage{graphicx}
\usepackage{color}
\usepackage{cite}
\usepackage{url}
\usepackage{indentfirst}
\usepackage[labelfont=bf,labelsep=period,justification=raggedright]{caption}
\usepackage{hyperref}
\usepackage[colorinlistoftodos]{todonotes}
\usepackage{tkz-fct}
\usepackage{tikz}
\usetikzlibrary{cd}
\usetikzlibrary{calc}
\usepackage{multicol}
\PassOptionsToPackage{dvipsnames,svgnames}{xcolor}
\usepackage{textcomp}
\usepackage{titling}
\usepackage{enumitem} 

\usepackage[T1]{fontenc}    
\usepackage{booktabs}       
\usepackage{nicefrac}       
\usepackage{microtype}      
\usepackage{lipsum}

\usepackage{arxiv}

\DeclareMathOperator{\SO}{SO}

\DeclareMathOperator{\Tr}{Tr}

\DeclareMathOperator{\Var}{Var}
\DeclareMathOperator{\init}{init}
\DeclareMathOperator{\feat}{feat}

\newcommand{\norm}[1]{\left\lVert#1\right\rVert}
\newcommand{\opnorm}[1]{\left\lVert#1\right\rVert_{\text{op}}}

\newcommand{\diag}{\text{diag}}
\newcommand{\Unif}{\text{Unif}}
\newcommand{\eigmax}{\lambda_{\text{max}}}
\newcommand{\eigmin}{\lambda_{\text{min}}}

\newcommand{\MP}{\mu_{\text{MP}}}

\newcommand{\iid}{\overset{\mathrm{iid}}{\sim}}

\newcommand{\Lpconverge}{\overset{L^p}{\longrightarrow}}
\newcommand{\Lconverge}{\overset{L^2}{\longrightarrow}}
\newcommand{\asconverge}{\overset{\text{a.s.}}{\longrightarrow}}
\newcommand{\Pconverge}{\overset{\mathbb{P}}{\longrightarrow}}
\newcommand{\Leq}{\overset{L^2}{\sim}}

\newcommand{\ndlim}{\overset{n, D \nearrow \infty}{\longrightarrow}}
\newcommand{\cl}{c_\lambda}
\newcommand{\alim}{a_\infty}
\newcommand{\blim}{b_\infty}
\newcommand{\atlim}{\tilde{a}_\infty}
\newcommand{\btlim}{\tilde{b}_\infty}
\newcommand{\bcl}{\bar{c}_\lambda}
\newcommand{\halpha}{\hat{\alpha}}
\newcommand{\aseq}{\overset{\text{a.s.}}{\sim}}

\newcommand{\ws}{w_\star}
\newcommand{\what}{\hat{w}_\star}

\newcommand{\EE}{\mathbb{E}}

\newcommand{\PP}{\mathbb{P}}

\newcommand{\RR}{\mathbb{R}}
\renewcommand{\SS}{\mathbb{S}}
\newcommand{\ZZ}{\mathbb{Z}}

\newcommand{\gen}{\mathcal{G}}
\newcommand{\bias}{\mathcal{B}}
\newcommand{\variance}{\mathcal{V}}

\newcommand{\Bt}{\tilde{\mathcal{B}}}
\newcommand{\Vt}{\tilde{\mathcal{V}}}
\newcommand{\Gt}{\tilde{\mathcal{G}}}

\newcommand{\kl}{k_\ell}
\newcommand{\bkl}{\bar{k}_\ell}
\newcommand{\flk}{\hat{f}_{\lambda;\kl}}

\newcommand{\finit}{\hat{f}_{\lambda; \init}}
\newcommand{\ffeat}{\hat{f}_{\lambda; \feat}}

\newcommand{\Kf}{K_{\feat}^\ell}

\newcommand{\wL}{\mathbf{w}^L}
\newcommand{\wl}{\mathbf{W}^\ell}
\newcommand{\weight}{\mathbf{W}}

\def\multiset#1#2{\ensuremath{\left(\kern-.3em\left(\genfrac{}{}{0pt}{}{#1}{#2}\right)\kern-.3em\right)}}

\theoremstyle{plain}
\newtheorem{thm}{Theorem}
\newtheorem{cor}[thm]{Corollary}

\newtheorem{prop}[thm]{Proposition}
\newtheorem{lemma}[thm]{Lemma}

\newtheorem{assump}{Assumption}

\theoremstyle{definition}
\newtheorem{defn}[thm]{Definition}

\theoremstyle{remark}
\newtheorem{rem}[thm]{Remark}

\numberwithin{equation}{section}
\numberwithin{thm}{section}

\begin{document}
\title{Investigating the Early Signatures of Feature Learning in Deep Linear Networks}
\author{Shaunak Bhandarkar \\
        shaunak@princeton.edu \\
        Princeton University \\ 
        \and \textbf{Boris Hanin} \\
        bhanin@princeton.edu \\
        Princeton University}

\date{\today}

\maketitle

\begin{abstract}
Deep neural networks learn data-dependent features, a capacity widely credited for their empirical success. However, explicitly characterizing what features emerge --- and when they improve generalization --- is challenging, since training is nonlinear and nonconvex and the data is often anisotropic. Here, we give a closed-form account of feature learning for a deep linear network in a mean-field parameterization, trained by one gradient step from initialization on high-dimensional data carrying a low-dimensional (spiked) signal. In the infinite-width mean-field limit, one step updates each layer's feature kernel by a target-aligned rank-one spike, amplified by both the learning rate and the layer depth. 
We next assess how these features impact prediction through kernel ridge regression (KRR), finding that
KRR on the learned feature kernel yields scalar inflation of the KRR estimator at initialization --- an ``anti-shrinkage'' mechanism that reinflates task-relevant directions suppressed by ridge, while also amplifying noise. To study when this inflation improves generalization, in the overparameterized proportional limit we derive closed-form expressions for the bias, variance, and generalization errors of these estimators in terms of key macroscopic parameters --- notably the spike strength and spike-target alignment.
Crucially, we find that a one-step generalization advantage vanishes for isotropic inputs and appears when both the spike strength and an effective signal-to-noise ratio are sufficiently large, demonstrating that a successful first gradient step both requires and benefits from low-dimensional task structure. 
\end{abstract}

\section{Introduction}

A defining trait of deep neural networks is their ability to learn data-dependent features within their internal layers \cite{damian2022neural}. This ability marks a salient departure from their predecessors --- kernel models --- which utilize a fixed set of features and only learn how to weight them:
\begin{equation}    \label{eq:kernel_model}
    z(x; \theta) = \sum_{j=1}^N W_j \phi_j(x)
\end{equation}
Here, $\theta = (W_1, W_2, \dots, W_N)$ is a collection of regression weights and $\phi_1, \phi_2, \dots, \phi_N$ are basis functions that transform the input $x \in \RR^D$ into the \emph{features} $(\phi_j(x))_{1 \leq j \leq N}$, giving rise to a \emph{feature map} $x\mapsto(\phi_1(x),\dots,\phi_N(x))$. 
A kernel model learns the $W_j$ while
holding this map fixed; a deep network additionally adapts the map to the
data through training.
This process of \emph{feature learning} is widely
credited for the empirical success of deep networks \cite{he2016deepresidual, brown2020languagemodels, allenzhu2019resnet, ghorbani2020neural}, yet it
is precisely what places them outside the reach of kernel theory. 
We therefore ask: what data-dependent features does a deep neural network learn, and when do such features generalize better than fixed, data-independent ones?

Answers to these questions are complicated by two issues.
First, the learning dynamics of deep neural networks are nonlinear and nonconvex, making closed-form descriptions of feature learning limited. 
A large body of work addresses this issue by passing to an infinite-width limit ($N \to \infty$), in which aspects of the network dynamics collapse onto deterministic objects.
The $\mu P$ \cite{yang2021tensor} and
mean-field \cite{mei2018meanfield} network parameterizations support stable feature
learning in this limit,
and have been used to analyze the training dynamics and convergence properties of deep networks \cite{mei2018meanfield, sirignano2020lln, sirignano2020clt, chizat2018global, rotskoff2018parameters, bordelon2022dmft}. Within these descriptions, the dynamical mean-field theory (DMFT) \cite{bordelon2022dmft, bordelon2023learningrule} characterizes the full training trajectory in terms of a reduced set of deterministically evolving kernel order parameters, which include the network's feature kernels at each layer.
This analysis is powerful, but
it characterizes feature learning \emph{implicitly} through self-consistent
equations for order parameters, rather than explicitly computing the features learned under gradient descent. Additionally, the DMFT typically involves fixing the training dataset \cite{bordelon2022dmft, bordelon2023dynamics, bordelon2024depth_muP, bordelon2025deeplinear}, leaving open settings in which the number of data points $n$ and data dimension $D$ grow in addition to the width $N$ (which \cite{bordelon2025deeplinear} explores for deep linear networks).

\begin{figure}[t]
  \centering
  \includegraphics[width=0.8\textwidth]{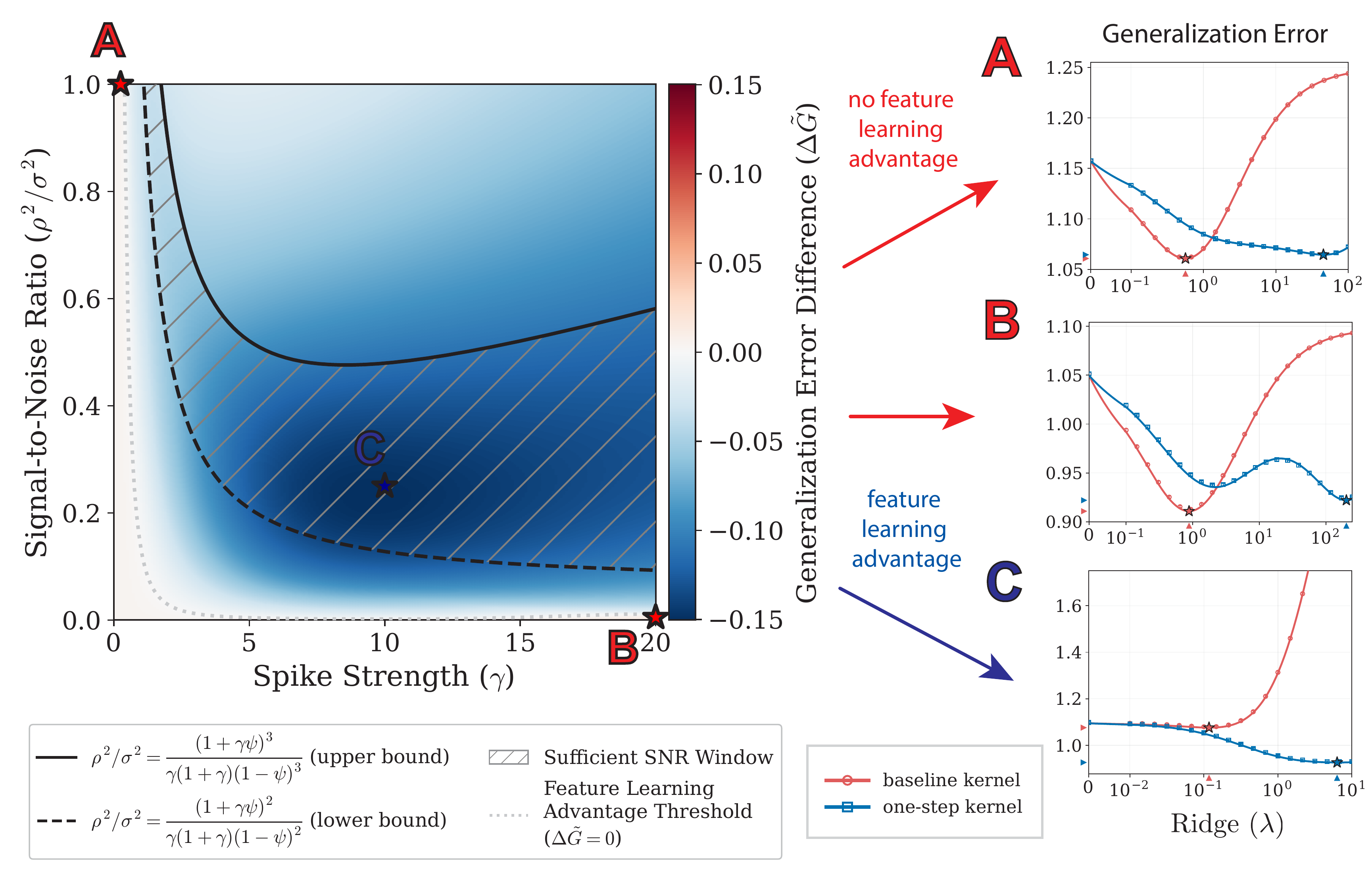}
  \caption{A phase diagram of the predicted minimal asymptotic generalization error (over ridge strengths $\lambda \geq 0$) arising from learned one-step layer-$\ell$ features (with effective update strength $\kl := \frac{\eta^2 \ell (\ell + 1)(2 \ell + 1)}{6D} \to 10$) relative to baseline features at initialization. Here, $n/D \to \psi = 0.2$ and the noise level is $\sigma^2 = 1$. Blue denotes a feature learning advantage, and red denotes the opposite. The shaded region marks an ``SNR window'' in which a global feature learning advantage provably holds: certain one-step ridge estimators ($\kl > 0$) can outperform \emph{any} baseline ridge estimator (see Sec. \ref{subsec:FLA}). To the right, generalization error curves over $\lambda$ are shown for one-step ($\kl \to 10$; blue) and baseline (red) estimators for selected values of $\gamma$ and the SNR ($\rho^2/\sigma^2$). \textbf{A.} $\gamma = 0.25$, $\text{SNR} = 1.0$. \textbf{B.} $\gamma = 20$, $\text{SNR} = 0.0049$. \textbf{C.} $\gamma = 10$, $\text{SNR} = 0.25$. Dots mark empirical simulations (with SEM; $D = 5000$, $n = 1000$) and lines indicate theory. Stars show location of the minimum generalization error attainable by either model. See Appendix \ref{app:experiments} for more details.}
  \label{fig:global_FLA_heatmap}
\end{figure}

The second issue is that the usefulness of any learned features depends on the underlying data distribution.
While prior theoretical approaches often use \emph{isotropic} inputs \cite{damian2022neural, bietti2022learning, mousavi-hosseini2023neural, ba2022onestep, dandi2024onegiantstep}, real-world data often contains low-dimensional structure embedded in otherwise high-dimensional observations \cite{ba2023learninglowdim, mousavi2023gradientbased}. A canonical model for studying such structure is a \emph{spiked matrix model} \cite{ba2023learninglowdim, mousavi2023gradientbased, johnstone2001distribution, baik2005phase}, in which a rank-one signal is embedded in high-dimensional noise. For regression tasks, this model is governed by two key parameters: the \emph{spike strength} ($\gamma \geq 0$) of this rank-one signal, and the \emph{spike-target alignment} ($\rho \in [-1, 1]$) between the signal and ground truth regression directions. 
Altogether, inputs are sampled as $x_i \iid N(0, I_D + \gamma vv^T)$ and targets as $y_i =
\frac{1}{\sqrt{D}} \ws^T x_i + \epsilon_i$, with $\epsilon_i \iid N(0, \sigma^2)$ and $\ws$ fixed so that $\frac{1}{\sqrt{D}}\ws^T v = \rho$.

In light of these issues, here we develop a simplified framework for describing feature learning and when it is useful, making three key modeling choices that afford closed-form analysis while preserving the phenomenon of interest. First, we study deep \emph{linear} networks with $L \geq 1$ layers and width $N$, in a mean-field parameterization, for which optimization remains nonlinear, nonconvex, as well as genuinely depth-dependent \cite{saxe2014exact, arora2019convergence}. 
Second, for a fixed dataset of $D$-dimensional inputs $X = [x_1, \dots, x_n] \in \RR^{D \times n}$ and scalar targets $y = [y_1, \dots, y_n] \in \RR^n$, we take a
\emph{single} gradient descent (GD) step with step size $\eta$ on the empirical mean squared error loss from initialization. As a result, writing $\theta_t$ for the network's weights after $t$ steps of GD ($t \in \{0, 1\}$), this approach allows us to exactly calculate each layer's \emph{feature kernel}
\begin{equation}    \label{eq:intro_feature_kernel}
    \Phi^\ell_t := \left( \frac{h^\ell(x_i;\theta_t) \cdot h^\ell(x_j;\theta_t)}{N} \right)_{1 \leq i, j \leq n} \in \RR^{n \times n}, \qquad 1 \leq \ell \leq L,
\end{equation}
in the limit of $N \to \infty$ (with $n$ and $D$ fixed), both before and after one step, and to compare it with the baseline \emph{input kernel} $K_x := \frac{X^T X}{D} \in \RR^{n \times n}$. Here, $h^\ell(x;\theta) \in \RR^N$ denotes the layer-$\ell$ representation of the input $x \in \RR^D$.
Third, to study when the features given by the limiting layer-$\ell$ one-step feature kernel are useful, we sample data from a
spiked matrix model and subsequently take $n, D \to \infty$; this enables us to compute (in closed-form) the generalization error arising from these features under kernel ridge
regression (for any ridge strength $\lambda \geq 0$) in the overparameterized proportional limit
$n/D \to \psi \in (0,1)$,
and to identify settings of $\gamma$ and $\rho$ that lead to better generalization after one step.

Together, these choices allow us to discover how one gradient step updates the layer-$\ell$ feature kernel, how this change propagates to the
resulting kernel ridge estimator, and for which data distributions this update lowers the
generalization error. 
Ultimately, we find that one gradient step within a deep linear network causes each of its layers to acquire a target-aligned low-rank spike, setting it up to improve generalization for data distributions with target-aligned low-dimensional structure (i.e. $\gamma$ and $\rho^2 / \sigma^2$ are sufficiently large), and to fail to do so if this structure is missing (Fig. \ref{fig:global_FLA_heatmap}).

Our approach is reminiscent of the one-step analysis of Ba et al. (2022) \cite{ba2022onestep}, in which one gradient step in a nonlinear network with one hidden layer ($L = 1$) causes the first-layer weights to acquire a target-aligned rank-one spike. Moreover, ridge regression on the updated hidden-state activations leads to meaningful improvement in generalization for a sufficiently large step. We find similar results in our work (see our contributions below). 
Unlike Ba et al., however, we do not consider nonlinearities. Nonetheless, this simplification precisely allows us to capture (in closed-form) feature learning in deeper networks $(L > 1)$ \emph{and} isolate the role of structure in the data in shaping a generalization advantage.

\subsection{Our Contributions}
Our core findings and analyses can be summarized as follows.

\paragraph{1. One step imprints a target-aligned rank-one spike on every
feature kernel, amplified by learning rate and depth.} 
For a fixed dataset $\left\{(x_i, y_i)\right\}_{i=1}^n$, let $y = [y_1, \dots, y_n] \in \RR^n$ denote the stacked targets. Then, as $N \to \infty$, the layer-$\ell$ feature kernel $\Phi^\ell$ concentrates to the input kernel $K_x$ at initialization and, after one GD step, to 
\begin{equation}
    \Kf := K_x + \frac{\eta^2 d_\ell}{n^2} K_x yy^T K_x
\end{equation}
where $d_\ell := \frac{\ell (\ell + 1)(2 \ell + 1)}{6}$ (Thm. \ref{thm:feature_kernel_limits}; see Fig. \ref{fig:one_step_kernels_finite_width}). Here, $\Kf$ differs from $K_x$ by a rank-one matrix that reflects input-target alignment, consistent with prior work \cite{ba2022onestep}.
Crucially, the strength of this update
grows with the step size $\eta$ \emph{and} with the layer index $\ell$. Thus, for a fixed step size, feature evolution after one step is more dramatic in deeper layers.

\paragraph{2. One step of feature learning inflates estimation.} 
Letting $\ffeat$ and $\finit$ denote the kernel ridge regression (KRR) estimators (with fixed ridge strength $\lambda \geq 0$) arising from $\Kf$ and $K_x$, respectively, we find that
\begin{equation}
    \ffeat(x) = c_\lambda \finit(x) \qquad \forall x \in \RR^D
\end{equation}
where $\cl \geq 1$ is a ridge- and data-dependent constant that uniformly inflates estimation (Prop. \ref{prop:inflated_estimator}); inflation vanishes at $\lambda = 0$ ($\cl = 1$). 
Although KRR at nonzero ridge $\lambda > 0$ applied to $K_x$ non-uniformly \emph{shrinks} estimation relative to the unregularized interpolator $\hat{f}_{0;\init}$, subsequent uniform inflation by $\cl$ has the effect of re-inflating estimation along high-variance directions of the empirical covariance $\frac1n XX^T$ while preserving net shrinkage along low-variance ones (Prop. \ref{prop:feat_learning_estimation_geometry}).
In the overparameterized proportional limit and in the spiked data setting, $\cl$ depends on $\lambda$ and 
\begin{equation}
    \kl := \frac{\eta^2 d_\ell}{D} = \frac{\eta^2 \ell (\ell + 1) (2 \ell + 1)}{6D},
\end{equation}
a parameter that sets the natural scale of feature learning (see Fig. \ref{fig:c_lambda_asymptotic}). Indeed, if $\eta^2 \ell^3 \ll D$, then asymptotically $c_\lambda \to 1$. However, $\eta^2 \ell^3 \asymp D$ results in genuine inflation (Thm. \ref{thm:inflation_constant}).
Thus, genuine inflation requires nonzero ridge and either performing a large gradient step --- reminiscent of \cite{ba2022onestep} --- \emph{or} sufficient depth. 
Since a one-step KRR estimator is pinned by
$(\lambda, \kl)$, we write it as $\flk$ throughout, with $\kl = 0$ recovering the
baseline ridge estimator $\hat f_{\lambda;\init}$.

\paragraph{3. For data coming from a spiked population model, we derive and analyze closed-form asymptotic formulas for bias, variance, and generalization error.} In the
proportional limit $n/D \to \psi \in (0,1)$ and where the task is noisy linear regression $\left( y_i = \frac{1}{\sqrt{D}}\ws^T x_i + \epsilon_i, \text{ } \epsilon_i \iid N(0, \sigma^2) \right)$, we derive these exact formulas for $\flk$ (Thm. \ref{thm:bias_variance_gen_error}; see Fig. \ref{fig:gen_error_sweep}).
We decompose the generalization error into \emph{bias}, which measures how well the average predictor fits the targets, and \emph{variance}, which measures the estimator's sensitivity to the specific dataset. Since all estimators coincide when $\lambda = 0$, we ask how feature learning affects bias and variance when mild ridge is added (Prop. \ref{prop:bias_variance_lambda_derivatives}; Fig. \ref{fig:bias_variance}):
\begin{enumerate}[label=(\Alph*)]
    \item To first order in $\lambda$ around $\lambda = 0$, the bias of the baseline estimator $\hat{f}_{\lambda;0}$ increases and its variance decreases.

    \item To first order in $\lambda$ around $\lambda = 0$, the effect of adding feature learning ($0 < \kl = \Theta_D(1)$) is to \emph{decrease} the bias (of $\flk$) relative to that of $\hat{f}_{\lambda;0}$, and to \emph{increase} the variance (of $\flk$) relative to that of $\hat{f}_{\lambda;0}$. 
\end{enumerate}
Altogether, ridge and feature learning act as opposing forces along the bias-variance axis. Intuitively, ridge alone shrinks the estimator, increasing bias, but improves robustness to noise in the dataset, decreasing variance. Adding
feature learning-induced \emph{uniform} inflation corrects shrinkage on average at the cost of amplifying noise along task-irrelevant directions.
The question is when this bias-for-variance trade is worth making.

\paragraph{4. A feature learning advantage exists when the data carries a sufficiently strong, target-aligned spike, and vanishes in the absence of a spike or target-alignment} (see Fig. \ref{fig:global_FLA_heatmap})\textbf{.} Since the inflation effect arising from one gradient step is only nontrivial at nonzero ridge strengths, we begin by studying when having a nonzero feature learning strength ($\kl = \Theta_D(1)$) produces a lower generalization error than baseline ridge estimators ($\kl = 0$) locally near $\lambda = 0$ (see Fig. \ref{fig:local_FLA}).
We next ask when there exists $(\lambda', \kl')$ with $\kl' > 0$ such that $\hat{f}_{\lambda';\kl'}$ asymptotically lowers the generalization error compared to the baseline estimators $\hat{f}_{\lambda;0}$ globally across \emph{all} $\lambda \geq 0$. 
For either of the above questions, a sufficient condition is the window
\begin{equation}    \label{eq:intro_SNR_window}
    \frac{(1+\gamma\psi)^2}{\gamma(1+\gamma)(1-\psi)^2}
    \;<\;\frac{\rho^2}{\sigma^2}\;<\;
    \frac{(1+\gamma\psi)^3}{\gamma(1+\gamma)(1-\psi)^3},
\end{equation}
in which $\rho^2/\sigma^2$ emerges as an \emph{effective signal-to-noise ratio}
(SNR; Thm. \ref{thm:global_FLA}). As $\gamma$ grows the upper bound diverges and the lower bound descends to
$\mathrm{SNR}_{\mathrm{crit}} = \psi^2/(1-\psi)^2$, so for a strong enough spike this condition collapses to an SNR lower bound set by the aspect ratio $\psi$.
On the other hand, for isotropic inputs ($\gamma = 0$) or for a task-orthogonal spike ($\rho = 0$), we prove that no choice of $(\lambda,
\kl)$ beats the optimally tuned baseline ridge estimator
(Thm. \ref{thm:no_isotropic_FLA}; see Fig. \ref{fig:isotropic_gen_error}). 
Collectively, these results resemble a deep linear network-based
analog of the famous BBP phase transition \cite{baik2005phase, baik2006eigenvalues}:
the ``spiked'' one-step kernel meaningfully picks up the ``spike'' in the data once
the spike strength and SNR are sufficiently large.

\subsection{Related Works}

\paragraph{Deep linear networks.} 
Despite their lack of nonlinearities, deep
linear networks possess a nonconvex loss landscape \cite{baldi1989neural} and
depth-dependent training dynamics \cite{saxe2014exact, arora2018optimization},
making them an analytically tractable testbed for theories of learning
\cite{fukumizu1998matrixricatti, bordelon2022dmft, braun2022exact,
domine2025lazyrich}.
Notably, Saxe et al. (2014) \cite{saxe2014exact} show that
when weights are suitably ``decoupled'', modes of the input--output correlation
are acquired sequentially in order of their singular values, so that depth gives
rise to ``stage-like'' transitions in learning. Closer to our question, Atanasov
et al. (2022) \cite{atanasov2022neural} show that at small initialization a deep
linear network's neural tangent kernel (NTK) can align with the target early in
training. In a similar vein, we make the earliest such alignment explicit: after a single step, each layer's feature kernel acquires a target-aligned rank-one component.

\paragraph{One-step analyses.}
A key approach for explicitly characterizing the
features a network acquires is to study the features learned after a \emph{single}
GD step from initialization. For two-layer nonlinear networks, Ba et al.
(2022) \cite{ba2022onestep} show that one GD step imprints a rank-one ``spike'' on
the first-layer weights that aligns with the linear component of the target, and
that ridge regression on the resulting features can strictly improve
generalization over kernel estimators with no learned features once the step
size is large enough. Subsequent work refines this picture: larger learning-rate
scalings can imprint multiple spikes, each capturing a higher-order component of the
target \cite{moniri2024theory}; the one-GD-step network can be mapped to a
\emph{spiked random features} model, for which test error asymptotics may be derived \cite{cui2024asymptotics}; and learning \emph{multi-index}
targets requires characteristic batch sizes and numbers of GD steps
\cite{dandi2024onegiantstep}. These analyses treat \emph{two-layer}
nonlinear networks with \emph{isotropic} inputs, and typically express the generalization
error through self-consistent equations that must be solved numerically. 
In this paper, we study linear networks but account for depth and a particular class of anisotropic inputs, and we carry out our analyses in closed form.

\paragraph{Spiked matrix models.}
The spiked population model we study stems from Johnstone (2001)
\cite{johnstone2001distribution}, and its defining feature is the BBP transition
\cite{baik2005phase}: the top sample eigenvalue separates from the bulk only
once the spike strength exceeds a critical value set by the aspect ratio. Using this model, prior work has shown that feature learning can exploit a spike, a property that depends both on the spike strength and the spike-target alignment \cite{ba2023learninglowdim, mousavi2023gradientbased}. In this paper, we make this
dependence quantitative for a deep linear network.

\paragraph{High-dimensional ridge regression.} 
Ridge regularization classically serves to \emph{shrink} estimation, mitigating
overfitting in order to lower test error \cite{hoerl1970ridge}. Using tools from random matrix theory --- notably the
Marcenko--Pastur law --- prior work has derived asymptotics for the
generalization error of ridge regression in the proportional regime and
characterized the optimal ridge strength \cite{dobriban2018highdim}. However, this shrinkage prescription can fail for heavily overparameterized settings ($D \gg n$) in which the
population covariance is anisotropic and the true regression direction is aligned
with its top eigendirections \cite{hastie2022surprises}. 
For example, for a spiked covariance, the optimal ridge strength can even be negative \cite{kobak2020optimalridge}.
More recently, Freeman (2025) \cite{freeman2025shrinkage} shows that scalar \emph{inflation} of
the minimum $\ell_2$-norm interpolator ($\lambda = 0$) lowers
the generalization error for sufficiently anisotropic covariances when $D \gg n$. These ideas challenge the classical wisdom that shrinkage enhances prediction, suggesting that \emph{inflating} estimation can improve test error in certain cases. In this paper, we
arrive at inflation from a different angle. Rather than being prescribed, the
inflation factor $\cl$ is produced by one GD step; in effect, the
deep linear network discovers an anti-shrinkage correction on its own.

\section{Problem Setup} \label{sec:setup}

\paragraph{Notation.} Throughout this paper, we let $\{(x_i, y_i)\}_{i=1}^n$ denote a dataset of $n$ input-target pairs, where $x_i \in \RR^D$ and $y_i \in \RR$. Define $X := \left[x_1, x_2, \dots, x_n \right] \in \RR^{D \times n}$, $y := [y_1, \dots, y_n] \in \RR^n$, and $\epsilon := \left[\epsilon_1, \dots, \epsilon_n \right] \in \RR^n$. $\opnorm{\cdot}$ denotes the operator norm of a matrix, and $\Lpconverge$ and $\asconverge$ refer to $L^p$ and almost sure convergence, respectively. Additionally, $\MP^{\psi}$ denotes the Marcenko-Pastur (MP) law with ratio $\psi$, i.e. the limiting spectral distribution of $\frac{1}{D} Z^T Z$ as $n/D \to \psi$, where $Z \in \RR^{D \times n}$ has i.i.d. standard Gaussian entries. For a given $\psi$, we let $g(z) = \int \frac{1}{x-z} \MP^{\psi}(dx)$ denote the Stieltjes transform of the MP law; it is well known that $g$ admits a closed-form description (see Appendix Sec. \ref{supp:setup}).

\subsection{Deep Linear Network Training Procedure}

We consider a depth-$L$ linear network of width $N$ in the feature learning
parameterization of Bordelon \& Pehlevan (2022) \cite{bordelon2022dmft}, which
maps an input $x \in \RR^D$ through $L$ hidden layers to a scalar output:
\begin{equation}    \label{eq:deep_linear_net}
    z(x;\theta)=\tfrac1N\,\mathbf{w}^L\!\cdot h^L(x;\theta),
    \quad
    h^{\ell+1}(x;\theta)=\tfrac{1}{\sqrt N}\mathbf{W}^{\ell}h^{\ell}(x;\theta),
    \quad
    h^{1}(x;\theta)=\tfrac{1}{\sqrt D}\mathbf{W}^{0}x .
\end{equation}
The trainable parameters are $\theta := (\weight^0, \dots, \weight^{L-1},
\wL)$, where $\weight^0 \in \RR^{N \times D}$, $\wl \in \RR^{N \times N}$ for
$1 \leq \ell \leq L-1$, and $\wL \in \RR^N$; at initialization, every entry of each weight matrix is
drawn i.i.d.\ from $\mathcal{N}(0,1)$.

Fix an arbitrary dataset $\{(x_\mu, y_\mu)\}_{\mu=1}^n$ with $x_\mu \in \RR^D$
and $y_\mu \in \RR$, collected into $y = [y_1, \dots, y_n] \in \RR^n$ and
$X = [\,x_1 \; \cdots \; x_n\,] \in \RR^{D \times n}$. For each layer $\ell$,
time step $t$, and pair of inputs $x, x'$, the layer-$\ell$ \emph{feature kernel}
\begin{equation}    \label{eq:feature_kernel_def}
    \Phi^\ell_t(x, x') := \frac{h^\ell(x;\theta_t) \cdot h^\ell(x';\theta_t)}{N_\ell},
    \qquad
    N_\ell := \begin{cases} D, & \ell = 0 \\ N, & \ell \geq 1 \end{cases}
\end{equation}
records the similarity structure between any pair of inputs at layer $\ell$.
Over the training set we write $\Phi^\ell_{\mu \nu; t} := \Phi^\ell_t(x_\mu,
x_\nu)$ so that $\Phi^\ell_t \in \RR^{n \times n}$, and the convention $h^0(x) := x$ recovers the input kernel $K_x :=
\Phi^0 = \tfrac1D X^\top X$. For any pair of inputs $x, x' \in \RR^D$ (not necessarily in the training set), we also write $K_x(x, x') = \frac1D x \cdot x'$.

We train the network by a \emph{single} GD step on the empirical mean squared
error loss $\mathcal{L} = \frac{1}{2n} \sum_{\mu=1}^n (z(x_\mu;\theta) - y_\mu)^2$:
\begin{equation}    \label{eq:weights_gd_update}
\begin{aligned}
\mathbf{w}_{t=1}^L
&= \mathbf{w}_{t=0}^L
+ \frac{\eta}{n} \sum_\mu h_{\mu}^L \Delta_{\mu}, \\
\mathbf{W}_{t=1}^\ell
&= \mathbf{W}_{t=0}^\ell
+ \frac{\eta}{n\sqrt{N_\ell}} \sum_\mu \Delta_{\mu}
\, g^{\ell+1}_{\mu}
\, {h_{\mu}^\ell}^{\!\top}, \qquad 0 \leq \ell \leq L-1
\end{aligned}
\end{equation}
Here $\eta \geq 0$ is the learning rate\footnote{In the mean-field parameterization of Bordelon \& Pehlevan (2022) \cite{bordelon2022dmft} (which we follow), the learning rate is $\eta N$: $\theta_{t+1} = \theta_t - \eta N \left.\nabla_{\theta}\mathcal{L}\right|_{\theta_t}$. For simplicity, here we choose to refer to the ``learning rate'' as $\eta$, keeping the dependence on $N$ implicit.}, and every quantity on the right-hand
side is evaluated at initialization: $h_\mu^\ell := h^\ell(x_\mu; \theta_{t=0})$ denotes the layer-$\ell$ hidden state, $\Delta_\mu := -n\partial \mathcal{L} /
\partial z |_{z(x_\mu;\theta_{t=0})} = y_\mu - z(x_\mu;\theta_{t=0})$ is the
output error signal, and the gradient vectors $g^\ell_\mu = N \frac{\partial z(x_\mu, \theta_{t=0})}{\partial h^\ell} \in \RR^N$ --- obtained by
backpropagation --- satisfy the recursion
\begin{equation}
\label{eqn:gd_gradient_recursion}
    g_\mu^\ell = \frac{1}{\sqrt{N}} {\mathbf{W}_{t=0}^\ell}^\top g_\mu^{\ell+1},
    \qquad
    g_\mu^L = \mathbf{w}_{t=0}^L.
\end{equation}
Fixing $n$, $D$, $L$, and $\eta$, we will compute $\Phi^\ell_{t=0}$ and
$\Phi^\ell_{t=1}$, for each $1 \leq \ell \leq L$, for arbitrary pairs of inputs (as in Eq. \ref{eq:feature_kernel_def}) as well as over the training dataset, as $N \to \infty$.

\subsection{Kernel Ridge Regression Analysis}

To characterize the effect of one learning step on estimation, we perform kernel
ridge regression (KRR) on the training data using the $N \to \infty$ limits of
$\Phi^\ell_{t=0}$ and $\Phi^\ell_{t=1}$. 
Recall that KRR with the kernel $K$ solves 
\begin{equation} \label{eqn:RKHS_objective}
\min_{f \in \mathcal{H}_K} \sum_{i=1}^{n} (y_i - f(x_i))^2 + \lambda ||f||_{\mathcal{H}_K}^2\end{equation} 
Here, $\mathcal{H}_K$ is the \emph{reproducing kernel Hilbert space} (RKHS) generated by the kernel $K$. Letting $K(x, X) = [K(x, x_1), \dots, K(x, x_n)] \in \RR^{1 \times n}$ and $K(X, X) = (K(x_i, x_j))_{1 \leq i, j \leq n} \in \RR^{n \times n}$, the optimal ridge estimator is 
\begin{equation}    \label{eq:KRR_estimator}
\hat{f}(x) = K(x, X)\left( K(X, X) + \lambda I_n\right)^{-1} y.
\end{equation}

Writing $\ffeat$
and $\finit$ for the estimators arising from the $N \to \infty$ limits of $\Phi^\ell_{t=1}$ and $\Phi^\ell_{t=0}$, respectively, we ask how these estimators differ and how feature learning interacts with ridge to shape the predictions of $\ffeat$.

\subsection{Calculation of Generalization Error}

To determine when one step of feature learning improves generalization, we
specialize to a data-generating process $\mathcal{D}$, stated
precisely in Assumption \ref{assump:data_gen_process}. In particular, inputs are sampled from a spiked matrix model with a fixed \emph{spike strength} $\gamma \geq 0$, which tunes the amount of
low-dimensional (i.e.\ rank-one) structure in the input covariance $\Sigma = I_D
+ \gamma vv^\top$, and a fixed \emph{spike-target alignment} $\rho =
\tfrac{1}{\sqrt D} \ws^\top v \in [-1,1]$.

We write $f_\star(x) := \tfrac{1}{\sqrt D} \ws^\top x$ for the target function and
$\bar{f}(x) := \EE_{\mathcal{D}}[\hat{f}(x)]$ for the average estimator across
realizations of the training data. Then, the generalization error $\gen(\hat{f})$ of $\hat{f}$ across test
inputs $x \sim \mathcal{N}(0, \Sigma)$ and realizations of the training dataset decomposes into the bias $\bias(\hat{f})$ and variance $\variance(\hat{f})$ (see Lemma \ref{lemma:bias_variance_decomp}):
\begin{equation}    \label{eq:gen_error}
\mathcal{G}(\hat f) = \EE_{x \sim \mathcal{N}(0,\Sigma)} \EE_{\mathcal{D}}
\big[ (\hat f(x) - f_\star(x))^2 \big] = \mathcal{B}(\hat f) + \mathcal{V}(\hat f),
\end{equation}
\begin{equation}    \label{eq:bias_variance}
\mathcal{B}(\hat{f}) = \EE_{x \sim \mathcal{N}(0,\Sigma)}\big[ (\bar f(x) - f_\star(x))^2 \big],
\qquad
\mathcal{V}(\hat{f}) = \EE_{x \sim \mathcal{N}(0,\Sigma)} \EE_{\mathcal{D}}
\big[ (\hat f(x) - \bar f(x))^2 \big]
\end{equation}
Under Assumption \ref{assump:data_gen_process}, we compare the generalization errors for $\ffeat$ and $\finit$ in
the proportional asymptotic limit.

\begin{assump}  \label{assump:data_gen_process}
\leavevmode
    \begin{enumerate}
        \item \textbf{Overparameterized proportional limit.} $n, D \to \infty$
        with $D > n+1$ and $\frac{n}{D} \to \psi$ for a fixed $\psi \in (0, 1)$.

        \item \textbf{Spiked population model.} Inputs $x_i \iid N(0, I_D +
        \gamma vv^T)$ for a fixed $\gamma \geq 0$ and fixed unit vector $v \in
        \SS^{D-1}$.

        \item \textbf{Noisy high-dimensional regression.} Targets $y_i =
        \frac{1}{\sqrt{D}} \ws^T x_i + \epsilon_i$ with $\epsilon_i \iid N(0,
        \sigma^2)$, where $\ws \in \sqrt{D}\,\SS^{D-1}$ is fixed and chosen such that
        $\frac{1}{\sqrt{D}} \ws^T v = \rho$ for a fixed $\rho \in [-1, 1]$.

        \item \textbf{Asymptotics.} Each of $\sigma$, $\lambda$, $\gamma$, and
        $\rho$ is a fixed constant independent of $D$ as $n, D \to \infty$.
    \end{enumerate}
\end{assump}

\section{Calculating the Features Learned from One Step of Gradient Descent}    \label{sec:onestep_GD_features}

\subsection{Calculating the feature kernels}  \label{subsec:calculating_feature_kernels}

Starting with a generic dataset $\{(x_i, y_i)\}_{i=1}^n$,
we start by describing the limiting behavior of the feature kernels $\Phi^\ell_{t}$ for each $1 \leq \ell \leq L$ and $t \in \{0, 1\}$ as $N \to \infty$. Define the \emph{depth coefficient} $d_\ell := \frac{\ell (\ell + 1) (2\ell + 1)}{6}$.

\begin{thm}    \label{thm:feature_kernel_limits}
    Let $x, x' \in \RR^D$ be any pair of inputs, not necessarily in the dataset. 
    For each $1 \leq \ell \leq L$ and any $p \geq 1$, the following results hold in the limit as $N \to \infty$:
    \begin{enumerate}
        \item $\Phi^\ell_{t=0}(x, x') \Lpconverge \frac{x \cdot x'}{D}$. In particular, over the dataset, $\Phi^\ell_{t=0} \Lpconverge K_x$.

        \item $\Phi^\ell_{t=1}(x, x') \Lpconverge \frac{x \cdot x'}{D} + \frac{\eta^2 d_\ell}{n^2} \frac{x^T X}{D} yy^T \frac{X^T x'}{D}$. In particular, over the dataset, $\Phi^\ell_{t=1} \Lpconverge K_x + \frac{\eta^2 d_\ell}{n^2} K_x yy^T K_x$.
    \end{enumerate}
\end{thm}
We prove this theorem in Appendix \ref{app:calculating_Phi_ell} (see Fig. \ref{fig:one_step_kernels_finite_width}).
Intuitively, in a deep linear network at initialization, the layer-$\ell$ features are a random linear projection of the input into $N$-dimensional space. Despite this randomness, the intrinsic \emph{correlational} structure of the inputs, as reflected by $K_x$, is still preserved in these features. 
Furthermore, after one GD step, the layer-$\ell$ feature kernel acquires information about the targets. The $N \to \infty$ limit for $\Phi^\ell_{t=1}$ may be written as
\begin{equation}    \label{eq:K_feat}
    K_{\feat}^\ell := K_x + \frac{\eta^2 d_\ell}{n^2} K_x yy^T K_x = \frac1D X^T \Lambda X
\end{equation}
where $\Lambda := I_D + \frac{\eta^2 d_\ell}{n^2}\frac1D X y y^T X^T$ is a rank-one perturbation of the identity matrix that captures input-target alignment, as reflected by the $Xy$ terms.
Crucially, this update is amplified by the learning rate $\eta$ and the layer index $\ell$. This supports the notion that feature learning is more prominent with a larger learning step and within deeper layers. 

\begin{figure}[t]
  \centering
  \includegraphics[width=\textwidth]{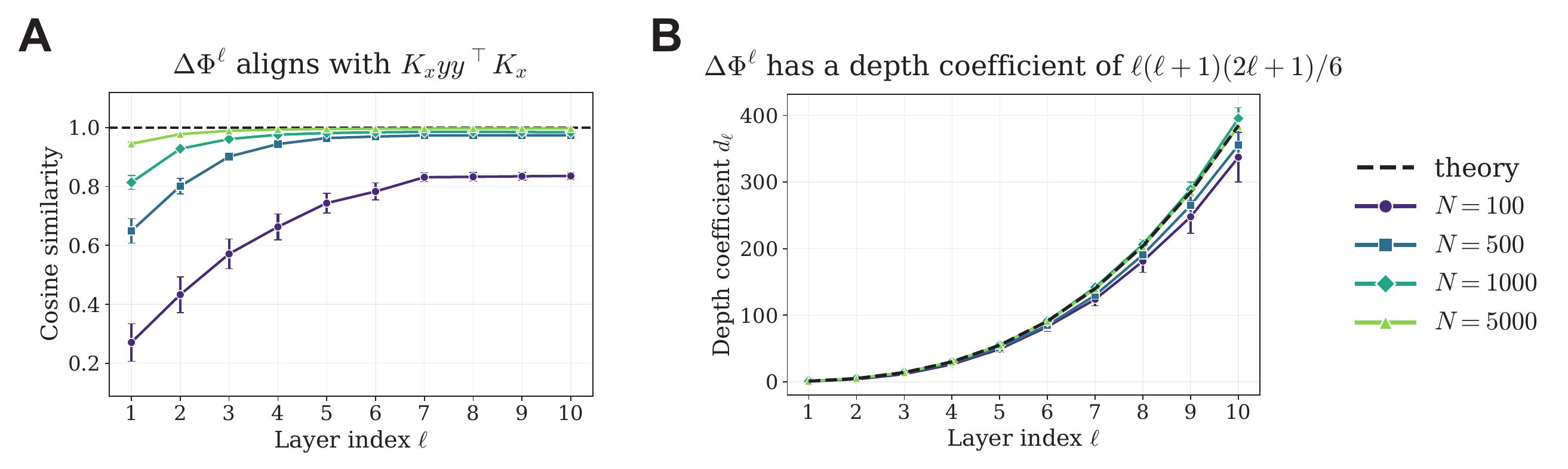}
  \caption{Empirical validation that one GD step in a deep linear network asymptotically updates $\Phi^\ell$ by the rank-one spike $\frac{\eta^2 d_\ell}{n^2} K_x yy^T K_x$. Empirical simulations are shown for various widths $N$ (with SEM) and dashed lines show theory. Here, $L = 10$, $D = 30$, $n = 10$, and $\eta = 0.5$. \textbf{A.} The feature kernel update at each layer $\ell$ structurally matches the rank-one matrix $K_x yy^T K_x$ at large $N$. \textbf{B.} Empirical estimation of the depth coefficient $d_\ell$ matches theory at large $N$.}
  \label{fig:one_step_kernels_finite_width}
\end{figure}

\subsection{Kernel ridge regression with the feature kernels}   \label{subsec:KRR_estimator}

Since $\Phi^\ell_{t=0} \Lpconverge K_x$ and $\Phi^\ell_{t=1} \Lpconverge \Kf$, we next calculate the optimal ridge estimators arising from the kernels $K_x$ and $\Kf$ under kernel ridge regression (KRR). 
Fixing a ridge strength $\lambda \geq 0$, the baseline estimator (using Eq. \ref{eq:KRR_estimator}) is 
\begin{equation}    \label{eq:baseline_estimator}
    \finit(x) = K_x(x, X)(K_x + \lambda I_n)^{-1} y = \frac{x^T X A^{-1} y}{D} = \frac{1}{\sqrt{D}} w_{\init}^T x
\end{equation}
where $A := K_x + \lambda I_n$ and $w_{\init} := \frac{1}{\sqrt{D}} X A^{-1} y$. 
We next characterize the one-step feature learning estimator $\ffeat$.
\begin{prop}    \label{prop:inflated_estimator}
    For any fixed $\lambda \geq 0$, the estimator $\ffeat$ is given by
    \begin{equation}    \label{eq:feature_learning_estimator}
        \ffeat(x) = \left( 1 + \frac{\lambda \frac{\eta^2 d_\ell}{n^2} y^T K_x A^{-1} y}{1 + \frac{\eta^2 d_\ell}{n^2} y^T K_x A^{-1} K_x y} \right) \finit(x).
    \end{equation}
    In particular, defining the dataset-dependent constant
    \begin{equation}    \label{eq:c_lambda}
        \cl := 1 + \frac{\lambda \frac{\eta^2 d_\ell}{n^2} y^T K_x A^{-1} y}{1 + \frac{\eta^2 d_\ell}{n^2} y^T K_x A^{-1} K_x y}
    \end{equation}
    it follows that $\ffeat$ is an inflation of the estimator $\finit$ and that $\ffeat(x) = \frac{1}{\sqrt{D}} w_{\feat}^T x$, where $w_{\feat} = \cl w_{\init}$.
\end{prop}

We prove this proposition in Appendix \ref{app:inflated_estimation_proof}.
Note that $y^T K_x A^{-1} y = y^T K_x^{\frac12} A^{-1} K_x^{\frac12} y \geq 0$ and $1 + \frac{\eta^2 d_\ell}{n^2} y^T K_x A^{-1} K_x y \geq 1$ since the matrices $K_x^{\frac12} A^{-1} K_x^{\frac12}$ and $K_x A^{-1} K_x$ are symmetric positive semidefinite. Thus, the KRR estimator $\ffeat$ obtained from the one-step features is a \emph{uniform inflation} of the baseline ridge estimator $\finit$.

To further analyze $\cl$, we write $K_x = U S' U^T$ as the eigen-decomposition for $K_x$, where $S' = \diag((s_i')_{i=1}^n)$ contains the eigenvalues of $K_x$, which are all nonnegative. Then, the eigenvalues of $\lambda K_x A^{-1}$ are $\left( \frac{\lambda s_i'}{\lambda + s_i'} \right)_{i=1}^n$. In the limit of $\lambda \downarrow 0$, these eigenvalues go to $0$ and so $\lambda y^T K_x A^{-1} y$ tends to $0$. In turn, this sends $c_\lambda \to 1$, and $\ffeat$ and $\finit$ coincide. When $\lambda > 0$, $\lambda y^T K_x A^{-1} y > 0$ if $y$ is not in the null space of $K_x$; in this case, inflation is \emph{strict} ($\cl > 1$).

Additionally, noting that $d_\ell \asymp \ell^3$, Eq. \ref{eq:c_lambda} indicates that $\cl$ asymptotically depends on the scaling behavior of  $\eta^2 \ell^3$. To see why, suppose that each target $y_i = \Theta_n(1)$ so that $||y||_2^2 = \Theta(n)$. If $\eta^2 \ell^3 = o(n)$, then for any $\lambda > 0$, it will be the case that 
\begin{equation}
    1 \leq \cl \leq 1 + \lambda \frac{\eta^2 d_\ell}{n} \frac{y^T y}{n} = 1 + o_n(1).
\end{equation}
Here, we have applied the crude bound
\begin{equation}
    \cl = 1 + \frac{\lambda \frac{\eta^2 d_\ell}{n^2} y^T K_x A^{-1} y}{1 + \frac{\eta^2 d_\ell}{n^2} y^T K_x A^{-1} K_x y} \leq 1 + \frac{\lambda \frac{\eta^2 d_\ell}{n^2} y^T I_n y}{1} = 1 + \lambda \frac{\eta^2 d_\ell}{n} \cdot \frac{y^T y}{n}.
\end{equation}
It will follow that $\cl \overset{n \nearrow \infty}{\longrightarrow} 1$ asymptotically, resulting in a feature learning estimator $\ffeat$ that coincides with $\finit$. Altogether, this suggests that $\eta^2 \ell^3$ must scale at least as $n$ in order to obtain genuine inflation of $\ffeat$ relative to $\finit$; this hints at the scaling $\eta^2 \ell^3 \asymp D$ in the proportional regime. We rigorously address this fact in Section \ref{sec:generalization_error}.

\subsection{The tension between inflation and ridge regression} \label{subsec:feature_geometry}

It is a curious fact that $\ffeat$ is a $\lambda$-dependent inflation of $\finit$, given that the role of ridge is to shrink $\finit$ relative to the unregularized interpolator $\hat{f}_{0;\text{init}}$. At first glance, $\cl$ appears to undo the effect of ridge regularization. However, because ridge-based shrinkage is \emph{not} uniform across different directions, relative to the interpolating estimator $\hat{f}_{0;\text{init}}$ the composition of uniform inflation with ridge (as in $\ffeat$) re-inflates estimation along high-variance directions while maintaining net shrinkage along low-variance directions.

\begin{prop}    \label{prop:feat_learning_estimation_geometry}
    Suppose that $D \geq n$, and let $\frac{1}{\sqrt{n}} X = VS^{\frac12}U^T$ be the (truncated) singular value decomposition of $\frac{1}{\sqrt{n}} X$ (with $U \in \RR^{n \times n}$ and $V \in \RR^{D \times n}$), where $S = \diag\left((s_i)_{i=1}^n \right) \in \RR^{n \times n}$ with $s_1 \geq \dots \geq s_n > 0$. 
    Define $\alpha := U^T y$ as the coordinates of $y$ in the eigenbasis given by $U$, and let $\hat{\alpha} := \frac{\alpha}{||\alpha||_2}$. Additionally, define $\bar{s} := \sum_{j=1}^n s_j \halpha_j^2$. Finally, for any vector $x \in \RR^D$, write $x = V \beta + x_\perp$, where ${\beta} \in \RR^n$ are coordinates for the eigenvectors in $V$ and $x_\perp$ is the component of $x$ in the orthogonal complement of $\text{Ran}(X) := \text{span}\left( \left\{ x_1, \dots, x_n \right\} \right)$. Then, for $\lambda \ll \frac{n}{D} s_n$,
    \begin{equation}
        \ffeat(x) = \left[ \frac{1}{\sqrt{n}} \sum_{i=1}^n \left( 1 + \frac{D}{n} \left( \frac{1}{ \frac{1}{\frac{\eta^2 d_\ell}{D} \frac{\norm{\alpha}_2^2}{n}} + \bar{s}} - \frac{1}{s_i} \right) \lambda \right) \frac{\alpha_i \beta_i}{\sqrt{s_i}} \right] + o(\lambda).
    \end{equation}
\end{prop}

This proposition is proven in Appendix \ref{app:feat_learning_geometry}. To interpret this result, consider a regime where $\eta^2 \ell^3$ is large enough that $\frac{\eta^2 d_\ell}{D} \frac{\norm{\alpha}_2^2}{n} \gg 1$. In this case, the $i$th term in the above sum is approximately
\begin{equation}
    \left( 1 + \frac{D}{n}\left( \frac{1}{ \bar{s}} - \frac{1}{s_i} \right) \lambda \right) \frac{\alpha_i \beta_i}{\sqrt{n s_i}}.
\end{equation}
Moreover, since $\beta_i$ gives the coordinate of $x$ along the $i$th eigenvector $v_i$ of $V$ (i.e. $i$th column of $V$), the coefficient
\begin{equation}
    1 + \frac{D}{n}\left( \frac{1}{ \bar{s}} - \frac{1}{s_i} \right) \lambda
\end{equation}
provides a measure of how estimation along the $i$th principal component (PC) of the data gets distorted by $\lambda$.

Since $\sum_{j=1}^n \hat{\alpha}_j^2 = 1$, the $\bar{s} = \sum_{j=1}^n s_j \hat{\alpha}_j^2$ is an average of the $n$ nonzero eigenvalues of $\frac1n XX^T$, weighted by the alignment of the targets $y$ with each eigendirection of $K_x$. In particular, for each eigenvalue $s_i \geq \bar{s}$, 
\begin{equation}    \label{eq:ridge_inflation_eig_term}
    \frac{D}{n}\left( \frac{1}{ \bar{s} } - \frac{1}{s_i} \right)  \geq 0
\end{equation}
which precisely corresponds to ``undoing'' shrinkage along the $i$th PC. For each eigenvalue $s_i$ that is smaller than $\bar{s}$, the ridge effect dominates, resulting in net shrinkage along that direction. For example, if $K_x$ has distinct eigenvalues and $y$ is perfectly aligned with the top eigen-direction of $K_x$, then $\halpha = [1, 0, \dots, 0]^T \in \RR^n$; consequently, Eq. \ref{eq:ridge_inflation_eig_term} holds for the top eigendirection of $\frac1n XX^T$, and other eigendirections will undergo net shrinkage.

Altogether, for sufficiently small $\lambda$, the one-step feature learning estimator $\ffeat$ primarily reverses shrinkage along the top eigenmodes of $\frac1n XX^T$ and genuinely shrinks estimation along smaller-eigenvalue directions of $\frac1n XX^T$ relative to the interpolator $\hat{f}_{0;\text{init}}$. Moreover, such an effect is realized when $\eta^2 \ell^3$ is sufficiently large.

\section{Linking Learned Features to Generalization}   \label{sec:generalization_error}

Having explicitly computed the estimators $\finit$ and $\ffeat$ in terms of the baseline and one-step kernels $K_x$ and $\Kf$, respectively, we now turn our focus to calculating the generalization errors arising from these estimators. For this analysis, we specialize to the setting described in Assumption \ref{assump:data_gen_process}, and we first define important intermediate quantities.

\begin{defn}
    For any fixed $\lambda \geq 0$, let $\tau_1 := \psi(1 - \lambda g(-\lambda))$ and $\tau_2 := \psi(g(-\lambda) - \lambda g'(-\lambda))$, where $g$ is the Stieltjes transform of $\MP^{\psi}$. We define four (closed-form) variables relevant to calculating the asymptotic generalization error:
    \begin{enumerate}
        \item $\alim := \tau_1$.

        \item $\blim := \frac{\gamma \tau_1 (1 - \tau_1)}{1 + \gamma \tau_1}$.

        \item $\atlim := \tau_2$.
        
        \item $\btlim := \frac{(\gamma + 1) \tau_2}{(1 + \gamma \tau_1)^2} - \tau_2$.
        
    \end{enumerate}
\end{defn}

\subsection{A careful look at \texorpdfstring{$\cl$}{the inflationary constant}}    \label{subsec:cl_asymptotics}

First, we characterize $\cl$ in the overparameterized proportional limit. Defining $\kl := \frac{\eta^2 d_\ell}{D}$, we state the following result:

\begin{thm} \label{thm:inflation_constant}
Suppose that Assumption \ref{assump:data_gen_process} holds and that $\lim_{n, D \to \infty} k_\ell = \bkl$ for some nonnegative constant $\bkl$ independent of $D$ and $n$. Then, as $n, D \to \infty$, $\cl$ converges almost surely to a nonrandom constant $\bcl$ that is at least $1$:
         \begin{equation}
    \cl \asconverge 1 + \frac{\frac{\lambda}{\psi} \bkl \left( 1 + \sigma^2 + \gamma \rho^2 - \lambda \left( \frac{\alim + \blim \rho^2}{\psi} + \sigma^2 g(-\lambda) \right) \right)}{1 + \bkl \left( 1 + \frac{1+\sigma^2}{\psi} + \gamma \rho^2 \left( 2 + \gamma + \frac{1}{\psi} \right) - \frac{\lambda}{\psi}\left( 1 + \sigma^2 + \gamma \rho^2 \right) + \frac{\lambda^2}{\psi} \left( \frac{\alim + \blim \rho^2}{\psi} + \sigma^2 g(-\lambda) \right) \right)}.
\end{equation}
In particular, when $k_\ell = o_D(1)$, $\cl \asconverge 1$, and $\bcl > 1$ when both $\bkl > 0$ (meaning that $\eta^2 \ell^3 \asymp D$) and $\lambda > 0$.
\end{thm}

\begin{figure}[t]
  \centering
  \includegraphics[width=0.8\textwidth]{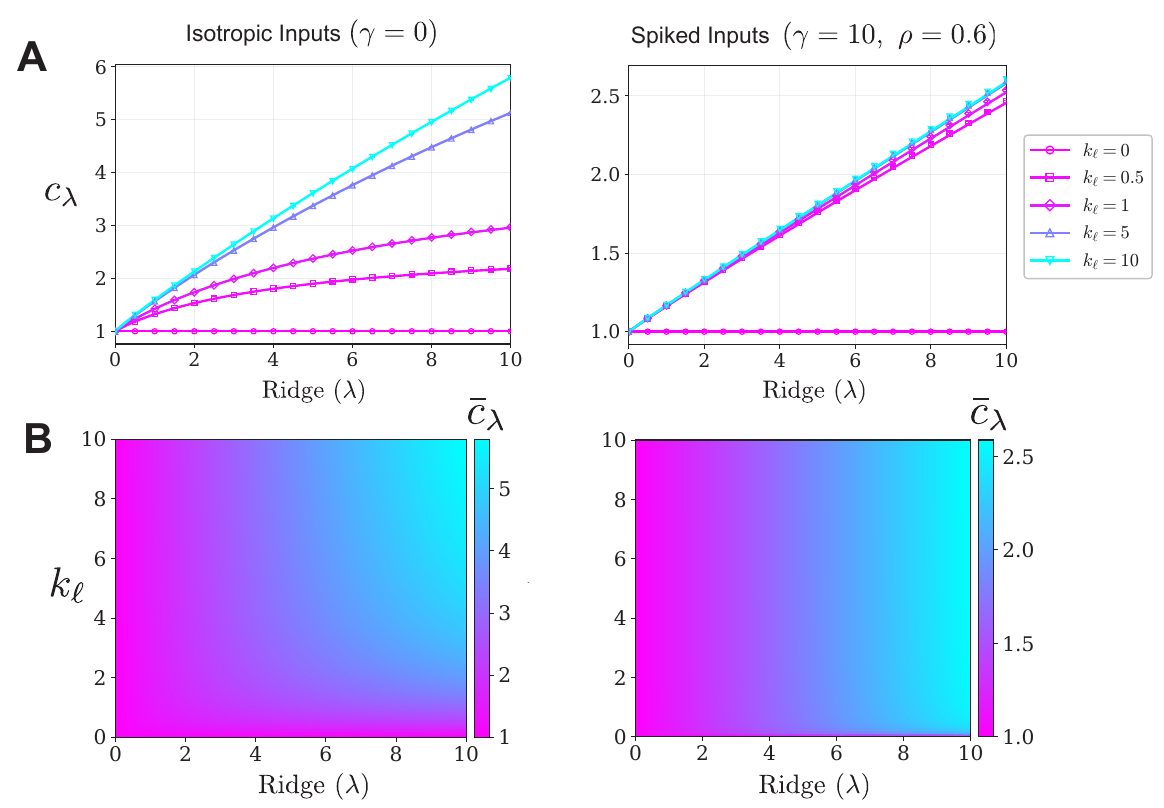}
  \caption{Asymptotics of $\cl$ across ridge strengths ($\lambda$) and feature learning strengths ($\kl$); here, $\psi = \sigma = 0.5$. \textbf{A.} $\cl$ across $\lambda$ for different choices of $\kl$, both for isotropic data (left) and spiked data (right). For the choice of spiked data (right), even small $\kl$ saturates $\cl$. Dots represent empirical simulations (with SEM; $D = 1000$, $n = 500$) and lines indicate theory ($\bcl$). \textbf{B.} Heatmaps of theoretically predicted $\bcl$ values, shown for isotropic (left) and spiked (right) data.}
  \label{fig:c_lambda_asymptotic}
\end{figure}

A proof of this theorem is provided in Appendix Sec. \ref{app:cl_thm_proof}.
Theorem \ref{thm:inflation_constant} demonstrates that $\bcl$ depends on $\lambda$ and $k_\ell$. Indeed, genuine inflation requires that $\lambda > 0$ and, crucially, that $k_\ell$ is at the right scale, i.e. $\Theta_D(1)$. Consequently, we view $\kl$ as the effective strength of the feature learning update. Ensuring that $\kl$ is at this critical scale can be achieved by 
taking a large learning rate (e.g. $\eta \asymp \sqrt{D}$, $\ell \asymp 1$), a sufficiently deep layer (e.g. $\ell \asymp \sqrt[3]{D}$, $\eta \asymp 1$), or some combination of both (e.g. $\eta \asymp D^{\frac14}$, $\ell \asymp D^{\frac16}$). The dependence of $\cl$ on $\lambda$ and $\kl$ is illustrated in Fig. \ref{fig:c_lambda_asymptotic}.

Additionally, straightforward calculation shows that
\begin{equation}    \label{eq:dc_dlambda}
    \left. \frac{\partial \bcl}{\partial \lambda} \right|_{\lambda = 0} = \frac{\frac{1}{\psi} \bkl \left( 1 + \sigma^2 + \gamma \rho^2  \right)}{1 + \bkl \left( 1 + \frac{1+\sigma^2}{\psi} + \gamma \rho^2 \left( 2 + \gamma + \frac{1}{\psi} \right) \right)} 
\end{equation}
which is positive for all $\bkl > 0$, indicating that applying mild ridge regularization $\lambda \ll 1$ causes $\bcl$ to increase. Also, for any fixed $\lambda > 0$, $\bcl$ is an increasing function in the effective feature learning strength, $\bkl$ (see Prop. \ref{prop:bcl_properties}).

\subsection{Calculating the bias, variance, and generalization error}   \label{subsec:bias_variance_gen_error_expressions}

We now write closed-form equations for the bias, variance, and generalization error of $\ffeat$. For ease of notation, here we assume that $k_\ell = O_D(1)$ is a fixed constant (independent of $D$ and $n$), so that we may write $k_\ell$ in place of $\bkl$.
Observe that the space of all possible one-step KRR estimators is characterized by the parameter space $(\lambda, k_\ell)$, with $(\lambda, 0)$ precisely corresponding to $\finit$, the baseline estimator with ridge strength $\lambda$. Thus, we calculate the asymptotic generalization error of any such estimator $\hat{f}_{\lambda;k_\ell}$ (corresponding to $(\lambda, k_\ell)$) in terms of $\lambda$ and $k_\ell$.

Using Prop. \ref{prop:inflated_estimator}, the one-step feature learning estimator corresponding to $(\lambda, k_\ell)$ is given by
\begin{equation}    \label{eq:f_lambda_kl_defn}
    \begin{aligned}
        \hat{f}_{\lambda;k_\ell}(x) = \cl \frac{x^T X A^{-1} y}{D} = \frac{\cl}{D} \left( \frac{1}{\sqrt{D}} \ws^T X + \epsilon^T \right) A^{-1} X^T x \\ 
         = \frac{\cl}{\sqrt{D}} w_\star^T \left(\frac1D X A^{-1} X^T \right) x + \frac{\cl}{D} \epsilon^T A^{-1} X^T x.
    \end{aligned}
\end{equation}
In particular, the bias, variance, and generalization error for $\hat{f}_{\lambda;k_\ell}$ may be written in closed-form in terms of $\alim$, $\blim$, $\atlim$, $\btlim$, and $\bcl$ (each of which admits closed form descriptions themselves):

\begin{thm} \label{thm:bias_variance_gen_error}
Fix any $\lambda \geq 0$ and $k_\ell \geq 0$. 
Under Assumption \ref{assump:data_gen_process}, we asymptotically obtain the following limiting formula for the bias, variance, and generalization error for the estimator $\hat{f}_{\lambda;k_\ell}$:
\begin{equation}    \label{eq:bias_limit}
 \bias(\hat{f}_{\lambda;k_\ell}) \ndlim (1 - \alim \bcl)^2 + \rho^2 \blim \bcl \left(\blim \bcl - 2 (1 - \alim \bcl) \right) + \gamma \rho^2 (1 - (\alim + \blim)\bcl)^2.    
\end{equation}

\begin{equation}    \label{eq:variance_limit}
    \variance(\hat{f}_{\lambda;k_\ell}) \ndlim \bcl^2 \left( \alim + \blim \rho^2 - \lambda \left(\atlim + \btlim \rho^2 \right) - \alim^2 - \rho^2 \left( 2\alim \blim + \blim^2 \right) + \sigma^2 \atlim \right).
\end{equation}

    \begin{multline}    \label{eq:gen_error_limit}
    \mathcal{G}(\hat{f}_{\lambda;k_\ell}) \ndlim 1 - 2\bcl \alim - 2 \rho^2 \blim \bcl + \gamma \rho^2 (1 - \bcl( \alim + \blim ))^2 + \alim \bcl^2 \\+ \rho^2 \blim \bcl^2 - \lambda \bcl^2 (\atlim + \btlim \rho^2) + \bcl^2 \sigma^2 \atlim.
\end{multline}

\end{thm}

\begin{figure}[t]
  \centering
  \includegraphics[width=\textwidth]{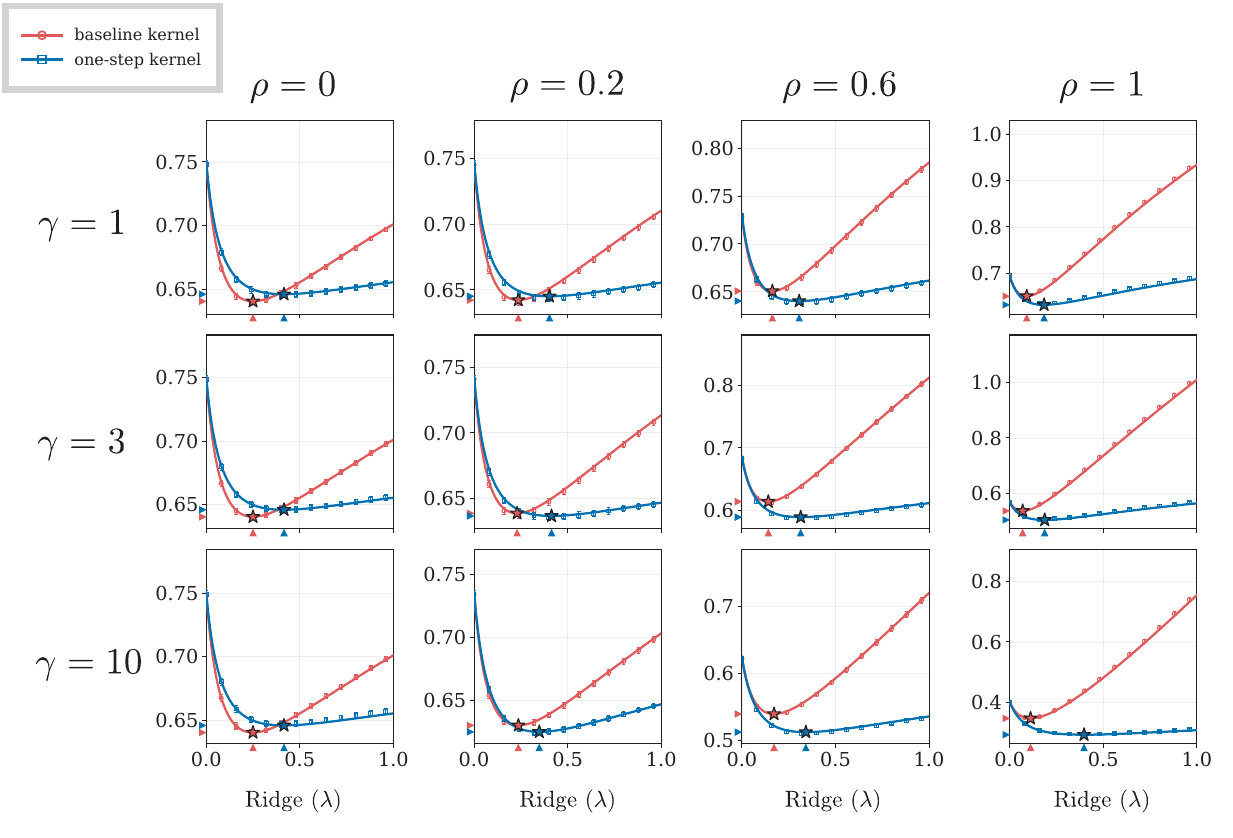}
  \caption{Comparison between the asymptotic generalization curves over ridge strengths ($\lambda$) arising from KRR with a one-step kernel with $\kl = 10$ (blue) versus that of KRR with the baseline kernel $K_x$ (red), shown across a range of $(\gamma, \rho)$ pairs. Here, $\psi = \sigma = 0.5$. Markers represent empirical simulations (with SEM; $D = 1000$, $n = 500$), and solid lines indicate theory. Stars indicate the minimum generalization error attained by either kernel.}
  \label{fig:gen_error_sweep}
\end{figure}

We refer to the limiting formulas for bias, variance, and generalization error as $\Bt(\flk)$, $\Vt(\flk)$, and $\Gt(\flk)$, respectively. This theorem is proven in Appendix Sec. \ref{app:bias_var_generror_formula_proof}. In Fig. \ref{fig:gen_error_sweep}, we demonstrate empirical validation of the predicted asymptotic generalization error $\Gt$ for one-step as well as baseline estimators across a range of ridge strengths ($\lambda$), spike strengths ($\gamma$), and spike-target alignments ($\rho$). For sufficiently large $\gamma$ and $\rho$, the lowest generalization error comes from a one-step estimator. Moreover, the minimum generalization error over $\lambda$ for the one-step estimator is consistently attained at a larger ridge strength than that of the baseline estimator (Fig. \ref{fig:gen_error_sweep}); the best one-step estimator applies more ridge-based shrinkage than the best baseline estimator, but subsequently inflates estimation by $\cl$. 

Note that when $\lambda = 0$, $\bcl = 1$ and so $\Bt(\hat{f}_{0;k_\ell}) = \Bt(\hat{f}_{0;0})$, $\Vt(\hat{f}_{0;k_\ell}) = \Vt(\hat{f}_{0;0})$, and $\Gt(\hat{f}_{0;k_\ell}) = \Gt(\hat{f}_{0;0})$ for any $k_\ell > 0$. Consequently, in the absence of ridge, the respective bias, variance, and generalization errors of the estimators $\left\{\hat{f}_{0;k_\ell} : k_\ell \geq 0\right\}$ coincide, as we have found before. Thus, to understand when feature learning yields genuinely distinct performance, we carry out a perturbative analysis of the bias and variance around $\lambda = 0$. 

In the absence of feature learning ($\bcl = 1$), there is extensive work on the classical bias-variance tradeoff in linear regression. In particular, to first order in $\lambda$ around $\lambda = 0$, the shrinkage in estimation caused by $\lambda$ typically results in the bias increasing and the variance decreasing \cite{intro-statistical-learning}. Intuitively, ridge regression favors smaller, ``simpler'' regression weights --- which penalizes the bias --- that make the estimator more robust to the particular realization of the dataset, thereby lowering the variance. In our present setting, we establish the following perturbative result around $\lambda = 0$:
\begin{prop}    \label{prop:bias_variance_lambda_derivatives}
    With the bias and variance of $\hat{f}_{\lambda; k_\ell}$ as in Theorem \ref{thm:bias_variance_gen_error} and under Assumption \ref{assump:data_gen_process}, the following two results hold:
    \begin{equation}    \label{eq:bias_lambda_derivative}
        \left. \frac{\partial \Bt(\hat{f}_{\lambda; k_\ell})}{\partial \lambda} \right|_{\lambda = 0} = 2 \psi \left( 1 - \rho^2 + \frac{\rho^2 (1 + \gamma)^2}{(1 + \gamma \psi)^3} \right) 
        - 2 \psi (1 - \psi) \left( 1 - \rho^2 + \frac{ \rho^2 \left( 1 + \gamma \right)^2}{(1 + \gamma \psi)^2} \right)  \left. \frac{\partial \bcl}{\partial \lambda} \right|_{\lambda = 0} 
    \end{equation}

    \begin{multline}    \label{eq:variance_lambda_derivative}
        \left. \frac{\partial \Vt(\hat{f}_{\lambda; k_\ell})}{\partial \lambda} \right|_{\lambda = 0} = - 2 \psi \left(1 - \rho^2 + \frac{ \rho^2 (1+\gamma)}{(1 + \gamma \psi)^3} + \frac{\sigma^2}{(1 - \psi)^3} \right) \\
        + \left( 2 \psi (1 - \psi) \left( 1 - \rho^2 + \frac{ \rho^2 (1 + \gamma)}{(1 + \gamma \psi)^2} \right) + \frac{2\psi \sigma^2}{1 - \psi} \right) \left. \frac{\partial \bcl}{\partial \lambda} \right|_{\lambda = 0}.
    \end{multline}
    
    In particular, when $k_\ell 
    = 0$, $\left. \frac{\partial \Bt(\hat{f}_{\lambda; 0})}{\partial \lambda} \right|_{\lambda = 0} > 0$ and $\left. \frac{\partial \Vt(\hat{f}_{\lambda; 0})}{\partial \lambda} \right|_{\lambda = 0} < 0$.
    
    Furthermore, for any fixed $k_\ell > 0$,
    \begin{equation}
        \left. \frac{\partial \Bt(\hat{f}_{\lambda; k_\ell})}{\partial \lambda} \right|_{\lambda = 0} - \left. \frac{\partial \Bt(\hat{f}_{\lambda; 0})}{\partial \lambda} \right|_{\lambda = 0} < 0
    \end{equation}
    and 
    \begin{equation}
        \left. \frac{\partial \Vt(\hat{f}_{\lambda; k_\ell})}{\partial \lambda} \right|_{\lambda = 0} - \left. \frac{\partial \Vt(\hat{f}_{\lambda; 0})}{\partial \lambda} \right|_{\lambda = 0} > 0.
    \end{equation}
\end{prop}

\begin{figure}[t]
  \centering
  \includegraphics[width=\textwidth]{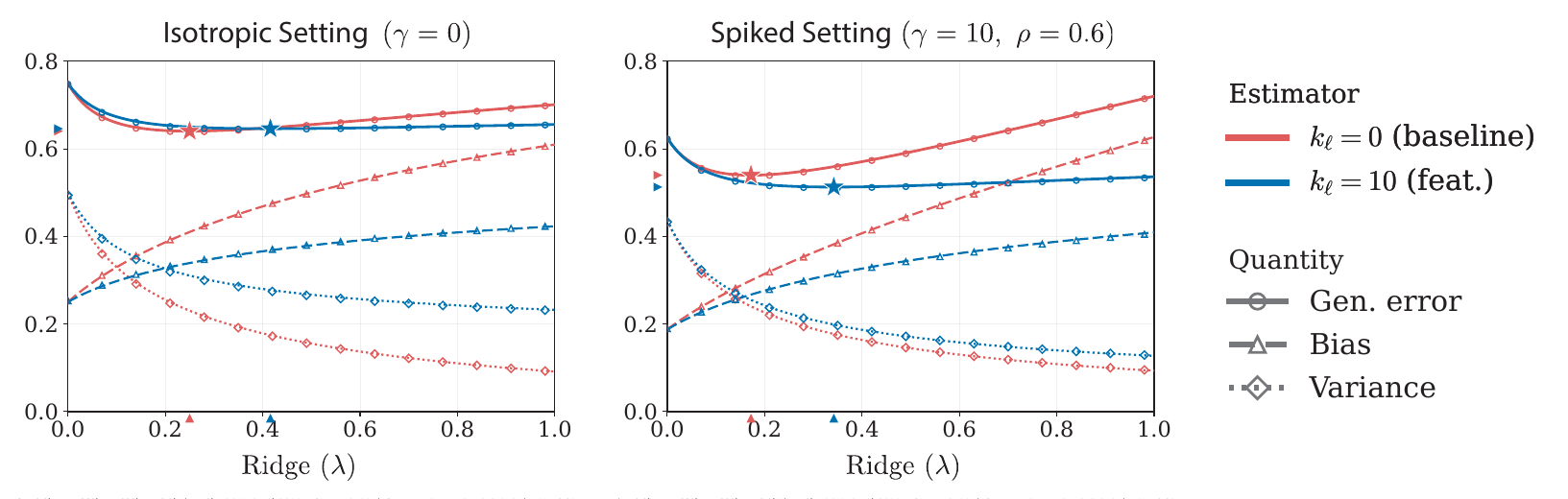}
  \caption{Decomposition of generalization error into bias and variance, for baseline and one-step estimators, as a function of ridge ($\lambda$). Here, $\psi =\sigma = 0.5$, and $\kl = 10$ for the feature learning estimators. Left: isotropic inputs ($\gamma = 0$). Right: spiked, target-aligned inputs ($\gamma = 10$, $\rho = 0.6$). Empirical simulations shown as markers ($D = 1000$, $n = 500$) and theory shown as lines. Stars mark the minimum generalization error attained by either type of estimator. }
  \label{fig:bias_variance}
\end{figure}

This result is proven in Appendix Sec. \ref{app:bias_variance_lambda_sensitivity}. Thus, when $k_\ell = 0$, we recover the expected behavior under the classical bias-variance tradeoff, with bias increasing and variance decreasing near $\lambda = 0$. 
However, when feature learning is present ($k_\ell > 0$), its effect is to facilitate a ``bias-for-variance trade'' (Fig. \ref{fig:bias_variance}). That is, for $\lambda \ll 1$, $\Bt(\hat{f}_{\lambda; k_\ell}) < \Bt(\hat{f}_{\lambda; 0})$ and $\Vt(\hat{f}_{\lambda; k_\ell}) > \Vt(\hat{f}_{\lambda; 0})$. We interpret this result as a key consequence of the interaction between ridge and feature learning: by counteracting ridge-based shrinkage, feature learning-induced inflation drives dataset-averaged estimation to become more accurate, but at the cost of amplifying estimation sensitivity to the specific realization of the data.

\subsection{Identifying when a feature learning advantage holds --- and when it does not}   \label{subsec:FLA}

We now turn to analyzing when the one step of feature learning asymptotically attains a lower generalization error than that of any baseline estimator, as well as when it does not. Fixing the ``macroscopic'' parameters $\psi$, $\gamma$, $\rho$, and $\sigma$, the generalization error arising from a given estimator $\hat{f}_{\lambda;k_\ell}$ will systematically vary as a function of the ridge strength $\lambda$ and the feature learning update strength $k_\ell$.
Here, we seek to identify the settings (characterized by the value of $\psi$, $\sigma$, $\gamma$, and $\rho$) in which feature learning-based estimation reliably lowers generalization error compared to any baseline estimator:
\begin{equation}    \label{eq:global_FLA}
    \inf_{\lambda \geq 0, k_\ell > 0} \Gt\left(\hat{f}_{\lambda; k_\ell} \right) < \inf_{\lambda \geq 0} \Gt\left(\hat{f}_{\lambda; 0} \right)
\end{equation}

When this inequality holds, we will say that there is a \emph{global feature learning advantage} (FLA). 
To develop an intuition for when a global FLA holds, we first identify settings in which one step of feature learning results in a lower generalization error \emph{locally} around $\lambda = 0$. Interestingly, the sufficient conditions we identify for a local FLA also turn out to be sufficient conditions for a global FLA. In effect, within these settings, the emergence of an FLA locally around $\lambda = 0$ extends to a local FLA around \emph{any} $\lambda \geq 0$; altogether, this results in a global FLA.
Finally, we rigorously characterize two key settings in which a global FLA \emph{does not} hold: the setting of \emph{isotropic} inputs ($\gamma = 0$) and the setting of a task-orthogonal spike ($\rho = 0$).

\subsubsection{Conditions for a local feature learning advantage}   \label{subsubsec:local_FLA}

Since the effects of feature learning only become visible when $\lambda$ is nonzero, we begin by asking when feature learning with fixed strength $k_\ell > 0$ produces a generalization advantage near $\lambda = 0$. Recalling that $\Gt\left(\hat{f}_{0; k_\ell} \right) = \Gt\left(\hat{f}_{0; 0} \right)$, we say that a \emph{local feature learning advantage} holds when, for some $\kl > 0$,
\begin{equation}    \label{eq:local_FLA_condition}
     \left. \frac{\partial}{\partial \lambda} \right|_{\lambda = 0}\Gt\left(\hat{f}_{\lambda; k_\ell} \right) < 0 \text{ and } \left. \frac{\partial}{\partial \lambda} \right|_{\lambda = 0}\Gt\left(\hat{f}_{\lambda; k_\ell} \right) < \left. \frac{\partial}{\partial \lambda} \right|_{\lambda = 0}\Gt\left(\hat{f}_{\lambda; 0} \right).
\end{equation}
Furthermore, recall that Prop. \ref{prop:bias_variance_lambda_derivatives} can be used to study the bias and variance --- and thus generalization error --- near $\lambda = 0$. 
Consequently, as stated in Prop. \ref{prop:bias_variance_lambda_derivatives}, Eqs. \ref{eq:bias_lambda_derivative} and \ref{eq:variance_lambda_derivative} may be added to obtain
\begin{equation}    \label{eq:gen_error_lambda_derivative}
    \left. \frac{\partial}{\partial \lambda} \right|_{\lambda = 0}\Gt\left(\hat{f}_{\lambda; k_\ell} \right) = 2\psi \left( \frac{ \rho^2 \gamma (1 + \gamma)}{(1 + \gamma \psi)^3} - \frac{\sigma^2}{(1-\psi)^3} \right) - 2 \psi \left( \frac{ \rho^2 (1 - \psi) \gamma (1 + \gamma)}{(1 + \gamma \psi)^2} - \frac{\sigma^2 }{1 - \psi} \right) \left. \frac{\partial \bcl}{\partial \lambda} \right|_{\lambda = 0}.
\end{equation}

Since $\left. \frac{\partial \bcl}{\partial \lambda} \right|_{\lambda = 0} \geq 0$ for $k_\ell \geq 0$ with equality if and only if $k_\ell = 0$, we state two conditions (based on Eq. \ref{eq:gen_error_lambda_derivative}) that collectively imply a local feature learning advantage:
\begin{enumerate}
    \item $2\psi \left( \frac{ \rho^2 \gamma (1 + \gamma)}{(1 + \gamma \psi)^3} - \frac{\sigma^2}{(1-\psi)^3} \right) < 0$. This is equivalent to the condition that
    \begin{equation}
    \label{eqn:SNR_init_ridge_upper_bound}
    \frac{\rho^2}{\sigma^2} < \frac{(1 + \gamma \psi)^3}{\gamma (1 + \gamma) (1 - \psi)^3}
    \end{equation}
    and it implies that $\left. \frac{\partial}{\partial \lambda} \right|_{\lambda = 0}\Gt\left(\hat{f}_{\lambda; 0} \right) < 0$. Under this condition, ridge on its own is able to lower the generalization error.

    \item $2 \psi \left( \frac{ \rho^2 (1 - \psi) \gamma (1 + \gamma)}{(1 + \gamma \psi)^2} - \frac{\sigma^2 }{1 - \psi} \right) > 0$. This is equivalent to the condition that 
    \begin{equation} \label{eqn:SNR_ridge_lower_bound}
    \frac{\rho^2}{\sigma^2} > \frac{(1 + \gamma \psi)^2}{\gamma (1 + \gamma) (1 - \psi)^2}
    \end{equation}
    and it is both a necessary and sufficient condition for $\left. \frac{\partial}{\partial \lambda} \right|_{\lambda = 0}\Gt\left(\hat{f}_{\lambda; k_\ell} \right) < \left. \frac{\partial}{\partial \lambda} \right|_{\lambda = 0}\Gt\left(\hat{f}_{\lambda; 0} \right)$. Moreover, because $\left. \frac{\partial \bcl}{\partial \lambda} \right|_{\lambda = 0}$ is increasing in $k_\ell$, this condition implies that $\left. \frac{\partial}{\partial \lambda} \right|_{\lambda = 0}\Gt\left(\hat{f}_{\lambda; k_\ell} \right) < \left. \frac{\partial}{\partial \lambda} \right|_{\lambda = 0}\Gt\left(\hat{f}_{\lambda; \kl'} \right)$ for any $\kl > \kl' \geq 0$. This means that the generalization error locally decreases even more in the presence of mild ridge when the feature learning strength is increased.
\end{enumerate}

\begin{figure}[t]
  \centering
  \includegraphics[width=0.7\textwidth]{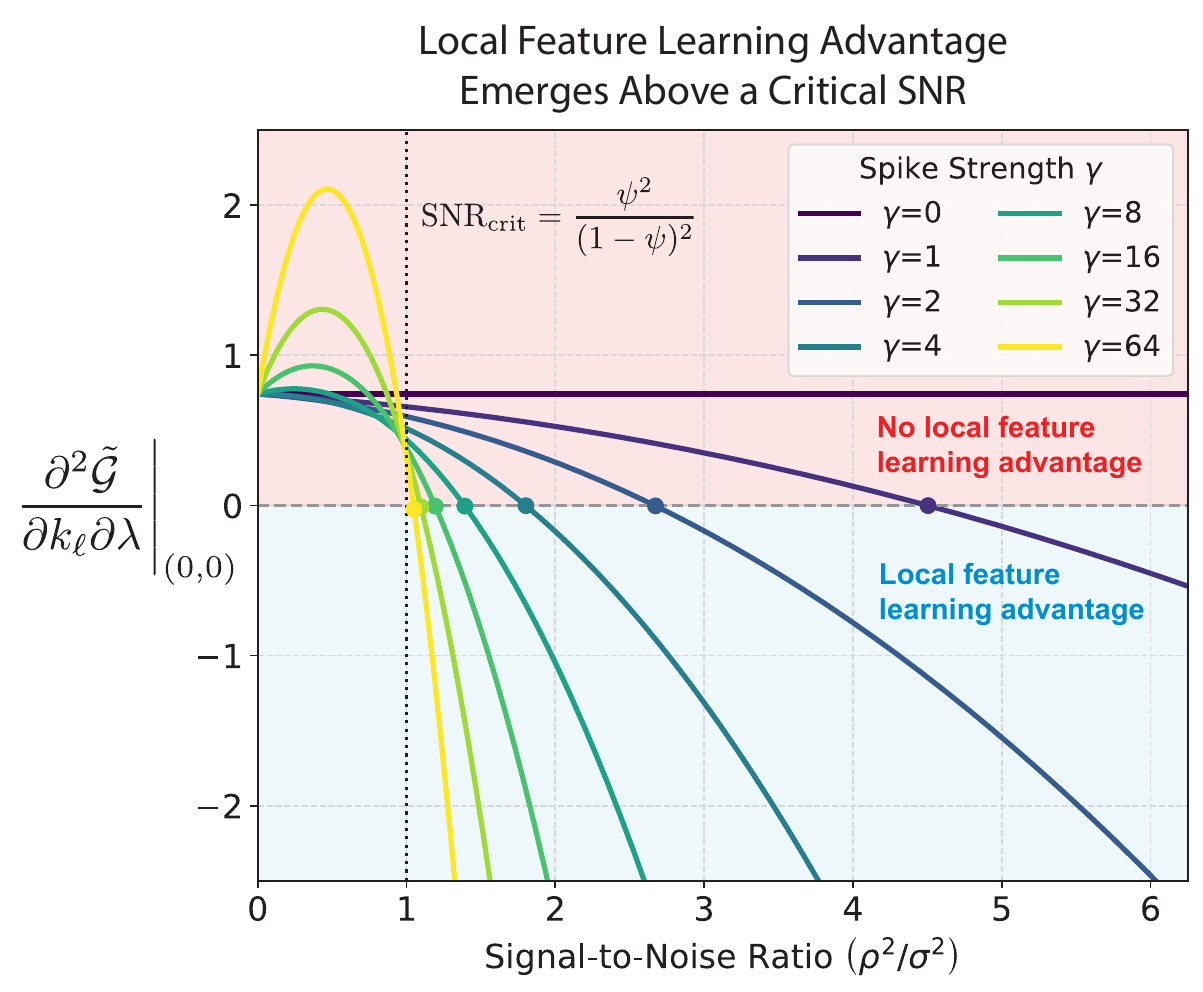}
  \caption{A diagram of $\left. \frac{\partial^2 \tilde{\mathcal{G}}}{\partial k_\ell \partial \lambda} \right|_{(0, 0)}$
  as a function of the SNR. 
  For various nonzero spike strengths, this mixed partial becomes negative for sufficiently large SNRs, implying that $\left. \frac{\partial}{\partial \lambda} \right|_{\lambda = 0}\Gt\left(\hat{f}_{\lambda; k_\ell} \right) < \left. \frac{\partial}{\partial \lambda} \right|_{\lambda = 0}\Gt\left(\hat{f}_{\lambda; 0} \right)$ for $\kl > 0$ (see Eq. \ref{eq:gen_error_lambda_derivative}). As $\gamma$ grows large, the critical SNR at which a local advantage appears descends to $\text{SNR}_{\text{crit}} := \frac{\psi^2}{(1 - \psi)^2}$. Here, $\psi = 0.5$ and $\sigma = 0.4$; as a result, $\max_\rho \rho^2/\sigma^2 = 6.25$ and $\min_{\gamma > 0} \frac{(1 + \gamma \psi)^3}{\gamma (1 + \gamma) (1 - \psi)^3} = 6 \sqrt{3} \approx 10.39$, meaning that Eq. \ref{eqn:SNR_init_ridge_upper_bound} automatically holds. Thus, in this setting, the sign of $\left. \frac{\partial^2 \tilde{\mathcal{G}}}{\partial k_\ell \partial \lambda} \right|_{(0, 0)}$ indicates whether a local FLA holds. See Appendix \ref{app:experiments} for more details regarding this mixed partial.}
  \label{fig:local_FLA}
\end{figure}

Altogether, if the condition
\begin{equation}    \label{eq:local_FLA_window}
    \frac{(1 + \gamma \psi)^2}{\gamma (1 + \gamma) (1 - \psi)^2} < \frac{\rho^2}{\sigma^2} < \frac{(1 + \gamma \psi)^3}{\gamma (1 + \gamma) (1 - \psi)^3}
\end{equation}
holds, then a local FLA will hold; moreover, if 
\begin{equation}    \label{eq:SNR_too_low_no_local_FLA}
    \frac{\rho^2}{\sigma^2} < \frac{(1 + \gamma \psi)^2}{\gamma (1 + \gamma) (1 - \psi)^2}
\end{equation}
then, in fact, a local baseline (``no learning'') advantage holds. 
Here, $\frac{\rho^2}{\sigma^2}$ may be interpreted as an \emph{effective signal-to-noise ratio} (SNR) that governs how clear spike-target alignment is compared to the noise in the targets. Ultimately, a local FLA requires that the SNR surpasses the threshold $\frac{(1 + \gamma \psi)^2}{\gamma (1 + \gamma) (1 - \psi)^2}$. Moreover, this threshold may be lowered by increasing $\gamma$, with a limiting value of $\text{SNR}_{\text{crit}} = \frac{\psi^2}{(1 - \psi)^2}$. On the other hand, from setting $\sigma^2 > 0$ and taking $\gamma \downarrow 0$, it is evident that a local FLA does not hold in the setting of isotropic inputs. We illustrate these ideas in Fig. \ref{fig:local_FLA}.

\subsubsection{From local to global: conditions for a global feature learning advantage}  \label{subsubsec:global_FLA}

Strikingly, the same SNR window that implies a local FLA (Eq. \ref{eq:local_FLA_window}) also implies that a \emph{global} FLA holds. Intuitively, this result (stated below) underscores the importance of having a local advantage around $\lambda = 0$: within the sufficient SNR window set by Eq. \ref{eq:local_FLA_window}, the local FLA around $\lambda = 0$ automatically extends to a global FLA over all $\lambda \geq 0$.

\begin{thm} \label{thm:global_FLA}
    Suppose that $\gamma > 0$ and $\sigma^2 > 0$ and that Assumption \ref{assump:data_gen_process} holds. Moreover, assume that the signal-to-noise ratio satisfies the inequality
    \begin{equation}    \label{eq:global_FLA_inequality}
        \frac{(1 + \gamma \psi)^2}{\gamma (1+\gamma) (1-\psi)^2} < \frac{\rho^2}{\sigma^2} < \frac{(1 + \gamma \psi)^3}{\gamma (1 + \gamma) (1 - \psi)^3}. 
    \end{equation}
    Then, the following results hold:
    \begin{enumerate}
        \item $\left. \frac{\partial}{\partial \lambda} \right|_{\lambda = 0}\Gt\left(\hat{f}_{\lambda; k_\ell=0} \right) < 0$, implying that for all $\lambda > 0$ suitably small, $\Gt\left(\hat{f}_{\lambda; 0} \right) < \Gt\left(\hat{f}_{0; 0} \right)$.

        \item For any $k_\ell > 0$, $\left. \frac{\partial}{\partial \lambda} \right|_{\lambda = 0}\Gt\left(\hat{f}_{\lambda; k_\ell} \right) < \left. \frac{\partial}{\partial \lambda} \right|_{\lambda = 0}\Gt\left(\hat{f}_{\lambda; 0} \right)$. This implies that for any $k_\ell > 0$, there exists some constant $\lambda^{(\kl)} > 0$ such that the inequality $\Gt\left(\hat{f}_{\lambda; k_\ell} \right) < \Gt\left(\hat{f}_{\lambda; 0}\right)$ holds for all $\lambda \in \left(0, \lambda^{(k_\ell)}\right)$.

        \item For each $\lambda > 0$, $\left. \frac{\partial}{\partial k_\ell} \right|_{k_\ell = 0} \Gt\left(\hat{f}_{\lambda; k_\ell} \right) < 0$. This means that for any $\lambda > 0$, it is possible to find some $k_\ell^{(\lambda)} > 0$ such that $\Gt\left(\hat{f}_{\lambda; k_\ell}\right) < \Gt\left( \hat{f}_{\lambda; 0} \right)$ for all $k_\ell \in (0, k_\ell^{(\lambda)})$.

        \item $\inf_{\lambda \geq 0} \Gt(\hat{f}_{\lambda;0})$ is attained at some finite ridge strength $\lambda_\star > 0$.
    \end{enumerate}
Results 1 and 2 together imply that a local FLA holds, and Results 3 and 4 together imply that a global FLA holds.
\end{thm}

\begin{rem}
    The upper bound for $\rho^2/\sigma^2$ in Ineq. \ref{eq:global_FLA_inequality} serves as a way to ensure that $\Gt(\hat{f}_{\lambda;0})$ does not attain a global minimum at $\lambda = 0$. In this case, the lower bound for $\rho^2/\sigma^2$ then provides a condition under which feature learning ($\kl > 0$) can improve generalization relative to $\hat{f}_{\lambda;0}$ at nonzero ridge strengths $\lambda$.
\end{rem}

A proof of Theorem \ref{thm:global_FLA} is provided in Appendix Sec. \ref{app:FLA_proof}.
To interpret this result, note that if $\sigma^2 > 0$ and $\gamma \uparrow \infty$, the left bound of Ineq. \ref{eq:global_FLA_inequality} descends to $\text{SNR}_{\text{crit}} = \frac{\psi^2}{(1-\psi)^2}$ and the right bound diverges, making a global FLA readily attainable past the threshold set by $\text{SNR}_{\text{crit}}$. (Note that there may be SNRs outside of the window set by Ineq. \ref{eq:global_FLA_inequality} for which a global FLA holds.) As a result, having a larger spike strength $\gamma$ naturally accommodates a wider range of SNRs. On the other hand, we may also interpret Theorem \ref{thm:global_FLA} in the following manner: for any fixed SNR that exceeds $\text{SNR}_{\text{crit}}$, it is possible to find a suitably large spike strength $\gamma$ so that the global FLA holds. Altogether, a combination of a sufficiently strong spike and a sufficiently large SNR produces a global FLA (Fig. \ref{fig:global_FLA_heatmap}).

\subsubsection{No feature learning advantage in the isotropic setting}  \label{subsubsec:isotropic_no_FLA}

\begin{figure}[t]
  \centering
  \includegraphics[width=0.8\textwidth]{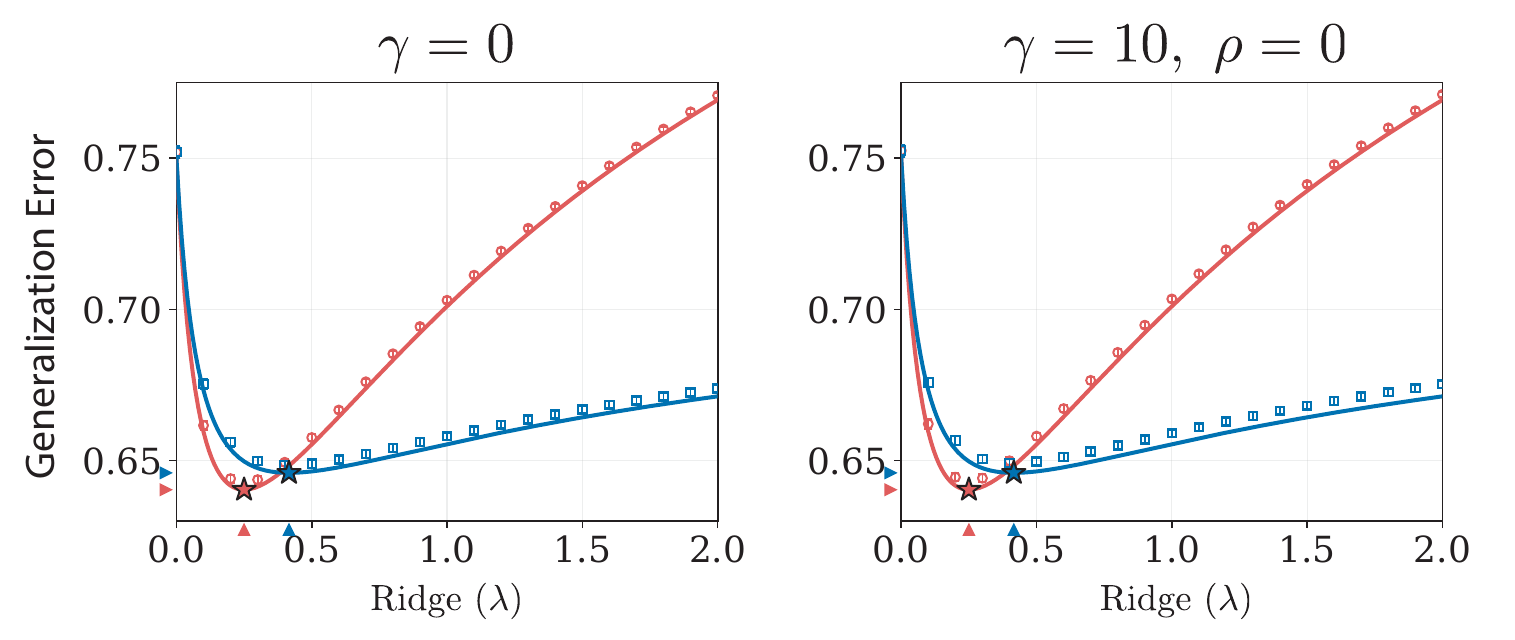}
  \caption{Comparison of asymptotic generalization error curves over $\lambda$ for one-step estimators ($\kl = 10$) and baseline estimators, shown for a setting where the inputs are isotropic (left) and a setting where the spike and target directions are orthogonal (right). Here, $\psi = \sigma = 0.5$. Markers represent empirical simulations (with SEM; $D = 3000$, $n = 1500$) and solid lines indicate theory. Stars indicate the minimum generalization error attained by either kernel. }
  \label{fig:isotropic_gen_error}
\end{figure}

Previously, we have shown that when $\gamma$ and $\frac{\rho^2}{\sigma^2}$ are each taken to be suitably large, a global FLA holds per Ineq. \ref{eq:global_FLA_inequality}. Notably, this inequality will fail to hold when $\gamma \downarrow 0$ or $\rho = 0$. Here, we rigorously prove that a global FLA fails to hold when $\gamma = 0$ --- which corresponds to a setting in which inputs are isotropic (i.e. drawn i.i.d. from $N(0, I_D)$) --- or when $\rho = 0$ --- which corresponds to a setting in which the spike direction is orthogonal to the target direction.
In particular, among all possible combinations of $(\lambda, \kl) \in \RR^2_{\geq 0}$,
we show that there exists a baseline estimator that attains the lowest possible generalization error.

\begin{thm} \label{thm:no_isotropic_FLA}
    Suppose that $\gamma = 0$ or $\rho = 0$, with $\psi \in (0, 1)$, and that Assumption \ref{assump:data_gen_process} holds. Then, if $\sigma^2 > 0$, the following results hold:
    \begin{enumerate}
        \item When viewed as a function in $\lambda$, $\Gt(\hat{f}_{\lambda; 0})$ is strictly decreasing for $\lambda \in [0, \sigma^2)$ and strictly increasing for $\lambda \in (\sigma^2, \infty)$. In particular, $$\inf_{\lambda \geq 0} \Gt(\hat{f}_{\lambda;0}) = \Gt(\hat{f}_{\sigma^2;0})$$ 
        and this minimum is uniquely attained when $\lambda = \sigma^2$.

        \item For any $\lambda, \kl \geq 0$, $$\Gt(\hat{f}_{\lambda;\kl}) \geq \Gt(\hat{f}_{\sigma^2;0})$$
        with equality if and only if $(\lambda, \kl) = (\sigma^2, 0)$.
    \end{enumerate}

    If $\gamma = 0$ or $\rho = 0$ in the noiseless regime ($\sigma^2 = 0$), then $\Gt(\hat{f}_{\lambda; 0})$ strictly increases over $\lambda \geq 0$, thereby attaining a unique minimum when $\lambda = 0$. Moreover, $\inf_{\lambda \geq 0, \kl \geq 0} \Gt(\hat{f}_{\lambda;\kl}) = \Gt(\hat{f}_{0;0})$, and this minimum is attained if and only if $\lambda = 0$ (with no restrictions on $\kl$).
\end{thm}

A proof of this result is provided in Appendix \ref{app:no_isotropic_FLA}.
Ultimately, whenever $\gamma = 0$ or $\rho = 0$, the lowest generalization error comes from a baseline estimator --- namely one where the ridge strength is tuned to $\lambda = \sigma^2$. 
Although inflation arising from one step of feature learning could partially undo the shrinkage effects of over-regularization (i.e. for large $\lambda$), it cannot outperform the optimally tuned ridge-only baseline (see Fig. \ref{fig:isotropic_gen_error}). 

We also remark that the asymptotic generalization error, as given by Eq. \ref{eq:gen_error_limit}, is the same regardless of whether $\gamma = 0$ or $\rho = 0$, since all of the $\gamma$ and $\rho$ dependencies appear together; in this sense, $\gamma$ and $\rho$ may be seen as two sides of the same coin. 
The upshot of Theorem \ref{thm:no_isotropic_FLA} is that a feature learning advantage requires a genuine spike that is at least partially aligned with the target direction.

\section{Discussion}

In this paper, we studied the earliest signatures of feature learning in a deep linear network: the change in the layer-$\ell$ feature kernel under a single gradient step, the effect of this change on estimators obtained by kernel ridge regression (KRR), as well as the generalization error arising from these estimators under KRR. 

Three aspects of our analysis are worth emphasizing. First, the effective strength of one step of feature learning is governed by both the learning rate $\eta$ (as demonstrated by prior works) and the layer index $\ell$, highlighting in particular that utilizing depth is another mechanism for increasing the strength of feature learning without needing to scale the learning rate. Second, our work highlights the tension between the uniform inflation arising from one gradient step and the shrinkage arising from ridge regularization, which act as opposing forces along the bias-variance tradeoff; ridge tilts the balance towards lowering variance and inflation tilts the balance towards lowering the bias. Third, we find that a feature learning advantage (at the level of one step) is not possible unless genuine task-relevant low-rank structure is present in the data. This suggests that gradient descent is especially well-suited to learning structured low-dimensional information embedded in otherwise high-dimensional data, even early in training.

We now highlight key limitations of our work, alongside important future directions.

\paragraph{Going beyond linearity.} Our closed-form analysis is afforded, in part, by the linearity of our deep networks. Carrying out similar analyses with deep nonlinear networks could provide a richer description of the linear and higher-order nonlinear modes of target functions that can be efficiently learned early in training. 
As a salient point of contrast with our present work, Ba et al. (2022) \cite{ba2022onestep} find that for two-layer \emph{nonlinear} networks, features learned under one gradient step can improve generalization in a setting with isotropic inputs, thereby removing the need for low-rank structure in the inputs.
Thus, future work would involve characterizing the feature learning advantages (and limitations) that arise from nonlinear computation.

\paragraph{Going beyond one step.} An important simplification in our analysis is to study a single step of gradient descent, because it makes the learned feature kernel $\Phi^\ell_{t}$ exactly computable. Nonetheless, a natural next step is to study how $\Phi^\ell_t$ evolves across several gradient steps, how it depends on learning rate and depth, and how it might further widen the feature learning advantage. Such an analysis would come closer to the long training-time setting studied by Bordelon \& Pehlevan (2022) \cite{bordelon2022dmft} while providing greater transparency of what features are learned.

\paragraph{Going beyond one spike.}
Here, we studied data distributions in which the inputs have one ``spike'' direction. A natural extension is to study data distributions consisting of multiple ``spikes'', or in which the target functions are multi-index models \cite{dandi2024onegiantstep}. This would raise new questions as to whether distinct task-relevant directions can be learned within a single gradient step or whether this requires multiple steps.

\paragraph{Studying the predictive power of one gradient step.} The main limitation of a one-step analysis is that it omits a description of learning later in training. However, it remains to explore whether the feature learning phenomena observed at the level of one step meaningfully predict longer-term learning behaviors. In particular, establishing a link between early signatures of learning and longer-term generalization error could inform optimal architectural and training paradigms for learning particular data distributions.

\paragraph{Acknowledgments}
SB was supported by an NIH training grant (T32MH065214) and is currently supported by a DoD National Defense Science and Engineering Graduate (NDSEG) fellowship. This material is based upon work supported by the Air Force Office of Scientific Research
under award number FA9550-25-C-B010 in the amount of \$144,500.
BH is supported by a 2024 Sloan Fellowship in Mathematics, NSF CAREER grant DMS-2143754, NSF grant DMS-2133806, and DARPA AIQ grant (HR001124S0029).

\bibliographystyle{unsrt}  
\bibliography{references} 

\section*{Appendix}
\appendix

\renewcommand{\thefigure}{S\arabic{figure}}
\setcounter{figure}{0}

\renewcommand{\thetable}{S\arabic{table}}
\setcounter{table}{0}

\renewcommand{\theequation}{\thesection\arabic{equation}}

\paragraph{AI usage.} In carrying out this research, we used AI to help run numerical experiments and generate figures. We also used it to explore possible strategies for proving certain results (e.g. Thm. \ref{thm:feature_kernel_limits}, Thm. \ref{thm:no_isotropic_FLA}). Finally, we used AI to verify the correctness of our proofs and overall prose. 

\section{Experimental Details}  \label{app:experiments}

All figures presented in the main text were generated using the NumPy, SciPy, and Matplotlib libraries in Python. All experiments were either run locally or on a high-performance computing cluster composed of CPU nodes supporting serial and parallel workloads. All code will be made publicly available within a GitHub repository upon publication. 

Here, we present further details of the experiments and analyses used to generate each of the figures in the main text.

\paragraph{A note regarding asymptotic limits.} As described in the paper, our methodology is to first take the $N \to \infty$ limit and next to take the $n, D \to \infty$ (proportional) limit. Thus, in generating figures other than Fig. \ref{fig:one_step_kernels_finite_width} (which provides numerical validation for Thm. \ref{thm:feature_kernel_limits} for large, finite $N$), we assume that the $N \to \infty$ limit has already been taken, giving rise to the inflationary constant $\cl$. In these figures, we focus on empirically validating the results we obtain under the second limit, where $n/D \to \psi \in (0, 1)$ and the data is sampled from a spiked population model (as stated in Assumption \ref{assump:data_gen_process}).

\paragraph{Validation of one-step feature kernel update at finite width (Fig. \ref{fig:one_step_kernels_finite_width}).} Here, we elaborate on the procedure used to generate Fig. \ref{fig:one_step_kernels_finite_width}. We first generated a single dataset $\{(x_i, y_i)\}_{i=1}^n$, with $D = 30$ and $n = 10$, where $x_i \iid N(0, I_D)$ and $y_i = \frac{1}{\sqrt{D}} \ws^T x_i + \epsilon_i$ for some randomly sampled vector $\ws \in \sqrt{D} \SS^{D-1}$, and where $\epsilon_i \iid N(0, \sigma^2)$. Here, we set $\sigma = 0.3$. For any given width $N \in \{100, 500, 1000, 5000\}$, across each of $50$ independent trials, we randomly initialized a deep linear network as described in Sec. \ref{sec:setup}, with $L = 10$. In each trial, we then simulated one step of gradient descent on the aforementioned dataset, with step size $\eta = 0.5$, from that trial's random initialization. Subsequently, across each $\ell \in \{1, 2, \dots, 10\}$, we calculated $\Delta \Phi^\ell := \Phi^\ell_{t=1} - \Phi^\ell_{t=0}$. We computed the \emph{cosine similarity} between $\Delta \Phi^\ell$ and $K_x yy^T K_x$ as
\begin{equation}    \label{eq:cosine_sim}
     \frac{\Tr\left( \left(\Delta \Phi^\ell\right) K_x yy^T K_x \right)}{\norm{\Delta \Phi^\ell}_F \norm{K_x yy^T K_x}_F}.
\end{equation}
Additionally, noting the theoretical prediction $\Delta \Phi^\ell \Lconverge \frac{\eta^2 d_\ell}{n^2} K_x yy^T K_x$ as $N \to \infty$, we next empirically estimated the depth coefficient $d_\ell$ as
\begin{equation}    \label{eq:depth_coeff}
    \hat{d}_\ell = \frac{\Tr\left( \left(\Delta \Phi^\ell\right) K_x yy^T K_x \right)}{\norm{K_x yy^T K_x}_F^2} \cdot \frac{n^2}{\eta^2}.
\end{equation}
For each chosen width $N$, we plot the trial-averaged cosine similarity measure (in Fig. \ref{fig:one_step_kernels_finite_width}A) and the empirically estimated depth coefficient (in Fig. \ref{fig:one_step_kernels_finite_width}B) across the layer index $\ell$.

\paragraph{Generalization error curves (Figs. \ref{fig:global_FLA_heatmap}, \ref{fig:gen_error_sweep}, \ref{fig:bias_variance}, \ref{fig:isotropic_gen_error}).} Here, we describe the procedure used to find the location of the minimum asymptotic generalization error over ridge strength ($\lambda$), marked as stars in each of the indicated figures in the paper. For any fixed values of $\psi$, $\gamma$, $\rho$, $\sigma$, and $\kl$, we compute the asymptotic generalization error $\tilde{\mathcal{G}}(\flk)$ across $m = 200$ values of $\lambda$ on a log-spaced grid between $0.001$ and $1000$ as well as at $\lambda = 0$. Based on preliminary exploration, we chose $\lambda = 1000$ as a reasonable upper bound beyond which we do not expect $\Gt(\flk)$ to have a minimum. Write the collection of all of these values of $\lambda$ as $0 = \lambda_0 < \lambda_1 < \dots < \lambda_m = 1000$. We then collect every $i \in \{0, 1, \dots, m\}$ such that $\lambda_i$ results in a smaller asymptotic generalization error than its immediate neighbors on the grid; we record each such $i$ because the asymptotic generalization error can be non-monotonic in $\lambda$ (e.g. see Fig. \ref{fig:global_FLA_heatmap}B). We also mark the lowest generalization error value attained over these $\lambda_i$. Then, for each of these identified $i$, we apply scipy.optimize.minimize\_scalar to find the minimum asymptotic generalization error in the range $\left[\lambda_{\max(0, i-1)}, \lambda_{\min(i+1, m)} \right]$. Finally, we return the smallest asymptotic generalization error value obtained across all of these scipy minimizations, as well as over the original grid points $\lambda_i$.

\paragraph{SNR heatmap (Fig. \ref{fig:global_FLA_heatmap}).} This figure shows a heatmap of the quantity 
\begin{equation}
    \Delta \Gt(\gamma, \text{SNR}) := \min_{\lambda \in [0, 1000]} \Gt(\hat{f}_{\lambda;\kl=10}) - \min_{\lambda \in [0, 1000]} \Gt(\hat{f}_{\lambda;\kl=0})
\end{equation}
across a grid of values of $\gamma$ and the SNR. We also show a dotted gray contour line to indicate where $\Delta \Gt(\gamma, \text{SNR}) = 0$. Additionally, we visualize the upper and lower bound curves from Thm. \ref{thm:global_FLA} between which a global feature learning advantage provably holds, as given by Eq. \ref{eq:global_FLA_inequality}.

\paragraph{Empirical calculation of inflation constant, bias, variance, and generalization error (Figs.~\ref{fig:global_FLA_heatmap}, \ref{fig:c_lambda_asymptotic}--\ref{fig:bias_variance}, \ref{fig:isotropic_gen_error}).} All empirical calculations were averaged across $100$ independent draws of the dataset. Given $n$, $D$, $\gamma$, $\rho$, and $\sigma$, first, a random unit $v \in \SS^{D-1}$ was sampled, allowing for calculation of the spiked covariance matrix $\Sigma = I_D + \gamma vv^T$. Additionally, a random vector $\ws \in \sqrt{D} \SS^{D-1}$ was sampled such that $\frac{1}{\sqrt{D}} \ws^T v = \rho$. Subsequently, in each of the $100$ independent draws of the data, $n$ input-target pairs $\left\{(x_i, y_i)\right\}_{i=1}^n$ were sampled (i.i.d.) according to the data-generating process listed in Assumption \ref{assump:data_gen_process} (and using the same, fixed $v$ and $\ws$ across these $100$ realizations). Each of these datasets was used to produce an estimate of the quantity of interest (e.g. asymptotic generalization error, etc.); in our figures, we show the average value across the $100$ realizations of the training data. 

For any given realization of the data, $X \in \RR^{D \times n}$ (inputs) and $y \in \RR^n$ (targets); any fixed value of $k_\ell \geq 0$; and any ridge strength $\lambda \geq 0$, we calculate $w_{\init}$, $\cl$, and $w_{\feat} = \cl w_{\init}$, as described in Sec. \ref{subsec:KRR_estimator} of the paper. To calculate $\cl$, we note that $\frac{\eta^2 d_\ell}{n^2} = k_\ell \frac{D}{n^2}$; $k_\ell$ abstracts away the specific step size and the layer index used. We then calculate the generalization error of the estimator $\hat{f}(x) = \frac{1}{\sqrt{D}}w^T x$, with $w \in \{w_{\init}, w_{\feat}\}$ over this particular dataset as
\begin{equation}
    \hat{\mathcal{G}}_{X, y, \lambda} := \frac{(w - \ws)^T \Sigma (w - \ws)}{D} = \EE_{x \sim N(0, \Sigma)} \left[ \left( \hat{f}(x) - f_\star(x) \right)^2 \right]
\end{equation}
where $f_\star(x) := \frac{1}{\sqrt{D}} \ws^T x$. Taking an average over $100$ draws of the dataset then yields an estimate of the generalization error, as defined in Eq. \ref{eq:gen_error}.

Next, to estimate the asymptotic bias, we first calculated the dataset-averaged vectors $\bar{w}_{\init}$ and $\bar{w}_{\feat}$ (over $100$ dataset draws), where $\bar{w} = \frac{1}{100}\sum_{i=1}^{100} w^{(i)}$ (and $w^{(i)}$ is the estimate from the $i$th dataset draw). Subsequently, for each $w \in \{w_{\init}, w_{\feat}\}$, we estimated the bias as 
\begin{equation}
    \hat{\mathcal{B}}_{X, y, \lambda} := \frac{(\bar{w} - \ws)^T \Sigma (\bar{w} - \ws)}{D} = \EE_{x \sim N(0, \Sigma)} \left[ \left( {f_b}(x) - f_\star(x) \right)^2 \right].
\end{equation}
Here, ${f_b}(x) := \frac{1}{\sqrt{D}} \bar{w}^T x$ is an estimate of $\EE_{\mathcal{D}}[\hat{f}(x)]$; as a result, our empirical calculation of bias lines up with the theoretical calculation in Eq. \ref{eq:bias_variance}.

Finally, to estimate the asymptotic variance, first let $w^{(1)}, \dots, w^{(100)}$ denote the empirically calculated regression weights that all correspond to the baseline estimator or to the one-step estimator; let $\hat{f}^{(i)}(x) := \frac{1}{\sqrt{D}} {w^{(i)}}^T x$. We then estimated the asymptotic variance as
\begin{equation}
    \begin{aligned}
        \hat{\mathcal{V}}_{X, y, \lambda} := \left(\frac{1}{100} \sum_{i=1}^{100}\frac{{w^{(i)}}^T \Sigma w^{(i)}}{D}\right) - \frac{\bar{w}^T \Sigma \bar{w}}{D} = \left( \frac{1}{100} \sum_{i=1}^{100} \EE_{x \sim N(0, \Sigma)}\left[ \left( \hat{f}^{(i)}(x) \right)^2  \right]\right) - \EE_{x \sim N(0, \Sigma)} \left[ {f_b}(x)^2 \right] \\
        = \left( \frac{1}{100} \sum_{i=1}^{100} \EE_{x \sim N(0, \Sigma)}\left[ \left( \hat{f}^{(i)}(x) \right)^2  \right]\right) - \EE_{x \sim N(0, \Sigma)} \left[ \left( \frac{1}{100} \sum_{j=1}^{100} \hat{f}^{(j)}(x) \right)^2 \right].
    \end{aligned}
\end{equation}
This provides an estimate for the quantity 
$$\EE_{\mathcal{D}}\EE_{x \sim N(0, \Sigma)}\left[ \hat{f}(x)^2 \right] - \EE_{x \sim N(0, \Sigma)} \left[ \EE_{\mathcal{D}} \left[ \hat{f}(x)  \right]^2 \right]$$
$$ = \EE_{x \sim N(0, \Sigma)} \left[ \EE_{\mathcal{D}}\left[ \hat{f}(x)^2 \right] - \EE_{\mathcal{D}} \left[ \hat{f}(x)  \right]^2  \right]$$
$$ = \EE_{x \sim N(0, \Sigma)} \EE_{\mathcal{D}} \left[ \left( \hat{f}(x) - \EE_{\mathcal{D}} \left[ \hat{f}(x)  \right] \right)^2  \right] = \variance(\hat{f})$$
which lines up with our definition of the variance (Eq. \ref{eq:bias_variance}).

\paragraph{Calculation of the mixed partial (Fig. \ref{fig:local_FLA}).} For completeness, here we explicitly calculate the mixed partial $\left. \frac{\partial^2 \tilde{\mathcal{G}}}{\partial k_\ell \partial \lambda} \right|_{(0, 0)}$. Starting from Eq. \ref{eq:gen_error_lambda_derivative}, we obtain
\begin{equation}
    \begin{aligned}
        \left. \frac{\partial^2 \tilde{\mathcal{G}}}{\partial k_\ell \partial \lambda} \right|_{(0, 0)} = - 2 \psi \left( \frac{ \rho^2 (1 - \psi) \gamma (1 + \gamma)}{(1 + \gamma \psi)^2} - \frac{\sigma^2 }{1 - \psi} \right) \left. \frac{\partial^2 \bcl}{\partial \kl \partial \lambda} \right|_{(\lambda, \kl) = (0, 0)} \\
        = - 2 \psi \left( \frac{ \rho^2 (1 - \psi) \gamma (1 + \gamma)}{(1 + \gamma \psi)^2} - \frac{\sigma^2 }{1 - \psi} \right) \left(\frac{1}{\psi} (1 + \sigma^2 + \gamma \rho^2) \right) \\
        = \frac{-2 (1 + \sigma^2 + \gamma \rho^2) \sigma^2 \gamma (1+\gamma) (1-\psi)}{(1 + \gamma \psi)^2} \left( \frac{\rho^2}{\sigma^2} - \frac{(1+ \gamma \psi)^2}{\gamma (1 + \gamma) (1 - \psi)^2} \right).
    \end{aligned}
\end{equation} 
Fig. \ref{fig:local_FLA} plots this quantity as a function of the SNR $\rho^2/\sigma^2$ across a range of values of $\gamma$ when $\psi = 0.5$ and $\sigma = 0.4$ are fixed.
In particular, $\left. \frac{\partial^2 \tilde{\mathcal{G}}}{\partial k_\ell \partial \lambda} \right|_{(0, 0)} < 0$ precisely implies that $\frac{\rho^2}{\sigma^2} > \frac{(1+ \gamma \psi)^2}{\gamma (1 + \gamma) (1 - \psi)^2}$, with the converse being true when $\gamma, \sigma^2 > 0$. 

\section{Some Intermediate Results}

\subsection{Results from Linear Algebra}

\begin{lemma}   \label{lemma:orthogonal_symmetry}
    (a) Suppose that $M \in \RR^{N \times N}$ is a matrix satisfying the following property: for any orthogonal matrix $U \in \RR^{N \times N}$, the equation $M = UMU^T$ holds. Then, $M = \lambda I_N$, for some $\lambda \in \RR$.

    (b) Suppose that $M \in \RR^{N \times N}$ is a matrix and $v \in \SS^{N-1} \subset \RR^N$ is a unit vector that collectively satisfy the following property: for any orthogonal matrix $U \in \RR^{N \times N}$ that fixes $v$ (i.e. $Uv = v$), the equation $M = UMU^T$ holds. Then, there exist $\lambda_1, \lambda_2 \in \RR$ such that $M = \lambda_1 I_N + \lambda_2 vv^T$.
\end{lemma}

\begin{proof}
    (a) First, consider any unit vector $w \in \SS^{N-1}$. Since $UM = MU$ for all orthogonal matrices $U$, this identity holds for all orthogonal matrices that fix $w$. In particular, for any orthogonal matrix $U$ that fixes $w$, we have that $Mw = MUw = UMw$, i.e. $U$ also fixes $Mw$. We claim that $Mw = \lambda_w w$, for some $\lambda_w \in \RR$. Indeed, we may first write the decomposition $Mw = \lambda_w w + w_\perp$, where $\lambda_w \in \RR$ and $w_\perp$ is some vector in $\left(\text{span}(\{w\})\right)^\perp$. Then, if $UMw = Mw$ for all orthogonal matrices $U$ that fix $w$, we obtain
    $$\lambda_w w + w_\perp = Mw = UMw = U \left( \lambda_w w + w_\perp \right) = \lambda_w w + U w_\perp.$$ This implies that for all orthogonal matrices $U$ that fix $w$, $w_\perp = U w_\perp$. If $w_\perp \neq 0$, then there exists some orthogonal matrix $U_0$ that fixes $w$ but does \emph{not} fix $w_\perp$. Indeed, identifying $\left(\text{span}(\{w\})\right)^\perp$ with $\RR^{N-1}$ and noting that the orthogonal matrices on $\RR^{N-1}$ act transitively on $\SS^{N-2}$, any matrix in $\RR^{N \times N}$ that fixes $w$ and acts as an orthogonal matrix on $\left(\text{span}(\{w\})\right)^\perp \cong \RR^{N - 1}$ that shifts $w_\perp$ will suffice. For such a matrix $U_0$, this would mean that $U_0 w_\perp \neq w_\perp$. Consequently, we must have that $w_\perp = 0$. Thus, we have found that $Mw = \lambda_w w$; that is, for any unit vector $w \in \SS^{N - 1}$, there exists some $\lambda_w \in \RR$ such that $Mw = \lambda_w w$. In particular, every unit vector in $\RR^N$ is an eigenvector of $M$.

    Now, we leverage the fact that the group of orthogonal matrices on $\RR^N$ acts transitively on $\SS^{N-1}$. In particular, for any unit vectors $w_1, w_2 \in \SS^{N-1}$, there exists some orthogonal matrix $U$ such that $Uw_1 = w_2$. Then, $$\lambda_{w_2} w_2 = M w_2 = M U w_1 = UM w_1 = U (\lambda_{w_1} w_1) = \lambda_{w_1} w_2.$$ Consequently, we find that $\lambda_{w_1} = \lambda_{w_2}$. Thus, every unit vector in $\RR^N$ is an eigenvector of $M$ with the same eigenvalue, which implies that $M$ is a scalar multiple of $I_N$.

    (b) Because $MU = UM$ for all orthogonal matrices $U$ that fix $v$, our argument from part (a) shows that $Mv = \lambda_v v$, for some $\lambda_v \in \RR$. Observe also that transposing this identity yields $U^T M^T = M^T U^T$ for all orthogonal matrices $U$ that fix $v$. Moreover, since $U^T$ is the inverse of $U$, $U$ fixing $v$ is equivalent to $U^T$ fixing $v$; thus, we also have that $U M^T = M^T U$ for all orthogonal matrices $U$ fixing $v$. Following the same argument as in (a) again yields $M^T v = \lambda_v' v$, for some $\lambda_v' \in \RR$.
    
    Now, consider any $w \in W := \left(\text{span}(\{v\})\right)^\perp$. Note then that for any orthogonal matrix $U$ that fixes $v$, $$\left\langle Uw, v \right\rangle = \left\langle Uw, Uv \right\rangle = \left\langle w, v \right\rangle = 0$$ meaning that $Uw \in W$. Moreover, 
    $$\left\langle Mw, v \right\rangle = w^T M^T v = \lambda_v' w^T v = 0$$ meaning that $Mw \in W$ as well. As a result, $M$ may be viewed as a block diagonal operator on $\text{span}(\{v\}) \oplus W$. In particular, the identity $MU = UM$ implies that $M |_W U|_W = U|_W M|_W$ for any orthogonal matrix $U$ that fixes $v$; furthermore, the set of such matrices corresponds to operators that fix $\text{span}(\{v\})$ and act as any orthogonal matrix on $W$. Thus, identifying $W$ with $\RR^{N-1}$, we have that $M|_W V = V M|_W$ for all orthogonal matrices $V \in \RR^{(N-1) \times (N-1)}$. By part (a), this implies that $M|_W = \lambda_1 I_N|_W$, for some $\lambda_1 \in \RR$. Combining this result with the fact that $Mv = \lambda_v v$ and defining $\lambda_2 := \lambda_v - \lambda_1$, we find that $M = \lambda_1 I_N + \lambda_2 vv^T$.
\end{proof}

\subsection{Results from Probability Theory}

\begin{defn}    \label{defn:aseq_Leq}
    For two sequences of random variables $(X_n)_{n \geq 1}$ and $(Y_n)_{n \geq 1}$, we say that $X_n \aseq Y_n$ if $X_n - Y_n \asconverge 0$ as $n \to \infty$.
    Similarly, if $(X_n)_{n \geq 1}$ and $(Y_n)_{n \geq 1}$ are sequences of square-integrable random variables (i.e. in $L^2$), we say that $X_n \Leq Y_n$ if $X_n - Y_n \Lconverge 0$. Additionally, let $\Pconverge$ denote convergence in probability.
\end{defn}

\begin{prop}    \label{prop:as_p_convergence}
    Suppose that $(X_n)_{n \geq 1}$ and $(Y_n)_{n \geq 1}$ are sequences of random variables defined in a probability space $(\Omega, \mathcal{F}, \mu)$ such that $X_n \asconverge x$ and $Y_n \asconverge y$ as $n \to \infty$, where $x$ and $y$ are deterministic constants. Then, the following statements hold:

    (a) $X_n + Y_n \asconverge x + y$.

    (b) $X_n Y_n \asconverge xy$.

    (c) If $y \neq 0$, then $\frac{X_n}{Y_n} \asconverge \frac{x}{y}$.

    Moreover, analogous results as (a)-(c) hold for convergence in probability, provided that $X_n \Pconverge x$ and $Y_n \Pconverge y$.
\end{prop}

\begin{rem}
    A key implication of this result is that almost sure convergence, or convergence in probability, to a deterministic constant is preserved under any standard algebraic operation (i.e. addition, subtraction, multiplication, division), provided that there is no division by zero involved. For instance, if $X_n \asconverge x$ and $Y_n \asconverge y \geq 0$, then $\frac{X_n}{1 + Y_n} \asconverge \frac{x}{1 + y}$. Observe that no assumptions of independence between $X_n$ and $Y_n$ are needed.
\end{rem}

For a proof of these statements for both forms of convergence, see van der Vaart (1998) \cite{Vaart_1998}, Theorems 2.3 and 2.7. This result is a consequence of the Continuous Mapping Theorem. 

\begin{lemma}   \label{lemma:borel_cantelli_as_convergence}
    Suppose $(X_n)_{n \geq 1}$ is a sequence of random variables defined on the probability space $(\Omega, \mathcal{F}, \mu)$. Let $x \in \RR$ be a nonrandom constant. Suppose that, for any $\epsilon > 0$, the sum $$\sum_{n \geq 1} \PP(|X_n - x| \geq \epsilon)$$ converges. Then, $X_n \asconverge x$.
\end{lemma}

\begin{proof}
    This result follows readily from the first Borel-Cantelli lemma. For any $\epsilon > 0$, define the events $E_n \subset \Omega$ by $E_n := \{\omega \in \Omega : |X_n(\omega) - x| \geq \epsilon\}$. Then, since $\PP(|X_n - x| \geq \epsilon) = \mu(E_n)$, the given assumption amounts to $\sum_{n \geq 1} \mu(E_n) < \infty$. As a result, the first Borel-Cantelli lemma implies that $\mu(\lim \sup_{n \to \infty} E_n) = 0$ where $$\lim \sup_{n \to \infty} E_n = \bigcap_{n \geq 1} \bigcup_{k \geq n} E_k$$ denotes the set of all $\omega \in \Omega$ that lie in infinitely many $E_n$. This means that there exists a set $\Omega_\epsilon \subset \Omega$ with $\mu(\Omega_\epsilon) = 1$ such that for each $\omega \in \Omega_\epsilon$, there exists some $N(\omega) \in \ZZ_{+}$ such that $|X_n(\omega) - x| < \epsilon$ for all $n \geq N(\omega)$. Now, fix any sequence $(\epsilon_i)_{i \geq 1}$ with $\epsilon_i \to 0$, and define $\tilde{\Omega} := \bigcap_{i \geq 1} \Omega_{\epsilon_i}$. Because $\mu(\Omega_{\epsilon_i}) = 1$ and $\tilde{\Omega}$ is a countable intersection, it follows that $\mu\left(\tilde{\Omega}\right) = 1$. By definition, for each $\omega \in \tilde{\Omega}$ and any $\epsilon > 0$, there is some $j \in \ZZ_{+}$ such that $\epsilon_j \leq \epsilon$; accordingly, there exists some $N_j(\omega) \in \ZZ_{+}$ such that $|X_n(\omega) - x| < \epsilon_j \leq \epsilon$ for all $n \geq N_j(\omega)$. This precisely implies that $\lim_{n \to \infty} X_n(\omega) = x$ for all $\omega \in \tilde{\Omega}$, which implies that $X_n \asconverge x$ because 
    $$\PP\left( \left\{ \omega \in \Omega : \lim_{n \to \infty} X_n(\omega) = x \right\} \right) = 1$$
    owing to $\mu\left(\tilde{\Omega}\right) = 1$. 
\end{proof}

\begin{prop}    \label{prop:generalized_DCT}
    Suppose that $(X_n)_{n \geq 1}$ is a sequence of random variables defined in a probability space $(\Omega, \mathcal{F}, \mu)$, such that $X_n \geq 0$ for each $n \geq 1$ and $X_n \asconverge x$ where $x$ is a nonrandom constant. Suppose that $(Y_n)_{n \geq 1}$ is another sequence of random variables in this probability space such that $Y_n \asconverge y$, $Y_n \geq X_n$ for each $n \geq 1$, $\EE[Y_n]$ exists for each $n \geq 1$, and $\lim_{n\to\infty}\EE[Y_n] = y$. Then, for each $n \geq 1$, $\EE[X_n]$ exists, and $\lim_{n\to\infty}\EE[X_n] = x$.
\end{prop}

\begin{proof}
    First, $\EE[X_n]$ exists for each $n \geq 1$ because $0 \leq X_n \leq Y_n$ and $\EE[Y_n]$ exists.
    Since $X_n \geq 0$ and $X_n \asconverge x$, Fatou's Lemma implies that 
    $$x = \lim\inf_{n\to\infty} X_n = \EE\left[ \lim\inf_{n\to\infty} X_n \right] \leq \lim\inf_{n \to \infty} \EE\left[ X_n \right].$$
    On the other hand, since $Y_n - X_n \geq 0$ for each $n \geq 1$ and $Y_n - X_n \asconverge y-x$, Fatou's Lemma implies that
    $$y-x = \EE\left[ \lim\inf_{n\to\infty} \left( Y_n - X_n \right) \right] \leq \lim\inf_{n\to\infty} \EE\left[ Y_n - X_n \right]$$
    $$ = \lim\inf_{n\to\infty} \EE[Y_n] - \lim\sup_{n\to\infty} \EE[X_n] = y - \lim\sup_{n\to\infty} \EE[X_n]$$
    $$ \implies \lim\sup_{n\to\infty} \EE[X_n] \leq x.$$
    Here, we have used the fact that $y = \lim_{n\to\infty} \EE[Y_n] = \lim\inf_{n\to\infty} \EE[Y_n]$. Altogether, we have found that
    $$x \leq \lim\inf_{n \to \infty} \EE\left[ X_n \right] \leq \lim\sup_{n \to \infty} \EE\left[ X_n \right] \leq x$$
    thus establishing that $\lim_{n\to\infty}\EE[X_n]$ exists and is equal to $x$.
\end{proof}

\begin{thm}[Vitali's Convergence Theorem] \label{thm:vitali}
    Suppose that $(X_n)_{n \geq 1}$ is a sequence of random variables defined in a probability space $(\Omega, \mathcal{F}, \mu)$ and that $X$ is another random variable in this space such that $X_n \Pconverge X$. Then, $X_n \overset{L^1}{\longrightarrow} X$ if and only if the $(X_n)_{n \geq 1}$ are uniformly integrable. 
\end{thm}

\begin{proof}
    See Dembo (2023) \cite{dembo2023probability}, Theorem 1.3.49.
\end{proof}

\begin{cor} \label{cor:Lp_convergence_Vitali}
    Suppose that $X_n \asconverge X$ in the probability space $(\Omega, \mathcal{F}, \mu)$. Suppose, moreover, that each $X_n \in L^p(\Omega, \mathcal{F}, \mu)$ for some $p \geq 1$. If there exists some $k > 0$ and $K \in \ZZ_+$ such that $\sup_{n \geq K} \EE\left[ |X_n|^{p + k} \right] < \infty$, then $\EE[ |X_n|^p] \to \EE[|X|^p]$ and $\EE[X_n] \to \EE[X]$.
\end{cor}

\begin{proof}
    Since $X_n \asconverge X$, it follows that $X_n \Pconverge X$ and that $|X_n|^p \Pconverge |X|^p$ (e.g. by van der Vaart (1998) \cite{Vaart_1998}, Theorems 2.3 and 2.7). Now, define $E := \sup_{n \geq K} \EE\left[ |X_n|^{p + k} \right]$. By Fatou's Lemma,
    $$\EE[|X|^{p+k}] \leq \lim \inf_{n \to \infty} \EE[|X_n|^{p+k}] \leq E.$$
    Thus, for all $n \geq K$, $X_n \in L^{p+k}(\Omega, \mathcal{F}, \mu)$ and $X \in L^{p+k}(\Omega, \mathcal{F}, \mu)$, meaning that these variables are also in $L^{p'}(\Omega, \mathcal{F}, \mu)$ for any $1 \leq p' \leq p+k$.
    
    Note that the $(|X_n|^p)_{n \geq K}$ are uniformly integrable because for any $M > 0$, $$|X_n|^p \mathds{1}_{\{|X_n| > M\}} \leq \frac{|X_n|^{p+k}}{M^k} \mathds{1}_{\{|X_n| > M\}} \leq \frac{|X_n|^{p+k}}{M^k}.$$ Thus, the uniform integrability of $|X_n|^p$ follows from the fact that 
    $$\lim_{M \to \infty} \left( \sup_{n \geq K} \int |X_n|^p \mathds{1}_{\{|X_n| > M\}} d\mu  \right) \leq \lim_{M \to \infty} \frac{1}{M^k} \left( \sup_{n \geq K} \int |X_n|^{p+k}  d\mu  \right)$$
    $$ = \lim_{M \to \infty} \frac{1}{M^k} \left( \sup_{n \geq K} \EE\left[ |X_n|^{p+k} \right] \right) = \lim_{M \to \infty} \frac{E}{M^k} = 0.$$
    As a result, Vitali's Convergence Theorem (Thm. \ref{thm:vitali}) implies that $|X_n|^p \overset{L^1}{\longrightarrow} |X|^p$, which means that 
    $$\left| \EE[|X_n|^p] - \EE[|X|^p] \right| \leq \EE\left[ \left| |X_n|^p - |X|^p \right| \right]  \overset{n \uparrow \infty}{\longrightarrow} 0.$$
    Thus, $\EE[|X_n|^p] \to \EE[|X|^p]$.
    
    Next, to see why $\EE[X_n] \to \EE[X]$, observe that $|X_n - X| \asconverge 0$ and that
    $$\sup_{n \geq K} \EE\left[ |X_n - X|^{p+k} \right] \leq \sup_{n \geq K} \EE\left[ |2 \max\left(|X_n|, |X| \right)|^{p+k} \right] \leq 2^{p+k} \left( \sup_{n \geq K} \EE[|X_n|^{p+k}] + \sup_{n \geq K} \EE[|X|^{p+k}] \right) < \infty$$
    where we have used the aforementioned facts that $X_n, X \in L^{p+k}(\Omega, \mathcal{F}, \mu)$.
    Thus, applying the first part of this corollary with $p' = 1$ and $k':= p+k-1 > 0$ and $X_n$ and $X$ replaced by $X_n - X$ and $0$, respectively, yields $\EE[|X_n-X|] = \EE[|X_n - X|^{p'}] \to \EE[0] = 0$ because $\sup_{n \geq K} \EE[|X_n - X|^{p' + k'}] = \sup_{n \geq K} \EE[|X_n - X|^{p+k}] < \infty$. Consequently,
    $$\left| \EE[X_n] - \EE[X] \right| \leq \EE\left[ \left| X_n - X \right| \right] \overset{n \uparrow \infty}{\longrightarrow} 0.$$
    Thus, $\EE[X_n] \to \EE[X]$.
\end{proof}

\begin{cor}     \label{cor:expectation_factorizes_Lp}
    Suppose that $X_n \asconverge x$ and $Y_n \asconverge y$ in the probability space $(\Omega, \mathcal{F}, \mu)$, where $x$ and $y$ are nonrandom constants. Suppose, moreover, that each $X_n, Y_n \in L^{4p}(\Omega, \mathcal{F}, \mu)$ for some $p \geq 1$. If $\sup_{n \geq K} \EE\left[ |X_n|^{4p} \right] < \infty$ and $\sup_{n \geq K} \EE\left[ |Y_n|^{4p} \right] < \infty$ for some $K \in \ZZ_+$, then $\EE[|X_n Y_n|^p] \to |xy|^p$ and $\EE[X_n Y_n] \to xy$. 
\end{cor}

\begin{proof}
    Define $E_X := \sup_{n \geq K} \EE\left[ |X_n|^{4p} \right]$ and $E_Y := \sup_{n \geq K} \EE\left[ |Y_n|^{4p} \right]$. 
    By the Cauchy-Schwarz inequality, $$\sup_{n \geq K} \EE[|X_n Y_n|^{2p}] \leq E_X^{1/2} E_Y^{1/2} < \infty.$$
    Since $X_n Y_n \asconverge xy$ (Prop. \ref{prop:as_p_convergence}), Cor. \ref{cor:Lp_convergence_Vitali} implies that $\EE[|X_n Y_n|^p] \to |xy|^p$ and that $\EE[X_n Y_n] \to xy$.
\end{proof}

\begin{defn}    \label{defn:uniform_Lp_control}
    We say that a sequence of random variables $(X_n)_{n \geq 1}$ defined on the probability space $(\Omega, \mathcal{F}, \mu)$ is moment controlled if there exists $K \in \ZZ_+$ such that $\sup_{n \geq K} \EE[|X_n|^p] < \infty$ for all $p \geq 1$. Moreover, we say that $(X_n)_{n \geq 1}$ is \emph{quotient controlled} if $X_n > 0$ almost everywhere and $X_n^{-1}$ is moment controlled.
\end{defn}

\begin{prop}    \label{prop:uniform_Lp_control}
    Suppose that $(X_n)_{n \geq 1}$ and $(Y_n)_{n \geq 1}$ are sequences of random variables defined on the probability space $(\Omega, \mathcal{F}, \mu)$ such that $\sup_{n \geq K} \EE[|X_n|^p] < \infty$ and $\sup_{n \geq K} \EE[|Y_n|^p] < \infty$ for all $p \geq 1$ and some fixed $K \geq 1$. Then, the following results hold:

    (a) $\sup_{n \geq K} \EE[|X_n + Y_n|^p] < \infty$ for all $p \geq 1$.

    (b) $\sup_{n \geq K} \EE[|X_n Y_n|^p] < \infty$ for all $p \geq 1$.

    (c) If there exists a nonrandom constant $\delta > 0$ such that $Y_n > \delta$ almost everywhere, then $\sup_{n \geq K} \EE[|\frac{X_n}{Y_n}|^p] < \infty$.

    (d) If $Y_n > 0$ almost everywhere and $\sup_{n \geq K} \EE[|Y_n|^{-p}] < \infty$ for all $p \geq 1$, then $\sup_{n \geq K} \EE[|\frac{X_n}{Y_n}|^p] < \infty$.
    
\end{prop}

\begin{proof}
    (a) Note that $$||X_n + Y_n||_{L^p}^p \leq \left(||X_n||_{L^p} + ||Y_n||_{L^p}\right)^p \leq 2^{p-1}\left( \EE[|X_n|^p] + \EE[|Y_n|^p] \right)$$
    where we have used the $C_r$-inequality:
    $|x+y|^r \leq 2^{r-1}(|x|^r + |y|^r)$, for $r \geq 1$, which straightforwardly follows from convexity of the function $x \mapsto |x|^r$. As a result, for any $p \geq 1$,
    $$\sup_{n \geq K} \EE[|X_n + Y_n|^p] = \sup_{n \geq K} ||X_n + Y_n||_{L^p}^p \leq 2^{p-1} \left( \sup_{n \geq K} \EE[|X_n|^p] + \sup_{n \geq K} \EE[|Y_n|^p] \right) < \infty.$$

    (b) By the Cauchy-Schwarz inequality, for any $p \geq 1$,
    $$\sup_{n \geq K} \EE[|X_n Y_n|^p] \leq \sup_{n \geq K} \left( \EE[|X_n |^{2p}]^{\frac12} \EE[|Y_n |^{2p}]^{\frac12} \right) \leq \left( \sup_{n \geq K} \EE[|X_n |^{2p}] \right)^{\frac12} \left( \sup_{n \geq K} \EE[|Y_n |^{2p}] \right)^{\frac12} < \infty.$$

    (c) Since $Y_n > \delta > 0$ almost everywhere, $0 < \frac{1}{Y_n} < \frac{1}{\delta}$ almost everywhere. Then, for any $p \geq 1$, 
    $$\sup_{n \geq K} \EE\left[\left|\frac{1}{Y_n}\right|^p\right] \leq \frac{1}{\delta^p} < \infty.$$
    As a result, for any $p \geq 1$, $$\sup_{n \geq K} \EE\left[ \left| \frac{X_n}{Y_n} \right|^p \right] = \sup_{n \geq K} \EE\left[ \left|X_n \cdot \frac{1}{Y_n}\right|^p \right] < \infty$$
    using our result from (b). 

    (d) Since $Y_n > 0$ almost everywhere, $\frac{1}{Y_n}$ is well-defined (and positive) almost everywhere and it moreover has uniformly bounded $p$th moments for any $p \geq 1$. Consequently, the same argument as in (b) applied to $X_n$ and $\frac{1}{Y_n}$ shows that $\sup_{n \geq K} \EE[|\frac{X_n}{Y_n}|^p] < \infty$.
\end{proof}

\begin{lemma}   \label{lemma:multiplying_adding_L2_limits}
    Suppose that $(X_n)_{n \geq 1}$ and $(Y_n)_{n \geq 1}$ are each moment controlled sequences, and let $x, y \in \RR$ be nonrandom constants. If $X_n \Lconverge x$, then $X_n \Lpconverge x$ for all $p \geq 1$. Moreover, if $X_n \Lconverge x$ and $Y_n \Lconverge y$, then $X_n + Y_n \Lconverge x + y$ and $X_n Y_n \Lconverge xy$. Finally, if $X_n \Lconverge 0$, then $X_n Y_n \Lconverge 0$ as well.
\end{lemma}

\begin{proof}
    First, if $X_n \Lconverge x$, then $X_n - x \Lconverge 0$ and so $\EE[|X_n - x|^p] \leq \EE\left[ \left( X_n - x \right)^2 \right]^{p/2} \to 0$ for any $1 \leq p < 2$. For $p > 2$, we use the elementary interpolation inequality over $L^p$ space (a consequence of Hölder's inequality): there exists some $\theta \in (0, 1)$ such that $\frac1p = \frac{1 - \theta}{2} + \frac{\theta}{2p}$, and for this $\theta$,
    $$\norm{X_n - x}_{L^p} \leq ||X_n - x||_{L^2}^{1-\theta} ||X_n - x||_{L^{2p}}^{\theta}.$$
    Since $||X_n - x||_{L^2} \to 0$ by assumption and $||X_n - x||_{L^{2p}}$ is uniformly bounded (as $X_n$ is moment controlled), it follows that $\norm{X_n - x}_{L^p} \to 0$, thus proving that $X_n \Lpconverge x$ when $p > 2$ as well.

    As a result, when $X_n$ and $Y_n$ are both moment controlled, with $X_n \Lconverge x$ and $Y_n \Lconverge y$, then $X_n + Y_n$ and $X_n Y_n$ are also moment controlled (Prop. \ref{prop:uniform_Lp_control}). Moreover, using the triangle inequality,
    $$\norm{X_n + Y_n - (x+y)}_{L^2} \leq \norm{X_n - x}_{L^2} + \norm{Y_n - y}_{L^2} \to 0$$
    $$\norm{X_n Y_n - xy}_{L^2} = \norm{X_n(Y_n - y) + y (X_n - x)}_{L^2} \leq \norm{X_n (Y_n - y)}_{L^2} + |y| \cdot \norm{X_n - x}_{L^2} \to 0$$
    where we have noted that $$\norm{X_n (Y_n - y)}_{L^2} = \EE\left[ X_n^2 (Y_n - y)^2 \right]^{1/2} \leq \EE\left[ X_n^4\right]^{1/4} \EE\left[ (Y_n - y)^4 \right]^{1/4} = \norm{X_n}_{L^4} \cdot \norm{Y_n - y}_{L^4}$$
    $$ \leq \left( |x| + \norm{X_n - x}_{L^4} \right) \cdot \norm{Y_n - y}_{L^4} \to 0.$$
    Here, we have applied the Cauchy-Schwarz inequality and noted that $X_n \Lpconverge x$ and $Y_n \Lpconverge y$ for all $p \geq 1$ (e.g. $p = 4$). Finally, if $X_n \Lconverge 0$ with $Y_n$ moment controlled, then
    $$\norm{X_nY_n}_{L^2} = \EE\left[ X_n^2 Y_n^2 \right]^{1/2} \leq \EE\left[ X_n^4 \right]^{1/4} \EE\left[ Y_n^4 \right]^{1/4} \to 0.$$
    Here, we have used that $X_n \Lpconverge 0$ for $p = 4$ (by the first part of the lemma) and that $Y_n$ is moment controlled.
\end{proof}

\begin{lemma}   \label{lemma:cvg_in_prob_to_Lp}
    Suppose that $(X_n)_{n \geq 1}$ is a sequence of random variables such that $X_n \Pconverge x$, for some nonrandom constant $x \in \RR$, and that $(X_n)_{n \geq 1}$ is moment controlled. Then, $X_n \Lpconverge x$ for any $p \geq 1$.
\end{lemma}

\begin{proof}
    Because $X_n \Pconverge x$, it is the case that $\lim_{n\to\infty}\PP(|X_n - x| > \epsilon) = 0$ for any fixed $\epsilon > 0$. Then,
    $$\EE\left[ |X_n - x|^p \right] = \EE\left[ |X_n - x|^p \mathds{1}_{\{ |X_n - x| \leq \epsilon \}} \right] + \EE\left[ |X_n - x|^p \mathds{1}_{\{ |X_n - x| > \epsilon \}} \right]$$
    $$ \leq \epsilon^p + \EE\left[ |X_n - x|^{2p} \right]^{1/2} \EE\left[ \mathds{1}_{\{ |X_n - x| > \epsilon \}} \right]^{1/2}$$
    $$ = \epsilon^p + \PP(|X_n - x| > \epsilon)^{1/2} \cdot \EE\left[ |X_n - x|^{2p} \right]^{1/2} \overset{n \uparrow \infty}{\longrightarrow} \epsilon^p.$$
    Here, we have noted that $X_n$ is moment controlled and that $X_n \Pconverge x$. Thus, 
    $$\lim\sup_{n\to\infty} \EE\left[ |X_n - x|^p \right] \leq \epsilon^p$$
    for any $\epsilon > 0$, implying that $\lim_{n \to \infty} \EE\left[ |X_n - x|^p \right] = 0$, i.e. $X_n \Lpconverge x$.
\end{proof}

\subsection{Some Facts about \texorpdfstring{$L^2$}{L2} Convergence}

\begin{lemma}   \label{lemma:gaussian_vector_concentration}
    Suppose that $x, y \in \RR^N$ are sampled iid from $\mathcal{N}(0, I_N)$. Then, as $N \to \infty$, $\frac1N ||x||_2^2 \overset{L^2}{\longrightarrow} 1$ and $\frac{x^T y}{N} \overset{L^2}{\longrightarrow} 0$.
\end{lemma}

\begin{proof}
    This is a standard concentration result. Note first that $\EE\left[ \frac1N ||x||_2^2 \right] = \frac1N \sum_{i=1}^N \EE[x_i^2] = \frac1N \sum_{i=1}^N 1 = 1$, and that 
    $$\EE\left[ \left( \frac1N ||x||_2^2 \right)^2 \right] = \frac{1}{N^2} \sum_{i, j = 1}^N \EE[x_i^2 x_j^2] = \frac{1}{N^2} \sum_{i=1}^N \EE[x_i^4] + \frac{1}{N^2} \sum_{i \neq j} \EE[x_i^2] \EE[x_j^2]$$
    $$ = \frac{1}{N^2} \sum_{i=1}^N 3 + \frac{1}{N^2} \sum_{i \neq j} 1 = \frac{3N}{N^2} + \frac{N(N-1)}{N^2} = 1 + \frac2N$$ meaning that $\Var\left(  \frac1N ||x||_2^2\right) = O\left( \frac1N \right)$, thereby establishing that $\frac1N ||x||_2^2 \overset{L^2}{\longrightarrow} 1$. 

    Second, note that $\EE\left[ \frac{x^T y}{N} \right] = 0$, so $$\Var\left( \frac{x^T y}{N} \right) = \EE\left[ \left(\frac{x^T y}{N}\right)^2 \right] = \frac{1}{N^2} \sum_{i, j=1}^N \EE[x_i y_i x_j y_j] = \frac{1}{N^2} \sum_{i, j=1}^N \EE[x_i x_j] \EE[y_i y_j]$$
    $$ = \frac{1}{N^2} \sum_{i, j=1}^N \delta_{ij}^2 = \frac1N$$ implying that $\frac{x^T y}{N} \overset{L^2}{\longrightarrow} 0$ as $N \to \infty$.
\end{proof}

\begin{lemma}   \label{lemma:weight_matrix_proj_concentration}
    (a) Suppose that $a, b \in \RR^q$ are fixed vectors and that $W \in \RR^{p \times q}$ is a random matrix, each of whose entries is iid standard Gaussian. Then, as $p \to \infty$, $$\frac{\left( \frac{1}{\sqrt{q}}Wa \right)^T \left( \frac{1}{\sqrt{q}}Wb \right)}{p} \overset{L^2}{\longrightarrow} \frac{a^T b}{q}.$$
    
    (b) Suppose that for each $p \geq 1$, $a_p, b_p \in \RR^p$ are random vectors (which are not necessarily independent), such that as $p \to \infty$ we have that $$\frac{a_p^T a_p}{p} \overset{L^2}{\longrightarrow} a_0, \text{ } \frac{b_p^T b_p}{p} \overset{L^2}{\longrightarrow} b_0, \text{ and } \frac{a_p^T b_p}{p} \overset{L^2}{\longrightarrow} c_0$$ where $a_0$, $b_0$, and $c_0$ are all deterministic constants independent of $p$. 
    Suppose that $W \in \RR^{p \times p}$ is a random matrix, each of whose entries is iid standard Gaussian and independent from $a_p$ and $b_p$. Then, as $p \to \infty$, 
    $$\frac{\left( \frac{1}{\sqrt{p}}Wa_p \right)^T \left( \frac{1}{\sqrt{p}}Wb_p \right)}{p} \overset{L^2}{\longrightarrow} c_0.$$

\end{lemma} 

\begin{proof}
    (a) First, note that $\frac{\left( \frac{1}{\sqrt{q}}Wa \right)^T \left( \frac{1}{\sqrt{q}}Wb \right)}{p} = \frac{a^T \left( \frac1p W^T W\right) b}{q}$, and 
    $$\EE\left[ \frac{a^T \left( \frac1p W^T W\right) b}{q} \right] = \frac{1}{pq}\sum_{i, j = 1}^{q} \sum_{k = 1}^p a_i b_j \EE[W_{ki} W_{kj}] = \frac{1}{pq}\sum_{i, j = 1}^{q} \sum_{k = 1}^p a_i b_j \delta_{ij}$$
    $$ = \frac{1}{pq}\sum_{i = 1}^{q} \sum_{k = 1}^p a_i b_i = \frac{a^T b}{q}.$$ 
    Next, we consider the second moment:
    $$ \EE\left[ \left( \frac{a^T \left( \frac1p W^T W\right) b}{q} \right)^2 \right] = \frac{1}{p^2 q^2} \sum_{i, j, k, l = 1}^{q} \sum_{r, s = 1}^p a_i b_j a_k b_l \EE[W_{ri} W_{rj} W_{sk} W_{sl}].$$
    By Wick's theorem, we have that $$\EE[W_{ri} W_{rj} W_{sk} W_{sl}] = \EE[W_{ri} W_{rj}] \EE[ W_{sk} W_{sl}] + \EE[W_{ri} W_{sk}] \EE[ W_{rj} W_{sl}] + \EE[W_{ri} W_{sl}] \EE[ W_{rj} W_{sk}]$$
    $$ = \delta_{ij}\delta_{kl} + \delta_{rs}\delta_{ik}\delta_{jl} + \delta_{rs}\delta_{il}\delta_{jk}.$$ 
    Thus, we find that
    $$\EE\left[ \left( \frac{a^T \left( \frac1p W^T W\right) b}{q} \right)^2 \right] = \left(\frac{1}{p^2 q^2} \sum_{i, k = 1}^{q} \sum_{r, s = 1}^p a_i b_i a_k b_k \right) + \left( \frac{1}{p^2 q^2} \sum_{i, j = 1}^{q} \sum_{r = 1}^p a_i^2 b_j^2 \right) + \left( \frac{1}{p^2 q^2} \sum_{i, j = 1}^{q} \sum_{r = 1}^p a_i b_i a_j b_j \right)$$
    $$ = \left(\frac{a^T b}{q}\right)^2 + \frac1p \left( \frac{a^T a}{q} \right) \left( \frac{b^T b}{q} \right) + \frac1p \left( \frac{a^T b}{q} \right)^2.$$ 
    In particular, as $p \to \infty$, we find that 
    $$\EE\left[ \left( \frac{a^T \left( \frac1p W^T W\right) b}{q} - \frac{a^T b}{q} \right)^2 \right] = \EE\left[ \left( \frac{a^T \left( \frac1p W^T W\right) b}{q} \right)^2 \right] - \left( \frac{a^T b}{q} \right)^2$$
    $$ = \frac1p \left( \frac{a^T a}{q} \right) \left( \frac{b^T b}{q} \right) + \frac1p \left( \frac{a^T b}{q} \right)^2 \to 0.$$ Therefore, $\frac{\left( \frac{1}{\sqrt{q}}Wa \right)^T \left( \frac{1}{\sqrt{q}}Wb \right)}{p} \overset{L^2}{\longrightarrow} \frac{a^T b}{q}.$

    (b) For shorthand, we simply refer to $a_p$ and $b_p$ by $a$ and $b$, and note that $p$ is taken to $\infty$. We proceed by a similar argument as in (a). First, note that 
    $$\EE \left[ \frac{\left( \frac{1}{\sqrt{p}}Wa \right)^T \left( \frac{1}{\sqrt{p}}Wb \right)}{p} \right] = \frac{1}{p^2} \sum_{i, j = 1}^{p} \sum_{k = 1}^p \EE[a_i b_j] \EE[W_{ki} W_{kj}] = \frac{1}{p^2}\sum_{i, j = 1}^{p} \sum_{k = 1}^p \EE[a_i b_j] \delta_{ij}$$
    $$ = \frac{1}{p} \sum_{i=1}^p \EE[a_i b_i] = \EE\left[ \frac{a^T b}{p} \right] \overset{p \nearrow \infty}{\longrightarrow} c_0$$ where to obtain this latter limit we have noted that $L^2$ convergence to a deterministic constant implies convergence in expectation to that same constant. Next, we consider the second moment (and, again, rely on Wick contractions):
    $$\EE\left[ \left( \frac{a^T \left( \frac1p W^T W\right) b}{p} \right)^2 \right] = \frac{1}{p^4} \sum_{i, j, k, l = 1}^{p} \sum_{r, s = 1}^p \EE[a_i b_j a_k b_l] \EE[W_{ri} W_{rj} W_{sk} W_{sl}]$$
    $$ = \left(\frac{1}{p^4} \sum_{i, k = 1}^{p} \sum_{r, s = 1}^p \EE[a_i b_i a_k b_k] \right) + \left( \frac{1}{p^4} \sum_{i, j = 1}^{p} \sum_{r = 1}^p \EE[a_i^2 b_j^2] \right) + \left( \frac{1}{p^4} \sum_{i, j = 1}^{p} \sum_{r = 1}^p \EE[a_i b_i a_j b_j] \right)$$
    $$ = \EE\left[\left(\frac{a^T b}{p}\right)^2\right] + \frac1p \EE\left[\left( \frac{a^T a}{p} \right) \left( \frac{b^T b}{p} \right)\right] + \frac1p \EE\left[\left( \frac{a^T b}{p} \right)^2\right]$$
    $$ \leq \left( 1 + \frac1p \right) \EE\left[\left(\frac{a^T b}{p}\right)^2\right] + \frac1p \sqrt{\EE\left[ \left(\frac{a^T a}{p}\right)^2 \right] \EE\left[ \left(\frac{b^T b}{p}\right)^2 \right]}.$$
    Here, we have used the Cauchy-Schwarz inequality to deduce that $\EE\left[\left( \frac{a^T a}{p} \right) \left( \frac{b^T b}{p} \right)\right] \leq \sqrt{\EE\left[ \left(\frac{a^T a}{p}\right)^2 \right] \EE\left[ \left(\frac{b^T b}{p}\right)^2 \right]}$. Altogether, we find that 
    $$\EE\left[ \left( \frac{a^T \left( \frac1p W^T W\right) b}{p} - c_0 \right)^2 \right] = \EE\left[ \left( \frac{a^T \left( \frac1p W^T W\right) b}{p} \right)^2 \right] - 2 \EE \left[ \frac{\left( \frac{1}{\sqrt{p}}Wa \right)^T \left( \frac{1}{\sqrt{p}}Wb \right)}{p} \right] c_0 + c_0^2$$
    $$ = \EE\left[ \left( \frac{a^T \left( \frac1p W^T W\right) b}{p} \right)^2 \right] - 2 \EE \left[ \frac{a^T b}{p} \right] c_0 + c_0^2$$
    $$ \leq \left( 1 + \frac1p \right) \EE\left[\left(\frac{a^T b}{p}\right)^2\right] + \frac1p \sqrt{\EE\left[ \left(\frac{a^T a}{p}\right)^2 \right] \EE\left[ \left(\frac{b^T b}{p}\right)^2 \right]} - 2 \EE \left[ \frac{a^T b}{p} \right] c_0 + c_0^2$$
    $$ \leq  \EE\left[\left(\frac{a^T b}{p} - c_0 \right)^2 \right] + \frac1p \EE\left[\left(\frac{a^T b}{p}\right)^2\right] + \frac1p \sqrt{\EE\left[ \left(\frac{a^T a}{p}\right)^2 \right] \EE\left[ \left(\frac{b^T b}{p}\right)^2 \right]}.$$ This latter quantity tends to $0$ as $p \to \infty$ because $\frac{a^T b}{p} \overset{L^2}{\longrightarrow} c_0$ and $$\lim_{p \to \infty}\EE\left[\left(\frac{a^T b}{p}\right)^2\right] + \sqrt{\EE\left[ \left(\frac{a^T a}{p}\right)^2 \right] \EE\left[ \left(\frac{b^T b}{p}\right)^2 \right]} = c_0^2 + a_0b_0 = O_p(1)$$ owing to the fact that each of $\frac{a^Ta}{p}$, $\frac{b^T b}{p}$, and $\frac{a^T b}{p}$ converges in $L^2$ to $a_0$, $b_0$, and $c_0$, respectively. We conclude that $\frac{a^T \left( \frac1p W^T W\right) b}{p} \overset{L^2}{\longrightarrow} c_0$.
\end{proof}

\begin{cor} \label{cor:weight_matrix_proj_concentration}
    Suppose that $x_a, x_b \in \RR^q$ are fixed vectors. Moreover, for a fixed integer $k \geq 1$, suppose that $W^{0} \in \RR^{p \times q}$ and $W^{1}, \dots, W^{k-1} \in \RR^{p \times p}$ are all random matrices, each of whose entries is i.i.d. standard Gaussian. For each $i = 1, 2, \dots, k$, define $z_\mu^{i}$ by $z_\mu^1 := \frac{1}{\sqrt{q}}W^0 x_\mu$ and $z_\mu^{i+1} = \frac{1}{\sqrt{p}}W^i z_\mu^i$, where $\mu \in \{a, b\}$. Then, for each $i = 1, 2, \dots, k$, $\frac{{z_a^i}^T z_b^i}{p} \overset{L^2}{\longrightarrow} \frac{x_a^T x_b}{q}$ as $p \to \infty$.
\end{cor}

\begin{proof}
    This result readily follows by induction, using Lemma \ref{lemma:weight_matrix_proj_concentration}. First, $\frac{{z_a^1}^T z_b^1}{p} \overset{L^2}{\longrightarrow} \frac{x_a^T x_b}{q}$, $\frac{{z_a^1}^T z_a^1}{p} \overset{L^2}{\longrightarrow} \frac{x_a^T x_a}{q}$, and $\frac{{z_b^1}^T z_b^1}{p} \overset{L^2}{\longrightarrow} \frac{x_b^T x_b}{q}$ as $p \to \infty$ by Lemma \ref{lemma:weight_matrix_proj_concentration}(a). Now, by induction, assuming that $\frac{{z_a^i}^T z_b^i}{p} \overset{L^2}{\longrightarrow} \frac{x_a^T x_b}{q}$, $\frac{{z_a^i}^T z_a^i}{p} \overset{L^2}{\longrightarrow} \frac{x_a^T x_a}{q}$, and $\frac{{z_b^i}^T z_b^i}{p} \overset{L^2}{\longrightarrow} \frac{x_b^T x_b}{q}$ (as $p \to \infty$) for some $i \in \{1, 2, \dots, k-1\}$, then Lemma \ref{lemma:weight_matrix_proj_concentration}(b) precisely implies that $\frac{{z_a^{i+1}}^T z_b^{i+1}}{p} \overset{L^2}{\longrightarrow} \frac{x_a^T x_b}{q}$, $\frac{{z_a^{i+1}}^T z_a^{i+1}}{p} \overset{L^2}{\longrightarrow} \frac{x_a^T x_a}{q}$, and $\frac{{z_b^{i+1}}^T z_b^{i+1}}{p} \overset{L^2}{\longrightarrow} \frac{x_b^T x_b}{q}$ as $p \to \infty$. This completes the induction.
\end{proof}

\subsection{Results from Random Matrix Theory}

\begin{defn}
    For $m, n \in \ZZ_+$, let $Z \in \RR^{m \times n}$ be a matrix such that $Z_{ij} \iid \mathcal{N}(0, 1)$ for each $1 \leq i \leq m, \text{ } 1 \leq j \leq n$. We say that the sample covariance matrix $W := \frac{1}{m} Z^T Z \in \RR^{n \times n}$ is a \emph{Wishart matrix} with aspect ratio $\frac{n}{m}$. 

    Note that $W' := \frac1n ZZ^T$ is also a Wishart matrix, with aspect ratio $\frac{m}{n}$.
\end{defn}

\begin{thm} [Marcenko-Pastur Theorem]   \label{thm:MP_thm}
    Fix some $q \in (0, \infty)$, define increasing sequences $(n_k)_{k \geq 1}$ and $(m_k)_{k \geq 1}$ of integers such that $\lim_{k \to \infty} \frac{n_k}{m_k} = q$. For $Z_k \in \RR^{m_k \times n_k}$ a random matrix whose entries are each iid standard Gaussians, define $W_k := \frac{1}{m_k} Z_k^T Z_k \in \RR^{n_k \times n_k}$ to be the corresponding Wishart matrix. For each $k$, define $\mu_k := \frac{1}{n_k} \sum_{i=1}^{n_k} \delta_{\lambda_i(W_k)}$ to be the empirical spectral distribution (ESD) of the eigenvalues of $W_k$. Then, as $k \to \infty$, $\mu_k$ weakly almost surely converges to a deterministic measure $\MP^{q}$ known as the Marcenko-Pastur law. This means that for any bounded continuous function $f: \RR \to \RR$, 
    $$\frac{1}{n_k} \sum_{i=1}^{n_k} f(\lambda_i(W_k)) = \mu_k(f) \asconverge \mu(f) = \int f(x) \MP^{q}(dx).$$
\end{thm}

\begin{proof}
    See \cite{marchenko1967distribution} and \cite{silverstein1995strong}.
\end{proof}

\begin{defn}    \label{defn:finite_stieltjes_transform}
    For a random square matrix $W \in \RR^{n \times n}$, define the \emph{Stieltjes transform} of $W$ as $g_W(z) =  \Tr\left( \frac1n (W - zI_n)^{-1} \right)$. Note that $g_W$ is well-defined over $z \in \RR \setminus \sigma(W)$, where $\sigma(W)$ denotes the set of eigenvalues of $W$.
\end{defn}

In particular, the Stieltjes transform of a Wishart matrix will appear in our subsequent calculations. We now state some important results about the Marcenko-Pastur (MP) law and its Stieltjes transform.
\begin{prop}    \label{prop:properties_of_MP_and_Stieltjes}
Using the same notation as in Theorem \ref{thm:MP_thm}, the following results hold:
\begin{enumerate}
    \item The first two moments of the MP law are given as follows:
    \begin{align*}
        \int x \MP^{q}(dx) = 1 \\
        \int x^2 \MP^{q}(dx) = 1 + q. \\
    \end{align*}
    If $q \in (0, 1)$, then the first two inverse moments of the MP law are given as follows:
    \begin{align*}
        \int \frac1x \MP^{q}(dx) = \frac{1}{1-q} \\
        \int \frac{1}{x^2} \MP^{q}(dx) = \frac{1}{(1 - q)^3}. \\
    \end{align*}

    \item For the Wishart matrix $W_k = \frac{1}{m_k} Z_k^T Z_k$, in the asymptotic limit of $n_k, m_k \to \infty$ with $\frac{n_k}{m_k} \to q \in (0, \infty)$, the Stieltjes transform of $W_k$ converges (pointwise) to the closed-form function
    \begin{equation}    \label{eqn:stieltjes_closed_form_defn}
        {g}_{W}(z) = \frac{1 - q - z + \sqrt{z - \lambda_-}\sqrt{z - \lambda_+}}{2 q z}
    \end{equation}
    where $\sqrt{\cdot}$ is defined using its principal branch in the complex plane. Here, $\lambda_- = (1 - \sqrt{q})^2$ and $\lambda_+ = (1 + \sqrt{q})^2$ are the left and right edges, respectively, of the \emph{bulk} of the MP distribution.

    \item Suppose that $q \in (0, 1)$. Then, the Stieltjes transform $g_W(z)$ is smooth over all $z \in (-\infty, 0]$, i.e. $g_W \in C^{\infty}((-\infty, 0])$.

    \item For any function $f(z, x)$ that is integrable against the MP law for each fixed $z \leq 0$, for which $\frac{\partial f}{\partial z}$ exists for all $z \leq 0$, and for which $\left| \frac{\partial f}{\partial z}(z, x) \right| \leq h(x)$ over all $z \leq 0$ and almost all $x$ for some function $h$ that is integrable against the MP law, differentiation may be done under the integral for all $z \leq 0$:
    $$\frac{\partial}{\partial z} \int f(z, x) \MP^{q}(dx) = \int \frac{\partial f}{\partial z}(z, x) \MP^{q}(dx). $$ In particular, when $q \in (0, 1)$, this result holds when $f(z, x) = \frac{1}{x - z}$, which corresponds to the Stieltjes transform $g_W(z) = \int \frac{1}{x - z} \MP^{q}(dx)$. More generally, for $q \in (0, 1)$, it holds for any function $f(z, x) = \frac{1}{(x-z)^j}$ for any fixed $j \geq 1$, meaning that each subsequent derivative of $g_W$ can be computed by differentiation under the integral.
\end{enumerate}
\end{prop}

\begin{proof}
\leavevmode
    \begin{enumerate}
        \item See Tulino \& Verdu (2004), Section 2.3.3 \cite{tulino_verdu_rmt_2004}.

        \item See Bai \& Silverstein (2010), Lemma 3.11 \cite{bai2010spectral}.

        \item This follows immediately from Eq. \ref{eqn:stieltjes_closed_form_defn}, which indicates that $g_W(z)$ is smooth outside of $[\lambda_-, \lambda_+]$. Since $\lambda_- > 0$ for $q \in (0, 1)$, it follows that $g_W(z)$ is smooth over all $z \in (-\infty, 0]$.

        \item This is a standard differentiation-under-the-integral result that holds for general measures; see Folland, Thm 2.27 \cite{folland1999real}.
    \end{enumerate}
\end{proof}

\begin{thm} \label{thm:max_min_MP_eigvals}
    Using the same notation as in Theorem \ref{thm:MP_thm}, it is the case that $$\eigmax(W_k) \asconverge \left( 1 + \sqrt{q} \right)^2$$
    as $k \to \infty$.
    Moreover, if $m_k > n_k$, then 
    $$\eigmin(W_k) \asconverge \left( 1 - \sqrt{q} \right)^2$$
    as $k \to \infty$.
\end{thm}

\begin{proof}
    See Bai \& Yin (1993) \cite{bai1993edgeeigvalsMP}.
\end{proof}

\begin{rem}
    Note that if $m_k < n_k$, then $W_k$ will always have $n_k - m_k$ zero eigenvalues, and so 
    $$\eigmin(W_k) \asconverge 0.$$
\end{rem}

\begin{cor} \label{cor:unbounded_f_MP_convergence}
    Let $f \in C([0, \infty))$ be any continuous (not necessarily bounded) function. Then, with the same notation as in Theorem \ref{thm:MP_thm}, it is the case that
    $$\frac{\Tr(f(W_k))}{n_k} \asconverge \int f(x) \MP^q(dx).$$
\end{cor}

\begin{proof}
    As in the statement of Theorem \ref{thm:MP_thm}, let $\mu_k = \frac{1}{n_k} \sum_{i=1}^{n_k} \delta_{\lambda_i(W_k)}$ denote the ESD of $W_k$. Because $W_k$ is symmetric positive-semidefinite, $\lambda_i(W_k) \geq 0$ for each $i = 1, \dots, n_k$. Let $\lambda_{+} := \left( 1 + \sqrt{q} \right)^2$. As a convenient shorthand, we simply refer to $\lambda_i(W_k)$ by $\lambda_i$ and to $\eigmax(W_k)$ by $\eigmax$. These eigenvalues are random variables defined over some probability space $(\Omega, \mathcal{F}, \nu)$.

    Now, for any $\epsilon > 0$, pick some $M_\epsilon > \lambda_+ + \epsilon$. By Theorem \ref{thm:max_min_MP_eigvals}, there exists a full-measure set $\Omega_{1, \epsilon} \subset \Omega$ such that for all $\omega \in \Omega_{1, \epsilon}$, there exists $k_{1, \epsilon}(\omega) \in \ZZ_+$ such that $\eigmax(W_k(\omega)) < \lambda_+ + \epsilon < M_\epsilon$ for all $k \geq k_{1, \epsilon}(\omega)$. Define the \emph{truncated} function $f_{M_\epsilon} \in C_b([0, \infty))$ by
    $$f_{M_\epsilon}(x) := \begin{cases}
        f(x) & 0 \leq x \leq M_\epsilon \\
        f(M_\epsilon) & x > M_\epsilon
    \end{cases}.$$
    Then, since $\eigmax(W_k(\omega)) \leq M_\epsilon$ for all $k \geq k_{1, \epsilon}(\omega)$, it follows that 
    $$\frac{\Tr(f(W_k(\omega)))}{n_k} = \int f d\mu_k = \int f_{M_\epsilon} d \mu_k$$
    for all $k \geq k_{1, \epsilon}(\omega)$, for each $\omega \in \Omega_{1, \epsilon}$.
    Furthermore, owing to the fact that $f_{M_\epsilon}$ is bounded and continuous, Theorem \ref{thm:MP_thm} implies that $$\int f_{M_\epsilon} d \mu_k \asconverge \int f_{M_\epsilon} d\MP^q = \int f d\MP^q.$$ (The latter equality follows from the fact that $\MP^q$ is supported within $[0, \lambda_+]$.) This means that there exists a full-measure set $\Omega_{2, \epsilon}$ so that for any $\omega \in \Omega_{2, \epsilon}$, there exists $k_{2, \epsilon}(\omega) \in \ZZ_+$ such that 
    $$\left| \int f_{M_\epsilon} d \mu_k - \int f d \MP^q \right| < \epsilon$$ for all $k \geq k_{2, \epsilon}(\omega)$. 
    
    Now, further define $K_{\epsilon}(\omega) := \max\left(\left\{ k_{1, \epsilon}(\omega), k_{2, \epsilon}(\omega) \right\}\right)$.
    In particular, taking a countable sequence $(\epsilon_i)_{i \geq 1}$ with $\epsilon_i \to 0$, define the full-measure set $\tilde{\Omega} := \bigcap_{i \geq 1}\Omega_{1, \epsilon_i} \cap \Omega_{2, \epsilon_i}$.
    Then, for any $\epsilon > 0$, we may find some $j \in \ZZ_+$ such that $\epsilon_j < \epsilon$. Because $\tilde{\Omega} \subset \Omega_{1, \epsilon_j} \cap \Omega_{2, \epsilon_j}$, we find that 
    $$\left| \frac{\Tr(f(W_k(\omega)))}{n_k} - \int f d \MP^q \right| = \left| \int f_{M_{\epsilon_j}} d \mu_k - \int f d \MP^q \right| < \epsilon_j < \epsilon$$
    for all $k \geq K_{\epsilon_j}(\omega)$ and any $\omega \in \tilde{\Omega}$. This precisely means that $$\frac{\Tr(f(W_k))}{n_k} \asconverge \int f(x) \MP^q(dx).$$
\end{proof}

\begin{cor} \label{cor:unbounded_f_MP_convergence_zero_case}
    Let $f \in C((0, \infty))$ be any continuous (not necessarily bounded) function. Using the same notation as in Theorem \ref{thm:MP_thm}, suppose that $m_k > n_k$, meaning that $W_k$ has full rank almost surely, and that $\lim_{k \to \infty} \frac{n_k}{m_k} = q \in (0, 1)$. Then, it is the case that
    $$\frac{\Tr(f(W_k))}{n_k} \asconverge \int f(x) \MP^q(dx).$$
\end{cor}

\begin{proof}
    We proceed just like in the proof of Cor. \ref{cor:unbounded_f_MP_convergence}. Let $\lambda_{-} := (1 - \sqrt{q})^2$ and, as a shorthand, we refer to $\eigmin(W_k)$ as $\eigmin$. For any $\epsilon \in (0, \lambda_{-})$, pick some $M_\epsilon > \lambda_{+} + \epsilon$, and some $L_\epsilon \in (0, \lambda_{-} - \epsilon)$. Because $m_k > n_k$, Theorem \ref{thm:max_min_MP_eigvals} implies that there exists a full measure set $\Omega_{1, \epsilon} \subset \Omega$ such that for all $\omega \in \Omega_{1, \epsilon}$ there exists $k_{1, \epsilon}(\omega) \in \ZZ_+$ so that $\eigmax(W_k(\omega)) < \lambda_+ + \epsilon < M_\epsilon$ and $\eigmin(W_k(\omega)) > \lambda_- - \epsilon > L_\epsilon$ for all $k \geq k_{1, \epsilon}(\omega)$. Define the truncated function $f_\epsilon \in C_b([0, \infty))$ by 
    $$f_{\epsilon}(x) := \begin{cases}
        f(L_\epsilon) & 0 \leq x < L_\epsilon \\
        f(x) & L_\epsilon \leq x \leq M_\epsilon \\
        f(M_\epsilon) & x > M_\epsilon
    \end{cases}.$$
    For all $\omega \in \Omega_{1, \epsilon}$ it is the case that for all $k \geq k_{1, \epsilon}(\omega)$,
    $$\frac{\Tr(f(W_k(\omega)))}{n_k} = \int f d\mu_k = \int f_{\epsilon} d \mu_k.$$
    Again, owing to the fact that $f_{\epsilon}$ is bounded and continuous, Theorem \ref{thm:MP_thm} implies that
    $$\int f_{\epsilon} d \mu_k \asconverge \int f_{\epsilon} d\MP^q = \int f d\MP^q.$$
    This means that there exists a full measure subset $\Omega_{2, \epsilon} \subset \Omega$ so that for any $\omega \in \Omega_{2, \epsilon}$ there exists $k_{2, \epsilon}(\omega) \in \ZZ_+$ such that 
    $$\left| \int f_{\epsilon} d \mu_k - \int f d \MP^q \right| < \epsilon$$
    for all $k \geq k_{2, \epsilon}(\omega)$.
    
    Now, further define $K_{\epsilon}(\omega) := \max\left(\left\{ k_{1, \epsilon}(\omega), k_{2, \epsilon}(\omega) \right\}\right)$.
    Taking a countable sequence $(\epsilon_i)_{i \geq 1} \subset (0, \lambda_-)$ with $\epsilon_i \to 0$, define the full-measure set $\tilde{\Omega} := \bigcap_{i \geq 1}\Omega_{1, \epsilon_i} \cap \Omega_{2, \epsilon_i}$.
    Then, for any $\epsilon \in (0, \lambda_-)$, we may find some $j \in \ZZ_+$ such that $\epsilon_j < \epsilon$. Because $\tilde{\Omega} \subset \Omega_{1, \epsilon_j} \cap \Omega_{2, \epsilon_j}$, we find that
    $$\left| \frac{\Tr(f(W_k(\omega)))}{n_k} - \int f d \MP^q \right| = \left| \int f_{\epsilon_j} d \mu_k - \int f d \MP^q \right| < \epsilon_j < \epsilon$$
    for all $k \geq K_{\epsilon_j}(\omega)$ and any $\omega \in \tilde{\Omega}$. It follows that $\frac{\Tr(f(W_k(\omega)))}{n_k} \asconverge \int f(x) \MP^q(dx)$.
\end{proof}

\begin{lemma}\label{lemma:max_MP_eigval_bound}
Let $(m_k)_{k\ge1}$ be a strictly increasing sequence and $(n_k)_{k\ge1}$ a nondecreasing sequence, both of positive integers,
with $n_k/m_k \to q \in [0,\infty)$, and set $\tilde{q} := \sup_{k \geq 1} n_k/m_k < \infty$.
Let $Z_k \in \RR^{m_k\times n_k}$ have iid $\mathcal N(0,1)$ entries and $W_k := \frac{1}{m_k}Z_k^T Z_k \in \RR^{n_k\times n_k}$.
Then for every $p \geq 1$ there exists a constant $C_{p,\tilde{q}}<\infty$, depending only on $p$ and $\tilde{q}$, such that
$$\sup_{k\ge1}\EE[\eigmax(W_k)^p] \le C_{p,\tilde{q}}.$$
Additionally, if $q > 0$, then the analogous result holds for $W_k' := \frac{1}{n_k}Z_kZ_k^T$ with $\tilde{q}' := \sup_k m_k/n_k$ in place of $\tilde{q}$ and $C_{p, \tilde{q}'}$ in place of $C_{p, \tilde{q}}$. 
\end{lemma}

\begin{proof}
    It suffices to establish this lemma for $W_k$; the same argument with $m_k$ and $n_k$ interchanged will yield the analogous result for $W_k'$ when $q > 0$, because in this case, $m_k / n_k \to \frac1q$ and so $\tilde{q}'$ is finite.
    To proceed, we recall a classic result from high-dimensional probability (\cite{Vershynin_2026}; Thm. 4.4.3) that there exists an absolute constant $C > 0$ such that, for any $k \geq 1$,
    $$\PP\left( \frac{1}{\sqrt{m_k}} ||Z_k|| > C\left( \sqrt{\frac{n_k}{m_k}} + 1 \right) + u \right) \leq 2e^{-m_k u^2/C}.$$
    Because $\frac{n_k}{m_k} \to q$, it follows that $\tilde{q} = \sup_{k \geq 1} \frac{n_k}{m_k} < \infty$, meaning that 
\begin{equation}    \label{eq:vershynin_max_eigval_reformulation}
    \begin{aligned}
        \PP\left( \frac{1}{\sqrt{m_k}} ||Z_k|| > C\left( \sqrt{\tilde{q}} + 1 \right) + u \right) \leq \PP\left( \frac{1}{\sqrt{m_k}} ||Z_k|| > C\left( \sqrt{\frac{n_k}{m_k}} + 1 \right) + u \right) \\
        \leq 2e^{-m_k u^2/C}.
    \end{aligned}
\end{equation}
    Observe additionally that $\eigmax(W_k) = \left( \frac{1}{\sqrt{m_k}} ||Z_k|| \right)^2$. Then, 
    $$\EE\left[ \eigmax(W_k)^p \right] = \EE\left[ \left( \frac{1}{\sqrt{m_k}} ||Z_k|| \right)^{2p} \right] = \int_{0}^{\infty} 2p t^{2p-1} \PP\left( \frac{1}{\sqrt{m_k}} ||Z_k|| > t\right) dt$$
    $$ \leq \int_{0}^{C(\sqrt{\tilde{q}}+1) + 1} 2pt^{2p-1} dt + \int_{C(\sqrt{\tilde{q}}+1) + 1}^{\infty} 4pt^{2p-1} \exp\left( -\frac{m_k}{C} \left( t - C(\sqrt{\tilde{q}}+1) \right)^2 \right) dt$$
    $$ \leq \left( C(\sqrt{\tilde{q}}+1) + 1 \right)^{2p} + 4p \int_{-\infty}^{\infty} |t|^{2p-1} \exp\left( -\frac1C  \left( t - C(\sqrt{\tilde{q}}+1) \right)^2 \right) dt $$
    $$ \leq \left( C(\sqrt{\tilde{q}}+1) + 1 \right)^{2p} + 4p C' \EE[|\tau|^{2p-1}].$$
    Here, we define the random variable $\tau \sim \mathcal{N}\left( C(\sqrt{\tilde{q}}+1), \frac{C}{2} \right)$, and have defined an absolute constant $C'$ so that we may extract the $(2p-1)$st moment of $\tau$. Observe that this final quantity only depends on $p$ and $\tilde{q}$, and that it is moreover finite as all moments of a Gaussian random variable are finite. Consequently, it follows that $\EE\left[ \eigmax(W_k)^p \right] \leq C_{p, \tilde{q}}$, where $C_{p, \tilde{q}} := \left( C(\sqrt{\tilde{q}}+1) + 1 \right)^{2p} + 4p C' \EE[|\tau|^{2p-1}]$ only depends on $p$ and $\tilde{q}$.
\end{proof}

\begin{cor} \label{cor:Lp_control_of_NN_weights_and_states}
    With the same notation as in Lemma \ref{lemma:max_MP_eigval_bound}, $\opnorm{\frac{1}{\sqrt{m_k}} Z_k}$ is moment controlled. Moreover, suppose that $n_k \equiv n \in \ZZ_+$ is fixed, and let $x \in \RR^n$ be a vector with fixed entries. Then, for $h_k := \frac{1}{\sqrt{n}} Z_k x \in \RR^{m_k}$, $\norm{\frac{1}{\sqrt{m_k}}h_k}_2$ is moment controlled. 
\end{cor}

\begin{proof}
    Since $\eigmax(W_k)^r = \opnorm{\frac{1}{\sqrt{m_k}} Z_k}^{2r}$, Lemma \ref{lemma:max_MP_eigval_bound} implies that $\sup_{k\ge1} \EE\left[ \opnorm{\frac{1}{\sqrt{m_k}} Z_k}^p \right] < \infty$ for any $p \geq 2$. Since $$\EE\left[ \opnorm{\frac{1}{\sqrt{m_k}} Z_k}^p \right] \leq \EE\left[ \opnorm{\frac{1}{\sqrt{m_k}} Z_k}^{p'} \right]^{p/p'}$$
    for any $1 \leq p \leq p'$, it follows that $\opnorm{\frac{1}{\sqrt{m_k}} Z_k}$ is moment controlled. Similarly, in the case where $n_k \equiv n$, 
    $$\norm{\frac{1}{\sqrt{m_k}} h_k}_2 = \norm{\frac{1}{\sqrt{m_k}} Z_k \cdot \frac{x}{\sqrt{n}}}_2 \leq \opnorm{\frac{1}{\sqrt{m_k}} Z_k} \cdot \norm{\frac{x}{\sqrt{n}}}_2$$
    which is moment controlled because $\opnorm{\frac{1}{\sqrt{m_k}} Z_k}$ is.
\end{proof}

\begin{thm} \label{thm:min_singular_value_concentration}
    Define $n_k$, $m_k$, $Z_k$, and $W_k$ as in Lemma \ref{lemma:max_MP_eigval_bound}, with the added constraint that $m_k > n_k$ for all $k \geq 1$ and that $\lim_{k\to\infty} \frac{n_k}{m_k} = q \in (0, 1)$ and set $\tilde{q} := \sup_{k \geq 1} \frac{n_k}{m_k} < 1$. Then, there exists an absolute constant $c > 0$ and a constant $C_{q, \tilde{q}}$ that depends on $q$ and $\tilde{q}$ (and not on $k$) such that
    $$\PP\left( s_{\text{min}}\left( \frac{1}{\sqrt{m_k}} Z_k\right) < C_{q, \tilde{q}}\left(1 - \sqrt{q}\right) - u \right) \leq 2 \exp\left( -c m_k u^2 \right)$$
    for all $u \geq 0$ and all $k \geq 1$. (Here, $s_{\text{min}}$ denotes minimum singular value.)
\end{thm}

\begin{proof}
    For any two matrices $A, B \in \RR^{m_k \times n_k}$, Weyl's inequality (e.g. see Vershynin, Lemma 4.1.14 \cite{Vershynin_2026}) implies that $|s_{\text{min}}\left( A \right) - s_{\text{min}}\left( B \right)| \leq \opnorm{A-B} \leq \norm{A-B}_F$. In particular, $s_{\text{min}}:\RR^{m_k \times n_k} \to \RR$ is $1$-Lipschitz under the metric induced by the Frobenius norm. 
    Thus, viewing $Z_k \in \RR^{m_k \times n_k}$ as an $m_k n_k$-dimensional standard Gaussian vector, Gaussian Lipschitz concentration (Vershynin, Thm 5.2.3 \cite{Vershynin_2026}) yields
    $$\PP\left( s_{\text{min}}\left( Z_k\right) < \EE[s_{\text{min}}\left( Z_k\right)] - t \right) \leq 2 \exp(-ct^2)$$
    for some absolute constant $c > 0$. Moreover, it is the case that $\EE[s_{\text{min}}\left( Z_k\right)] \geq \sqrt{m_k} - \sqrt{n_k}$ \cite{rudelson2010nonasymptotic}. As a result, 
    $$\PP\left( s_{\text{min}}\left( Z_k\right) < \sqrt{m_k} - \sqrt{n_k} - t \right) \leq \PP\left( s_{\text{min}}\left( Z_k\right) < \EE[s_{\text{min}}\left( Z_k\right)] - t \right) \leq 2 \exp(-ct^2).$$
    Thus, normalizing by $\sqrt{m_k}$ and making the substitution $u := \frac{t}{\sqrt{m_k}}$, we find that 
    $$\PP\left( s_{\text{min}}\left( \frac{1}{\sqrt{m_k}} Z_k\right) < 1 - \sqrt{\frac{n_k}{m_k}} - u \right) \leq 2 \exp\left( -c m_k u^2 \right).$$
    Finally, since $m_k > n_k$ and $\frac{n_k}{m_k} \to q < 1$, it follows that $\tilde{q} < 1$. In turn, $\inf_{k \geq 1} \left( 1 - \sqrt{\frac{n_k}{m_k}}\right) > 0$. Thus, we may find a constant $C_{q, \tilde{q}} > 0$, dependent only on $q$ and $\tilde{q}$, such that $C_{q, \tilde{q}}(1 - \sqrt{q}) \leq \inf_{k \geq 1} \left( 1 - \sqrt{\frac{n_k}{m_k}} \right)$. For this constant $C_{q, \tilde{q}}$, it follows that
    $$\PP\left( s_{\text{min}}\left( \frac{1}{\sqrt{m_k}} Z_k\right) < C_{q, \tilde{q}}\left(1 - \sqrt{q}\right) - u \right) \leq \PP\left( s_{\text{min}}\left( \frac{1}{\sqrt{m_k}} Z_k\right) < 1 - \sqrt{\frac{n_k}{m_k}} - u \right) \leq 2 \exp\left( -c m_k u^2 \right).$$
\end{proof}

\begin{cor} \label{cor:min_singular_val_tail_bound}
    With the same notation as in Theorem \ref{thm:min_singular_value_concentration}, there exist constants $c', C' > 0$ that depend only on $q$ and $\tilde{q}$ (and not on $k$) such that
    $$\PP\left( s_{\text{min}}\left( \frac{1}{\sqrt{m_k}} Z_k\right) < C'\left(1 - \sqrt{q}\right)\right) \leq 2 \exp\left( -c' m_k \right).$$
\end{cor}

\begin{proof}
    Taking $u = \frac{C_{q, \tilde{q}}}{2}\left(1 - \sqrt{q} \right)$ in the statement of Theorem \ref{thm:min_singular_value_concentration} gives the result.
\end{proof}

\begin{thm} \label{thm:orthogonal_matrix_concentration}
    Let $h:SO(n) \to \RR$ be a Lipschitz function (with respect to the Frobenius norm) on the special orthogonal group, and let $X \sim \Unif(SO(n))$ be Haar distributed. Then, there exists an absolute constant $C > 0$ such that 
    $$||h(X) - \EE[h(X)]||_{\psi_2} \leq \frac{C ||h||_{\text{Lip}}}{\sqrt{n}}$$ 
    where $||\cdot||_{\psi_2}$ denotes the subgaussian norm. Consequently, we have the following high-probability tail bound:
    $$\PP\left( \left| h(X) - \EE[h(X)] \right| > t \right) \leq 2\exp\left({-\frac{C'nt^2}{||h||_{\text{Lip}}^2}}\right)$$
    for some absolute constant $C' > 0$.
\end{thm}

\begin{proof}
    See Vershynin, Thm 5.2.7 \cite{Vershynin_2026}.
\end{proof}

\begin{defn}
    We say that a nonnegative function $f \in C([0, \infty))$ is \emph{polynomially bounded} if there exist constants $C_f \geq 0$ and $s \geq 1$ such that $f(x) \leq C_f (x^s + 1)$ for all $x \in [0, \infty)$.
\end{defn}

Key examples of polynomially bounded functions are $f(x) = x^p$ for $p \geq 0$ and $f(x) = \frac{1}{(x + \lambda)^p}$ for $p > 0$ and some fixed $\lambda > 0$; indeed, the latter class of functions is bounded above by the constant $\lambda^{-p}$.

\begin{prop}    \label{prop:Lp_control}
    Let $f \in C([0, \infty))$ be a polynomially bounded function. Fix $q \in (0, \infty)$, and define $m_k$, $n_k$, $Z_k$, $W_k = \frac{1}{m_k} Z_k^T Z_k \in \RR^{n_k \times n_k}$, as in Lemma \ref{lemma:max_MP_eigval_bound}. For each $k \geq 1$, let $a_k, b_k \in \SS^{m_k - 1}$ be fixed unit vectors, $a_k', b_k' \in \SS^{n_k - 1}$ be fixed unit vectors, and let $\alpha_k \sim \mathcal{N}\left(0, \frac{1}{n_k} I_{n_k} \right)$ 
    be sampled independently from $Z_k$. Then, for any $p \geq 1$, the following statements are true:
    \begin{enumerate}
        \item $\sup_{k \geq 1} \EE\left[ \left| a_k'^T f(W_k) b_k' \right|^p \right] < \infty$.

        \item $\sup_{k \geq 1} \EE\left[ \left| \alpha_k^T f(W_k) \alpha_k \right|^p \right] < \infty$.

        \item $\sup_{k \geq 1} \EE\left[ \left| \frac{1}{\sqrt{m_k}} \alpha_k^T f(W_k) Z_k^T a_k \right|^p \right] < \infty$.

        \item $\sup_{k \geq 1} \EE\left[ \left| \frac{1}{m_k} a_k^T Z_k f(W_k) Z_k^T b_k \right|^p \right] < \infty$.
    \end{enumerate}

    Here, $f$ is applied spectrally. Altogether, each of the quantities $a_k'^T f(W_k) b_k'$, $\alpha_k^T f(W_k) \alpha_k$, $\frac{1}{\sqrt{m_k}} \alpha_k^T f(W_k) Z_k^T a_k$, and $\frac{1}{m_k} a_k^T Z_k f(W_k) Z_k^T b_k$ is moment controlled.
\end{prop}

\begin{proof}
    By definition, $f(x) \leq C_f(x^s + 1)$ for some constants $C_f \geq 0$ and $s \geq 1$ and over all $x \in [0, \infty)$. Moreover, $x \mapsto C_f(x^s + 1)$ is an increasing function on $[0, \infty)$. We start by establishing a few key results. Note that for any $p \geq 1$, $\norm{\frac{1}{\sqrt{m_k}} Z_k}^{2p} = \eigmax(W_k)^p$ and so Lemma \ref{lemma:max_MP_eigval_bound} implies that 
    \begin{equation}    \label{eq:Z_norm_moments}
        \EE\left[ \norm{\frac{1}{\sqrt{m_k}} Z_k}^{2p} \right] \leq C_{p, \tilde{q}}
    \end{equation}
    is uniformly bounded over $k \geq 1$.
    Next, note that for any $p \geq 1$, because $f$ is polynomially bounded,
    $$||f(W_k)||^p = \left(\eigmax(f(W_k))\right)^p = \max_{1 \leq i \leq n_k} \left(f\left( \lambda_i(W_k) \right)\right)^p \leq \max_{1 \leq i \leq n_k} \left(C_f \left( \lambda_i(W_k)^s + 1 \right)\right)^p$$ 
    $$ = \left(C_f \left( \eigmax(W_k)^s + 1 \right)\right)^p \leq C_f^p \left(2 \cdot \max\left( \eigmax(W_k)^s, 1 \right)\right)^p$$ $$ = (2C_f)^p \max\left( \eigmax(W_k)^{ps}, 1 \right) \leq (2C_f)^p \left( \eigmax(W_k)^{ps} + 1 \right).$$
    Since $s \geq 1$, $ps \geq p \geq 1$, and so Lemma \ref{lemma:max_MP_eigval_bound} implies that
    \begin{equation}    \label{eq:f_Wishart_moments}
        \EE\left[ ||f(W_k)||^{p} \right] \leq (2C_f)^{p} (C_{ps, \tilde{q}} + 1)
    \end{equation}
    is uniformly bounded over $k \geq 1$.
    Finally, consider the multivariate Gaussian variable $\alpha_k \sim \mathcal{N}\left(0, \frac{1}{n_k} I_{n_k} \right)$. Writing $\alpha_k = \frac{1}{\sqrt{n_k}}g_k$, where $g_k \sim \mathcal{N}(0, I_{n_k})$, the power-mean inequality yields 
    $$||\alpha_k||_2^{2p} = \left( \frac{1}{n_k} \sum_{i=1}^{n_k} (g_k)_i^2 \right)^{p} \leq \frac{1}{n_k} \sum_{i=1}^{n_k} (g_k)_i^{2p}.$$ 
    In turn, we find that 
    \begin{equation}    \label{eq:Gaussian_var_moments}
        \EE\left[ ||\alpha_k||_2^{2p} \right] \leq \EE\left[ \frac{1}{n_k} \sum_{i=1}^{n_k} (g_k)_i^{2p} \right] 
    \leq \EE\left[ \left| (g_k)_1^{2p} \right| \right] \leq C_p'
    \end{equation}
    for some constant $C_p' > 0$ that only depends on $p$. Here, we have noted that the $(g_k)_{i}$ are independent and are moreover standard Gaussian variables; consequently, the $2p$-th moment of such a random variable is some constant only depending on $p$. Consequently, we find that $\EE\left[ ||\alpha_k||_2^{2p} \right]$ is uniformly bounded above by $C_{p}'$. 
    
    Having established these preliminary results, we turn to studying the quantities of interest. First, consider the quantity ${a_k'}^T f(W_k) b_k'$ and take any $p \geq 1$. Because $\norm{a_k'}_2 = \norm{b_k'}_2 = 1$,
    $$\sup_{k \geq 1} \EE\left[ \left|{a_k'}^T f(W_k) b_k' \right|^p \right] \leq \sup_{k \geq 1} \EE\left[ \left(||a_k'||_2 \cdot ||f(W_k)|| \cdot ||b_k'||_2\right)^p \right]$$ 
    $$ \leq \sup_{k \geq 1} \EE\left[ ||f(W_k)||^{p} \right] \leq (2C_f)^p (C_{ps, \tilde{q}} + 1) < \infty $$ 
    where we have used Eq. \ref{eq:f_Wishart_moments} and noted that the constant $C_{ps, \tilde{q}} > 0$ only depends on $ps$ and $\tilde{q}$.

    Second, consider the quantity $\alpha_k^T f(W_k) \alpha_k$, where $\alpha_k \sim \mathcal{N}\left(0, \frac{1}{n_k} I_{n_k} \right)$. Observe that $$\EE\left[|\alpha_k^T f(W_k) \alpha_k|^p\right] \leq \EE\left[||f(W_k)||^p \cdot ||\alpha_k||_2^{2p} \right] \leq \left(\EE\left[ ||f(W_k)||^{2p} \right] \EE\left[ ||\alpha_k||_2^{4p} \right] \right)^{\frac12}.$$
    Thus, combining Eqs. \ref{eq:f_Wishart_moments} and \ref{eq:Gaussian_var_moments}, we conclude that $\sup_{k \geq 1} \EE\left[ \left| \alpha_k^T f(W_k) \alpha_k \right|^p \right] < \infty$.

    Third, we turn to the quantity $\frac{1}{\sqrt{m_k}} \alpha_k^T f(W_k) Z_k^T a_k$, for which the analysis is similar to the previous cases. Indeed, 
    $$\sup_{k \geq 1} \EE\left[ \left| \frac{1}{\sqrt{m_k}} \alpha_k^T f(W_k) Z_k^T a_k \right|^p \right] \leq \sup_{k \geq 1} \EE\left[||a_k||_2^p \cdot ||f(W_k)||^p \cdot \norm{\frac{1}{\sqrt{m_k}} Z_k}^p \cdot ||\alpha_k||_2^{p} \right]$$ 
    $$ = \sup_{k \geq 1} \EE\left[||f(W_k)||^p \cdot \norm{\frac{1}{\sqrt{m_k}} Z_k}^p \cdot ||\alpha_k||_2^{p} \right] \leq \sup_{k \geq 1} \left(\EE\left[ ||f(W_k)||^{2p} \norm{\frac{1}{\sqrt{m_k}} Z_k}^{2p} \right] \EE\left[ ||\alpha_k||_2^{2p} \right] \right)^{\frac12}$$
    $$ \leq \sup_{k \geq 1} \EE\left[ ||f(W_k)||^{4p}\right]^{\frac14} \EE\left[ \norm{\frac{1}{\sqrt{m_k}} Z_k}^{4p} \right]^{\frac14} \EE\left[ ||\alpha_k||_2^{2p} \right]^{\frac12} < \infty.$$
    Here, we have again applied the Cauchy-Schwarz inequality twice and used Eqs. \ref{eq:Z_norm_moments}, \ref{eq:f_Wishart_moments}, and \ref{eq:Gaussian_var_moments} to deduce that each of $\EE\left[ \norm{\frac{1}{\sqrt{m_k}} Z_k}^{4p} \right]$, $\EE\left[ ||f(W_k)||^{4p}\right]$, and $\EE\left[ ||\alpha_k||_2^{2p} \right]$ is uniformly bounded above. 

    Fourth, 
    $$\sup_{k \geq 1} \EE\left[ \left| \frac{1}{m_k} a_k^T Z_k f(W_k) Z_k^T b_k \right|^p \right] \leq \sup_{k \geq 1} \EE\left[ \norm{\frac{1}{\sqrt{m_k}} Z_k}^{2p} \norm{f(W_k)}^p  \right] $$
    $$ \leq \sup_{k \geq 1} \left( \EE\left[ \norm{\frac{1}{\sqrt{m_k}} Z_k}^{4p} \right] \EE\left[ \norm{f(W_k)}^{2p}  \right] \right)^{\frac12} \leq \left( \sup_{k\geq 1} \EE\left[ \norm{\frac{1}{\sqrt{m_k}} Z_k}^{4p} \right] \right)^{\frac12} \left( \sup_{k \geq 1} \EE\left[ \norm{f(W_k)}^{2p}  \right] \right)^{\frac12} < \infty $$
    where we have again used the Cauchy-Schwarz inequality as well as Eqs. \ref{eq:Z_norm_moments} and \ref{eq:f_Wishart_moments}.

\end{proof}

\begin{lemma}   \label{lemma:key_example_h_Lip}
    Let $a, b \in \RR^n$ be fixed unit vectors. For some fixed $B > 0$, let $F := \diag\left([ F_1, \dots, F_n ] \right) \in \RR^{n \times n}$ be a fixed diagonal matrix with $0 \leq F_i \leq B$ for each $1 \leq i \leq n$. Let $G := F - \frac{\Tr(F)}{n} I_n$ and define the function $h:SO(n) \to \RR$ by
    $$h(V) := {a}^T V G V^T b.$$
    Then, $h$ is 2$B$-Lipschitz with respect to the Frobenius norm. Moreover, for $V$ Haar distributed over $\SO(n)$, $\EE[h(V)] = 0$ and, for any $\epsilon \geq 0$,
    $$\PP\left( \left| h(V) \right| > \epsilon \right) \leq 2 \exp\left( - \frac{C}{B^2} n \epsilon^2 \right)$$
    for some absolute constant $C > 0$.
\end{lemma}

\begin{proof}
To see why $h$ is Lipschitz, observe that for any $V, V' \in \SO(n)$,
$$|h(V) - h(V')| = |{a}^T V G V^T b - {a}^T V' G {V'}^T b|$$
$$ = |{a}^T \left( V G V^T - V' G {V'}^T \right) b| \leq ||{a}||_2 ||V G V^T - V' G {V'}^T||_{\text{op}} ||b||_2$$
$$ = ||V G V^T - V' G {V'}^T||_{\text{op}} = ||V G (V - V')^T + (V - V') G {V'}^T||_{\text{op}}$$
$$ \leq ||V||_{\text{op}} ||G||_{\text{op}} ||V - V'||_{\text{op}} + ||V - V'||_{\text{op}} ||G||_{\text{op}} ||V'||_{\text{op}}$$
\begin{align}   \label{eq:h_Lipschitz}
= 2 ||G||_{\text{op}} ||V - V'||_{\text{op}} \leq 2 ||G||_{\text{op}} ||V - V'||_{F}.
\end{align}
Here, we have used the fact that $||{a}||_2 = ||b||_2 = 1$, that orthogonal matrices have unit operator norm, and that $\opnorm{M} \leq \norm{M}_F$ for any square matrix $M$. Moreover, since each $F_i \in [0, B]$ and $\frac{\Tr(F)}{n} = \frac{1}{n}\sum_{j=1}^n F_j \in [0, B]$,
\begin{align}   \label{eq:h_Lipschitz_norm}
    \opnorm{G} = \max_{1 \leq i \leq n} \left| F_i - \frac{\Tr(F)}{n} \right| \leq B.
\end{align}
Thus, $h$ is $2B$-Lipschitz with respect to the Frobenius norm, with $\norm{h}_{\text{Lip}} \leq 2B$. 

Next, if $V$ is Haar distributed in $\SO(n)$, $\EE[h(V)] = {a}^T \EE[V G V^T] b$. Since $UV \overset{D}{=} V$ for any $U \in \SO(n)$, we find that 
\begin{equation}    \label{eq:SO_n_commuting}
    U \EE[V G V^T] U^T = \EE[(UV) G (UV)^T] = \EE[V G V^T].
\end{equation}
Additionally, letting $R := \diag\left( \left[ 1, \dots, 1, -1 \right] \right) \in \RR^{n \times n}$ (with the only '$-1$' in the last diagonal entry), observe that $R^2 = I_n$ and that $RVR \overset{D}{=} V$ is also Haar distributed; indeed, for any $U_0 \in \SO(n)$,
$$U_0 (RVR) = R (R U_0 R) V R \overset{D}{=} RVR$$
because $R U_0 R \in \SO(n)$ and $V$ is Haar distributed, meaning that $R U_0 R V \overset{D}{=} V$.
Moreover, since $G$ is diagonal, $G$ commutes with $R$. Thus,
\begin{equation}    \label{eq:diag_R_commuting}
    \EE[VGV^T] = \EE[(RVR) G (RVR)^T] = R \EE[VGR^2V^T] R = R \EE[VGV^T] R.
\end{equation}
Furthermore, note that any orthogonal matrix $U' \in \RR^{n \times n}$ with $\det(U') = -1$ (i.e. not in $\SO(n)$) may be expressed as $U' = U'' R$ for some $U'' \in \SO(n)$. As a result, Eqs. \ref{eq:diag_R_commuting} and \ref{eq:SO_n_commuting} imply that
\begin{equation}    \label{eq:anti_SO_n_commuting}
    U' \EE[VGV^T] {U'}^T = U'' \left(R \EE[VGV^T] R\right) {U''}^T = U'' \EE[VGV^T] {U''}^T = \EE[VGV^T].
\end{equation}
Eqs. \ref{eq:SO_n_commuting} and \ref{eq:anti_SO_n_commuting} collectively imply that $\EE[VGV^T]$ commutes with any $n \times n$ orthogonal matrix, and so Lemma \ref{lemma:orthogonal_symmetry} implies that $\EE[V G V^T] = r I_{n}$ for some $r \in \RR$. Furthermore, we can solve for $r$:
$$r = \frac{\Tr(r I_{n})}{n} = \frac{1}{n} \Tr(\EE[V G V^T]) = \frac{1}{n} \Tr(G \EE[V^T V]) = \frac{1}{n} \Tr(G) = 0.$$ Here, we have noted that expectations commute with the trace operator, $V^T V = I_{n}$ (as $V$ is sampled from $\SO(n)$), and that $\Tr(G) = \Tr(F) - \frac{\Tr(F)}{n} \Tr(I_{n}) = 0$. Thus, we find that $\EE[V G V^T] = 0$, and so $\EE[h(V)] = 0$, as claimed. Since $\norm{h}_{\text{Lip}} \leq 2B$ is a constant that does not depend on $n$, Theorem \ref{thm:orthogonal_matrix_concentration} implies that 
\begin{equation}    \label{eq:h_tail_bound}
    \PP\left(|h(V)| > \epsilon \right) \leq 2\exp\left({-\frac{C}{B^2} n \epsilon^2}\right)
\end{equation}
where $C$ is an absolute constant.
\end{proof}

\begin{thm} [Bilinear Spectral Statistics of the Marcenko-Pastur Law]    \label{thm:spectral_statistics_MP}
Fix $q \in (0, \infty)$ and define $(m_k)_{k \geq 1}$, $(n_k)_{k \geq 1}$, $Z_k \in \RR^{m_k \times n_k}$ a matrix of iid standard Gaussian entries, and $W_k = \frac{1}{m_k} Z_k^T Z_k$ (as in Lemma \ref{lemma:max_MP_eigval_bound}), with $(m_k)_{k \geq 1}$ and $(n_k)_{k \geq 1}$ both strictly increasing and $\lim_{k \to \infty} \frac{n_k}{m_k} = q$. Let $f$ be a nonnegative continuous function that belongs to one of the following function classes:
\begin{enumerate}
    \item \textbf{Class 1 Functions:} $f \in C([0, \infty))$, with the added condition that $f$ is polynomially bounded.

    \item \textbf{Class 2 Functions:} $f \in C((0, \infty))$, with the added condition that $m_k > n_k$ for each $k \geq 1$, and $q < 1$.
\end{enumerate}
For any $k \geq 1$, let $f(W_k)$ denote the transformation of $W_k$ in which $f$ is applied spectrally. Then, the following results hold:

(a) Fix a constant $\rho_0 \in [-1, 1]$. For each $k \geq 1$, let $a_k', b_k' \in \SS^{n_k - 1}$ be fixed unit vectors such that ${a_k'}^T b_k' = \rho_0$. Then, 
$${a_k'}^T f(W_k) b_k' \asconverge \rho_0 \int f(x) \MP^{q}(dx).$$
Moreover, if $f$ is a Class 1 function, then $\lim_{k \to \infty} \EE[{a_k'}^T f(W_k) b_k'] = \rho_0 \int f(x) \MP^{q}(dx)$. 

(b) For each $k \geq 1$, define $\alpha_k \sim \mathcal{N}\left(0, \frac{1}{n_k}I_{n_k}\right)$ independently from $Z_k$. Then, 
$$\alpha_k^T f(W_k) \alpha_k \asconverge \int f(x) \MP^{q}(dx).$$

Moreover, if $f$ is a Class 1 function, then $\lim_{k \to \infty} \EE[\alpha_k^T f(W_{k}) \alpha_k] = \int f(x) \MP^{q}(dx)$. 

(c) For each $k \geq 1$, let $a_k \in \SS^{m_k - 1}$ be a fixed unit vector and sample $\alpha_k \sim \mathcal{N}\left(0, \frac{1}{n_k}I_{n_k}\right)$ independently from $Z_k$. Then,  
$$\frac{1}{\sqrt{m_k}}\alpha_k^T f(W_k) Z_k^T a_k \asconverge 0$$ as $k \to \infty$.
Moreover, if $f$ is a Class 1 function, then $\EE\left[\frac{1}{\sqrt{m_k}}\alpha_k^T f(W_k) Z_k^T a_k\right] = 0$ for each $k \geq 1$.

(d) Fix a constant $\rho_0 \in [-1, 1]$, and suppose that $m_k \geq n_k$. For each $k \geq 1$, let $a_k, b_k \in \SS^{m_k - 1}$ be fixed unit vectors such that ${a_k}^T b_k = \rho_0$. Then, 
$$\frac{1}{m_k} {a_k}^T Z_k f(W_k) Z_k^T b_k \asconverge \rho_0 q \int x f(x) \MP^{q}(dx).$$
Moreover, if $f$ is a Class 1 function, then $\lim_{k \to \infty} \EE\left[ \frac{1}{m_k} {a_k}^T Z_k f(W_k) Z_k^T b_k \right] = \rho_0 q \int x f(x) \MP^{q}(dx)$.    
\end{thm}

\begin{proof}
As before, let $\tilde{q} := \sup_{k \geq 1} n_k/m_k$.

First, if $f$ is a Class 1 function (i.e. $f$ is polynomially bounded) each of the quantities ${a_k'}^T f(W_k) b_k'$ (from (a)), $\alpha_k^T f(W_k) \alpha_k$ (from (b)), $\frac{1}{\sqrt{m_k}}\alpha_k^T f(W_k) Z_k^T a_k$ (from (c)), and $\frac{1}{m_k} {a_k}^T Z_k f(W_k) Z_k^T b_k$ (from (d)) is moment controlled, by Prop. \ref{prop:Lp_control}. Now, suppose that we are able to show that each of these quantities converges almost surely to its respective deterministic limit. Then, noting that if $\left(X_k\right)_{k \geq 1}$ is a moment controlled sequence with $X_k \asconverge x$ then $\EE[X_k] \to \EE[x] = x$ (Cor. \ref{cor:Lp_convergence_Vitali}), it follows that the expectation of each of these four quantities converges to its respective almost sure limit as $k \to \infty$. Thus, to complete our proof, it suffices to establish the aforementioned almost sure convergence results for these quantities, for functions $f$ within the two prescribed classes of functions. Before embarking on the proof, we state two key insights of our approach.

\textbf{Insight \#1.} We leverage the eigen-decomposition for the Wishart matrix $W_k = \frac{1}{m_k} Z_k^T Z_k$; in particular, when suitably defined, the eigenvectors and eigenvalues of $W_k$ are independent, with the eigenvectors being Haar distributed. To develop this idea formally, first suppose that $O$ is a Haar distributed $n_k \times n_k$ orthogonal matrix. Then, conditioned on $O$, $Z_k O \overset{D}{=} Z_k$. Because $O$ is Haar distributed, it follows that $Z_k O \overset{D}{=} Z_k$ and that $O^T W_k O \overset{D}{=} W_k$. Thus, $W_k$ has the same law as $Q_k \Lambda_k Q_k^T$ where $\Lambda_k := \diag\left( \left[ \lambda_1(W_k), \dots, \lambda_{n_k}(W_k) \right] \right)$ and $Q_k \sim \Unif(O(n_k))$, with $Q_k$ being independent of $\Lambda_k$. 
Furthermore, letting $R := \diag([1, \dots, 1, -1]) \in \RR^{n_k \times n_k}$ (with the only '$-1$' in the last diagonal entry) and defining the function $s(Q) := \mathds{1}_{\{\det Q = -1\}}$, we write $V_k := Q_k R^{s(Q_k)} \in \SO(n_k)$. Then, by construction, $V_k$ is independent of $\Lambda_k$ and $V_k F V_k^T = Q_k F Q_k^T$ for any diagonal matrix $F \in \RR^{n_k \times n_k}$ (note that $R$ commutes with $F$ and $R^2 = I_{n_k}$). Furthermore, we claim that $V_k$ is Haar distributed on $\SO(n_k)$. Indeed, for any $U \in \SO(n_k)$, $\det(UQ_k) = \det(Q_k)$, so $s(UQ_k) = s(Q_k)$ and $UV_k = (UQ_k) R^{s(U Q_k)} \overset{D}{=} Q_k R^{s(Q_k)} = V_k$ where we have noted that $Q_k$ is a Haar distributed orthogonal matrix (so that $UQ_k \overset{D}{=} Q_k$). Thus, $V_k \sim \Unif(\SO(n_k))$ is Haar distributed on $\SO(n_k)$. Altogether, $W_k \overset{D}{=} V_k \Lambda_k V_k^T$ and $f(W_k) \overset{D}{=} V_k F_k V_k^T$, where we have defined $F_k := f(\Lambda_k)$, which is $\Lambda_k$-measurable. Subsequently, we also use the fact that if $r: \RR^{n_k \times n_k} \to \RR$ is a continuous function, then $r(f(W_k)) \overset{D}{=} r(V_k F_k V_k^T)$.

\textbf{Insight \#2.} We apply a truncation argument: while the eigenvalues of $W_k$ are unbounded, the probability of the eigenvalues growing very large and, if $q < 1$ (with $m_k > n_k$), the probability of the eigenvalues getting very close to $0$, both decay exponentially in $m_k$.
Indeed, recalling Eq. \ref{eq:vershynin_max_eigval_reformulation} from the proof of Lemma \ref{lemma:max_MP_eigval_bound}, there is an absolute constant $C > 0$ such that, with $M_q := \left( C\left( \sqrt{\tilde{q}} + 1 \right) + 1 \right)^2$,
\begin{equation}    \label{eq:largest_eigval_tail_prob}
    \PP\left( \eigmax(W_k) > M_q \right) \leq 2e^{-m_k / C} \qquad \text{for all } k \geq 1.
\end{equation}
(See Vershynin, Thm. 4.4.3 \cite{Vershynin_2026}.) Moreover, define $L_q$ and $c_0'$ as follows. If $f$ is a Class 1 function, set $L_q := 0$ and $c_0' := 1$; if $f$ is a Class 2 function, so that $m_k > n_k$, then Cor. \ref{cor:min_singular_val_tail_bound} provides constants $L_q > 0$ and $c_0' > 0$, depending only on $q$ and $\tilde{q}$ (and not on $k$), such that
\begin{equation}    \label{eq:smallest_eigval_tail_prob}
    \PP\left( \eigmin(W_k) < L_q \right) \leq 2e^{-c_0' m_k} \qquad \text{for all } k \geq 1.
\end{equation}
For Class 1 functions, this inequality holds trivially, since $W_k$ is symmetric positive semi-definite. 

Furthermore, define the event
$$G_k := \left\{ L_q \leq \eigmin(W_k) \leq \eigmax(W_k) \leq M_q \right\},$$
and observe that $\mathds{1}_{G_k}$ is a measurable function in $\Lambda_k$. By Eqs. \ref{eq:largest_eigval_tail_prob} and \ref{eq:smallest_eigval_tail_prob}, 
\begin{equation}    \label{eq:truncation_bound}
    \PP(G_k^c) \leq \PP\left(\eigmax(W_k) > M_q\right) + \PP\left( \eigmin(W_k) < L_q \right) \leq 2e^{-m_k/C} + 2e^{-c_0' m_k}.
\end{equation}
By construction of $L_q$ and $M_q$ for Class 1 and Class 2 functions, in either case $f$ is continuous on $[L_q, M_q]$, so we may define the finite positive constants
\begin{equation}    \label{eq:lipschitz_B}
    B := 1 + \max_{x \in [L_q, M_q]} f(x) \qquad \text{and} \qquad \tilde{B} := 1 + \max_{x \in [L_q, M_q]} x f(x)
\end{equation}
which do not depend on $k$; we will use these constants when invoking Lemma \ref{lemma:key_example_h_Lip} as part of our proof.

Having laid out these key ideas, we now turn to proving (a), (b), (c), and (d). 

\paragraph{(a)} 

Noting that ${a_k'}^T b_k' = \rho_0$ for each $k$, we find that 
\begin{multline*}
    {a_k'}^T f(W_k) b_k' - \rho_0 \int f(x) \MP^{q}(dx) \\
    = {a_k'}^T \left( f(W_k) - \frac{\Tr(f(W_k))}{n_k} I_{n_k} \right) b_k' + \frac{\Tr(f(W_k))}{n_k} {a_k'}^T b_k' - \rho_0 \int f(x) \MP^{q}(dx)
\end{multline*}
\begin{equation}
    = {a_k'}^T \left( f(W_k) - \frac{\Tr(f(W_k))}{n_k} I_{n_k} \right) b_k' + \rho_0 \left( \frac{\Tr(f(W_k))}{n_k} - \int f(x) \MP^{q}(dx) \right) = r_{1, k} + \rho_0 r_{2, k}
\end{equation}
where we have defined $r_{1, k} := {a_k'}^T \left( f(W_k) - \frac{\Tr(f(W_k))}{n_k} I_{n_k} \right) b_k'$ and $r_{2, k} := \frac{\Tr(f(W_k))}{n_k} - \int f(x) \MP^{q}(dx)$. Note that $r_{2, k} \asconverge 0$, either by Cor. \ref{cor:unbounded_f_MP_convergence} if $f$ is a Class 1 Function, or by Cor. \ref{cor:unbounded_f_MP_convergence_zero_case} if $f$ is a Class 2 Function. Thus, it suffices to show that $r_{1, k} \asconverge 0$. 

To show $r_{1, k} \asconverge 0$, we will derive tail probability estimates for $r_{1, k}$ and then use Lemma \ref{lemma:borel_cantelli_as_convergence} to establish almost sure convergence to $0$. To proceed, we must show that, for any $\epsilon > 0$,
\begin{equation}
    \sum_{k \geq 1} \PP(|r_{1, k}| > \epsilon) < \infty.
\end{equation}
Since $W_k \overset{D}{=} V_k \Lambda_k V_k^T$, observe that $r_{1,k} \overset{D}{=} {a_k'}^T V_k \left( F_k - \frac{\Tr(F_k)}{n_k} I_{n_k} \right) V_k^T b_k' =: h_{F_k}(V_k)$.
Here, we have defined $h_F(V) := {a_k'}^T V ( F - \frac{\Tr(F)}{n_k} I_{n_k} ) V^T b_k'$ for $V \in \SO(n_k)$ and any diagonal matrix $F$.
We now apply a simple union bound to obtain good tail control over the eigenvalues of $W_k$: 
\begin{multline}    \label{eqn:r_1k_tail_bound}
    \PP(|r_{1, k}| > \epsilon) = \PP(|h_{F_k}(V_k)| > \epsilon) \leq \PP\left(G_k^c \right) + \PP\left(|h_{F_k}(V_k)| > \epsilon, \text{ } G_k \right) \\
    \leq 2e^{-m_k / C} + 2e^{-c_0' m_k} + \PP\left(|h_{F_k}(V_k)| > \epsilon, \text{ } G_k\right).
\end{multline}
On the event $G_k$, the diagonal entries of $F_k$ lie in $[0, B]$. Moreover, $F_k$ is $\Lambda_k$-measurable and $\mathds{1}_{G_k}$ is a measurable function in $\Lambda_k$, so $V_k$ is independent of $F_k$ and $G_k$. Thus, we proceed by conditioning on $\Lambda_k$:
$$\PP\left( \left| h_{F_k}(V_k) \right| > \epsilon, \text{ } G_k \right) = \EE\left[ \mathds{1}_{G_k} \mathds{1}_{\{|h_{F_k}(V_k)| > \epsilon\}} \right]$$
$$ = \EE\left[ \left. \EE\left[ \mathds{1}_{G_k} \mathds{1}_{\{|h_{F_k}(V_k)| > \epsilon\}} \right| \Lambda_k  \right] \right]  = \EE\left[ \mathds{1}_{G_k}  \left. \EE\left[\mathds{1}_{\{|h_{F_k}(V_k)| > \epsilon\}} \right| \Lambda_k  \right] \right]$$
$$ = \EE\left[ \mathds{1}_{G_k} \, \PP_{V \sim \Unif(\SO(n_k))}\left( |h_{F_k}(V)| > \epsilon \right) \right] \leq 2 e^{-C' n_k \epsilon^2}.$$
Here, we have invoked Lemma \ref{lemma:key_example_h_Lip}, which implies that 
\begin{equation}
    \PP_{V \sim \Unif(\SO(n_k))}\left( |h_{F}(V)| > \epsilon \right) \leq 2 e^{-C'' n_k \epsilon^2 / B^2}
\end{equation}
for any diagonal matrix $F \in \RR^{n_k \times n_k}$ whose diagonal entries lie in $[0, B]$ (which is the case for $F_k$ on the event $G_k$), where $C'' > 0$ is a suitable absolute constant. We have also defined $C' := C''/B^2$ as a fixed constant independent of $k$.
Thus, returning to Eq. \ref{eqn:r_1k_tail_bound}, we find that 
\begin{equation}
    \PP(|r_{1, k}| > \epsilon) \leq 2e^{-m_k/C} + 2e^{-c_0'm_k} + 2e^{-C' n_k \epsilon^2}.
\end{equation}
As a result, summation yields
$$\sum_{k \geq 1} \PP(|r_{1, k}| > \epsilon) \leq 2\sum_{k \geq 1} e^{-m_k/C} + 2\sum_{k \geq 1} e^{-c_0' m_k} + 2\sum_{k \geq 1} e^{-C' n_k \epsilon^2}$$
$$ \leq 2\sum_{m \geq 1} e^{-m/C} + 2\sum_{m \geq 1} e^{-c_0' m} + 2\sum_{n \geq 1} e^{-C' n \epsilon^2} < \infty$$
where we have noted that each of the latter sums is a convergent geometric series. We conclude that $r_{1, k} \asconverge 0$ (by Lemma \ref{lemma:borel_cantelli_as_convergence}), which completes our proof.

\paragraph{(b)} 

Note that for $\alpha_k \sim \mathcal{N}(0, I_{n_k}/n_k)$, we may decompose $\alpha_k$ as $\alpha_k = \norm{\alpha_k}_2 \cdot \halpha_k$, where the random variables $\halpha_k \sim \Unif(\SS^{n_k - 1})$ and $\norm{\alpha_k}_2$ are independent. We begin by writing 
\begin{multline*}
    \alpha_k^T f(W_k) \alpha_k - \int f(x) \MP^{q}(dx) \\
    = \alpha_k^T \left(f(W_k) - \frac{\Tr(f(W_k))}{n_k} I_{n_k} \right) \alpha_k + \frac{\Tr(f(W_k))}{n_k} \left( \norm{\alpha_k}_2^2 - 1 \right) + \left( \frac{\Tr(f(W_k))}{n_k} - \int f(x) \MP^{q}(dx) \right)
\end{multline*}
\begin{multline}
    = \norm{\alpha_k}_2^2 \cdot \halpha_k^T \left(f(W_k) - \frac{\Tr(f(W_k))}{n_k} I_{n_k} \right) \halpha_k + \frac{\Tr(f(W_k))}{n_k} \left( \norm{\alpha_k}_2^2 - 1 \right) \\ + \left( \frac{\Tr(f(W_k))}{n_k} - \int f(x) \MP^{q}(dx) \right).
\end{multline}

Recall from Cors. \ref{cor:unbounded_f_MP_convergence} and \ref{cor:unbounded_f_MP_convergence_zero_case} that $\frac{\Tr(f(W_k))}{n_k} - \int f(x) \MP^{q}(dx) \asconverge 0$. Consequently, $\frac{\Tr(f(W_k))}{n_k} \asconverge \int f(x) \MP^{q}(dx)$. Furthermore, $\norm{\alpha_k}_2^2 - 1 \asconverge 0$; to see why, observe that $\norm{\alpha_k}_2^2 - 1 = \sum_{i=1}^{n_k} \left( \frac{1}{n_k}(g_k)_i^2 - \frac{1}{n_k} \right)$ is a sum of mean-zero sub-exponential random variables, where $g_k = \sqrt{n_k} \alpha_k \sim \mathcal{N}(0, I_{n_k})$. As a result, the Bernstein inequality (Vershynin, Cor. 2.9.2 \cite{Vershynin_2026} with each $a_i = \frac{1}{n_k}$ and $X_i := (g_k)_i^2 - 1$) implies that
$$\PP\left( \left| \norm{\alpha_k}_2^2 - 1 \right| > \epsilon \right) \leq 2 \exp\left(-c n_k \min\left(\epsilon, \epsilon^2 \right) \right)$$
for any $\epsilon > 0$, and where $c > 0$ is an absolute constant. As a result,
$$\sum_{k \geq 1} \PP\left( \left| \norm{\alpha_k}_2^2 - 1 \right| > \epsilon \right) \leq 2 \sum_{k \geq 1} e^{-c n_k \min(\epsilon, \epsilon^2)} \leq 2 \sum_{n \geq 1} e^{-c n \min(\epsilon, \epsilon^2)} < \infty$$
because the latter sum is a convergent geometric series. Thus, by Lemma \ref{lemma:borel_cantelli_as_convergence}, we conclude that $\norm{\alpha_k}_2^2 - 1 \asconverge 0$.
Altogether, $\norm{\alpha_k}_2^2 \asconverge 1$ and $\frac{\Tr(f(W_k))}{n_k} \left( \norm{\alpha_k}_2^2 - 1 \right) \asconverge 0$. 

Thus, to show that $\alpha_k^T f(W_k) \alpha_k - \int f(x) \MP^{q}(dx) \asconverge 0$, it only remains to show that 
$$\halpha_k^T \left(f(W_k) - \frac{\Tr(f(W_k))}{n_k} I_{n_k} \right) \halpha_k \asconverge 0.$$ 
Now, define $\tilde{r}_{1, k} := \halpha_k^T \left(f(W_k) - \frac{\Tr(f(W_k))}{n_k} I_{n_k} \right) \halpha_k$. Then, $ \tilde{r}_{1,k} \overset{D}{=}  {\halpha_k}^T V_k\left( F_k - \frac{\Tr(F_k)}{n_k} I_{n_k} \right) V_k^T \halpha_k =: h_{F_k, \halpha_k}(V_k)$. Here, we have defined $h_{F, \alpha}(V) := \alpha^T V\left( F - \frac{\Tr(F)}{n_k} I_{n_k} \right) V^T \alpha$ for $V \in \SO(n_k)$ and any diagonal matrix $F \in \RR^{n_k \times n_k}$ and unit vector $\alpha \in \SS^{n_k - 1}$. Much like as in (a), we proceed by conditioning on $\Lambda_k$ and $\halpha_k$:
$$\PP\left( \left| h_{F_k, \halpha_k}(V_k) \right| > \epsilon, \text{ } G_k \right) = \EE\left[ \mathds{1}_{G_k} \mathds{1}_{\{|h_{F_k, \halpha_k}(V_k)| > \epsilon\}} \right]$$
$$ = \EE\left[ \left. \EE\left[ \mathds{1}_{G_k} \mathds{1}_{\{|h_{F_k, \halpha_k}(V_k)| > \epsilon\}} \right| \Lambda_k, \halpha_k  \right] \right]  = \EE\left[ \mathds{1}_{G_k}  \left. \EE\left[\mathds{1}_{\{|h_{F_k, \halpha_k}(V_k)| > \epsilon\}} \right| \Lambda_k, \halpha_k  \right] \right]$$
$$ = \EE\left[ \mathds{1}_{G_k} \, \PP_{V \sim \Unif(\SO(n_k))}\left( |h_{F_k, \halpha_k}(V)| > \epsilon \right) \right] \leq 2 e^{-C' n_k \epsilon^2}.$$
Here, we have again invoked Lemma \ref{lemma:key_example_h_Lip} as in (a), which implies that 
\begin{equation}
    \PP_{V \sim \Unif(\SO(n_k))}\left( |h_{F, \alpha}(V)| > \epsilon \right) \leq 2 e^{-C'' n_k \epsilon^2 / B^2}
\end{equation}
for any unit vector $\alpha \in \SS^{n_k - 1}$ and any diagonal matrix $F \in \RR^{n_k \times n_k}$ whose diagonal entries lie in $[0, B]$ (note that $B$ does not depend on $k$), and where $C'' > 0$ is an absolute constant.
Proceeding as in (a), we find that
\begin{multline}    \label{eqn:tilde_r_1k_tail_bound}
    \PP(|\tilde{r}_{1, k}| > \epsilon) =  \PP(|h_{F_k, \halpha_k}(V_k)| > \epsilon) \leq \PP\left(G_k^c \right) + \PP\left(|h_{F_k, \halpha_k}(V_k)| > \epsilon, \text{ } G_k \right) \\
    \leq 2e^{-m_k / C} + 2e^{-c_0' m_k} + 2 e^{-C'' n_k \epsilon^2 / B^2}.
\end{multline}
As a result, $\sum_{k \geq 1} \PP\left( |\tilde{r}_{1, k}| > \epsilon \right) < \infty$ for any fixed $\epsilon > 0$ (as in (a)). Thus, Lemma \ref{lemma:borel_cantelli_as_convergence} implies that $\tilde{r}_{1, k} \asconverge 0$, as desired.

\paragraph{(c)} 
First, if $f$ is a Class 1 function, then $\frac{1}{\sqrt{m_k}} \alpha_k^T f(W_k) Z_k^T a_k$ is moment controlled (Prop. \ref{prop:Lp_control}), so the independence of $\alpha_k$ and $Z_k$ together with $\EE[\alpha_k] = 0$ gives $\EE\left[ \frac{1}{\sqrt{m_k}} \alpha_k^T f(W_k) Z_k^T a_k \right] = 0$ for each $k \geq 1$.

Now, we must show that $r_{3, k} := \frac{1}{\sqrt{m_k}} \alpha_k^T f(W_k) Z_k^T a_k \asconverge 0$. Applying a union bound yields
\begin{equation}    \label{eqn:r_3k_tail_bound}
    \PP(|r_{3, k}| > \epsilon) \leq \PP\left(G_k^c \right) + \PP\left(|r_{3, k}| > \epsilon, \text{ } G_k \right)
    \leq 2e^{-m_k / C} + 2e^{-c_0' m_k} + \PP\left(|r_{3, k}| > \epsilon, \text{ } G_k \right).
\end{equation}

Now, observe that, when conditioned on any $Z_k$ such that $\eigmax(W_k) \leq M_q$ and $\eigmin(W_k) \geq L_q$, 
\begin{equation}
    r_{3, k} | Z_k \sim \mathcal{N}\left(0, \frac{1}{m_k n_k} a_k^T Z_k f(W_k)^2 Z_k^T a_k \right)
\end{equation}
which comes from the fact that $\alpha_k \sim \mathcal{N}\left(0, \frac{1}{n_k} I_{n_k}\right)$. Moreover, letting $\sigma_k^2 := \frac{1}{m_k n_k} a_k^T Z_k f(W_k)^2 Z_k^T a_k$, we find that
$$\sigma_k^2 \leq \frac{1}{n_k} \norm{a_k}_2 \opnorm{\frac{1}{\sqrt{m_k}} Z_k} \opnorm{{f(W_k)}}^2 \opnorm{\frac{1}{\sqrt{m_k}} Z_k} \norm{a_k}_2$$
$$ = \frac{1}{n_k} \opnorm{\frac{1}{\sqrt{m_k}} Z_k}^2 \opnorm{f(W_k)}^2 = \frac{1}{n_k} \eigmax(W_k) \left(\max_{1 \leq i \leq n_k} f(\lambda_i(W_k)) \right)^2$$ 
$$ \leq \frac{M_q}{n_k} \left( \max_{x \in [L_q, M_q]} f(x) \right)^2 \leq \frac{M_q B^2}{n_k}.$$
Noting that $G_k$ is $Z_k$-measurable ($\eigmin$ and $\eigmax$ are measurable functions) and using standard tail probability bounds for Gaussian random variables ($\left. \PP\left( |r_{3, k}| > \epsilon \text{ } \right| \text{ } Z_k \right) \leq 2 e^{-c \epsilon^2 / \sigma_k^2}$), we thus find that for any $\epsilon > 0$,
$$\PP\left( |r_{3,k}| > \epsilon, \, G_k \right) = \EE\left[ \mathds{1}_{G_k} \mathds{1}_{\{|r_{3, k}| > \epsilon\}} \right]$$
$$ = \EE\left[ \left. \EE\left[ \mathds{1}_{G_k} \mathds{1}_{\{|r_{3, k}| > \epsilon\}} \right| Z_k  \right] \right]  = \EE\left[ \mathds{1}_{G_k}  \left. \EE\left[\mathds{1}_{\{|r_{3, k}| > \epsilon\}} \right| Z_k  \right] \right]$$
$$ = \EE\left[ \mathds{1}_{G_k} \, \left. \PP\left( |r_{3, k}| > \epsilon \text{ } \right| \text{ } Z_k \right) \right] \leq \EE\left[ \mathds{1}_{G_k} \, 2 e^{-c \epsilon^2 / \sigma_k^2} \right] \leq 2 e^{-c' n_k \epsilon^2}$$
where $c >0$ is an absolute constant and $c' := \frac{c}{M_q B^2} > 0$ is a constant that does not depend on $k$.
In turn, Eq. \ref{eqn:r_3k_tail_bound} yields the inequality
\begin{equation}
    \PP(|r_{3, k}| > \epsilon) \leq 2e^{-m_k / C} + 2e^{-c_0' m_k} + 2 e^{ - {c' n_k \epsilon^2} }.
\end{equation}
As before, summation across $k$ then yields
$$\sum_{k \geq 1} \PP\left( |r_{3, k}| > \epsilon \right) \leq 2\sum_{k \geq 1} e^{-m_k/C} + 2\sum_{k \geq 1} e^{-c_0' m_k} + 2\sum_{k \geq 1} e^{-c' n_k \epsilon^2} < \infty.$$
By Lemma \ref{lemma:borel_cantelli_as_convergence}, we conclude that $r_{3, k} \asconverge 0$, as desired.

\paragraph{(d)} 
Here, we extend Insight \#1, as articulated at the outset of this proof.
 Letting $S_k \in \RR^{m_k \times n_k}$ be a diagonal matrix whose entries are the singular values of $ \frac{1}{\sqrt{m_k}} Z_k$, we first claim that $\frac{1}{\sqrt{m_k}} Z_k$ has the same law as $P_k S_k Q_k^T$ where $P_k \sim \Unif(O(m_k))$ and $Q_k \sim \Unif(O(n_k))$.
 As articulated in our description of Insight \# 1, this is because if $O \sim \Unif(O(n_k))$ and $O' \sim \Unif(O(m_k))$ are sampled independently and are each independent of $Z_k$, then $Z_k O \overset{D}{=} Z_k$ and $O' Z_k \overset{D}{=} Z_k$. Just as in Insight \#1, we may define $U_k := P_k {R'}^{s(P_k)} \in \SO(m_k)$ where $R' := \diag\left( \left[ 1, \dots, 1, -1 \right] \right) \in \RR^{m_k \times m_k}$, and show that $U_k$ is Haar distributed on $\SO(m_k)$ and independent of $\Lambda_k$, just like how $V_k$ is Haar distributed on $\SO(n_k)$ and independent of $\Lambda_k$. 
 Altogether, $W_k \overset{D}{=} V_k (S_k^T S_k) V_k^T$, $f(W_k) \overset{D}{=} V_k F_k V_k^T$ with $F_k := f(S_k^T S_k)$, and
 \begin{equation}
     \frac{1}{m_k} Z_k f(W_k) Z_k^T \overset{D}{=} U_k F_k' U_k^T, \qquad F_k' := S_k F_k S_k^T = \diag\left( \left[ \lambda_1 f(\lambda_1), \dots, \lambda_{n_k} f(\lambda_{n_k}), 0, \dots, 0 \right]\right) \in \RR^{m_k \times m_k}
 \end{equation}
where $\lambda_i := \lambda_i(W_k)$. 

Noting that $a_k^T b_k = \rho_0$, we may write
\begin{multline*}
    \frac{1}{m_k} a_k^T Z_k f(W_k) Z_k^T b_k - \rho_0 q \int x f(x) \MP^{q}(dx) \\
    = a_k^T \left( \frac{Z_k f(W_k) Z_k^T}{m_k} - \frac{\Tr\left( Z_k f(W_k) Z_k^T \right)}{m_k^2} I_{m_k}\right) b_k  + \rho_0 \left( \frac{\Tr\left( Z_k f(W_k) Z_k^T \right)}{m_k^2} - q\int x f(x) \MP^{q}(dx) \right) 
\end{multline*}
\begin{equation}
    = r_{4, k} + \rho_0 r_{5, k}
\end{equation}
where we have made the definitions
\begin{equation}
    r_{4, k} := a_k^T \left( \frac{Z_k f(W_k) Z_k^T}{m_k} - \frac{\Tr\left( Z_k f(W_k) Z_k^T \right)}{m_k^2} I_{m_k}\right) b_k
\end{equation}
\begin{equation}
    r_{5, k} := \frac{\Tr\left( Z_k f(W_k) Z_k^T \right)}{m_k^2} - q\int x f(x) \MP^{q}(dx).
\end{equation}
Note that 
$$\frac{\Tr\left( Z_k f(W_k) Z_k^T \right)}{m_k^2} = \frac{n_k}{m_k} \frac{\Tr\left(\frac{1}{m_k}Z_k f(W_k) Z_k^T \right)}{n_k} = \frac{n_k}{m_k} \frac{\Tr(F_k')}{n_k} = \frac{n_k}{m_k} \frac{\Tr(\tilde{f}(W_k))}{n_k}$$
where $\tilde{f}(x) := x f(x)$. Furthermore, $\tilde{f}$ belongs to the same class of functions as $f$ (i.e. Class 1 or Class 2). Thus, invoking Cor. \ref{cor:unbounded_f_MP_convergence} (for Class 1 functions) or Cor. \ref{cor:unbounded_f_MP_convergence_zero_case} (for Class 2 functions), we find that $\frac{\Tr(\tilde{f}(W_k))}{n_k} \asconverge \int x f(x) \MP^{q}(dx)$. Since $\frac{n_k}{m_k} \to q$, it follows that
$$r_{5, k} = \frac{n_k}{m_k} \frac{\Tr(\tilde{f}(W_k))}{n_k} - q \int x f(x) \MP^{q}(dx) \asconverge 0.$$
Thus, to finish our proof, it suffices to show that $r_{4, k} \asconverge 0$, which we do by again leveraging the Borel-Cantelli argument from Lemma \ref{lemma:borel_cantelli_as_convergence}. 
First, observe that ${r}_{4,k} \overset{D}{=}  {a_k}^T U_k\left( F_k' - \frac{\Tr(F_k')}{m_k} I_{m_k} \right) U_k^T b_k =: h_{F_k'}(U_k)$. Here, we have defined $h_{F}(U) := a_k^T U\left( F - \frac{\Tr(F)}{m_k} I_{m_k} \right) U^T b_k$ for $U \in \SO(m_k)$ and any diagonal matrix $F \in \RR^{m_k \times m_k}$.
As in previous parts, for any $\epsilon > 0$, we apply a union bound:
\begin{multline}    \label{eqn:r_4k_tail_bound}
    \PP(|r_{4, k}| > \epsilon) = \PP(|h_{F_k'}(U_k)| > \epsilon) \leq \PP\left(G_k^c \right) + \PP\left(|h_{F_k'}(U_k)| > \epsilon, \text{ } G_k \right) \\
    \leq 2e^{-m_k / C} + 2e^{-c_0' m_k} + \PP\left(|h_{F_k'}(U_k)| > \epsilon, \text{ } G_k\right).
\end{multline}
Much like as in (a), we proceed by conditioning on $\Lambda_k$, noting that $F_k'$ is $\Lambda_k$-measurable and that $\mathds{1}_{G_k}$ is a measurable function in $\Lambda_k$:
$$\PP\left( \left| h_{F_k'}(U_k) \right| > \epsilon, \text{ } G_k \right) = \EE\left[ \mathds{1}_{G_k} \mathds{1}_{\{|h_{F_k'}(U_k)| > \epsilon\}} \right]$$
$$ = \EE\left[ \left. \EE\left[ \mathds{1}_{G_k} \mathds{1}_{\{|h_{F_k'}(U_k)| > \epsilon\}} \right| \Lambda_k \right] \right] = \EE\left[ \mathds{1}_{G_k}  \left. \EE\left[\mathds{1}_{\{|h_{F_k'}(U_k)| > \epsilon\}} \right| \Lambda_k \right] \right]$$
$$ = \EE\left[ \mathds{1}_{G_k} \, \PP_{U \sim \Unif(\SO(m_k))}\left( |h_{F_k'}(U)| > \epsilon \right) \right] \leq 2 e^{-C'' m_k \epsilon^2 / \tilde{B}^2}.$$
Here, we have again invoked Lemma \ref{lemma:key_example_h_Lip}, which implies that 
\begin{equation}
    \PP_{U \sim \Unif(\SO(m_k))}\left( |h_{F}(U)| > \epsilon \right) \leq 2 e^{-C'' m_k \epsilon^2 / \tilde{B}^2}
\end{equation}
for any fixed diagonal matrix $F \in \RR^{m_k \times m_k}$ whose diagonal entries lie in $[0, \tilde{B}]$ (note that $F_k'$ has $m_k - n_k$ zero entries), and where $C'' > 0$ is an absolute constant.
From Eq. \ref{eqn:r_4k_tail_bound}, we thus find that
\begin{equation}    \label{eqn:r_4k_tail_bound_final}
    \PP(|{r}_{4, k}| > \epsilon)
    \leq 2e^{-m_k / C} + 2e^{-c_0' m_k} + 2 e^{-C'' m_k \epsilon^2 / \tilde{B}^2}.
\end{equation}
As a result, $\sum_{k \geq 1} \PP\left( |{r}_{4, k}| > \epsilon \right) < \infty$ for any fixed $\epsilon > 0$ (as in (a)). Thus, Lemma \ref{lemma:borel_cantelli_as_convergence} implies that ${r}_{4, k} \asconverge 0$, as desired.

\end{proof}

\section{Calculating the Features Learned after One Step of Gradient Descent}   \label{app:calculating_onestep_features}

\subsection{Calculating the feature kernels --- Proof of Theorem \ref{thm:feature_kernel_limits}} \label{app:calculating_Phi_ell}

In this section, we prove Thm. \ref{thm:feature_kernel_limits}, as stated in the main text. To proceed, we analyze the network at initialization and subsequently after performing one gradient step. In addition to the setup provided in Sec. \ref{sec:setup} of the paper, for each $1 \leq \ell \leq L$ we also define the \emph{gradient kernel} at initialization as ${G}_{\mu \nu}^\ell = \frac{g_\mu^\ell \cdot {g}_\nu^\ell}{N}$. Also, we define ${G}_{\mu \nu}^{L+1} := \frac{g_\mu^{L+1} \cdot g_\nu^{L+1}}{N} = 1$.
Here, we have defined $g_{\mu}^{L+1} := \sqrt{N} \in \RR$ by convention. Note that, in a deep linear network, the gradient $g^\ell_\mu$ is actually the same across all inputs $x_\mu$; thus, we frequently replace the term $g^\ell_\mu$ with the input-independent term $g^\ell$ defined by the same recursion $g^\ell = \frac{1}{\sqrt{N}} {W^\ell}^T g^{\ell + 1}$ with $g^{L+1} = \sqrt{N}$ (where we have defined $W^L := \left(w^L\right)^T$).

A key insight we use in establishing the desired $L^p$ convergence is to reduce to the case of $L^2$ convergence. Because all quantities we study have well-behaved moments (see Lemma \ref{lemma:everything_moment_bounded} and Remark \ref{rem:sums_products_L2_concentrate}  below), it will automatically follow from Lemma \ref{lemma:multiplying_adding_L2_limits} that $L^2$ convergence upgrades to $L^p$ convergence for any $p \geq 1$. We start by analyzing relevant quantities of the deep linear network at initialization and subsequently examine quantities arising in the one gradient step calculation.

\textbf{Initialization.} Noting that each entry of $\weight^\ell$ (for $\ell = 0, 1, \dots, L-1$) is an i.i.d. standard Gaussian random variable, Corollary \ref{cor:weight_matrix_proj_concentration} implies that for any pair of inputs $x, x' \in \RR^D$ (not necessarily in the training dataset), 
\begin{equation}    \label{eq:init_feature_kernel_concentration}
    \Phi_{t=0}^\ell(x, x') \overset{L^2}{\longrightarrow} \frac{x \cdot x'}{D}
\end{equation}
for each $\ell = 1, 2, \dots, L$ as $N \to \infty$. Since each training example $x_\mu \in \RR^D$ is a fixed vector, it further follows that
\begin{equation}
    \Phi^\ell_{t=0} \Lconverge K_x \in \RR^{n \times n}
\end{equation}
over the training dataset (recall that $(K_x)_{\alpha \beta} = \frac{x_\alpha \cdot x_\beta}{D}$ for any training inputs $x_\alpha, x_\beta$).
Once we establish that $\Phi^\ell_{t=0}(x, x')$ is moment controlled for any fixed inputs $x, x' \in \RR^D$ (see Lemma \ref{lemma:everything_moment_bounded} below), Lemma \ref{lemma:multiplying_adding_L2_limits} will imply that, in fact, $\Phi^\ell_{t=0}(x, x') \Lpconverge \frac{x \cdot x'}{D}$ and $\Phi^\ell_{t=0} \Lpconverge K_x$, establishing the first part of Theorem \ref{thm:feature_kernel_limits}.

Next, it will be helpful to compute the network output at initialization as well as the initial gradient kernels. First, we note that for any training example $x_\mu$, the output of the neural network $z(x_\mu, \theta_{t=0})$ at initialization concentrates to $0$ (in $L^2$) in the limit of $N \to \infty$. To see why, observe that at initialization, $\EE[z(x_\mu, \theta_{t=0})] = 0$ and 
$$\EE[z(x_\mu, \theta_{t=0})^2] = \frac{1}{N^2} \EE[{h_\mu^L}^T \mathbf{w}^L_{t=0} {\mathbf{w}^L_{t=0}}^T h_\mu^L]$$
$$ = \frac{1}{N^2} \EE\left[ \Tr\left(\wL_{t=0} {\wL_{t=0}}^T h_\mu^L {h_\mu^L}^T \right) \right]$$
$$ = \frac{1}{N^2} \Tr \left( \EE\left[ \wL_{t=0} {\wL_{t=0}}^T \right] \EE\left[ h_\mu^L {h_\mu^L}^T \right] \right)$$
$$ = \frac{1}{N^2} \Tr \left( \EE\left[ h_\mu^L {h_\mu^L}^T \right] \right) = \frac1N \EE\left[ \frac{h_\mu^L \cdot h_\mu^L}{N} \right] $$
$$ = \frac1N \EE\left[ \Phi_{\mu \mu; t=0}^L \right] = \frac{(K_x)_{\mu \mu}}{N} \overset{N \nearrow \infty}{\longrightarrow} 0.$$ 
Here, we have noted that 
$$\EE\left[ \Phi^L_{\mu \mu; t=0} \right] = \EE\left[ \frac1N {h_\mu^{L-1}}^T \left(\frac{1}{N_{L-1}} {W^{L-1}}^T W^{L-1} \right) h_\mu^{L-1} \right]$$
$$ = \EE\left[ \left. \EE\left[ \frac{1}{N_{L-1}} {h_\mu^{L-1}}^T \left(\frac1N {W^{L-1}}^T W^{L-1} \right) h_\mu^{L-1} \text{ } \right| \text{ } W^0, \dots, W^{L-2} \right] \right]$$
$$ = \EE\left[ \frac{1}{N_{L-1}} {h_\mu^{L-1}}^T \EE\left[\frac1N {W^{L-1}}^T W^{L-1} \right] h_\mu^{L-1} \right]$$
$$ = \EE\left[ \frac{1}{N_{L-1}} {h_\mu^{L-1}}^T h_\mu^{L-1} \right] = \EE\left[ \Phi^{L-1}_{\mu \mu; t=0} \right]$$
where we have noted that $\EE\left[\frac1N {W^{L-1}}^T W^{L-1} \right] = I_{N_{L-1}}$ since $W^{L-1} \in \RR^{N \times N_{L-1}}$ has iid standard Gaussian entries. Thus, by proceeding inductively, we find that 
$$\EE\left[ \Phi^L_{\mu \mu; t=0} \right] = \EE\left[ \Phi^0_{\mu \mu; t=0} \right] = (K_x)_{\mu \mu}$$
as claimed.

Consequently, 
\begin{equation}    \label{eq:Delta_mu_cvg}
    \Delta_\mu = - n \left. \frac{\partial \mathcal{L}}{\partial z(x_\mu;\theta)}\right|_{z(x_\mu;\theta_{t=0})} = y_\mu - z(x_\mu; \theta_{t=0}) \Lconverge y_\mu.
\end{equation}

We next address the gradient kernels at the start of training, where gradients are computed via GD on the mean-squared error loss over the training dataset. We inductively compute gradient kernels for earlier layers (going backward through the layers).
As a base case, observe that because $g^L = w^L \sim N(0, I_N)$, $\frac{{g}^{L} \cdot {g}^{L}}{N} \overset{L^2}{\longrightarrow} 1$ by Lemma \ref{lemma:gaussian_vector_concentration}. By induction, if $\frac{{g}^{\ell + 1} \cdot {g}^{\ell + 1}}{N} \overset{L^2}{\longrightarrow} 1$ for some $\ell \in \{1, 2, \dots, L-1\}$, then Lemma \ref{lemma:weight_matrix_proj_concentration}(b) precisely implies that $\frac{{g}^{\ell} \cdot {g}^{\ell}}{N} \overset{L^2}{\longrightarrow} 1$ as well. Altogether, at initialization, we find that 
\begin{equation}    \label{eq:g_dot_g}
    \frac{g^\ell \cdot g^\ell}{N} \overset{L^2}{\longrightarrow} 1
\end{equation}
for all layers $1 \leq \ell \leq L$ as $N \to \infty$.

\textbf{One Step.} We now begin our analysis of the feature kernels $\Phi^\ell$ after one gradient step. For ease of notation, we simply refer to the layer $\ell$ weight matrix at initialization $\wl_{t=0}$ by $\wl$. Note that for a fixed dataset the input kernel at initialization is given by $\Phi_{t=0}^0(x, x') = K_x(x, x') = \frac{x \cdot x'}{D}$. To calculate the feature and gradient kernels after one learning step, we proceed inductively: we calculate the feature kernel for successively deeper layers and we calculate the gradient kernel for successively shallower layers (i.e. going backward).

For $t \in \{0, 1\}$, we use $h_{\mu; t}^\ell$ and $h^\ell_{t}(x)$ to denote the hidden state at layer $\ell$ for the training input $x_\mu$ and the generic input $x$, respectively. We write $g^\ell$ as the gradient at initialization. Doing one learning step means that the layer $\ell$ weight matrix $\wl$ will have the update 
\begin{equation}    \label{eq:weight_update_GD}
    \Delta \wl = \frac{\eta}{n \sqrt{N_\ell}} \sum_{\mu=1}^n \Delta_\mu g^{\ell + 1} {h_{\mu; 0}^\ell}^T
\end{equation}
where $\eta \geq 0$ is the learning rate and $N_\ell = \begin{cases}
    N, \ell \geq 1 \\
    D, \ell = 0
\end{cases}$.
Then, for any vector $x \in \RR^D$ and $\ell \geq 1$, $$h_{t=1}^\ell(x) = \frac{1}{\sqrt{N_{\ell - 1}}}\left(\weight^{\ell - 1} + \Delta \weight^{\ell - 1} \right) h_{ t=1}^{\ell - 1}(x) = \frac{1}{\sqrt{N_{\ell - 1}}}\weight^{\ell - 1} h_{ t=1}^{\ell-1}(x) + \frac{1}{\sqrt{N_{\ell-1}}} \Delta \weight^{\ell - 1} h_{t=1}^{\ell - 1}(x).$$
Substituting our expression for $\Delta W^\ell$ then yields the hidden state update
\begin{equation}    \label{eqn:discrete_hidden_state_update}
    h_{t=1}^\ell(x) = \frac{1}{\sqrt{N_{\ell - 1}}}\weight^{\ell - 1} h_{t=1}^{\ell-1}(x) + \frac{\eta}{n} \sum_{\mu=1}^n \Delta_\mu \frac{h_{ t=1}^{\ell - 1}(x) \cdot h_{\mu; t=0}^{\ell - 1}}{N_{\ell - 1}} g^{\ell}.
\end{equation}
Since $h^\ell_{t=0}(x) = \frac{1}{\sqrt{N_{\ell-1}}} W^{\ell-1} h^{\ell-1}_{t=0}(x)$, defining $\Delta h^\ell(x) := h^\ell_{t=1}(x) - h^\ell_{t=0}(x)$ it follows that
\begin{equation}    \label{eq:Delta_h_def}
    \Delta h^\ell(x) = \frac{1}{\sqrt{N_{\ell-1}}} W^{\ell-1} \Delta h^{\ell - 1}(x) + f^\ell(x) g^{\ell},
\end{equation}
where we have defined 
\begin{equation}    \label{eq:fl_def}
    f^\ell(x) := \frac{\eta}{n} \sum_{\mu=1}^n \Delta_\mu \frac{h_{t=1}^{\ell - 1}(x) \cdot h_{\mu; t=0}^{\ell - 1}}{N_{\ell - 1}}.
\end{equation}
Furthermore, define 
\begin{equation}
    F(x) := \frac{\eta}{n} \sum_{\mu = 1}^n y_\mu K_x(x, x_\mu) = \frac{\eta}{n} \sum_{\mu=1}^n y_\mu \frac{x \cdot x_\mu}{D}.
\end{equation}

Additionally, for each $1 \leq \ell \leq L$ define the vectors $r^\ell \in \RR^N$ recursively by $r^1 := g^1$ (when $\ell = 1$) and
\begin{equation}    \label{eq:rl_recursion}
    r^\ell := \frac{1}{\sqrt{N}}W^{\ell -1} r^{\ell - 1} + g^\ell
\end{equation}
for $\ell \geq 2$. For each $1 \leq \ell \leq L$, observe that there exists a symmetric matrix $B_\ell \in \RR^{N \times N}$ that only depends on $W^0, \dots, W^{\ell - 1}$ such that $r^\ell = B_\ell g^\ell$. Indeed, when $\ell = 1$, $B_1 = I_N$ suffices and is symmetric. Proceeding inductively, if such a symmetric matrix $B_\ell$ exists for some $1 \leq \ell \leq L-1$, then 
$$r^{\ell + 1} = \frac{1}{\sqrt{N}}W^{\ell} r^{\ell} + g^{\ell+1} = \frac{1}{\sqrt{N}}W^{\ell} B_\ell g^\ell + g^{\ell+1} = \frac{1}{N}W^{\ell} B_\ell {W^{\ell}}^T g^{\ell + 1} + g^{\ell+1}$$
\begin{equation}    \label{eq:Bl_recursion}
    = \left(I_N + \frac{1}{N}W^{\ell} B_\ell {W^{\ell}}^T \right) g^{\ell + 1}
\end{equation}
and $I_N + \frac1N W^\ell B_\ell {W^{\ell}}^T$ is symmetric given that $B_\ell$ is.
Thus, we may recursively define $B_{\ell + 1} = I_N + \frac{1}{N}W^{\ell} B_\ell {W^{\ell}}^T$, which yields the identity $r^\ell = B_\ell g^\ell$ for each $1 \leq \ell \leq L$. We first note the following useful lemma:

\begin{lemma}   \label{lemma:everything_moment_bounded}
    For any fixed vector $x \in \RR^D$, define the quantity 
    \begin{equation}
        M := 1 + \sum_{j=1}^{L-1} \opnorm{\frac{1}{\sqrt{N}} W^j} + \frac{\norm{w^L}_2}{\sqrt{N}} + \sum_{\mu=1}^n \frac{\norm{h^1_{\mu;t=0}}_2}{\sqrt{N}} + \frac{\norm{h^1_{t=0}(x)}_2}{\sqrt{N}}.
    \end{equation}
    Then, $\sup_{N} \EE M^p < \infty$ for each $p \geq 1$, that is, $M$ is moment controlled with respect to $N$. Consequently, each of $\Delta_\mu$, $f^\ell(x)$, $f^\ell(x_\mu)$, $\opnorm{B_\ell}$, and $\frac{\norm{z}_2}{\sqrt{N}}$ for $z \in \left\{ h^\ell_{\mu;t=0}, h^\ell_{t=0}(x),  h^\ell_{\mu;t=1}, h^\ell_{t=1}(x), r^\ell, g^\ell : 1 \leq \ell \leq L \right\}$, is moment controlled over all $1 \leq \mu \leq n$.
\end{lemma}

\begin{proof}
    $M$ is uniformly moment controlled by Cor. \ref{cor:Lp_control_of_NN_weights_and_states}. The resulting aforementioned quantities contain sums and products of the terms $\frac{1}{\sqrt{N}} W^j$ ($1 \leq j \leq L-1$), $\frac{w^L}{\sqrt{N}}$, and $\frac{h^1_{t=0}(\tilde{x})}{\sqrt{N}}$ for $\tilde{x} \in \{x, x_\mu\}$, meaning that each of the aforementioned quantities may be bounded above by a polynomial in $M$, thereby implying that it is moment controlled, because $M$ is. First, for $\tilde{x} \in \{x, x_\mu\}$, 
    \begin{equation}    \label{eq:h_bounded}
        \begin{aligned}
            \frac{\norm{h^\ell_{t=0}(\tilde{x})}_2}{\sqrt{N}} = \frac{\norm{\frac{1}{\sqrt{N_{\ell-1}}} W^{\ell - 1}h^{\ell-1}_{t=0}(\tilde{x})}_2}{\sqrt{N}} \leq \opnorm{\frac{1}{\sqrt{N_{\ell-1}}} W^{\ell - 1}}\frac{\norm{h^{\ell-1}_{t=0}(\tilde{x})}_2}{\sqrt{N}}  \\
            \leq M \frac{\norm{h^{\ell-1}_{t=0}(\tilde{x})}_2}{\sqrt{N}} \leq \dots \leq M^\ell.
        \end{aligned}
    \end{equation}
    Similarly,
    \begin{equation}    \label{eq:g_bounded}
        \begin{aligned}
            \frac{\norm{g^\ell}_2}{\sqrt{N}} = \frac{\norm{\frac{1}{\sqrt{N}} {W^{\ell}}^T g^{\ell+1}}_2}{\sqrt{N}} \leq \opnorm{\frac{1}{\sqrt{N}} W^{\ell}}\frac{\norm{g^{\ell+1}}_2}{\sqrt{N}} \\
            \leq M \frac{\norm{g^{\ell+1}}_2}{\sqrt{N}} \leq \dots \leq M^{L+1-\ell}.
        \end{aligned}
    \end{equation}
    The result for $r^\ell$ readily follows from the result for $g^\ell$ along with the recursive definition for $r^\ell$ (Eq. \ref{eq:rl_recursion}).
    Furthermore, noting that $|z(x_\mu;\theta_{t=0})| = \left| \frac{w^L}{\sqrt{N}} \cdot \frac{h^L_{\mu;t=0}}{\sqrt{N}} \right| \leq \frac{\norm{w^L}_2}{\sqrt{N}} \cdot \frac{\norm{h^L_{\mu;t=0}}_2}{\sqrt{N}} \leq M^{L+1}$ (using Eq. \ref{eq:h_bounded}),
    $$|\Delta_\mu| \leq |y_\mu| + M^{L+1}.$$
    Next, $\opnorm{B_1} = 1$ and for $\ell > 1$, $$\opnorm{B_\ell} \leq 1 + M^2 \opnorm{B_{\ell-1}}$$
    by Eq. \ref{eq:Bl_recursion}, which establishes moment control for $\opnorm{B_\ell}$. 
    Finally, moment control of $f^\ell(\tilde{x})$ and $\frac{\norm{h^\ell_{t=1}(\tilde{x})}_2}{\sqrt{N}}$ is readily established by induction. Indeed, when $\ell = 1$, $\left|f^\ell(\tilde{x}) \right| \leq \frac{\eta}{n} \sum_{\mu=1}^n |\Delta_\mu| \left|\frac{\tilde{x} \cdot x_\mu}{D} \right|$ is bounded above by a polynomial in $M$ because $\Delta_\mu$ is. Similarly, $\frac{\norm{h^1_{t=1}(\tilde{x})}_2}{\sqrt{N}}$ is moment controlled because $\frac{h^1_{t=1}(\tilde{x})}{\sqrt{N}}$
    involves sums and products of $h^1_{t=0}(\tilde{x})$, $h^1_{\mu;t=0}$, $g^1$, and $\Delta_\mu$, each of which is moment controlled. Moment control for $\ell > 1$ then readily follows by induction from Eqs. \ref{eq:Delta_h_def} and \ref{eq:fl_def}.
\end{proof}

\begin{rem} \label{rem:sums_products_L2_concentrate}
    Combining Lemma \ref{lemma:everything_moment_bounded} with Lemma \ref{lemma:multiplying_adding_L2_limits}, we effectively have that sums and products of quantities that individually concentrate to nonrandom constants in $L^2$ also concentrate to nonrandom constants in $L^2$. Moreover, for any such moment controlled quantities $X, Y$, if $X \Lconverge 0$, Lemma \ref{lemma:multiplying_adding_L2_limits} implies that $XY \Lconverge 0$ as well.
\end{rem}

For any fixed vectors $x, x' \in \RR^D$, observe that $$\left| \Phi^\ell_{t=0}(x, x') \right| = \left| \frac{h^\ell_{t=0}(x) \cdot h^\ell_{t=0}(x')}{N} \right| \leq \frac{\norm{h^\ell_{t=0}(x)}_2}{\sqrt{N}} \cdot \frac{\norm{h^\ell_{t=0}(x')}_2}{\sqrt{N}}$$ is moment controlled by Lemma \ref{lemma:everything_moment_bounded} and because we have established that $\Phi^\ell_{t=0}(x, x') \Lconverge \frac{x \cdot x'}{D}$, a key consequence arising from Lemma \ref{lemma:multiplying_adding_L2_limits} is that $\Phi^\ell_{t=0}(x, x') \Lpconverge \frac{x \cdot x'}{D}$ for any $p \geq 1$. In particular, over the training dataset, we find that $\Phi^\ell_{t=0} \Lpconverge K_x$. This establishes the first part of Theorem \ref{thm:feature_kernel_limits}. 

We next state the following relevant lemmas:

\begin{lemma}   \label{lemma:gl_conditioning}
    For each $1 \leq \ell \leq L$, let $\mathcal{H}_\ell$ denote the sigma algebra generated by all of $W^0, \dots, W^{L-1}, w^L$ except for the weight matrix at layer $\ell$. Then, $g^\ell | \mathcal{H}_\ell \sim N\left( 0, \frac{\norm{g^{\ell+1}}_2^2}{N} I_N \right)$. Moreover, for any $u \in \RR^N$ that is $\mathcal{H}_\ell$-measurable,
    $$\EE\left[ \left( \frac{g^\ell \cdot u}{N} \right)^2 \right] \leq \frac1N \EE\left[ \left(\frac{\norm{g^{\ell+1}}_2^2}{N} \right)^2 \right]^{1/2} \EE\left[ \left( \frac{\norm{u}_2^2}{N} \right)^2 \right]^{1/2}.$$
    In particular, because $\frac{\norm{g^{\ell+1}}_2^2}{N} \Lconverge 1$, if $\frac{1}{\sqrt{N}}\norm{u}_2$ is moment controlled, then $\frac{g^\ell \cdot u}{N} \Lconverge 0$.
\end{lemma}

\begin{proof}
    When $\ell = L$, $g^L = w^L$ is independent of the sigma algebra $\mathcal{H}_L$, so $g^L |\mathcal{H}_L$ follows the same law as $g^L = w^L \sim N(0, I_N)$, which agrees with our convention that $g^{L+1} := \sqrt{N}$ so that $\frac{\norm{g^{L+1}}_2^2}{N} = 1$. For $1 \leq \ell \leq L-1$, since $g^\ell = \frac{1}{\sqrt{N}}{W^\ell}^T g^{\ell + 1}$ and $g^{\ell+1}$ is $\mathcal{H}_\ell$-measurable, $g^\ell | \mathcal{H}_\ell \sim N\left(0, \frac{\norm{g^{\ell+1}}_2^2}{N} I_N \right)$ owing to the Gaussian weights $W^\ell$. Moreover, for any $u \in \RR^N$ that is $\mathcal{H}_\ell$-measurable,
    $$\EE\left[ \left( \frac{g^\ell \cdot u}{N} \right)^2 \right] = \EE\left[\left. \EE\left[ \left( \frac{g^\ell \cdot u}{N} \right)^2 \right| \mathcal{H}_\ell \right]\right] = \frac{1}{N^2}\EE\left[ u^T \left.\EE\left[ g^\ell {g^\ell}^T \right| \mathcal{H}_\ell\right] u \right]$$
    $$ = \frac{1}{N} \EE\left[ \frac{\norm{g^{\ell+1}}_2^2}{N} \cdot \frac{\norm{u}_2^2}{N} \right] \leq \frac1N \EE\left[ \left(\frac{\norm{g^{\ell+1}}_2^2}{N} \right)^2 \right]^{1/2} \EE\left[ \left( \frac{\norm{u}_2^2}{N} \right)^2 \right]^{1/2}$$
    where we have used the Cauchy-Schwarz inequality. Since we established earlier that $\frac{\norm{g^{\ell+1}}_2^2}{N} \Lconverge 1$, it follows that if $\frac{1}{\sqrt{N}}\norm{u}_2$ is moment controlled, then $\EE\left[ \left( \frac{\norm{u}_2^2}{N} \right)^2 \right]^{1/2}$ remains uniformly bounded above as $N \to \infty$, and so 
    $$\lim_{N \to \infty} \EE\left[ \left( \frac{g^\ell \cdot u}{N} \right)^2 \right] = 0$$
    precisely meaning that $\frac{g^\ell \cdot u}{N} \Lconverge 0$.
\end{proof}

\begin{lemma}[Gaussian Conditioning] \label{lemma:gaussian_conditioning}
    Define $\mathcal{H}_\ell$ as in Lemma \ref{lemma:gl_conditioning}. For any $\ell \in \{1, \dots, L-1\}$, define $\Pi_{\ell} := I_N - \frac{1}{\norm{g^{\ell+1}}_2^2}g^{\ell+1}{g^{\ell+1}}^T$. Then, for any $u, u' \in \RR^N$ that are measurable in the sigma algebra generated by $\mathcal{H}_\ell$ and $g^{\ell}$, 
    \begin{equation}
        \frac{\frac{1}{\sqrt{N}} W^{\ell} u \cdot \frac{1}{\sqrt{N}} W^{\ell} u'}{N} = \frac{N}{\norm{g^{\ell+1}}_2^2} \cdot \frac{g^\ell \cdot u}{N} \cdot \frac{g^\ell \cdot u'}{N} + \frac{\Pi_\ell W^\ell u \cdot \Pi_\ell W^\ell u'}{N^2}.
    \end{equation}
    Moreover, 
    \begin{equation}
        \EE\left[ \left( \frac{\Pi_\ell W^\ell u \cdot \Pi_\ell W^\ell u'}{N^2} - \frac{u \cdot u'}{N}\right)^2 \right] \leq \frac2N \EE\left[ \left( \frac{\norm{u}_2^2}{N} \right)^2 \right]^{1/2} \EE\left[ \left( \frac{\norm{u'}_2^2}{N} \right)^2 \right]^{1/2}.
    \end{equation}
    In particular, if $\frac{1}{\sqrt{N}}\norm{u}_2$ and $\frac{1}{\sqrt{N}}\norm{u'}_2$ are moment controlled (with respect to $N$), then
    $$\frac{\Pi_\ell W^\ell u \cdot \Pi_\ell W^\ell u'}{N^2} \Leq \frac{u \cdot u'}{N}.$$
\end{lemma}

\begin{proof}
    Let $\hat{g}^{\ell+1} := \frac{g^{\ell+1}}{\norm{g^{\ell + 1}}_2}$. 
    Then, $\Pi_\ell = I_N - \hat{g}^{\ell+1} (\hat{g}^{\ell+1})^T$ is a projection operator 
    and $$W^\ell = \hat{g}^{\ell+1} (\hat{g}^{\ell+1})^T W^\ell + \Pi_\ell W^\ell$$
    $$\implies \frac{1}{\sqrt{N}} W^\ell u = \hat{g}^{\ell+1} \left( \frac{1}{\sqrt{N}} (\hat{g}^{\ell+1})^T W^\ell u\right) + \frac{1}{\sqrt{N}} \Pi_\ell W^\ell u$$
    $$ = \frac{\hat{g}^{\ell+1}}{\norm{g^{\ell+1}}_2} \left( \frac{1}{\sqrt{N}} {{g}^{\ell+1}}^T W^\ell u\right) + \frac{1}{\sqrt{N}} \Pi_\ell W^\ell u$$
    $$ = \frac{\hat{g}^{\ell+1}}{\norm{g^{\ell+1}}_2} \left( g^\ell \cdot u\right) + \frac{1}{\sqrt{N}} \Pi_\ell W^\ell u.$$
    Noting that $\Pi_\ell g^{\ell+1} = 0$, we thus obtain
    $$\frac{\frac{1}{\sqrt{N}} W^{\ell} u \cdot \frac{1}{\sqrt{N}} W^{\ell} u'}{N} = \frac1N \frac{1}{\norm{g^{\ell+1}}_2^2} (g^\ell \cdot u)(g^\ell \cdot u') + \frac{\Pi_\ell W^\ell u \cdot \Pi_\ell W^\ell u'}{N^2}$$
    $$ = \frac{N}{\norm{g^{\ell+1}}_2^2} \frac{g^\ell \cdot u}{N}\frac{g^\ell \cdot u'}{N} + \frac{\Pi_\ell W^\ell u \cdot \Pi_\ell W^\ell u'}{N^2}$$
    as claimed.
    Next, we turn to analyzing the second claim. First, conditioned on $\mathcal{H}_\ell$, $g^\ell = \frac{1}{\sqrt{N}} {W^\ell}^T g^{\ell + 1} \sim N\left(0, \frac{\norm{g^{\ell+1}}_2^2}{N} I_N \right)$ (note that $g^{\ell + 1}$ is $\mathcal{H}_\ell$-measurable) and $\Pi_\ell W^{\ell}$ is a Gaussian weight matrix (as $\Pi_\ell$ is fixed under this conditioning). Moreover, we claim that, under this conditioning, $g^\ell$ and $\Pi_\ell W^\ell$ are independent Gaussian random variables. Indeed, consider the covariance between any entries $\left( \Pi_\ell W^\ell\right)_{ij}$ and $(g^\ell)_k$:
    $$\left. \EE\left[  \left( \Pi_\ell W^\ell\right)_{ij} (g^\ell)_k \right| \mathcal{H}_\ell \right] = \frac{1}{\sqrt{N}} \sum_{r, s=1}^N (\Pi_\ell)_{ir} \EE\left[ (W^\ell)_{rj} (W^\ell)_{sk} \right] (g^{\ell+1})_s$$
    $$ = \frac{1}{\sqrt{N}} \sum_{r, s=1}^N (\Pi_\ell)_{ir} \delta_{rs} \delta_{jk} (g^{\ell+1})_s = \frac{\delta_{jk}}{\sqrt{N}} \sum_{r=1}^N (\Pi_\ell)_{ir} (g^{\ell+1})_r$$
    $$ = \frac{\delta_{jk}}{\sqrt{N}} \left( \Pi_\ell g^{\ell+1}\right)_i = 0$$
    since $\Pi_\ell$ is a projection onto the orthogonal complement of $\text{span}\left(\left\{ g^{\ell+1} \right\}\right)$. Thus, conditioned on $\mathcal{H}_\ell$, $g^\ell$ and $\Pi_\ell W^\ell$ are jointly Gaussian and uncorrelated, and hence independent. As a result, letting $\mathcal{J}_\ell$ denote the sigma algebra generated by $\mathcal{H}_\ell$ and $g^\ell$, $\Pi_\ell W^\ell | \mathcal{J}_\ell$ follows the same law as $\Pi_\ell W^\ell | \mathcal{H}_\ell$. In particular, 
    $$\left. \EE\left[ \frac{\Pi_\ell W^\ell u \cdot \Pi_\ell W^\ell u'}{N^2} \right| \mathcal{J}_\ell \right] = \frac{1}{N^2} u^T \EE\left[ {W^{\ell}}^T \Pi_\ell W^{\ell} \right] u'$$
    $$ = \frac{1}{N^2} \sum_{i, j, k, m=1}^N u_i (\Pi_\ell)_{jk} u'_m \EE\left[(W^\ell)_{ji} (W^{\ell})_{km} \right] = \frac{1}{N^2} \sum_{i, j, k, m=1}^N u_i (\Pi_\ell)_{jk} u'_m \delta_{jk} \delta_{im}$$
    $$ = \frac{1}{N^2} \sum_{i, j =1}^N u_i u_i' (\Pi_\ell)_{jj} = \frac{1}{N^2} (u \cdot u') \Tr(\Pi_\ell) = \frac{N-1}{N} \cdot \frac{u \cdot u'}{N}.$$
    Here, we have used the fact that $\Tr(\Pi_\ell) = N-1$. We may now use Wick's theorem to calculate the following quantity: 
    $$\left. \EE\left[ \left( \frac{\Pi_\ell W^\ell u \cdot \Pi_\ell W^\ell u'}{N^2} \right)^2 \right| \mathcal{J}_\ell \right] = \frac{1}{N^4} \EE \left[ \left( u^T {W^\ell}^T \Pi_\ell W^\ell u' \right)^2 \right]$$ 
    $$= \frac{1}{N^4}\sum_{i, j, k, m=1}^N \sum_{p, q, r, s=1}^N u_i (\Pi_\ell)_{jk} u'_m u_p (\Pi_\ell)_{qr} u'_s \EE \left[ W^\ell_{ji} W^{\ell}_{km} W^{\ell}_{qp} W^\ell_{rs} \right]$$
    $$= \frac{1}{N^4}\sum_{i, j, k, m=1}^N \sum_{p, q, r, s=1}^N u_i (\Pi_\ell)_{jk} u'_m u_p (\Pi_\ell)_{qr} u'_s (\delta_{jk}\delta_{im}\delta_{qr} \delta_{ps} + \delta_{jq}\delta_{ip}\delta_{kr} \delta_{ms} + \delta_{jr}\delta_{is}\delta_{kq} \delta_{mp})$$
    \begin{align*}
        = \frac{1}{N^4} \sum_{i, j, p, q = 1}^N u_i u_i' u_p u_p' (\Pi_\ell)_{jj} (\Pi_\ell)_{qq} + \frac{1}{N^4} \sum_{i, j, k, m=1}^N u_i^2 {u_m'}^2 (\Pi_{\ell})_{jk}^2 + \frac{1}{N^4} \sum_{i, j, k, m=1}^N u_i u_i' u_m u_m' (\Pi_\ell)_{jk}^2
    \end{align*}
    $$ = \frac{1}{N^2} \frac{(u \cdot u')^2}{N^2} \Tr(\Pi_\ell)^2 + \frac{1}{N^2} \frac{\norm{u}_2^2}{N} \frac{\norm{u'}_2^2}{N} \Tr(\Pi_\ell^2) + \frac{1}{N^2} \frac{(u\cdot u')^2}{N^2} \Tr(\Pi_\ell^2)$$
    $$ = \frac{(N-1)^2}{N^2} \frac{(u \cdot u')^2}{N^2} + \frac{N-1}{N^2} \frac{\norm{u}_2^2}{N} \frac{\norm{u'}_2^2}{N} + \frac{N-1}{N^2} \frac{(u \cdot u')^2}{N^2}.$$
    As a result, we find that
    \begin{multline*}
        \left. \EE\left[ \left( \frac{\Pi_\ell W^\ell u \cdot \Pi_\ell W^\ell u'}{N^2} - \frac{u\cdot u'}{N} \right)^2 \right| \mathcal{J}_\ell \right] = \left. \EE\left[ \left( \frac{\Pi_\ell W^\ell u \cdot \Pi_\ell W^\ell u'}{N^2} \right)^2 \right| \mathcal{J}_\ell \right] \\- 2 \left. \EE\left[ \frac{\Pi_\ell W^\ell u \cdot \Pi_\ell W^\ell u'}{N^2}  \right| \mathcal{J}_\ell \right] \frac{u \cdot u'}{N} + \frac{(u \cdot u')^2}{N^2}
    \end{multline*}
    \begin{align*}
        = \frac{(N-1)^2}{N^2} \frac{(u \cdot u')^2}{N^2} + \frac{N-1}{N^2} \frac{\norm{u}_2^2}{N} \frac{\norm{u'}_2^2}{N} + \frac{N-1}{N^2} \frac{(u \cdot u')^2}{N^2} 
        \\ - 2 \frac{N-1}{N} \frac{(u \cdot u')^2}{N^2} + \frac{(u \cdot u')^2}{N^2}
    \end{align*}
    $$ = \frac{N-1}{N^2} \frac{\norm{u}_2^2}{N} \frac{\norm{u'}_2^2}{N} + \frac{1}{N} \frac{(u \cdot u')^2}{N^2}$$
    $$ \leq \frac2N \frac{\norm{u}_2^2}{N} \frac{\norm{u'}_2^2}{N}$$
    where we have used the Cauchy-Schwarz inequality. In turn, the tower law of expectations yields
    $$\left. \EE\left[ \left( \frac{\Pi_\ell W^\ell u \cdot \Pi_\ell W^\ell u'}{N^2} - \frac{u\cdot u'}{N} \right)^2 \right. \right] = \EE\left[\left. \EE\left[ \left( \frac{\Pi_\ell W^\ell u \cdot \Pi_\ell W^\ell u'}{N^2} - \frac{u\cdot u'}{N} \right)^2 \right| \mathcal{J}_\ell \right]\right]$$ 
    $$\leq \frac{2}{N} \EE\left[ \frac{\norm{u}_2^2}{N} \frac{\norm{u'}_2^2}{N} \right] \leq \frac2N \EE\left[ \left( \frac{\norm{u}_2^2}{N} \right)^2 \right]^{1/2} \EE\left[ \left( \frac{\norm{u'}_2^2}{N} \right)^2 \right]^{1/2}$$
    where we have used the Cauchy-Schwarz inequality. In particular, if $\frac{1}{\sqrt{N}} \norm{u}_2$ and $\frac{1}{\sqrt{N}} \norm{u'}_2$ are moment controlled, then the latter term is $O(1/N)$, implying that
    $$\lim_{N \to \infty} \left. \EE\left[ \left( \frac{\Pi_\ell W^\ell u \cdot \Pi_\ell W^\ell u'}{N^2} - \frac{u\cdot u'}{N} \right)^2 \right. \right] = 0$$
    which precisely means that
    $$\frac{\Pi_\ell W^\ell u \cdot \Pi_\ell W^\ell u'}{N^2} \Leq \frac{u\cdot u'}{N}.$$
\end{proof}

We now apply these lemmas to prove the second part of Thm. \ref{thm:feature_kernel_limits}.

First, we claim that for any fixed vector $x \in \RR^D$ (not necessarily in the dataset),
\begin{equation}    \label{eq:h_dot_r}
    \frac{h^\ell_{t=0}(x) \cdot r^\ell}{N} \Lconverge 0
\end{equation}
for each $1 \leq \ell \leq L$. Indeed, note that $B_\ell$ and $h^\ell_{t=0}(x)$ are both $\mathcal{H}_\ell$-measurable and that $\frac{1}{\sqrt{N}}\norm{B_\ell h^\ell_{t=0}(x)}_2 = \norm{B_\ell \left( \frac{1}{\sqrt{N}} h^\ell_{t=0}(x) \right)}_2 \leq \opnorm{B_\ell} \norm{\frac{1}{\sqrt{N}} h^\ell_{t=0}(x) }_2$ is moment controlled by Lemma \ref{lemma:everything_moment_bounded}. Thus,
$$\frac{h^\ell_{t=0}(x) \cdot r^\ell}{N} = \frac{h^\ell_{t=0}(x) \cdot B_\ell g^\ell}{N} = \frac{g^\ell \cdot B_\ell h^\ell_{t=0}(x)}{N} \Lconverge 0$$
where we have applied Lemma \ref{lemma:gl_conditioning}.

Next, we claim that
\begin{equation}    \label{eq:g_dot_r}
    \frac{g^\ell \cdot r^\ell}{N} \Lconverge \ell
\end{equation}
for each $1 \leq \ell \leq L$. We proceed by induction. Indeed, when $\ell = 1$, $r^1 = g^1$, and we have shown that $\frac{g^1 \cdot g^1}{N} \Lconverge 1$. For $\ell > 1$, we use the recursive equation defining $r^\ell$:
$$\frac{g^\ell \cdot r^\ell}{N} = \frac{g^\ell \cdot g^\ell}{N} + \frac{g^\ell \cdot \frac{1}{\sqrt{N}}W^{\ell - 1} r^{\ell - 1}}{N} = \frac{g^\ell \cdot g^\ell}{N} + \frac{\frac{1}{\sqrt{N}}{W^{\ell - 1}}^T g^\ell \cdot r^{\ell - 1}}{N} $$
$$ = \frac{g^\ell \cdot g^\ell}{N} + \frac{g^{\ell - 1} \cdot r^{\ell - 1}}{N} \Leq 1 + \frac{g^{\ell - 1} \cdot r^{\ell - 1}}{N}.$$ As a result, induction readily yields that $\frac{g^\ell \cdot r^\ell}{N} \Lconverge \ell$ for each $1 \leq \ell \leq L$.

Next, we claim that 
\begin{equation}    \label{eq:r_dot_r}
    \frac{r^\ell \cdot r^\ell}{N} \Lconverge \sum_{j =1}^\ell j^2
\end{equation}
for each $1 \leq \ell \leq L$. When $\ell = 1$, $\frac{r^1 \cdot r^1}{N} = \frac{g^1 \cdot g^1}{N} \Lconverge 1 = 1^2$. Proceeding inductively, assuming the claimed result holds for some $1 \leq \ell \leq L-1$, we now use the recursive definition of $r^{\ell+1}$ to show the analogous result for $\ell +1$:
$$\frac{r^{\ell+1} \cdot r^{\ell+1}}{N} = \frac{\left(\frac{1}{\sqrt{N}}W^{\ell}r^{\ell} + g^{\ell+1} \right) \cdot \left(\frac{1}{\sqrt{N}}W^{\ell}r^{\ell} + g^{\ell+1} \right)}{N}$$
$$ = \frac{\frac{1}{\sqrt{N}}W^{\ell}r^{\ell} \cdot \frac{1}{\sqrt{N}}W^{\ell}r^{\ell}}{N} + 2 \frac{\frac{1}{\sqrt{N}}W^{\ell}r^{\ell} \cdot g^{\ell+1}}{N} + \frac{g^{\ell+1} \cdot g^{\ell+1}}{N}$$
$$ = \frac{\frac{1}{\sqrt{N}}W^{\ell}r^{\ell} \cdot \frac{1}{\sqrt{N}}W^{\ell}r^{\ell}}{N} + 2 \frac{r^{\ell} \cdot \frac{1}{\sqrt{N}}{W^{\ell}}^Tg^{\ell+1}}{N} + \frac{g^{\ell+1} \cdot g^{\ell+1}}{N}$$
$$ = \frac{\frac{1}{\sqrt{N}}W^{\ell}r^{\ell} \cdot \frac{1}{\sqrt{N}}W^{\ell}r^{\ell}}{N} + 2 \frac{r^{\ell} \cdot g^\ell}{N} + \frac{g^{\ell+1} \cdot g^{\ell+1}}{N}$$
\begin{align}   \label{eq:rl_inductive_intermediate}
     \Leq \frac{\frac{1}{\sqrt{N}}W^{\ell}r^{\ell} \cdot \frac{1}{\sqrt{N}}W^{\ell}r^{\ell}}{N} + 2\ell + 1
\end{align}
where we have applied our earlier results for $\frac{g^\ell \cdot g^\ell}{N}$ (Eq. \ref{eq:g_dot_g}) and $\frac{r^\ell \cdot g^\ell}{N}$ (Eq. \ref{eq:g_dot_r}).
Now, since $r^{\ell} = B_\ell g^{\ell}$, $r^\ell$ is measurable in the sigma algebra generated by $\mathcal{H}_\ell$ and $g^{\ell}$ and since $\frac{1}{\sqrt{N}}\norm{r^\ell}_2$ is moment controlled (Lemma \ref{lemma:everything_moment_bounded}), Lemma \ref{lemma:gaussian_conditioning} implies that
$$\frac{\frac{1}{\sqrt{N}}W^{\ell}r^{\ell} \cdot \frac{1}{\sqrt{N}}W^{\ell}r^{\ell}}{N} = \frac{1}{\frac{\norm{g^{\ell + 1}}_2^2}{N}} \cdot \left( \frac{g^\ell \cdot r^\ell}{N} \right)^2 + \frac{r^\ell \cdot r^\ell}{N} + R_\ell$$
where $R_\ell := \frac{\Pi_\ell W^\ell r^\ell \cdot \Pi_\ell W^\ell r^\ell}{N^2} - \frac{r^\ell \cdot r^\ell}{N} \Lconverge 0$. Recall from earlier that $\frac{g^\ell \cdot r^\ell}{N} \Lconverge \ell$ and $\frac{\norm{g^{\ell+1}}_2^2}{N} \Lconverge 1$; as a result, these quantities converge in probability to their respective limits as well. Similarly, our inductive hypothesis $\frac{r^\ell \cdot r^\ell}{N} \Lconverge \sum_{j=1}^\ell j^2$ and $R_\ell \Lconverge 0$ imply convergence in probability to their respective limits. As a result, Prop. \ref{prop:as_p_convergence} implies that 
$$\frac{\frac{1}{\sqrt{N}}W^{\ell}r^{\ell} \cdot \frac{1}{\sqrt{N}}W^{\ell}r^{\ell}}{N} \Pconverge \ell^2 + \sum_{j=1}^\ell j^2.$$
Returning to Eq. \ref{eq:rl_inductive_intermediate}, we thus find that $$\frac{r^{\ell + 1} \cdot r^{\ell + 1}}{N} \Pconverge \ell^2 + 2 \ell + 1 + \sum_{j = 1}^\ell j^2 = \sum_{j=1}^{\ell + 1} j^2.$$
Moreover, because $\frac{1}{\sqrt{N}}\norm{r^{\ell+1}}_2$ is moment controlled (Lemma \ref{lemma:everything_moment_bounded}), so is $\frac{r^{\ell + 1} \cdot r^{\ell + 1}}{N} = \left( \frac{1}{\sqrt{N}}\norm{r^{\ell+1}}_2 \right)^2$. In turn, Lemma \ref{lemma:cvg_in_prob_to_Lp} implies that, in fact, 
$$\frac{r^{\ell + 1} \cdot r^{\ell + 1}}{N} \Lconverge \sum_{j=1}^{\ell + 1} j^2.$$
This completes our induction, and establishes that $\frac{r^\ell \cdot r^\ell}{N} \Lconverge \sum_{j=1}^{\ell} j^2$ for each $1 \leq \ell \leq L$.

Finally, we inductively prove the following two results for any fixed vector $x \in \RR^D$ and each $1 \leq \ell \leq L$, as $N \to \infty$:
\begin{equation}    \label{eq:fl_to_Fl_cvg}
    f^\ell(x) \Lconverge F(x).
\end{equation}
\begin{equation}    \label{eq:Delta_h_cvg}
    \frac{\norm{\Delta h^\ell(x) - F(x) r^\ell}_2}{\sqrt{N}} \Lconverge 0.
\end{equation}
As a base case, when $\ell = 1$, Eqs. \ref{eq:Delta_h_def} and \ref{eq:fl_def} imply that
$$\Delta h^1(x) = f^1(x) g^1 = \frac{\eta}{n} \sum_{\mu = 1}^n \Delta_\mu \frac{x \cdot x_\mu}{D} g^1.$$
Since $r^1 = g^1$, we find that
$$ \frac{\norm{\Delta h^1(x) - F(x) r^1}_2}{\sqrt{N}}  =  \left| \frac{\eta}{n} \sum_{\mu = 1}^n (\Delta_\mu - y_\mu) \frac{x \cdot x_\mu}{D} \right| \frac{\norm{g^1}_2}{\sqrt{N}} \Lconverge 0$$
because $\Delta_\mu \Lconverge y_\mu$, all terms in the above expression are moment controlled (Lemma \ref{lemma:everything_moment_bounded}), and Remark \ref{rem:sums_products_L2_concentrate} implies that sums and products of $L^2$ convergent quantities also retain $L^2$ convergence. Similarly,
$$f^1(x) - F(x) = \frac{\eta}{n} \sum_{\mu = 1}^n (\Delta_\mu - y_\mu) \frac{x \cdot x_\mu}{D} \Lconverge 0.$$
Now, suppose that Eqs. \ref{eq:fl_to_Fl_cvg} and \ref{eq:Delta_h_cvg} hold for some $1 \leq \ell \leq L-1$. Then,
$$\left| f^{\ell+1}(x) - F(x) \right| = \left| \frac{\eta}{n} \sum_{\mu = 1}^n \left( \Delta_\mu \frac{h^{\ell}_{\mu;t=0} \cdot h^{\ell}_{t=1}(x)}{N} - y_\mu \frac{x \cdot x_\mu}{D}\right) \right|$$
$$ = \left| \frac{\eta}{n} \sum_{\mu = 1}^n \left( \Delta_\mu \frac{h^{\ell}_{\mu;t=0} \cdot h^{\ell}_{t=0}(x)}{N} - y_\mu \frac{x \cdot x_\mu}{D} \right) + \frac{\eta}{n} \sum_{\mu = 1}^n \Delta_\mu \frac{h^{\ell}_{\mu;t=0} \cdot \Delta h^{\ell}(x)}{N} \right|$$
\begin{multline*}
    \leq \frac{\eta}{n} \sum_{\mu = 1}^n \left| \left( \Delta_\mu  \frac{h^{\ell}_{\mu;t=0} \cdot h^{\ell}_{t=0}(x)}{N} - y_\mu \frac{x \cdot x_\mu}{D} \right) \right| + \frac{\eta}{n} |F(x)| \sum_{\mu = 1}^n \left| \Delta_\mu \frac{h^{\ell}_{\mu;t=0} \cdot r^\ell}{N} \right| \\ + \frac{\eta}{n} \sum_{\mu = 1}^n \left| \Delta_\mu \frac{h^{\ell}_{\mu;t=0} \cdot \left( \Delta h^\ell(x) - F(x) r^\ell \right)}{N} \right|
\end{multline*}
\begin{multline*}
    \leq \frac{\eta}{n} \sum_{\mu = 1}^n \left| \left( \Delta_\mu \frac{h^{\ell}_{\mu;t=0} \cdot h^{\ell}_{t=0}(x)}{N} - y_\mu \frac{x \cdot x_\mu}{D} \right) \right| + \frac{\eta}{n} |F(x)| \sum_{\mu = 1}^n \left| \Delta_\mu \frac{h^{\ell}_{\mu;t=0} \cdot r^\ell}{N} \right| \\+  \frac{\eta}{n} \sum_{\mu = 1}^n |\Delta_\mu| \frac{\norm{h^{\ell}_{\mu;t=0}}_2}{\sqrt{N}} \frac{\norm{\Delta h^\ell(x) - F(x) r^\ell}_2 }{\sqrt{N}} 
\end{multline*}
\begin{equation}    \label{eq:fl_to_Fl_inductive_step}
    \Lconverge 0
\end{equation}
where we have used the Cauchy-Schwarz inequality to bound $\left| \frac{h^{\ell}_{\mu;t=0} \cdot \left( \Delta h^\ell(x) - F(x) r^\ell \right)}{N} \right|$; we have also used the facts that $\Delta_\mu \Lconverge y_\mu$, $\frac{h^{\ell}_{\mu;t=0} \cdot h^{\ell}_{t=0}(x)}{N} \Lconverge \frac{x \cdot x_\mu}{D}$ for any $1 \leq \mu \leq n$, $\frac{h^{\ell}_{\mu;t=0} \cdot r^\ell}{N} \Lconverge 0$, and the inductive hypothesis that $\frac{\norm{\Delta h^\ell(x) - F(x) r^\ell}_2}{\sqrt{N}} \Lconverge 0$, all along with Remark \ref{rem:sums_products_L2_concentrate}. 
Next,
$$\frac{\norm{\Delta h^{\ell+1}(x) - F(x) r^{\ell+1}}_2}{\sqrt{N}} = \frac{\norm{\frac{1}{\sqrt{N}} W^{\ell} \Delta h^{\ell}(x) + f^{\ell+1}(x) g^{\ell+1} - F(x) \left( \frac{1}{\sqrt{N}} W^{\ell} r^\ell + g^{\ell + 1} \right)}_2}{\sqrt{N}}$$
$$ \leq \frac{\norm{\frac{1}{\sqrt{N}} W^\ell \left( \Delta h^\ell(x) - F(x) r^\ell \right)}_2}{\sqrt{N}} + \left| f^{\ell+1}(x) - F(x) \right| \frac{\norm{ g^{\ell+1}}_2}{\sqrt{N}}$$
$$ \leq \opnorm{\frac{1}{\sqrt{N}} W^\ell} \cdot \frac{\norm{\Delta h^\ell(x) - F(x) r^\ell}_2}{\sqrt{N}} + \left| f^{\ell+1}(x) - F(x) \right| \frac{\norm{ g^{\ell+1}}_2}{\sqrt{N}} \Lconverge 0$$
where we have used the inductive hypothesis that $\frac{\norm{\Delta h^\ell(x) - F(x) r^\ell}_2}{\sqrt{N}} \Lconverge 0$ and the fact that $f^{\ell+1}(x) - F(x) \Lconverge 0$ as established in Eq. \ref{eq:fl_to_Fl_inductive_step}, along with Remark \ref{rem:sums_products_L2_concentrate}.
This completes our induction and establishes Eqs. \ref{eq:fl_to_Fl_cvg} and \ref{eq:Delta_h_cvg} for each $1 \leq \ell \leq L$.

Finally, we piece together all of our results to find that for any fixed $x, x' \in \RR^D$,
$$\Phi^\ell_{t=1}(x, x') = \frac{h_{t=1}^\ell(x) \cdot h_{t=1}^\ell(x')}{N} = \frac{\left( h_{t=0}^\ell(x) + \Delta h^\ell(x) \right) \cdot \left( h_{t=0}^\ell(x') + \Delta h^\ell(x') \right)}{N}$$
$$ = \frac{h_{t=0}^\ell(x) \cdot h_{t=0}^\ell(x')}{N} + \frac{\Delta h^\ell(x) \cdot h_{t=0}^\ell(x')}{N} + \frac{h_{t=0}^\ell(x) \cdot \Delta h^\ell(x')}{N} + \frac{\Delta h^\ell(x) \cdot \Delta h^\ell(x')}{N}$$
\begin{equation}    \label{eq:expanding_one_step_kernel}
\begin{aligned}
     = \frac{h_{t=0}^\ell(x) \cdot h_{t=0}^\ell(x')}{N} + F(x) \frac{r^\ell \cdot h_{t=0}^\ell(x')}{N} + F(x') \frac{h_{t=0}^\ell(x) \cdot r^\ell}{N} + F(x) F(x') \frac{r^\ell \cdot r^\ell}{N} \\
     + \frac{\left( \Delta h^\ell(x) - F(x) r^\ell\right) \cdot h^\ell_{t=0}(x') }{N} + \frac{\left( \Delta h^\ell(x') - F(x')  r^\ell\right) \cdot h^\ell_{t=0}(x) }{N} \\
     + F(x') \frac{ \left( \Delta h^\ell(x) - F(x) r^\ell \right) \cdot r^\ell}{N} + F(x) \frac{ \left( \Delta h^\ell(x') - F(x') r^\ell \right) \cdot r^\ell}{N} \\+ \frac{ \left( \Delta h^\ell(x) - F(x) r^\ell \right) \cdot \left( \Delta h^\ell(x') - F(x') r^\ell \right)}{N}.
\end{aligned}
\end{equation}
Now, because $\frac{1}{\sqrt{N}}\norm{r^\ell}_2$ and $\frac{\norm{\Delta h^\ell(x_0) - F(x_0) r^\ell}_2}{\sqrt{N}}$ and $\frac{\norm{h_{t=0}^\ell(x_0)}_2}{\sqrt{N}}$, for $x_0 \in \{x, x'\}$, are all moment controlled (Lemma \ref{lemma:everything_moment_bounded}), and because $\frac{\norm{\Delta h^\ell(x_0) - F(x_0) r^\ell}_2}{\sqrt{N}} \Lconverge 0$ (Eq. \ref{eq:Delta_h_cvg}), it follows that all of the error terms in Eq. \ref{eq:expanding_one_step_kernel} involving $\Delta h^\ell(x) - F(x) r^\ell$ and $\Delta h^\ell(x') - F(x') r^\ell$ converge in $L^2$ to $0$. For example,
$$\left| \frac{\left( \Delta h^\ell(x) - F(x) r^\ell\right) \cdot h^\ell_{t=0}(x') }{N} \right| \leq   \frac{\norm{\Delta h^\ell(x) - F(x) r^\ell}_2 }{\sqrt{N}} \cdot \frac{\norm{h^\ell_{t=0}(x')}_2 }{\sqrt{N}} \Lconverge 0$$
where we have used the Cauchy-Schwarz inequality and Remark \ref{rem:sums_products_L2_concentrate}.
Therefore, using Eqs. \ref{eq:init_feature_kernel_concentration}, \ref{eq:h_dot_r}, and \ref{eq:r_dot_r}, as well as the definition of $F(x)$, we find that Eq. \ref{eq:expanding_one_step_kernel} simplifies to
\begin{equation}
    \begin{aligned}
        \Phi^\ell_{t=1}(x, x') \Leq \frac{h_{t=0}^\ell(x) \cdot h_{t=0}^\ell(x')}{N} + F(x) \frac{r^\ell \cdot h_{t=0}^\ell(x')}{N} + F(x') \frac{h_{t=0}^\ell(x) \cdot r^\ell}{N} + F(x) F(x') \frac{r^\ell \cdot r^\ell}{N} \\
        \Leq \frac{x \cdot x'}{D} + F(x) \cdot 0 + F(x') \cdot 0 + F(x) F(x') \sum_{j=1}^\ell j^2 \\
        = \frac{x \cdot x'}{D} + \left( \sum_{j=1}^\ell j^2 \right) \frac{\eta^2}{n^2} \sum_{\mu, \nu = 1}^n y_\mu y_\nu \frac{x \cdot x_\mu}{D} \frac{x_\nu \cdot x'}{D}.
    \end{aligned}
\end{equation}
Recognizing that $\sum_{j=1}^\ell j^2 = \frac{\ell(\ell+1)(2\ell+1)}{6} =: d_\ell$, it follows that 
\begin{equation}
    \Phi^\ell_{t=1}(x, x') \Lconverge \frac{x \cdot x'}{D} + \frac{\eta^2 d_\ell}{n^2} \sum_{\mu, \nu = 1}^n \frac{x \cdot x_\mu}{D} y_\mu y_\nu \frac{x_\nu \cdot x'}{D} = \frac{x \cdot x'}{D} + \frac{\eta^2 d_\ell}{n^2} \frac{x^T X}{D} yy^T \frac{X^T x'}{D}.
\end{equation}
(Recall that $X = \left[ x_1, \dots, x_n \right] \in \RR^{D \times n}$ is the matrix of stacked inputs over the dataset.)
When viewed across all pairs of training examples, we further obtain that
\begin{equation}
    \Phi^\ell_{t=1} \Lconverge K_x + \frac{\eta^2 d_\ell}{n^2} K_x yy^T K_x
\end{equation}
where $K_x \in \RR^{n \times n}$ is the input kernel over the dataset and $y = [y_1, \dots, y_n] \in \RR^n$.
In fact, because $$\left|\Phi^\ell_{t=1}(x, x') \right| = \left|\frac{h^\ell_{t=1}(x) \cdot h^\ell_{t=1}(x')}{N} \right| \leq \frac{\norm{h^\ell_{t=1}(x)}_2}{\sqrt{N}} \cdot \frac{\norm{h^\ell_{t=1}(x')}_2}{\sqrt{N}}$$ 
is moment controlled for any fixed $x, x' \in \RR^D$ (Lemma \ref{lemma:everything_moment_bounded} applied to $x$ and $x'$), Lemma \ref{lemma:multiplying_adding_L2_limits} implies that $L^2$ convergence may be upgraded to $L^p$ convergence for any $p \geq 1$ as well. Consequently, for any $p \geq 1$ and any fixed $x, x' \in \RR^D$,
\begin{equation}
    \Phi^\ell_{t=1}(x, x') \Lpconverge \frac{x \cdot x'}{D} + \frac{\eta^2 d_\ell}{n^2} \frac{x^T X}{D} yy^T \frac{X^T x'}{D}.
\end{equation}
In turn, over the training dataset and any $p \geq 1$, 
\begin{equation}
    \Phi^\ell_{t=1} \Lpconverge K_x + \frac{\eta^2 d_\ell}{n^2} K_x yy^T K_x.
\end{equation}

\subsection{One gradient step yields inflated estimation --- Proof of Prop. \ref{prop:inflated_estimator}}   \label{app:inflated_estimation_proof}

In this section, we provide a proof of Prop. \ref{prop:inflated_estimator}, as stated in the main text.

\begin{proof} [Proof of Prop. \ref{prop:inflated_estimator}]
    Recall that $\finit(x) = \frac{x^T X A^{-1} y}{D}$. Define $\omega_\ell := \frac{\eta^2 d_\ell}{n^2}$, where $d_\ell = \frac{\ell (\ell + 1)(2 \ell + 1)}{6}$.
Now, through kernel ridge regression, the optimal estimator for the kernel $\Kf = K_x + \omega_\ell K_x yy^T K_x$ is given by 
$$\hat{f}_{\feat}(x) = K_{\feat}^\ell(x, X) \left( K_{\feat}^\ell(X, X) + \lambda I_n \right)^{-1} y $$ 
$$= \left( \frac{x^T X}{D} + \omega_\ell \frac{x^T X}{D} yy^T K_x \right) \left( K_x + \omega_\ell K_x yy^T K_x + \lambda I_n \right)^{-1} y$$ 
$$ = \left( \frac{x^T X}{D} + \omega_\ell \frac{x^T X}{D} yy^T K_x \right) \left( A + \omega_\ell K_x yy^T K_x \right)^{-1} y.$$ 
Now, note that $K_x = A - \lambda I_n$ so that $K_x A^{-1} = I_n - \lambda A^{-1}$. In addition to using this identity, we use the Sherman-Morrison formula to simplify $\left( A + \omega_\ell K_x yy^T K_x \right)^{-1}$:
$$ \ffeat(x) = \left( \frac{x^T X}{D} + \omega_\ell \frac{x^T X}{D} yy^T K_x \right) \left( A^{-1} - \frac{\omega_\ell A^{-1} K_x yy^T K_x A^{-1}}{1 + \omega_\ell y^T K_x A^{-1} K_x y} \right) y$$ 

\begin{align*}
    = \frac{x^T X A^{-1} y}{D} - \frac{\omega_\ell}{1 + \omega_\ell y^T K_x A^{-1} K_x y} \frac{x^T X A^{-1} K_x y}{D} y^T K_x A^{-1} y \\
    + \omega_\ell \frac{x^T Xy}{D} y^T K_x A^{-1} y - \omega_\ell \frac{x^T X y}{D} \frac{\omega_\ell y^T K_x A^{-1} K_x y}{1 + \omega_\ell y^T K_x A^{-1} K_x y} y^T K_x A^{-1} y
\end{align*}

\begin{align*}
    = \frac{x^T X A^{-1} y}{D} - \frac{\omega_\ell}{1 + \omega_\ell y^T K_x A^{-1} K_x y} \frac{x^T X (I_n - \lambda A^{-1}) y}{D} y^T K_x A^{-1} y \\
    + \omega_\ell \frac{x^T Xy}{D} y^T K_x A^{-1} y - \omega_\ell \frac{x^T X y}{D} \frac{\omega_\ell y^T K_x A^{-1} K_x y}{1 + \omega_\ell  y^T K_x A^{-1} K_x y} y^T K_x A^{-1} y 
\end{align*}

\begin{align*}
    = \frac{x^T X A^{-1} y}{D} + \frac{\lambda \omega_\ell y^T K_x A^{-1} y}{1 + \omega_\ell y^T K_x A^{-1} K_x y} \frac{x^T X A^{-1} y}{D} - \omega_\ell \frac{x^T X y}{D} \frac{1}{1 + \omega_\ell y^T K_x A^{-1} K_x y} y^T K_x A^{-1} y\\
     + \omega_\ell \frac{x^T Xy}{D} y^T K_x A^{-1} y - \omega_\ell \frac{x^T X y}{D} \frac{\omega_\ell y^T K_x A^{-1} K_x y}{1 + \omega_\ell  y^T K_x A^{-1} K_x y} y^T K_x A^{-1} y 
\end{align*}

\begin{align*}
    = \left( 1 + \frac{\lambda \omega_\ell y^T K_x A^{-1} y}{1 + \omega_\ell y^T K_x A^{-1} K_x y} \right)\frac{x^T X A^{-1} y}{D} + \omega_\ell \frac{x^T Xy}{D} y^T K_x A^{-1} y  \\
    - \omega_\ell \frac{x^T X y}{D} \frac{1 + \omega_\ell y^T K_x A^{-1} K_x y}{1 + \omega_\ell  y^T K_x A^{-1} K_x y} y^T K_x A^{-1} y 
\end{align*}

$$ = \left( 1 + \frac{\lambda \omega_\ell y^T K_x A^{-1} y}{1 + \omega_\ell y^T K_x A^{-1} K_x y} \right)\frac{x^T X A^{-1} y}{D} = \cl \finit(x)$$
where $\cl := \left( 1 + \frac{\lambda \frac{\eta^2 d_\ell}{n^2} y^T K_x A^{-1} y}{1 + \frac{\eta^2 d_\ell}{n^2} y^T K_x A^{-1} K_x y} \right)$. 
This concludes our proof.
\end{proof}

\subsection{Analyzing the geometry of estimation under one step of feature learning --- Proof of Proposition \ref{prop:feat_learning_estimation_geometry}}   \label{app:feat_learning_geometry}

In this section, we prove Prop. \ref{prop:feat_learning_estimation_geometry}, as stated in the main text.

\begin{proof}[Proof of Prop. \ref{prop:feat_learning_estimation_geometry}]
Recall that
$$\ffeat(x) = \cl \finit(x) = \cl \cdot y^T (K_x + \lambda I_n)^{-1} \frac{X^T x}{D}.$$
To first order in $\lambda$, we have that
$$\cl = 1 + \frac{\lambda \frac{\eta^2 d_\ell}{n^2} y^T K_x A^{-1} y}{1 + \frac{\eta^2 d_\ell}{n^2} y^T K_x A^{-1} K_x y} = 1 + \frac{\lambda \frac{\eta^2 d_\ell}{n^2} y^T K_x K_x^{-1} y}{1 + \frac{\eta^2 d_\ell}{n^2} y^T K_x K_x^{-1} K_x y} + o(\lambda)$$
$$ = 1 + \frac{\lambda \frac{\eta^2 d_\ell}{n^2} y^T y}{1 + \frac{\eta^2 d_\ell}{n^2} y^T K_x y} + o(\lambda).$$
Furthermore, $(K_x + \lambda I_n)^{-1} = K_x^{-1} - \lambda K_x^{-2} + o(\lambda)$ for $$\lambda \ll \eigmin(K_x) = \frac{n}{D}\eigmin\left(\frac1n X^T X \right) = \frac{n}{D} s_n.$$ Consequently,
$$\ffeat(x) = \left( 1 + \frac{\lambda \frac{\eta^2 d_\ell}{n^2} y^T y}{1 + \frac{\eta^2 d_\ell}{n^2} y^T K_x y} \right) y^T \left( K_x^{-1} - \lambda K_x^{-2} \right) \frac{X^T x}{D} + o(\lambda)$$
$$ = y^T \left( \left( 1 + \frac{\lambda \frac{\eta^2 d_\ell}{n^2} y^T y}{1 + \frac{\eta^2 d_\ell}{n^2} y^T K_x y} \right) K_x^{-1} - \lambda K_x^{-2} \right) \frac{X^T x}{D} + o(\lambda)$$
\begin{equation}    \label{eq:ffeat_first_order_formula}
     = y^T K_x^{-1} \left( \left( 1 + \frac{\lambda \frac{\eta^2 d_\ell}{n^2} y^T y}{1 + \frac{\eta^2 d_\ell}{n^2} y^T K_x y} \right) I_n - \lambda K_x^{-1} \right) \frac{X^T x}{D} + o(\lambda).
\end{equation}
Now, since $y = U\alpha$, $y^T y = \norm{\alpha}_2^2$. Moreover, observe that $$K_x = \frac1D X^T X = \frac{n}{D} \frac1n X^T X = \frac{n}{D} U S U^T$$
and so $K_x^{-1} = \frac{D}{n} U S^{-1} U^T$.
As a result, $$y^T K_x y = \frac{n}{D}\alpha^T S \alpha = \frac{n}{D}\sum_{j=1}^n s_j \alpha_j^2.$$
Additionally, since $X^T x_\perp = 0$,
$$\frac{X^T x}{D} = \frac1D X^T V \beta = \frac{\sqrt{n}}{D} U S^{\frac12} \beta.$$ Thus, in SVD form, Eq. \ref{eq:ffeat_first_order_formula} may be written as
$$\ffeat(x) = \frac{D}{n}\alpha^T S^{-1} \left( \left( 1 + \frac{\lambda \frac{\eta^2 d_\ell}{n^2} \norm{\alpha}_2^2}{1 + \frac{\eta^2 d_\ell}{n^2} \frac{n}{D} \alpha^T S \alpha} \right) I_n - \lambda \frac{D}{n} S^{-1} \right) \frac{\sqrt{n}}{D} S^{\frac12} \beta + o(\lambda)$$
$$ = \frac{1}{\sqrt{n}} \alpha^T \left( \left( 1 + \frac{\lambda }{ \frac{1}{\frac{\eta^2 d_\ell}{n^2} \norm{\alpha}_2^2} + \frac{n}{D} \hat{\alpha}^T S \hat{\alpha}} \right) I_n - \lambda \frac{D}{n} S^{-1} \right) S^{-\frac12} \beta + o(\lambda)$$
$$ = \frac{1}{\sqrt{n}} \alpha^T \left( \left( 1 + \frac{D}{n} \cdot \frac{\lambda }{ \frac{1}{\frac{\eta^2 d_\ell}{D} \frac{\norm{\alpha}_2^2}{n}} + \hat{\alpha}^T S \hat{\alpha}} \right) I_n - \lambda \frac{D}{n} S^{-1} \right) S^{-\frac12} \beta + o(\lambda) $$
$$ = \left[ \frac{1}{\sqrt{n}} \sum_{i=1}^n \left( 1 + \frac{D}{n} \left( \frac{1}{ \frac{1}{\frac{\eta^2 d_\ell}{D} \frac{\norm{\alpha}_2^2}{n}} + \sum_{j=1}^n s_j \hat{\alpha}_j^2} - \frac{1}{s_i} \right) \lambda \right) \frac{\alpha_i \beta_i}{\sqrt{s_i}} \right] + o(\lambda).$$

\end{proof}

\section{Linking Learned Features to Generalization}

\subsection{Setup for spiked Gaussian model} \label{supp:setup}

To link the features learned under one step of gradient descent to generalization, we specialize to a setting in which the data is sampled from a spiked Gaussian model. In particular, fix $\psi \in (0, 1)$ and define the strictly increasing sequences of positive integers $(D_k)_{k \geq 1}$ and $(n_k)_{k \geq 1}$ so that $D_k > n_k$ for each $k \geq 1$ and $\lim_{k \to \infty} \frac{n_k}{D_k} = \psi$. For any pair of $D$ and $n$ (corresponding to some index $k \geq 1$), we consider the high-dimensional regression setting from Assumption \ref{assump:data_gen_process}, in which inputs $(x_i)_{i=1}^n$ are chosen so that $x_i \iid \mathcal{N}(0, \Sigma)$, where $\Sigma = I_D + \gamma vv^T$ for some fixed spike strength $\gamma \geq 0$ and fixed unit vector $v \in \SS^{D - 1}$. Moreover, for each $x_i$, the corresponding target label is given as $y_i = \frac{1}{\sqrt{D}} \ws^T x_i + \epsilon_i = \what^T x_i + \epsilon_i$, where $\ws \in \sqrt{D} \SS^{D - 1}$ is a fixed vector and $\what := \frac{1}{\sqrt{D}} \ws$, with $\what^T v =: \rho \in [-1, 1]$ as the spike-target alignment and $\epsilon_i \iid \mathcal{N}(0, \sigma^2)$ for a given noise level $\sigma^2 \geq 0$. Altogether, we may write $y = X^T \what + \epsilon$, where $y \in \RR^n$ is the vector of all targets, $X \in \RR^{D \times n}$ is a matrix whose columns are $x_1, \dots, x_n$, and $\epsilon \sim \mathcal{N}(0, \sigma^2 I_n)$. Moreover, we may write
\begin{equation}    \label{eq:spiked_Wishart_matrix}
    X = \Sigma^{\frac12} Z = (I_D + \zeta vv^T) Z
\end{equation}
where $\zeta := \sqrt{\gamma + 1} - 1$ and $Z \in \RR^{D \times n}$ is a matrix of iid standard Gaussian entries.

Recall that the input kernel $K_x \in \RR^{n \times n}$ is given as 
\begin{equation}    \label{eq:K_x}
    K_x = \frac{1}{D} X^T X = \frac1D Z^T \Sigma Z = \frac1D Z^T Z + \frac{\gamma}{D} Z^T v v^T Z
\end{equation} 
which constitutes a rank-1 perturbation of the Wishart matrix $K_0 := \frac{1}{D} Z^T Z \in \RR^{n \times n}$. Moreover, for any fixed ridge regularization strength $\lambda \geq 0$, recall that 
\begin{equation}    \label{eq:A_resolvent}
    A := K_x + \lambda I_n = A_0 + \frac{\gamma}{D} Z^T v v^T Z
\end{equation}
where $A_0 := K_0 + \lambda I_n$. Since $D > n$, $A$ and $A_0$ are invertible for all $\lambda \geq 0$. Using the Sherman-Morrison formula, we find that 
\begin{equation}    \label{eq:A_inv_formula}
    A^{-1} = A_0^{-1} - \frac{\frac{\gamma}{D} A_0^{-1} Z^T v v^T Z A_0^{-1}}{1 + \frac{\gamma}{D} v^T Z A_0^{-1} Z^T v}.
\end{equation}
Next, define $M_1 := \frac{1}{D} X A^{-1} X^T$ and $M_2 := \frac1D X A^{-2} X^T$. Note that $\opnorm{M_1} \leq 1$ for any $\lambda \geq 0$. Additionally, for any matrix $K \in \RR^{n \times n}$ (such as $K_x$ or $K_0$), we write
\begin{equation}
    g_K(z) := \frac{1}{n} \Tr \left( (K - zI_n)^{-1} \right)
\end{equation}
as the Stieltjes transform of $K$ (per Defn. \ref{defn:finite_stieltjes_transform}).

Our subsequent analyses involve the limiting Marcenko-Pastur (MP) law for the Wishart matrix $K_0$. In the limit as $n, D \to \infty$ with $n < D$ and $\frac{n}{D} \to \psi \in (0, 1)$, the empirical spectral distribution (ESD) of $K_0$ weakly almost surely converges to the MP law corresponding to the aspect ratio $\psi$, which we denote by $\MP^{\psi}$. For ease of notation, we define the function $g$ as the limiting Stieltjes transform under the MP law:
\begin{equation}
    g(z) = \int \frac{1}{x - z} \MP^{\psi}(dx).
\end{equation}
It is well known that $g(z)$ is analytic outside of the support of $\MP^{\psi}$ and that it may be written in closed-form (see Prop. \ref{prop:properties_of_MP_and_Stieltjes}):
\begin{equation}    \label{eqn:stieltjes_closed_form_defn_SUPP}
        {g}(z) = \frac{1 - \psi - z + \sqrt{z - \lambda_-}\sqrt{z - \lambda_+}}{2 \psi z}
\end{equation}
where $\sqrt{\cdot}$ is defined using its principal branch in the complex plane. Here, $\lambda_- = (1 - \sqrt{\psi})^2$ and $\lambda_+ = (1 + \sqrt{\psi})^2$ are the left and right edges, respectively, of the bulk of the Marcenko-Pastur distribution.

Moreover, for $z \leq 0$, the function $z \mapsto \frac{1}{x-z}$ is differentiable and, moreover, for $z \leq 0$ and $x \in \left[(1 - \sqrt{\psi})^2, (1 + \sqrt{\psi})^2 \right]$ (the support of the Marcenko-Pastur law), 
$$\left| \frac{\partial}{\partial z} \frac{1}{x-z} \right| = \left| \frac{1}{(x-z)^2} \right| \leq \frac{1}{\left(1 - \sqrt{\psi} \right)^4}$$ 
is uniformly bounded above. Altogether, differentiating under the integral (Prop. \ref{prop:properties_of_MP_and_Stieltjes}) yields
\begin{equation}
    g'(z) = \int \frac{1}{(x-z)^2} \MP^{\psi}(dx).
\end{equation}

In particular, we obtain the following useful relations for any fixed $\lambda \geq 0$: 
\begin{equation}    \label{eq:MP_Stieltjes}
    g(-\lambda) = \int \frac{1}{x+\lambda} \MP^{\psi}(dx)
\end{equation}
\begin{equation}    \label{eq:MP_Stieltjes_prime}
    g'(-\lambda) = \int \frac{1}{(x+\lambda)^2} \MP^{\psi}(dx).
\end{equation}
We remark that $g(0) = \frac{1}{1 - \psi}$ and $g'(0) = \frac{1}{(1 - \psi)^3}$ (Prop. \ref{prop:properties_of_MP_and_Stieltjes}). Moreover, by induction, we may repeat the same argument as detailed above for computing higher order derivatives of $g$. In general, we find that
\begin{equation}    \label{eq:MP_Stieltjes_higher_order_derivs}
    g^{(j)}(-\lambda) = (j!)\int \frac{1}{(x + \lambda)^{j+1}} \MP^{\psi}(dx)
\end{equation}
for each $j \in \ZZ_+$. Finally, for each $\lambda \geq 0$, we define the quantities $\tau_1 := \psi(1 - \lambda g(-\lambda))$ and $\tau_2 := \psi(g(-\lambda) - \lambda g'(-\lambda))$; these quantities appear in subsequent calculations.

\subsection{Analysis of relevant quantities}

We now apply Prop. \ref{prop:Lp_control} and Thm. \ref{thm:spectral_statistics_MP} within the context of our present setup:

\begin{cor} \label{cor:list_of_MP_quantities_lambda_zero}
     ($\lambda = 0$) As $n$ and $D$ are jointly taken to infinity with $\frac{n}{D} \to \psi \in (0, 1)$ and $n < D$, the following quantities converge almost surely to nonrandom constants:
    \begin{enumerate}
        \item $\frac{\epsilon^T K_0^{-j} \epsilon}{n}$ for any $j \geq 1$, which converges almost surely to $ \sigma^2 \int x^{-j} \MP^{\psi}(dx)$. Notably, when $j = 1$, this limit is $\sigma^2 g(0)$; when $j = 2$, this limit is $\sigma^2 g'(0)$.

        \item $\frac{\epsilon^T K_0^{-j} Z^T u}{D}$, for $u \in \{\what, v\}$ and $j \geq 1$, which converges almost surely to $0$.

        \item $\frac{u_1^T Z K_0^{-j} Z^T u_2}{D}$ for $u_1, u_2 \in \{\what, v\}$ and $j \geq 1$, which converges almost surely to $(u_1^T u_2) \psi \int x^{-(j-1)} \MP^{\psi}(dx)$. Notably, when $j = 1$, this limit is $(u_1^T u_2) \psi$; when $j = 2$, this limit is $(u_1^T u_2) \psi g(0) = \frac{(u_1^T u_2) \psi}{1 - \psi}$.
    \end{enumerate}
\end{cor}

\begin{proof}
Observe that $f(x) = x^{-j}$ is a Class 2 function for $j \geq 1$, and that $D > n$, meaning that Theorem \ref{thm:spectral_statistics_MP} applies.

\paragraph{1.} Since $\epsilon \sim \mathcal{N}(0, \sigma^2 I_n)$, we may write $\frac{\epsilon}{\sqrt{n}} = \sigma \alpha$ where $\alpha \sim \mathcal{N}\left(0, \frac{1}{n} I_n \right)$. Then,
$$\frac{\epsilon^T K_0^{-j} \epsilon}{n} = \sigma^2 \alpha^T K_0^{-j} \alpha \asconverge \sigma^2 \int x^{-j} \MP^{\psi}(dx)$$
where we have used Theorem \ref{thm:spectral_statistics_MP}(b).

\paragraph{2.} For any $u \in \{\what, v\}$, the result that
$$\frac{\epsilon^T K_0^{-j} Z^T u}{D} \asconverge 0$$
follows from Theorem \ref{thm:spectral_statistics_MP}(c).

\paragraph{3.} For any $u_1, u_2 \in \{\what, v\}$, the result that
$$\frac{u_1^T Z K_0^{-j} Z^T u_2}{D} \asconverge (u_1^T u_2) \psi \int x^{-(j-1)} \MP^{\psi}(dx)$$
follows from Theorem \ref{thm:spectral_statistics_MP}(d).

\end{proof}

\begin{cor} \label{cor:list_of_MP_quantities_lambda_positive}
    As $n$ and $D$ are jointly taken to infinity with $n < D$ and $\frac{n}{D} \to \psi \in (0, 1)$, the following quantities are both moment controlled and converge almost surely to nonrandom constants:
    \begin{enumerate}
        \item $u_1^T \left(\frac1n ZZ^T \right)^j u_2$, for $u_1, u_2 \in \{\what, v\}$ and $j \geq 1$, which converges almost surely to $$(u_1^T u_2) \int x^j \MP^{1/\psi}(dx).$$
        (Recall that $\what^T \what = v^T v = 1$ and $\what^T v = \rho$.) In particular, when $j = 1$, this limit is simply $u_1^T u_2$; when $j = 2$, this limit is $u_1^T u_2 \left(1 + \frac{1}{\psi} \right)$.

        \item $\frac{\epsilon^T f(K_0) \epsilon}{n}$, for any polynomially bounded function $f$, which converges almost surely to $\sigma^2 \int f(x) \MP^{\psi}(dx)$. Examples include $\frac{\epsilon^T \epsilon}{n}$ ($f(x) \equiv 1$) which converges a.s. to $\sigma^2$; $\frac{\epsilon^T K_0 \epsilon}{n}$ ($f(x) \equiv x$), which converges a.s. to $\sigma^2$; and $\frac{\epsilon^T A_0^{-j} \epsilon}{n}$ for any fixed $j \geq 1$ and $\lambda > 0$ ($f(x) \equiv \frac{1}{(x + \lambda)^j}$), which converges a.s. to $\sigma^2 \int \frac{1}{(x + \lambda)^j} \MP^{\psi}(dx)$.

        \item $\frac{\epsilon^T f(K_0) Z^T u}{D}$, for $u \in \{\what, v\}$ and $f$ polynomially bounded, which converges almost surely to $0$. Examples include $\frac{\epsilon^T Z^T u}{D}$ ($f(x) \equiv 1$), $\frac{\epsilon^T K_0 Z^T u}{D}$ ($f(x) \equiv x$), and $\frac{\epsilon^T A_0^{-j} Z^T u}{D}$ for any fixed $j$ and $\lambda > 0$ ($f(x) \equiv \frac{1}{(x+\lambda)^j}$).

        \item $\frac{u_1^T Z f(K_0) Z^T u_2}{D}$ for $u_1, u_2 \in \{\what, v\}$ and $f$ polynomially bounded, which converges almost surely to $(u_1^T u_2) \psi \int x f(x) \MP^{\psi}(dx)$. Fixing $\lambda > 0$, key examples of the above quantity include $\frac{u_1^T Z A_0^{-j} Z^T u_2}{D}$ for $j \geq 1$ ($f(x) \equiv \frac{1}{(x+\lambda)^j}$), which converges a.s. to $$(u_1^T u_2) \psi \int \frac{x}{(x+\lambda)^j} \MP^{\psi}(dx) = (u_1^T u_2) \psi \left( \int \frac{1}{(x+\lambda)^{j-1}} \MP^{\psi}(dx) - \lambda \int \frac{1}{(x+\lambda)^j} \MP^{\psi}(dx) \right).$$

        \item $\Tr\left(\frac1n f(K_0) \right)$, with $f$ polynomially bounded, which converges almost surely to $\int f(x) \MP^{\psi}(dx)$. Fixing any $\lambda > 0$, key examples of this quantity include $\Tr\left(\frac1n A_0^{-1} \right)$, which converges a.s. to $g(-\lambda)$, and $\Tr\left(\frac1n A_0^{-2} \right)$, which converges a.s. to $g'(-\lambda)$.
    \end{enumerate}
\end{cor}

\begin{proof}
    Observe that, for any fixed $\lambda > 0$ and any $j \geq 1$, the functions $f_1(x) = 1$, $f_2(x) = x^j$, and $f_3(x) = \frac{1}{(x+\lambda)^j}$ are all polynomially bounded (with $f_1$ and $f_3$ in fact being bounded above by constants). As a result, they are Class 1 functions and so Theorem \ref{thm:spectral_statistics_MP} applies. Additionally, the quantities listed in parts 1-4 are all moment controlled by Prop. \ref{prop:Lp_control}.

    \paragraph{1.} This result follows from Theorem \ref{thm:spectral_statistics_MP}(a), using the Class 1 function $f_2(x) = x^j$.

    \paragraph{2.} Since $\epsilon \sim \mathcal{N}(0, \sigma^2 I_n)$, we may write $\frac{\epsilon}{\sqrt{n}} = \sigma \alpha$ where $\alpha \sim \mathcal{N}\left(0, \frac{1}{n} I_n \right)$. Then, 
    $$\frac{\epsilon^T f(K_0) \epsilon}{n} = \sigma^2 \alpha^T f(K_0) \alpha \asconverge \sigma^2 \int f(x) \MP^{\psi}(dx)$$
    where we have applied Theorem \ref{thm:spectral_statistics_MP}(b).

    \paragraph{3.} This result follows from Theorem \ref{thm:spectral_statistics_MP}(c).

    \paragraph{4.} This result follows from Theorem \ref{thm:spectral_statistics_MP}(d).

    \paragraph{5.} First, the fact that $$\Tr\left(\frac1n f(K_0) \right) \asconverge \int f(x) \MP^{\psi}(dx)$$
    follows from Cor. \ref{cor:unbounded_f_MP_convergence}.
    Additionally, for $f$ polynomially bounded, we may write the inequality $f(x) \leq C_f (x^s + 1)$ for some $C_f > 0$ and $s \geq 1$, and for all $x \geq 0$. To see why $\Tr\left(\frac1n f(K_0) \right)$ is moment controlled, observe that
    $$\Tr\left(\frac1n f(K_0) \right) \leq \max_{1 \leq i \leq n} f(\lambda_i(K_0)) \leq \max_{1 \leq i \leq n} C_f (\lambda_i(K_0)^s + 1) $$
    $$ = C_f (\eigmax(K_0)^s + 1).$$
    As a result, for any $p \geq 1$,
    $$\EE\left[ \left( \Tr\left(\frac1n f(K_0) \right) \right)^p  \right] \leq C_f^p \EE\left[ (\eigmax(K_0)^s + 1)^p \right]$$
    $$ \leq C_f^p \EE\left[\left(2 \cdot \max\left( \eigmax(K_0)^s, 1 \right)\right)^p \right] = (2C_f)^p \EE\left[ \max\left( \eigmax(K_0)^{ps}, 1 \right) \right]$$ $$ \leq (2C_f)^p \left( \EE\left[ \eigmax(K_0)^{ps} \right] + 1 \right) \leq (2C_f)^p \left( C_{ps, \tilde{q}} + 1 \right).$$
    For the latter inequality, we have invoked Lemma \ref{lemma:max_MP_eigval_bound} (see this lemma for a definition of $\tilde{q}$). This establishes moment control, as claimed. Consequently, using the polynomially bounded functions $x \mapsto (x+\lambda)^{-1}$ and $x \mapsto (x+\lambda)^{-2}$ for any fixed $\lambda > 0$, we find that $\Tr\left(\frac1n A_0^{-1} \right)$ and $\Tr\left(\frac1n A_0^{-2} \right)$ have almost sure limits --- namely $g(-\lambda) = \int \frac{1}{x+\lambda} \MP^{\psi}(dx)$ and $g'(-\lambda) = \int \frac{1}{(x+\lambda)^2} \MP^{\psi}(dx)$, respectively --- and that these quantities are moment controlled.
    
\end{proof}

\begin{prop} \label{prop:spiked_Ainv_terms}
For any $\lambda \geq 0$, recall that 
$$A^{-1} = A_0^{-1} - \frac{\frac{\gamma}{D} A_0^{-1} Z^T v v^T Z A_0^{-1}}{1 + \frac{\gamma}{D} v^T Z A_0^{-1} Z^T v} $$
Then, for any $z_1, z_2 \in \left\{\frac{1}{\sqrt{D}}\epsilon, \frac{1}{\sqrt{D}}Z^T \what, \frac{1}{\sqrt{D}}Z^T v \right\}$, $ z_1^T A^{-1} z_2$ and $z_1^T A^{-2} z_2$ each converge almost surely to deterministic constants in the limit of $n, D \to \infty$ with $n < D$ and $\frac{n}{D} \to \psi \in (0, 1)$. Moreover, if $\lambda > 0$, each of $z_1^T A^{-1} z_2$ and $z_1^T A^{-2} z_2$ is moment controlled.
\end{prop}

\begin{proof}
    First, for any $z_1, z_2 \in \left\{\frac{1}{\sqrt{D}}\epsilon, \frac{1}{\sqrt{D}}Z^T \what, \frac{1}{\sqrt{D}}Z^T v \right\}$, 
    \begin{equation}    \label{eq:A_inv_contraction_formula}
        z_1^T A^{-1} z_2 = z_1^T A_0^{-1} z_2 - \frac{\gamma \left( \frac{1}{\sqrt{D}}z_1^T A_0^{-1} Z^T v \right) \left( \frac{1}{\sqrt{D}} v^T Z A_0^{-1} z_2 \right)}{1 + \gamma \frac{ v^T Z A_0^{-1} Z^T v}{D}}.
    \end{equation}
    In particular, $z_1^T A_0^{-1} z_2$, $\frac{ v^T Z A_0^{-1} Z^T v}{D}$, $\frac{1}{\sqrt{D}}z_1^T A_0^{-1} Z^T v$, and $\frac{1}{\sqrt{D}}z_2^T A_0^{-1} Z^T v$ each converge almost surely to deterministic constants, by Cor. \ref{cor:list_of_MP_quantities_lambda_zero} (if $\lambda = 0$) or by Cor. \ref{cor:list_of_MP_quantities_lambda_positive} (if $\lambda > 0$). Since the denominator term $1 + \gamma \frac{ v^T Z A_0^{-1} Z^T v}{D} \geq 1$ is uniformly bounded below, it follows that $z_1^T A^{-1} z_2$ converges almost surely to a deterministic constant. Moreover, Cor. \ref{cor:list_of_MP_quantities_lambda_positive} further implies that, for any fixed $\lambda > 0$ and any $z_1, z_2 \in \left\{\frac{1}{\sqrt{D}}\epsilon, \frac{1}{\sqrt{D}}Z^T \what, \frac{1}{\sqrt{D}}Z^T v \right\}$, each of $z_1^T A_0^{-1} z_2$, $\frac{1}{\sqrt{D}}z_1^T A_0^{-1} Z^T v$, and $\frac{1}{\sqrt{D}} v^T Z A_0^{-1} z_2$ is moment controlled. Furthermore, $1 + \gamma \frac{ v^T Z A_0^{-1} Z^T v}{D}$
    is quotient controlled. Thus, Prop. \ref{prop:uniform_Lp_control} implies that $z_1^T A^{-1} z_2$ is moment controlled for any fixed $\lambda > 0$.

    Next, we find that
    $$A^{-2} = \left( A_0^{-1} - \frac{\frac{\gamma}{D} A_0^{-1} Z^T v v^T Z A_0^{-1}}{1 + \frac{\gamma}{D} v^T Z A_0^{-1} Z^T v} \right)^2$$
    \begin{align*} 
        = A_0^{-2} - \frac{\frac{\gamma}{D} A_0^{-2} Z^T v v^T Z A_0^{-1}}{1 + \frac{\gamma}{D} v^T Z A_0^{-1} Z^T v} - \frac{\frac{\gamma}{D} A_0^{-1} Z^T v v^T Z A_0^{-2}}{1 + \frac{\gamma}{D} v^T Z A_0^{-1} Z^T v} + \frac{\frac{\gamma^2}{D} \left( A_0^{-1} Z^T v \right) \left( \frac1D v^T Z A_0^{-2} Z^T v \right) \left( v^T Z A_0^{-1} \right)}{\left( 1 + \frac{\gamma}{D} v^T Z A_0^{-1} Z^T v \right)^2}
    \end{align*}
    As a result,
    \begin{multline}    \label{eq:A_inv_squared_contractions}
        z_1^T A^{-2} z_2 = z_1^T A_0^{-2} z_2 - \frac{\gamma \left( \frac{1}{\sqrt{D}}z_1^T A_0^{-2} Z^T v \right) \left( \frac{1}{\sqrt{D}} v^T Z A_0^{-1} z_2 \right) }{1 + \frac{\gamma}{D} v^T Z A_0^{-1} Z^T v} - \frac{\gamma \left( \frac{1}{\sqrt{D}} z_1^T A_0^{-1} Z^T v \right) \left( \frac{1}{\sqrt{D}} v^T Z A_0^{-2} z_2 \right)}{1 + \frac{\gamma}{D} v^T Z A_0^{-1} Z^T v} \\
        + \frac{\gamma^2 \left( \frac{1}{\sqrt{D}} z_1^T A_0^{-1} Z^T v \right) \left( \frac1D v^T Z A_0^{-2} Z^T v \right) \left( \frac{1}{\sqrt{D}} v^T Z A_0^{-1} z_2 \right)}{\left( 1 + \frac{\gamma}{D} v^T Z A_0^{-1} Z^T v \right)^2}.
    \end{multline}
    Again, Cor. \ref{cor:list_of_MP_quantities_lambda_zero} and Cor. \ref{cor:list_of_MP_quantities_lambda_positive} collectively imply that $z_1^T A_0^{-2} z_2$, $\frac{ v^T Z A_0^{-j} Z^T v}{D}$, and $\frac{1}{\sqrt{D}}z^T A_0^{-j} Z^T v$ each converge almost surely to deterministic constants for $j \geq 1$ and $z, z_1, z_2 \in \left\{\frac{1}{\sqrt{D}}\epsilon, \frac{1}{\sqrt{D}}Z^T \what, \frac{1}{\sqrt{D}}Z^T v \right\}$. Because the denominator terms in Eq. \ref{eq:A_inv_squared_contractions} are uniformly bounded below by $1$, altogether it follows that $z_1^T A^{-2} z_2$ converges almost surely to a nonrandom constant, for each $z_1, z_2 \in \left\{\frac{1}{\sqrt{D}}\epsilon, \frac{1}{\sqrt{D}}Z^T \what, \frac{1}{\sqrt{D}}Z^T v \right\}$.
    Additionally, Cor. \ref{cor:list_of_MP_quantities_lambda_positive} further implies that, for any fixed $\lambda > 0$, $j \geq 1$, and any $z, z_1, z_2 \in \left\{\frac{1}{\sqrt{D}}\epsilon, \frac{1}{\sqrt{D}}Z^T \what, \frac{1}{\sqrt{D}}Z^T v \right\}$, each of $z_1^T A_0^{-2} z_2$, $\frac{1}{\sqrt{D}}z^T A_0^{-j} Z^T v$, and $\frac{1}{D} v^T Z A_0^{-j} Z^T v$ is moment controlled. Moreover, $1 + \gamma \frac{ v^T Z A_0^{-1} Z^T v}{D} \geq 1$
    is quotient controlled for any $j \geq 1$. Thus, Prop. \ref{prop:uniform_Lp_control} implies that $z_1^T A^{-2} z_2$ is moment controlled for any fixed $\lambda > 0$.
\end{proof}

\begin{cor} \label{cor:A_inv_MP_quantities}
    For $u_1, u_2 \in \{\what, v\}$, each of the following quantities converges almost surely to a deterministic constant (in the limit of $n, D \to \infty$ with $n < D$ and $\frac{n}{D} \to \psi \in (0, 1)$) and is moment controlled when $\lambda > 0$:

    \begin{enumerate}
        \item $\frac{\epsilon^T A^{-1} \epsilon}{n}$ and $\frac{\epsilon^T A^{-2} \epsilon}{n}$, which converge almost surely to $\sigma^2 g(-\lambda)$ and $\sigma^2 g'(-\lambda)$, respectively. Moreover, for any fixed $\lambda > 0$, $\EE\left[ \frac{\epsilon^T A^{-1} \epsilon}{n} \right] \ndlim \sigma^2 g(-\lambda)$ and $\EE\left[ \frac{\epsilon^T A^{-2} \epsilon}{n} \right] \ndlim \sigma^2 g'(-\lambda)$.

        \item $\frac{\epsilon^T A^{-j} X^T u_1}{D}$, for $j \in \{1, 2\}$, which converges almost surely to $0$. Moreover, for any $\lambda > 0$, this quantity is $0$ in expectation.

        \item $\frac{u_1^T X A^{-j} X^T u_2}{D}$ for $j \in \{1, 2\}$.
        
    \end{enumerate}
\end{cor}

\begin{proof}
    In this proof, we apply Eqs. \ref{eq:A_inv_contraction_formula} and \ref{eq:A_inv_squared_contractions}. Additionally, for any $u_1 \in \{\what, v\}$, 
    \begin{equation}    \label{eq:XT_a}
        \frac{1}{\sqrt{D}}X^T u_1 = \frac{1}{\sqrt{D}} Z^T \left( I_D + \zeta vv^T \right) u_1 = \frac{1}{\sqrt{D}} Z^T u_1 + (v^T u_1) \zeta \frac{1}{\sqrt{D}} Z^T v.
    \end{equation}
    We now turn to analyzing the quantities described above.
    
    \paragraph{1.} Using Eq. \ref{eq:A_inv_contraction_formula}, we find that
    $$\frac{\epsilon^T A^{-1} \epsilon}{n} = \frac{\epsilon^T A_0^{-1} \epsilon}{n} - \frac{\gamma \left( \frac{1}{\sqrt{nD}} \epsilon^T A_0^{-1} Z^T v \right)^2}{1 + \gamma v^T \left( \frac{Z A_0^{-1} Z^T}{D} \right) v}$$
    $$ = \frac{\epsilon^T A_0^{-1} \epsilon}{n} - \frac{1}{\frac{n}{D}} \cdot \frac{\gamma \left( \frac{1}{D} \epsilon^T A_0^{-1} Z^T v \right)^2}{1 + \gamma v^T \left( \frac{Z A_0^{-1} Z^T}{D} \right) v} \asconverge \sigma^2 g(-\lambda)$$
    where we have used the facts that $\frac{n}{D} \to \psi$, $\frac{\epsilon^T A_0^{-1} \epsilon}{n} \asconverge \sigma^2 g(-\lambda)$, $\frac{1}{D} \epsilon^T A_0^{-1} Z^T v \asconverge 0$, and $v^T \left( \frac{Z A_0^{-1} Z^T}{D} \right) v \asconverge (v^T v) \psi (1 - \lambda g(-\lambda)) = \tau_1$ for any fixed $\lambda \geq 0$ (Cors. \ref{cor:list_of_MP_quantities_lambda_zero} and \ref{cor:list_of_MP_quantities_lambda_positive}). Moreover, by Prop. \ref{prop:spiked_Ainv_terms}, $\frac{\epsilon^T A^{-1} \epsilon}{n}$ is moment controlled when $\lambda > 0$; thus, for $\lambda > 0$, $\EE\left[ \frac{\epsilon^T A^{-1} \epsilon}{n} \right] \ndlim \sigma^2 g(-\lambda)$ as well (Cor. \ref{cor:Lp_convergence_Vitali}).

    Similarly analyzing $\frac{\epsilon^T A^{-2} \epsilon}{n}$ via Eq. \ref{eq:A_inv_squared_contractions}, we find that
    \begin{multline*}   
         \frac{\epsilon^T A^{-2} \epsilon}{n} = \frac{\epsilon^T A_0^{-2} \epsilon}{n} - \frac{1}{\frac{n}{D}} \cdot \frac{2\gamma \left( \frac{1}{{D}} \epsilon^T A_0^{-2} Z^T v \right) \left( \frac{1}{{D}} v^T Z A_0^{-1} \epsilon \right) }{1 + \frac{\gamma}{D} v^T Z A_0^{-1} Z^T v} + \frac{1}{\frac{n}{D}} \cdot\frac{\gamma^2 \left( \frac{1}{{D}} \epsilon^T A_0^{-1} Z^T v \right)^2 \left( \frac1D v^T Z A_0^{-2} Z^T v \right)}{\left( 1 + \frac{\gamma}{D} v^T Z A_0^{-1} Z^T v \right)^2} \\
         \asconverge \sigma^2 g'(-\lambda).
    \end{multline*}
Here, we have used the facts that as $n, D \to \infty$ with $\frac{n}{D} \to \psi$, $\frac{\epsilon^T A_0^{-2} \epsilon}{n} \asconverge \sigma^2 g'(-\lambda)$, $\frac{1}{D} \epsilon^T A_0^{-j} Z^T v \asconverge 0$, and $v^T \left( \frac{Z A_0^{-1} Z^T}{D} \right) v \asconverge (v^T v) \psi (1 - \lambda g(-\lambda)) = \tau_1$ for any fixed $\lambda \geq 0$ and $j \geq 1$ (Cors. \ref{cor:list_of_MP_quantities_lambda_zero} and \ref{cor:list_of_MP_quantities_lambda_positive}). Moreover, by Prop. \ref{prop:spiked_Ainv_terms}, $\frac{\epsilon^T A^{-2} \epsilon}{n}$ is moment controlled when $\lambda > 0$, in which case $\EE\left[ \frac{\epsilon^T A^{-2} \epsilon}{n} \right] \ndlim \sigma^2 g'(-\lambda)$ as well (Cor. \ref{cor:Lp_convergence_Vitali}).

\paragraph{2.} First, at any finite $n$ and $D$, $\EE\left[ \frac{\epsilon^T A^{-j} X^T u_1}{D} \right] = 0$ trivially for $j \geq 1$ and any $\lambda > 0$, since $\EE[\epsilon] = 0$. Next, using Eq. \ref{eq:XT_a}, we find that, for $j \in \{1, 2\}$, 
$$\frac{\epsilon^T A^{-j} X^T u_1}{D} = \frac{\epsilon^T A^{-j} Z^T u_1}{D} + (v^T u_1) \zeta \frac{\epsilon^T A^{-j} Z^T v}{D}.$$
Thus, to show that $\frac{\epsilon^T A^{-j} X^T u_1}{D} \asconverge 0$, it suffices to show that, for any $u_1 \in \{\what, v\}$, $\frac{\epsilon^T A^{-j} Z^T u_1}{D} \asconverge 0$. Indeed, when $j = 1$, by Eq. \ref{eq:A_inv_contraction_formula}, 
$$\frac{\epsilon^T A^{-1} Z^T u_1}{D} = \frac{\epsilon^T A_0^{-1} Z^T u_1}{D} - \frac{\gamma \left( \frac{1}{{D}} \epsilon^T A_0^{-1} Z^T v \right) \left( \frac{1}{{D}} v^T Z A_0^{-1} Z^T u_1 \right)}{1 + \gamma \frac{ v^T Z A_0^{-1} Z^T v}{D}} \asconverge 0$$
using Cors. \ref{cor:list_of_MP_quantities_lambda_zero} and \ref{cor:list_of_MP_quantities_lambda_positive}. Moreover, for $\lambda > 0$,  $\frac{\epsilon^T A^{-j} X^T u_1}{D}$ is moment controlled (Prop. \ref{prop:spiked_Ainv_terms}). 

Next, when $j = 2$, Eq. \ref{eq:A_inv_squared_contractions} gives
\begin{multline*}   
        \frac{\epsilon^T A^{-2} Z^T u_1}{D} = \frac{\epsilon^T A_0^{-2} Z^T u_1}{D} - \frac{\gamma \left( \frac{1}{{D}}\epsilon^T A_0^{-2} Z^T v \right) \left( \frac{1}{{D}} v^T Z A_0^{-1} Z^T u_1 \right) }{1 + \frac{\gamma}{D} v^T Z A_0^{-1} Z^T v} - \frac{\gamma \left( \frac{1}{{D}} \epsilon^T A_0^{-1} Z^T v \right) \left( \frac{1}{{D}} v^T Z A_0^{-2} Z^T u_1 \right)}{1 + \frac{\gamma}{D} v^T Z A_0^{-1} Z^T v} \\
        + \frac{\gamma^2 \left( \frac{1}{{D}} \epsilon^T A_0^{-1} Z^T v \right) \left( \frac1D v^T Z A_0^{-2} Z^T v \right) \left( \frac{1}{{D}} v^T Z A_0^{-1} Z^T u_1 \right)}{\left( 1 + \frac{\gamma}{D} v^T Z A_0^{-1} Z^T v \right)^2}
\end{multline*}
$$ \asconverge 0$$
by applying Cors. \ref{cor:list_of_MP_quantities_lambda_zero} and \ref{cor:list_of_MP_quantities_lambda_positive}.
Similarly to before, $\frac{\epsilon^T A^{-2} Z^T u_1}{D}$ is moment controlled as well (Prop. \ref{prop:spiked_Ainv_terms}).

\paragraph{3.} Using Eq. \ref{eq:XT_a}, we find that
\begin{align*}
    \frac{u_1^T X A^{-j} X^T u_2}{D} = \frac{u_1^T Z A^{-j} Z^T u_2}{D} + (v^T u_2) \zeta \frac{u_1^T Z A^{-j} Z^T v}{D} + (v^T u_1) \zeta \frac{v^T Z A^{-j} Z^T u_2}{D} \\ + (v^T u_1)(v^T u_2) \zeta^2 \frac{v^T Z A^{-j} Z^T v}{D}
\end{align*}
which has an almost sure limit by Cors. \ref{cor:list_of_MP_quantities_lambda_zero} and \ref{cor:list_of_MP_quantities_lambda_positive} and is moment controlled by Prop. \ref{prop:spiked_Ainv_terms}.
    
\end{proof}

In part 3 of Cor. \ref{cor:A_inv_MP_quantities} above, we do not specify the exact almost sure limit for the quantity $\frac{u_1^T X A^{-j} X^T u_2}{D}$. We provide the specific almost sure limit as part of Cor. \ref{cor:M_1_M_2_expectation_terms}.

\begin{lemma}   \label{lemma:expected_inverse_wishart}
    Suppose that $D > n+1$. Recall that $Z \in \RR^{D \times n}$ is a random matrix consisting of iid standard Gaussian random variables, and that $K_0 = \frac{Z^T Z}{D}$ is a Wishart matrix. Then, 
    $$\EE\left[ K_0^{-1} \right] = \frac{D}{D - n - 1} I_n = \frac{1}{1 - \frac{n}{D} - \frac1D} I_n.$$
\end{lemma}

\begin{proof}
    First, observe that for any orthogonal matrix $V \in \RR^{n \times n}$, $Z \overset{D}{=} ZV$, meaning that 
    $$\EE\left[K_0^{-1}\right] = \EE\left[ \left( \frac{Z^T Z}{D} \right)^{-1} \right] = \EE\left[ \left( \frac{V^T Z^T Z V}{D} \right)^{-1} \right] = V^T \EE\left[ K_0^{-1}  \right] V.$$
    As a result, Lemma \ref{lemma:orthogonal_symmetry}(a) implies that $\EE\left[K_0^{-1}\right] = a_0 I_n$, for a suitable constant $a_0 \geq 0$. Taking the trace then gives
    $$a_0 = \frac{1}{n} \Tr \left( \EE\left[K_0^{-1}\right] \right) = \EE \left(\frac{1}{n} \Tr\left( K_0^{-1} \right) \right) = \EE\left[ \left( K_0^{-1} \right)_{11} \right] = D\EE\left[ \left( \left( Z^T Z \right)^{-1} \right)_{11} \right].$$
    Now, write $H := Z^T Z$ in block matrix form as
    $$H = \left[
\begin{array}{c|c}
H_{11} & H_{1, 2:n} \\
\hline
H_{2:n, 1} & H_{2:n, 2:n}
\end{array}
\right]$$
so that $H_{11}$ is a scalar, and $H_{2:n, 2:n} = \tilde{Z}^T \tilde{Z} \in \RR^{(n-1) \times (n-1)}$ is an unnormalized Wishart matrix arising from the matrix $\tilde{Z} \in \RR^{D \times (n-1)}$ that is obtained by removing the first column $z_1 \in \RR^D$ of $Z$. In particular, we may write $H_{11} = z_1^T z_1$, $H_{1, 2:n} = z_1^T \tilde{Z} \in \RR^{1 \times (n-1)}$, $H_{2:n, 1} = \tilde{Z}^T z_1 \in \RR^{(n-1) \times 1}$, and $H_{2:n, 2:n} = \tilde{Z}^T \tilde{Z} \in \RR^{(n-1) \times (n-1)}$.
Now, we apply the block matrix inversion formula:
$$\left( \left( Z^T Z \right)^{-1} \right)_{11} = \left[ H_{11} - H_{1, 2:n} H_{2:n, 2:n}^{-1} H_{2:n, 1} \right]^{-1}$$
$$ = \left( z_1^T z_1 - z_1^T \tilde{Z} \left( \tilde{Z}^T \tilde{Z} \right)^{-1} \tilde{Z}^T z_1 \right)^{-1} $$ 
$$= \left( z_1^T \left( I_D - \tilde{Z} \left( \tilde{Z}^T \tilde{Z} \right)^{-1} \tilde{Z}^T \right) z_1 \right)^{-1}. $$

Writing $\tilde{Z} = \tilde{U} \tilde{S}\tilde{V}^T$ as the truncated singular value decomposition of $\tilde{Z}$ (where $\tilde{U} \in \RR^{D \times (n-1)}$, $\tilde{S} \in \RR^{(n-1) \times (n-1)}$ is diagonal, and $\tilde{V} \in \RR^{(n-1) \times (n-1)}$), we find that $$\tilde{Z} \left( \tilde{Z}^T \tilde{Z} \right)^{-1} \tilde{Z}^T = \tilde{U}\tilde{S} \tilde{V}^T \left( \tilde{V} \tilde{S} \tilde{U}^T \tilde{U} \tilde{S} \tilde{V}^T \right)^{-1} \tilde{V} \tilde{S} \tilde{U}^T $$
$$ = \tilde{U}\tilde{S} \tilde{V}^T \left( \tilde{V} \tilde{S}^{-2} \tilde{V}^T \right) \tilde{V} \tilde{S} \tilde{U}^T$$
$$ = \tilde{U}\tilde{U}^T$$
where we have used the facts that $\tilde{U}^T \tilde{U} = \tilde{V}^T \tilde{V} = I_{n-1}$. 
In turn, $I_D - \tilde{Z} \left( \tilde{Z}^T \tilde{Z} \right)^{-1} \tilde{Z}^T = \hat{U}\hat{U}^T$, where $\hat{U} \in \RR^{D \times (D-n+1)}$ is a matrix whose $D-n+1 = D - (n-1)$ columns are orthonormal and span the orthogonal complement of the column space of $\tilde{U}$. Since $\hat{U}$ is a function of $\tilde{Z}$, it is independent of $z_1$. As a result, conditioned on $\hat{U}$, $\hat{U}^T z_1 \sim \mathcal{N}\left(0, \hat{U}^T\hat{U}\right) = \mathcal{N}\left(0, I_{D-n+1}\right)$. This means that, regardless of the realization of $\hat{U}$, $\hat{U}^T z_1 \sim \mathcal{N}\left(0, I_{D-n+1}\right)$, meaning that $\norm{\hat{U}^T z_1}_2^2 \sim \chi_{D-n+1}^2$, a chi-squared distribution with $D-n+1$ degrees of freedom. Let $k := D - n + 1$. Moreover, 
$$\left( \left( Z^T Z \right)^{-1} \right)_{11} = \left( z_1^T \left( I_D - \tilde{Z} \left( \tilde{Z}^T \tilde{Z} \right)^{-1} \tilde{Z}^T \right) z_1 \right)^{-1} = \left( z_1^T 
\hat{U} \hat{U}^T z_1 \right)^{-1}$$
$$ = \frac{1}{\norm{\hat{U}^T z_1}_2^2}$$
meaning that $\EE\left[ \left( \left( Z^T Z \right)^{-1} \right)_{11} \right]$ corresponds to integrating the function $f(z) = \frac1z$ over the given chi-squared distribution:
$$\EE\left[ \left( \left( Z^T Z \right)^{-1} \right)_{11} \right] = \int_{0}^\infty \frac1z \frac{z^{\frac{k}{2}-1} e^{-\frac{z}{2}}}{2^{\frac{k}{2}} \Gamma\left( \frac{k}{2} \right)} dz = \int_{0}^\infty \frac{z^{\frac{k}{2}-2} e^{-\frac{z}{2}}}{2^{\frac{k}{2}} \Gamma\left( \frac{k}{2} \right)} dz$$
where $\Gamma(x) := \int_0^\infty t^{x-1} e^{-t} dt$ denotes the gamma function, which is known to satisfy the functional equation $\Gamma(x+1) = x\Gamma(x)$ for $x > 0$. Consequently, 
$$\int_{0}^\infty \frac{z^{\frac{k}{2}-2} e^{-\frac{z}{2}}}{2^{\frac{k}{2}} \Gamma\left( \frac{k}{2} \right)} dz = \frac{1}{2\Gamma\left( \frac{k}{2} \right)} \int_0^\infty \frac12 \left( \frac{z}{2} \right)^{\frac{k}{2}-2} e^{-\frac{z}{2}} dz$$
$$ = \frac{1}{2\Gamma\left( \frac{k}{2} \right)} \cdot \int_0^\infty t^{\frac{k}{2}-2} e^{-t} dt = \frac{1}{2\Gamma\left( \frac{k}{2} \right)} \cdot \Gamma\left( \frac{k}{2} - 1 \right) $$
$$ = \frac{1}{2 \left( \frac{k}{2} - 1 \right)} = \frac{1}{k-2} = \frac{1}{D-n-1}$$
where we have used the substitution of variables $t = \frac{z}{2}$ and the fact that $\frac{k}{2} - 1 = \frac{D-n+1}{2} - 1 > 0$, which follows from our assumption that $D > n+1$. As a result,
$$a_0 = \frac{1}{n} \Tr\left( \EE\left[ K_0^{-1} \right] \right) = D \EE\left[ \left( \left( Z^T Z \right)^{-1} \right)_{11} \right] = \frac{D}{D-n-1}$$
which completes our proof.

\end{proof}

\begin{cor}   \label{cor:integrability_of_inverses_lambda_zero}
    Fix any $\gamma \geq 0$, and suppose that $\lambda = 0$, so that $A = K_x$. Then, for any fixed $n$ and $D$ such that $D > n+1$, the following results hold:
    \begin{enumerate}
        \item $\EE\left[ \eigmin(K_x)^{-1} \right] \leq \frac{nD}{D - n - 1}$.

        \item $\opnorm{M_1} \leq 1$ and $\opnorm{M_2} \leq \eigmin(K_x)^{-1}$, and $\EE[M_1]$ and $\EE[M_2]$ exist. Moreover, $\EE[\epsilon^T K_x^{-1} \epsilon]$ exists.

        \item $\opnorm{\frac{1}{\sqrt{D}} K_x^{-1} X^T} = \eigmin(K_x)^{-1/2}$, and $\EE\left[ \frac{1}{\sqrt{D}} \epsilon^T K_x^{-1} X^T \right] = 0$.

    \end{enumerate}
\end{cor}

\begin{proof}
    Observe that, for any $\gamma \geq 0$, $\Sigma \succeq I_D$, and so $K_x = \frac{1}{D} Z^T \Sigma Z \succeq K_0$, and $K_0 \succ 0$ almost surely when $D > n$. Then,
    $$\eigmin(K_x)^{-1} \leq \Tr(K_x^{-1}) \leq \Tr(K_0^{-1}).$$ Thus, by Lemma \ref{lemma:expected_inverse_wishart}, 
    $$\EE[\eigmin(K_x)^{-1}] \leq \Tr \EE[K_0^{-1}] = \frac{nD}{D - n - 1}.$$
    This proves (1). To prove (2), note that $\opnorm{M_1} = \opnorm{\frac1D X K_x^{-1} X^T} \leq 1$ and $\opnorm{M_2} = \opnorm{\frac1D X K_x^{-2} X^T} = \eigmin(K_x)^{-1}$. Thus, since $\EE[\eigmin(K_x)^{-1}]$ exists at finite $n$ and $D$ by part (1), it follows that $\EE[M_1]$ and $\EE[M_2]$ exist. Similarly, 
    $$\EE[\epsilon^T K_x^{-1} \epsilon] = \EE\left[ \Tr(K_x^{-1} \epsilon \epsilon^T) \right] = \sigma^2\EE[\Tr(K_x^{-1})] \leq \sigma^2\EE[\Tr(K_0^{-1})] = \sigma^2 \frac{nD}{D - n - 1}$$
    exists, by Lemma \ref{lemma:expected_inverse_wishart}. This proves (2). Additionally, $\opnorm{\frac{1}{\sqrt{D}}K_x^{-1} X^T} = \eigmin(K_x)^{-1/2}$ using the singular value decomposition of $\frac{1}{\sqrt{D}}X$, and $\EE\left[\eigmin(K_x)^{-1/2}\right]$ exists as a consequence of part (1). As a result, $\EE\left[ \frac{1}{\sqrt{D}}K_x^{-1} X^T \right]$ exists at finite $n$ and $D$, and since $X$ and $\epsilon$ are independent, 
    $$\EE\left[ \frac{1}{\sqrt{D}} \epsilon^T K_x^{-1} X^T \right] = \EE\left[ \epsilon\right]^T \EE\left[ \frac{1}{\sqrt{D}} K_x^{-1} X^T \right] = 0$$
    establishing part (3).
\end{proof}

\begin{cor} \label{cor:example_inverse_wishart_expectations_at_zero}
    With the same notation as in Lemma \ref{lemma:expected_inverse_wishart} and again assuming that $D > n+1$, the following results hold:
    $$\EE\left[ \frac{ZK_0^{-1} Z^T}{D} \right] = \frac{n}{D} I_D$$
    $$\EE\left[ \frac{Z K_0^{-2} Z^T}{D}\right] = \frac{\frac{n}{D}}{1- \frac{n}{D} - \frac1D} I_D.$$
    In particular, $\frac{n}{D} \to \psi$ and $\frac{\frac{n}{D}}{1 - \frac{n}{D} - \frac1D} \to \frac{\psi}{1-\psi}$ as $n, D \to \infty$ with $D > n+1$ and $\frac{n}{D} \to \psi$.
\end{cor}

\begin{proof}
    First, for fixed $D$ and $n$ with $D > n+1$, $\EE\left[\frac1D Z K_0^{-1} Z^T \right]$ and $\EE\left[\frac1D Z K_0^{-2} Z^T \right]$ exist by Cor. \ref{cor:integrability_of_inverses_lambda_zero}, noting that $M_1 = \frac1D Z K_0^{-1} Z^T$ and $M_2 = \frac1D Z K_0^{-2} Z^T$ when $\lambda = 0$ and $\gamma = 0$.
    
    Recall that $K_0 = \frac{Z^T Z}{D}$. Moreover, since $UZ \overset{D}{=} Z$ for any orthogonal matrix $U \in \RR^{D \times D}$, replacing $Z$ with $UZ$ yields that 
    $$U \frac{Z K_0^{-j} Z^T}{D} U^T \overset{D}{=} \frac{Z K_0^{-j} Z^T}{D} $$ for $j \in \{1, 2\}$. Thus, by Lemma \ref{lemma:orthogonal_symmetry}(a) it follows that $\EE\left[ \frac{Z K_0^{-j} Z^T}{D} \right] = a_j I_D$ for some $a_1, a_2 \geq 0$. In particular, 
    $$a_1 = \frac1D \Tr\left( \EE\left[ \frac{Z K_0^{-1} Z^T}{D} \right] \right) = \frac1D \EE\left[ \Tr\left( \frac{Z K_0^{-1} Z^T}{D} \right) \right] $$
    $$ = \frac1D \EE\left[ \Tr\left( K_0^{-1} \frac{ Z^T Z}{D} \right) \right] = \frac1D \EE\left[ \Tr\left( I_n \right) \right] = \frac{n}{D}.$$
    Turning to $a_2$, we find that
    $$a_2 = \frac1D \Tr\left( \EE\left[ \frac{Z K_0^{-2} Z^T}{D} \right] \right) = \frac1D \EE\left[ \Tr\left( \frac{Z K_0^{-2} Z^T}{D} \right) \right] $$
    $$ = \frac1D \EE\left[ \Tr\left( K_0^{-2} \frac{ Z^T Z}{D} \right) \right] = \frac1D \EE\left[ \Tr\left( K_0^{-1} \right) \right] = \frac{n}{D} \Tr\left( \frac1n \EE\left[ K_0^{-1} \right] \right)$$
    $$ = \frac{\frac{n}{D} \cdot D}{D - n - 1} = \frac{\frac{n}{D}}{1 - \frac{n}{D} - \frac1D}$$
    where we have applied Lemma \ref{lemma:expected_inverse_wishart}.
\end{proof}

\begin{prop}    \label{prop:expected_stieltjes_transform_at_zero}
    Recall that $X = \Sigma^{\frac12}Z \in \RR^{D \times n}$, where $Z \in \RR^{D \times n}$ is a random matrix of iid standard Gaussian random variables and $\Sigma = I_D + \gamma vv^T$. Additionally, recall that $K_x = \frac{X^T X}{D}$. Then, in the proportional asymptotic regime, in which $n +1 < D$ and $\frac{n}{D} \to \psi \in (0, 1)$, $\frac1n \Tr\left( K_x^{-1} \right) \asconverge g(0) = \frac{1}{1 - \psi}$ and
    $$\lim_{n, D \to \infty} \EE\left[ \frac1n \Tr(K_x^{-1}) \right] = g(0) = \frac{1}{1-\psi}$$
    where $g$ is the Stieltjes transform of the Marcenko-Pastur law $\MP^{\psi}$.
\end{prop}

\begin{proof}
    Letting $g_{K_x}$ denote the Stieltjes transform of $K_x$, note that $$g_{K_x}(0) = \frac1n \Tr\left(K_x^{-1} \right) = \frac1n \Tr\left( \left(\frac{Z^T (I_D + \gamma vv^T) Z}{D} \right)^{-1} \right) = \frac1n \Tr\left( \left(K_0 + \frac{\gamma}{D}Z^T vv^T Z\right)^{-1} \right)$$ 
    $$ = \frac1n \Tr\left( K_0^{-1} - \frac{\frac{\gamma}{D} K_0^{-1} Z^T vv^T Z K_0^{-1}}{1 + \frac{\gamma}{D} v^T Z K_0^{-1} Z^T v} \right)$$
    \begin{equation}    \label{eq:g_Kx_zero}
        = \Tr\left(\frac1n K_0^{-1} \right) -  \frac1n \frac{\gamma v^T \left( \frac{Z K_0^{-2} Z^T}{D} \right) v}{1 + \gamma v^T \left( \frac{Z K_0^{-1} Z^T}{D} \right) v}.
    \end{equation}
    For any fixed $n$ and $D$ (with $D > n + 1$), observe that $0 \leq g_{K_x}(0) \leq \Tr\left( \frac1n K_0^{-1} \right)$, which follows from the fact that $K_0^{-1}$ is a symmetric positive semi-definite (PSD) matrix and that $\frac{ZK_0^{-j}Z^T}{D}$ are similarly symmetric PSD for $j \in \{1, 2\}$. Moreover, $\Tr\left( \frac1n K_0^{-1} \right) \asconverge \int \frac1x \MP^{\psi}(dx) = g(0)$ by Cor. \ref{cor:unbounded_f_MP_convergence_zero_case} and $v^T \left( \frac{Z K_0^{-2} Z^T}{D} \right) v \asconverge \frac{\psi}{1-\psi}$ and $v^T \left( \frac{Z K_0^{-1} Z^T}{D} \right) v \asconverge \psi$ by Cor. \ref{cor:list_of_MP_quantities_lambda_zero}. These facts imply that $g_{K_x}(0) = \frac1n \Tr(K_x^{-1}) \asconverge g(0) = \frac{1}{1 - \psi}$ (Prop. \ref{prop:as_p_convergence}).

    Additionally, since $D > n+1$, Lemma \ref{lemma:expected_inverse_wishart} implies that 
    $$\EE\left[ \Tr\left( \frac1n K_0^{-1} \right) \right] = \frac1n \Tr\left( \EE\left[ K_0^{-1} \right] \right) = \frac1n \Tr\left( \frac{D}{D-n-1}I_n \right)$$ 
    $$ = \frac{D}{D-n-1} = \frac{1}{1 - \frac{n}{D} - \frac1D} \overset{n, D \nearrow \infty}{\longrightarrow} \frac{1}{1-\psi} = g(0).$$
    Thus, because $g_{K_x}(0)$ is nonnegative and bounded above by $\Tr\left( \frac1n K_0^{-1} \right)$ (see Eq. \ref{eq:g_Kx_zero}), in the limit as $n, D \to \infty$ with $\frac{n}{D} \to \psi$ (with $D > n+1$), Prop. \ref{prop:generalized_DCT} implies that $\EE\left[g_{K_x}(0)\right]$ exists and converges to the almost sure limit of $g_{K_x}(0)$ as $n, D \to \infty$. That is,
    $$\lim_{n, D \to \infty} \EE\left[ \frac1n \Tr(K_x^{-1}) \right] = g(0) = \frac{1}{1 - \psi}.$$
\end{proof}

In calculating the bias and variance, it becomes important to explicitly compute the terms $\EE\left[ M_1 \right]$ and $\EE\left[ M_2 \right]$, which exist at each fixed $n$ and $D$ such that $D > n+1$; when $\lambda = 0$, this follows from Cor. \ref{cor:integrability_of_inverses_lambda_zero}. When $\lambda > 0$, $\opnorm{M_1} \leq 1$ so $\EE[M_1]$ exists; moreover, the nonzero eigenvalues of $M_2$ are precisely $\left( \frac{\lambda_i(K_x)}{(\lambda_i(K_x) + \lambda)^2} \right)_{i=1}^n$ and so $\EE[M_2]$ exists since
\begin{equation}    \label{eq:M2_opnorm_lambda_positive}
    \opnorm{M_2} \leq \max_{1 \leq i \leq n} \frac{\lambda_i(K_x)}{(\lambda_i(K_x) + \lambda)^2} \leq \frac{1}{\lambda}.
\end{equation}

To compute $\EE[M_1]$ and $\EE[M_2]$, we leverage an important symmetry. Observe that for any orthogonal matrix $U \in \RR^{D \times D}$ and the Gaussian matrix $Z$, $UZ \overset{D}{=} Z$, i.e. $UZ$ and $Z$ share the same law. In particular, observe that for any orthogonal matrix $U$ that \emph{fixes} $v$, $$UX = U \Sigma^{\frac12}Z = U \left( I_D + \zeta vv^T \right) Z = UZ + \zeta Uvv^T Z = UZ + \zeta vv^T UZ$$
$$ = \Sigma^{\frac12} UZ \overset{D}{=} \Sigma^{\frac12} Z = X.$$
Thus, for any such matrix $U$, the transformation $X \mapsto UX$ preserves the law of the matrices $M_1 = \frac{1}{D} X A^{-1} X^T = \frac1D X \left( \frac{X^T X}{D} + \lambda I_n \right)^{-1} X^T$ and $M_2 = \frac{1}{D} X A^{-2} X^T = \frac1D X \left( \frac{X^T X}{D} + \lambda I_n \right)^{-2} X^T$. In particular, the transformation $X \mapsto UX$ transforms $M_1$ to $U M_1 U^T$ and $M_2$ to $U M_2 U^T$. Thus, for $j \in \{1, 2\}$, we find that 
$$\EE\left[ \frac1D X A^{-j} X^T \right] = \EE\left[ U \left( \frac1D X A^{-j} X^T \right) U^T \right] = U \left( \EE\left[ \frac1D X A^{-j} X^T \right] \right) U^T$$ for all orthogonal matrices $U \in \RR^{D \times D}$ that fix $v$. As a result, Lemma \ref{lemma:orthogonal_symmetry}(b) implies that
\begin{equation}    \label{eq:M_1}
    \EE\left[ M_1 \right] = aI_D + b vv^T
\end{equation}  
\begin{equation} \label{eq:M_2}
    \EE\left[ M_2 \right] = \tilde{a}I_D + \tilde{b} vv^T
\end{equation}
    for some constants $a, b, \tilde{a}, \tilde{b} \in \RR$ that implicitly depend on $n$, $D$, $\lambda$, and $\gamma$. Note that $a$ and $\tilde{a}$ are nonnegative because $M_1$ and $M_2$ are symmetric positive semi-definite matrices.

\begin{prop}   \label{prop:M_trace}
    Recall that $M_1 = \frac1D X A^{-1} X^T$ and $M_2 = \frac1D X A^{-2}X^T$. In the proportional asymptotic regime, in which $n +1 < D$ and $\frac{n}{D} \to \psi \in (0, 1)$,
    $\Tr\left( \frac1D M_1\right) \asconverge \tau_1 = \psi(1 - \lambda g(-\lambda))$ and $\Tr\left( \frac1D M_2\right) \asconverge \tau_2 = \psi(g(-\lambda) - \lambda g'(-\lambda))$. Moreover, both $\Tr\left( \frac1D M_1\right)$ and $\lambda \Tr\left( \frac1D M_2\right)$ are moment controlled. Finally, $\EE\left[ \Tr\left( \frac1D M_1\right)  \right] \ndlim \tau_1$ and $\EE\left[ \Tr\left( \frac1D M_2\right)  \right] \ndlim \tau_2$ for any fixed $\lambda \geq 0$.
    
\end{prop}

\begin{proof}
    Letting $g_{K_0}$ denote the Stieltjes transform of $K_0$, note first that Cor. \ref{cor:unbounded_f_MP_convergence} implies that $g_{K_0}(-\lambda) \asconverge g(-\lambda)$ if $\lambda > 0$ and Cor. \ref{cor:unbounded_f_MP_convergence_zero_case} implies that $g_{K_0}(0) \asconverge g(0)$ if $\lambda = 0$ (since $n < D$).
    Now, if $\lambda = 0$, then $A = K_x$ and so $$\Tr\left( \frac1D M_1\right) = \frac1D \Tr\left( A^{-1} \frac{X^T X}{D}\right) = \frac1D \Tr\left( A^{-1} K_x \right) = \frac1D \Tr(K_x^{-1} K_x) = \frac{n}{D} \overset{n, D \nearrow \infty}{\longrightarrow} \psi$$ which is precisely the value of $\tau_1(\lambda = 0)$, and $\frac{n}{D} < 1$ is trivially moment controlled. For $\lambda > 0$, we similarly have that
    $$\Tr\left( \frac1D M_1\right) = \frac1D \Tr\left( A^{-1} K_x \right) = \frac1D \Tr\left( A^{-1} (A - \lambda I_n) \right) = \frac{n}{D}\left(1 - \lambda \Tr\left( \frac1n  A^{-1}\right) \right).$$
    Recall that 
    $$A^{-1} = A_0^{-1} - \frac{\frac{\gamma}{D} A_0^{-1} Z^T v v^T Z A_0^{-1}}{1 + \frac{\gamma}{D} v^T Z A_0^{-1} Z^T v}$$
    Then, taking the trace and using the cyclic property of the trace yields
    $$\Tr\left( \frac1n A^{-1} \right) =  \Tr\left( \frac1n A_0^{-1} \right) - \frac1n\left( \frac{ \gamma v^T \left( \frac{Z A_0^{-2} Z^T}{D} \right) v }{1 + \gamma v^T \left( \frac{Z A_0^{-1} Z^T}{D} \right) v} \right).$$
    Recalling that $\Tr\left( \frac1n A_0^{-1} \right)$, $v^T \left( \frac{Z A_0^{-1} Z^T}{D} \right) v$, and $v^T \left( \frac{Z A_0^{-2} Z^T}{D} \right) v$ are moment controlled for $\lambda > 0$ (Cor. \ref{cor:list_of_MP_quantities_lambda_positive}), that each of these quantities converges almost surely to a nonrandom constant for $\lambda > 0$ (Cor. \ref{cor:list_of_MP_quantities_lambda_positive}), and that $1 + \gamma v^T \left( \frac{Z A_0^{-1} Z^T}{D} \right) v \geq 1$  is quotient controlled, it follows that $\Tr\left( \frac1n A^{-1} \right)$ is moment controlled (Prop. \ref{prop:uniform_Lp_control}) and that
    $$\Tr\left( \frac1n A^{-1} \right) \asconverge g(-\lambda)$$
    where we have noted that 
    $\frac{ \gamma v^T \left( \frac{Z A_0^{-2} Z^T}{D} \right) v }{1 + \gamma v^T \left( \frac{Z A_0^{-1} Z^T}{D} \right) v}$ converges almost surely to a nonrandom constant independent of $n$ and 
    $\Tr\left( \frac1n A_0^{-1} \right) \asconverge g(-\lambda)$ (Cor. \ref{cor:list_of_MP_quantities_lambda_positive}).
    Consequently, $\Tr\left( \frac1D M_1 \right) \asconverge \psi(1 - \lambda g(-\lambda)) = \tau_1$, and it is moreover moment controlled. Thus, combining our results for both $\lambda = 0$ and $\lambda > 0$, we find that $\Tr\left( \frac1D M_1 \right) \asconverge \tau_1$ and that $\Tr\left( \frac1D M_1 \right)$ is moment controlled. In turn, Cor. \ref{cor:Lp_convergence_Vitali} implies that $\EE\left[ \Tr\left( \frac1D M_1 \right) \right] \overset{n, D \nearrow \infty}{\longrightarrow} \psi\left(1 - \lambda g(-\lambda) \right) = \tau_1$.

    Next, we turn to studying $\Tr\left( \frac1D M_2 \right)$. Again, if $\lambda = 0$, then 
    $$\Tr\left( \frac1D M_2 \right) = \frac1D \Tr\left( A^{-2} \frac{X^T X}{D}\right) = \frac1D \Tr\left( A^{-2} K_x \right) = \frac1D \Tr(K_x^{-2} K_x)$$ $$= \frac{n}{D}  \Tr\left(\frac1n K_x^{-1}\right) \asconverge \psi g(0) = \frac{\psi}{1 - \psi} $$
    which is precisely the value of $\tau_2 = \psi(g(-\lambda) - \lambda g'(-\lambda))$ when $\lambda = 0$.  Here, we have applied Prop. \ref{prop:expected_stieltjes_transform_at_zero} to show that $\Tr\left(\frac1n K_x^{-1}\right) \asconverge g(0)$. Additionally, $\EE\left[\frac1n \Tr(K_x^{-1}) \right] \to \frac{1}{1 - \psi}$ (Prop. \ref{prop:expected_stieltjes_transform_at_zero}), so $\EE\left[ \Tr\left(\frac1D M_2 \right) \right] \ndlim \frac{\psi}{1 - \psi} = \tau_2(\lambda = 0)$.
    Moreover, when $\lambda = 0$, $\lambda \Tr\left( \frac1D M_2 \right) = 0$ (almost surely) is trivially moment controlled. 
    If $\lambda > 0$, then 
    $$\Tr\left( \frac1D M_2\right) = \frac1D \Tr\left( A^{-2} K_x \right) = \frac1D \Tr\left( A^{-2} (A - \lambda I_n) \right) = \frac{n}{D}\left(\Tr\left( \frac1n A^{-1} \right) - \lambda \Tr\left( \frac1n  A^{-2}\right) \right) $$
    $$ = \frac{n}{D} \left(  \Tr\left( A_0^{-1} - \frac{\frac{\gamma}{D} A_0^{-1} Z^T v v^T Z A_0^{-1}}{1 + \frac{\gamma}{D} v^T Z A_0^{-1} Z^T v} \right) - \lambda \Tr \left( \left( A_0^{-1} - \frac{\frac{\gamma}{D} A_0^{-1} Z^T v v^T Z A_0^{-1}}{1 + \frac{\gamma}{D} v^T Z A_0^{-1} Z^T v} \right)^2  \right)  \right)$$
    \begin{multline*}
        = \frac{n}{D} \left( \Tr\left( \frac1n A_0^{-1} \right) - \frac1n\left( \frac{ \gamma v^T \left( \frac{Z A_0^{-2} Z^T}{D} \right) v }{1 + \gamma v^T \left( \frac{Z A_0^{-1} Z^T}{D} \right) v}\right) \right. \\ \left.
        - \lambda \left( \Tr\left(\frac1n A_0^{-2} \right) - \frac2n\left( \frac{ \gamma v^T \left( \frac{Z A_0^{-3} Z^T}{D} \right) v }{1 + \gamma v^T \left( \frac{Z A_0^{-1} Z^T}{D} \right) v}\right) + \frac1n \left( \frac{ \gamma v^T \left( \frac{Z A_0^{-2} Z^T}{D} \right) v }{1 + \gamma v^T \left( \frac{Z A_0^{-1} Z^T}{D} \right) v}\right)^2 \right) \right)
    \end{multline*}
    $$ \asconverge \psi(g(-\lambda) - \lambda g'(-\lambda)) = \tau_2.$$
 This convergence result follows from Cor. \ref{cor:list_of_MP_quantities_lambda_positive}, which implies that as $n, D \to \infty$ with $\frac{n}{D} \to \psi$, it is the case that $\Tr\left( \frac1n A_0^{-1} \right) \asconverge \int \frac{1}{x+\lambda} \MP^{\psi}(dx) = g(-\lambda)$, $\Tr\left( \frac1n A_0^{-2} \right) \asconverge \int \frac{1}{(x+\lambda)^2} \MP^{\psi}(dx) = g'(-\lambda)$, and that $$v^T \left( \frac{Z A_0^{-j} Z^T}{D} \right) v \asconverge \psi \left( \int \frac{1}{(x+\lambda)^{j-1}} \MP^{\psi}(dx) - \lambda \int \frac{1}{(x+\lambda)^{j}} \MP^{\psi}(dx)  \right) = O_n(1)$$
 for $j \geq 1$.
Consequently, terms other than $\Tr\left( \frac1n A_0^{-1} \right)$ and $\Tr\left( \frac1n A_0^{-2} \right)$ almost surely converge to $0$. Finally, for any fixed $\lambda > 0$, each of $\Tr\left( \frac1n A_0^{-j} \right)$ and $v^T \left( \frac{Z A_0^{-j} Z^T}{D} \right) v$ is moment controlled for $j \geq 1$, and $1 + \gamma v^T \left( \frac{Z A_0^{-1} Z^T}{D} \right) v \geq 1$ is quotient controlled. Therefore, Prop. \ref{prop:uniform_Lp_control} implies that, for any fixed $\lambda > 0$, $\Tr \left(\frac1D M_2 \right)$ is moment controlled. Thus, by Cor. \ref{cor:Lp_convergence_Vitali}, it follows that
$\EE\left[ \Tr\left( \frac1D M_2 \right) \right] \ndlim \tau_2$ for $\lambda > 0$. Altogether, combining our results for $\lambda = 0$ and $\lambda > 0$, we find that $\EE\left[ \Tr\left( \frac1D M_2 \right) \right] \ndlim \tau_2$, as desired.
 
\end{proof}

\begin{prop}    \label{prop:M_v_contraction}
    With the hypotheses that $n + 1 < D$ and $\psi \in (0, 1)$, the following statements hold true for any fixed $\lambda \geq 0$ as $n, D \to \infty$ with $\frac{n}{D} \to \psi$:
    $$\EE\left[ v^T M_1 v \right] \ndlim \frac{(\gamma + 1) \tau_1}{1 + \gamma \tau_1}$$
    $$\EE\left[v^T M_2 v \right] \ndlim \frac{(\gamma + 1)\tau_2}{(1 + \gamma \tau_1)^2}.$$
\end{prop}

\begin{proof}
    Observe first that 
    \begin{equation}    \label{eq:ZT_v}
        X^T v = Z^T (I_D + \zeta vv^T) v = (1 + \zeta) Z^T v.
    \end{equation}

    As a result, we find that
    $$v^T M_1 v = \frac1D v^T X A^{-1} X^T v = \frac{(1 + \zeta)^2}{D} v^T Z \left( A_0^{-1} - \frac{\frac{\gamma}{D} A_0^{-1} Z^T v v^T Z A_0^{-1}}{1 + \frac{\gamma}{D} v^T Z A_0^{-1} Z^T v} \right) Z^T v$$
    $$ = (\gamma + 1) \left(\frac1D v^T Z A_0^{-1} Z^T v - \frac{\gamma \left(\frac1D v^T Z A_0^{-1} Z^T v \right)^2}{1 + \frac{\gamma}{D} v^T Z A_0^{-1} Z^T v} \right)$$
    $$ = \frac{(\gamma + 1) \frac1D v^T Z A_0^{-1} Z^T v }{1 + \frac{\gamma}{D} v^T Z A_0^{-1} Z^T v} \asconverge \frac{(\gamma + 1) \tau_1}{1 + \gamma \tau_1}$$
    for any fixed $\lambda \geq 0$, where we have applied Cors. \ref{cor:list_of_MP_quantities_lambda_zero} and \ref{cor:list_of_MP_quantities_lambda_positive} to show that $\frac1D v^T Z A_0^{-1} Z^T v \asconverge \tau_1 = \psi(1 - \lambda g(-\lambda))$.
    If $\lambda > 0$, then $\frac1D v^T Z A_0^{-1} Z^T v$ is moreover moment controlled (Cor. \ref{cor:list_of_MP_quantities_lambda_positive}) and so we readily find that $$\EE[v^T M_1 v] \ndlim \frac{(\gamma + 1) \tau_1}{1 + \gamma \tau_1}.$$
    When $\lambda = 0$, $A_0 = K_0$. Moreover,
    $$ 0 \leq v^T M_1 v \leq (\gamma + 1) \frac1D v^T Z K_0^{-1} Z^T v.$$
    Note that as $n, D \to \infty$ with $n+1 < D$ and $\frac{n}{D} \to \psi \in (0, 1)$, $\frac1D v^T Z K_0^{-1} Z^T v \asconverge \psi = \tau_1(\lambda = 0)$ (Cor. \ref{cor:list_of_MP_quantities_lambda_zero}) and that 
    $$\EE\left[ (\gamma + 1) \frac1D v^T Z K_0^{-1} Z^T v \right] = (\gamma + 1) v^T \EE\left[ \frac{Z K_0^{-1} Z^T}{D} \right] v = (\gamma + 1)\frac{n}{D} \ndlim (\gamma + 1)\psi$$
    where we have applied Cor. \ref{cor:example_inverse_wishart_expectations_at_zero} to show that $$\EE\left[ \frac{Z K_0^{-1} Z^T}{D} \right] = \frac{n}{D} I_D$$ for $n + 1 < D$.
    Thus, the sequence formed by $(\gamma + 1) \frac1D v^T Z K_0^{-1} Z^T v$ (as $n$ and $D$ are taken to infinity in the proportional asymptotic setting) dominates the sequence formed by $v^T M_1 v$, both of which are nonnegative. Consequently, Prop. \ref{prop:generalized_DCT} (a generalization of the dominated convergence theorem) implies that 
    $$\EE[v^T M_1 v] \ndlim \frac{(\gamma + 1) \tau_1}{1 + \gamma \tau_1}$$
    when $\lambda = 0$ as well.

    Next, we consider $v^T M_2 v$. In particular, observe first that
    $$A^{-1} X^T v = (1 + \zeta) \left( A_0^{-1} - \frac{\frac{\gamma}{D} A_0^{-1} Z^T v v^T Z A_0^{-1}}{1 + \frac{\gamma}{D} v^T Z A_0^{-1} Z^T v} \right) Z^T v$$
    $$ = (1 + \zeta) \left( A_0^{-1} Z^T v - \frac{\frac{\gamma}{D} v^T Z A_0^{-1} Z^T v}{1 + \frac{\gamma}{D} v^T Z A_0^{-1} Z^T v}\cdot  A_0^{-1} Z^T v \right)$$
    $$ = \frac{1 + \zeta}{ 1 + \frac{\gamma}{D} v^T Z A_0^{-1} Z^T v } \cdot A_0^{-1} Z^T v.$$
    As a result,
    $$v^T M_2 v  = \frac{1}{D} v^T X A^{-2} X^T v = \frac{1}{D} \norm{A^{-1} X^T v}_2^2$$
    $$ = \frac{(1 + \zeta)^2 \frac1D v^T Z A_0^{-2} Z^T v}{\left( 1 + \frac{\gamma}{D} v^T Z A_0^{-1} Z^T v \right)^2} \asconverge \frac{(\gamma + 1) \tau_2}{(1 + \gamma \tau_1)^2}$$
    for any fixed $\lambda \geq 0$, where we have applied Cors. \ref{cor:list_of_MP_quantities_lambda_zero} and \ref{cor:list_of_MP_quantities_lambda_positive} to show that $\frac1D v^T Z A_0^{-1} Z^T v \asconverge \tau_1$ and $\frac1D v^T Z A_0^{-2} Z^T v \asconverge \tau_2$. For any fixed $\lambda > 0$, Cor. \ref{cor:list_of_MP_quantities_lambda_positive} implies that $\frac1D v^T Z A_0^{-1} Z^T v$ and $\frac1D v^T Z A_0^{-2} Z^T v$ are moment controlled (and the denominator term of $v^T M_2 v$ is bounded below by $1$), implying that $v^T M_2 v$ is moment controlled (Prop. \ref{prop:uniform_Lp_control}). Consequently, we find that $$\EE[v^T M_2 v] \ndlim \frac{(\gamma + 1) \tau_2}{(1 + \gamma \tau_1)^2}$$
    for any fixed $\lambda > 0$. When $\lambda = 0$ (so that $A_0 = K_0$), note that
    $$ 0 \leq v^T M_2 v \leq (\gamma + 1) \frac1D v^T Z K_0^{-2} Z^T v.$$
    As $n, D \to \infty$ with $n + 1 < D$ and $\frac{n}{D} \to \psi \in (0, 1)$, it is the case that $\frac1D v^T Z K_0^{-2} Z^T v \asconverge \frac{\psi}{1-\psi} = \tau_2(\lambda = 0)$ (Cor. \ref{cor:list_of_MP_quantities_lambda_zero}) and that
    $$\EE\left[ (\gamma + 1) \frac1D v^T Z K_0^{-2} Z^T v \right] = (\gamma + 1) v^T \EE\left[ \frac{Z K_0^{-2} Z^T}{D} \right] v = \frac{(\gamma + 1) \frac{n}{D}}{1 - \frac{n}{D} - \frac1D} \ndlim \frac{(\gamma + 1) \psi}{1 - \psi}$$
    where we have applied Cor. \ref{cor:example_inverse_wishart_expectations_at_zero} to show that $$\EE\left[ \frac{Z K_0^{-2} Z^T}{D} \right] = \frac{\frac{n}{D}}{1 - \frac{n}{D} - \frac1D}I_D$$ for $n + 1 < D$.
    Thus, the sequence formed by $(\gamma + 1) \frac1D v^T Z K_0^{-2} Z^T v$ (as $n$ and $D$ are taken to infinity in the proportional asymptotic setting) dominates the sequence formed by $v^T M_2 v$, both of which are nonnegative.
    Consequently, Prop. \ref{prop:generalized_DCT} implies that 
    $$\EE[v^T M_2 v] \ndlim \frac{(\gamma + 1) \tau_2}{(1 + \gamma \tau_1)^2}$$
    when $\lambda = 0$ as well.
\end{proof}

Combining Prop. \ref{prop:M_trace} and Prop. \ref{prop:M_v_contraction}, we obtain the following limits:
\begin{align}
    a + \frac{b}{D} = \EE\left[\Tr\left(\frac1D M_1 \right)\right] \ndlim \tau_1 \\
    a + b = \EE\left[ v^T M_1 v \right] \ndlim \frac{(\gamma + 1) \tau_1}{1 + \gamma \tau_1} 
\end{align}
\begin{align}
    \tilde{a} + \frac{\tilde{b}}{D} = \EE\left[\Tr\left(\frac1D M_2 \right)\right] \ndlim \tau_2 \\
    \tilde{a} + \tilde{b} = \EE\left[ v^T M_2 v \right] \ndlim \frac{(\gamma + 1) \tau_2}{(1 + \gamma \tau_1)^2} 
\end{align}

Solving these equations for $a$, $b$, $\tilde{a}$, and $\tilde{b}$ yields the following:
\begin{align*}
    a = \frac{1}{D-1} \left( D \EE\left[\Tr\left(\frac1D M_1 \right)\right] - \EE\left[ v^T M_1 v \right] \right) \ndlim \tau_1 
\end{align*}
\begin{align*}
    b = \frac{D}{D-1} \left( \EE\left[ v^T M_1 v \right] - \EE\left[\Tr\left(\frac1D M_1 \right)\right] \right) \ndlim \frac{(\gamma + 1) \tau_1}{1 + \gamma \tau_1} - \tau_1 = \frac{\gamma \tau_1 (1 - \tau_1)}{1 + \gamma \tau_1}
\end{align*}
\begin{align*}
    \tilde{a} = \frac{1}{D-1} \left( D \EE\left[\Tr\left(\frac1D M_2 \right)\right] - \EE\left[ v^T M_2 v \right] \right) \ndlim \tau_2 
\end{align*}
\begin{align*}
    \tilde{b} = \frac{D}{D-1} \left( \EE\left[ v^T M_2 v \right] - \EE\left[\Tr\left(\frac1D M_2 \right)\right] \right) \ndlim \frac{(\gamma + 1) \tau_2}{(1 + \gamma \tau_1)^2} - \tau_2
\end{align*}

As a direct consequence of this result, we may state the following corollaries.
\begin{cor} \label{cor:a_b_atilde_btilde}
    For $M_1 = \frac1D X A^{-1} X^T$ and $M_2 = \frac1D X A^{-2} X^T$, it is the case that $\EE[M_1] = aI_D + b vv^T$ and $\EE[M_2] = \tilde{a} I_D + \tilde{b} vv^T$, where $a$, $b$, $\tilde{a}$, and $\tilde{b}$ are each $O_D(1)$ quantities with the following respective limits:
    \begin{equation}
        a \ndlim \tau_1 =: \alim 
    \end{equation}
    \begin{equation}
        b \ndlim \frac{\gamma \tau_1 (1 - \tau_1)}{1 + \gamma \tau_1} =: \blim
    \end{equation}
    \begin{equation}
        \tilde{a} \ndlim \tau_2 =: \atlim 
    \end{equation}
    \begin{equation}
        \tilde{b} \ndlim \frac{(\gamma + 1) \tau_2}{(1 + \gamma \tau_1)^2} - \tau_2 =: \btlim
    \end{equation}
Here, $\tau_1 = \psi(1 - \lambda g(-\lambda))$ and $\tau_2 = \psi(g(-\lambda) - \lambda g'(-\lambda))$.
\end{cor}

\begin{cor} \label{cor:M_1_M_2_expectation_terms}
    For any $u_1, u_2 \in \{\what, v\}$, in the proportional limit of $n, D \to \infty$ with $n + 1 < D$ and $\frac{n}{D} \to \psi \in (0, 1)$, the following results hold: 
    $$\EE[u_1^T M_1 u_2] \ndlim \alim (u_1^T u_2) + \blim (u_1^T v) (u_2^T v)$$
    $$\EE[u_1^T M_2 u_2] \ndlim \atlim (u_1^T u_2) + \btlim (u_1^T v) (u_2^T v).$$
    Moreover, for any fixed $\lambda > 0$, the following almost sure convergence results hold:
    $$u_1^T M_1 u_2 \asconverge \alim (u_1^T u_2) + \blim (u_1^T v) (u_2^T v)$$
    $$u_1^T M_2 u_2 \asconverge \atlim (u_1^T u_2) + \btlim (u_1^T v) (u_2^T v).$$
    
\end{cor}

\begin{proof}
    The convergence results for $\EE[u_1^T M_j u_2]$ (for $j \in \{1, 2\}$) follow directly from our computation of $\alim$, $\blim$, $\atlim$, and $\btlim$, as well as the facts that $\EE[M_1] = a I_D + b vv^T$ and $\EE[M_2] = \tilde{a} I_D + \tilde{b} vv^T$. Furthermore, for any fixed $\lambda > 0$, Cor. \ref{cor:A_inv_MP_quantities} implies that $u_1^T M_j u_2$ almost surely converges to a deterministic constant and is moment controlled. Consequently, the almost sure limit of $u_1^T M_j u_2$ is precisely the limit of $\EE[u_1^T M_j u_2]$ as $n, D \to \infty$ (Cor. \ref{cor:Lp_convergence_Vitali}). 
\end{proof}

We now use the results we have established in order to study the limiting behavior of the inflationary constant, $\cl$.

\subsection{Asymptotics of \texorpdfstring{$\cl$}{the inflationary constant}}   \label{app:cl_thm_proof}

In this section, we state and prove a slight extension of Theorem \ref{thm:inflation_constant}, as stated in the main text. From the following Thm. \ref{thm:inflation_constant_extended}, it will follow that Theorem \ref{thm:inflation_constant} holds.

First recall that whenever $D > n$, 
$$\cl - 1 = \frac{\lambda \frac{\eta^2 d_\ell}{n^2} y^T K_x A^{-1} y}{1 + \frac{\eta^2 d_\ell}{n^2} y^T K_x A^{-1} K_x y} = \frac{\lambda \frac{\eta^2 d_\ell}{n^2} y^T (A - \lambda I_n) A^{-1} y}{1 + \frac{\eta^2 d_\ell}{n^2} y^T K_x A^{-1} (A - \lambda I_n) y}$$
$$ = \frac{\lambda \frac{D}{n} \frac{\eta^2 d_\ell}{D} \left( \frac1n y^T y - \frac{\lambda}{n} y^T A^{-1} y \right)}{1 + \frac{\eta^2 d_\ell}{D} \frac{D}{n^2}\left( y^T K_x y - \lambda y^T K_x A^{-1} y \right)} = \frac{\lambda \frac{D}{n} \frac{\eta^2 d_\ell}{D} \left( \frac1n y^T y - \lambda \frac1n y^T A^{-1} y \right)}{1 + \frac{\eta^2 d_\ell}{D}\left( \frac{D}{n^2} y^T K_x y - \lambda \frac{D}{n} \frac{y^T y}{n} + \lambda^2 \frac{D}{n} \frac{y^T A^{-1} y}{n} \right)}$$
\begin{align}   \label{eq:cl_expanded_expression}
    = \frac{\lambda \frac{D}{n} k_\ell \left( \frac1n y^T y - \lambda \frac1n y^T A^{-1} y \right)}{1 + k_\ell \left( \frac{D}{n^2} y^T K_x y - \lambda \frac{D}{n} \frac{y^T y}{n} + \lambda^2 \frac{D}{n} \frac{y^T A^{-1} y}{n} \right)}.
\end{align}
Here, we have applied the definition $\kl := \frac{\eta^2 d_\ell}{D}$. Now, we state the following extended version of Theorem \ref{thm:inflation_constant}.

\begin{thm} \label{thm:inflation_constant_extended}
     Suppose that Assumption \ref{assump:data_gen_process} holds and that $\lim_{n, D \to \infty} k_\ell = \bkl$ for some nonnegative constant $\bkl$ independent of $D$ and $n$. Then, as $n, D \to \infty$, 
     the inflationary constant $\cl$ is moment controlled and converges almost surely to a nonrandom constant, $\bcl$, that is at least $1$. In particular,
         \begin{equation}
    \cl \asconverge 1 + \frac{\frac{\lambda}{\psi} \bkl \left( 1 + \sigma^2 + \gamma \rho^2 - \lambda \left( \frac{\alim + \blim \rho^2}{\psi} + \sigma^2 g(-\lambda) \right) \right)}{1 + \bkl \left( 1 + \frac{1+\sigma^2}{\psi} + \gamma \rho^2 \left( 2 + \gamma + \frac{1}{\psi} \right) - \frac{\lambda}{\psi}\left( 1 + \sigma^2 + \gamma \rho^2 \right) + \frac{\lambda^2}{\psi} \left( \frac{\alim + \blim \rho^2}{\psi} + \sigma^2 g(-\lambda) \right) \right)}.
\end{equation}
In particular, when $k_\ell = o_D(1)$, $\cl \asconverge 1$, and $\bcl > 1$ when $\bkl > 0$ (meaning that $\eta^2 \ell^3 \asymp D$) and $\lambda > 0$.
\end{thm}

In order to prove Theorem \ref{thm:inflation_constant_extended}, as well as establish further properties of $\bcl$, we state the following useful results (Lemma \ref{lemma:bcl_intermediate_quantities_converge} and Prop. \ref{prop:bcl_properties}).

\begin{lemma}   \label{lemma:bcl_intermediate_quantities_converge}
    Under Assumption \ref{assump:data_gen_process} and for any fixed $\lambda \geq 0$, each of $\frac{y^T y}{n}$, $\frac{\lambda y^T A^{-1} y}{n}$, and $\frac{D}{n^2} y^T K_x y$ has an almost sure limit and is moment controlled. In particular, the following results hold:
\begin{align}   \label{eq:pure_y_term}
    \frac{y^T y}{n} \asconverge 1 + \gamma \rho^2 + \sigma^2.
\end{align}

\begin{align}   \label{eq:y_Ainv_y_term}
    \frac{\lambda y^T A^{-1} y}{n} \asconverge \frac{\lambda}{\psi}(\alim + \blim \rho^2) + \lambda \sigma^2 g(-\lambda).
\end{align}

\begin{align}   \label{eq:y_Kx_y_term}
    \frac{D}{n^2} y^T K_x y \asconverge 1 + \frac{1+\sigma^2}{\psi} + \gamma\rho^2\left(2 + \gamma + \frac{1}{\psi} \right).
\end{align}

\end{lemma}

\begin{proof}
We proceed by analyzing each of $\frac{y^T y}{n}$, $\frac{\lambda y^T A^{-1} y}{n}$, and $\frac{D}{n^2} y^T K_x y$. Recall that 
$X = \Sigma^{\frac12} Z = (I_D + \zeta vv^T) Z$, where $Z \in \RR^{D \times n}$ is a matrix of iid standard Gaussian random variables and $\zeta = \sqrt{\gamma + 1} - 1$. Also recall that $\rho = \what^T v = \frac{1}{\sqrt{D}}\ws^T v$.
Since $y = \frac{1}{\sqrt{D}}X^T w_\star + \epsilon$, 
$$\frac{y^T y}{n} = \frac1D w_\star^T \left(\frac1n XX^T \right) w_\star + \frac{2}{n} \epsilon^T \left( \frac{1}{\sqrt{D}} X^T w_\star \right) + \frac{\epsilon^T \epsilon}{n}.$$
First, $\frac{\epsilon^T \epsilon}{n} \asconverge \sigma^2$ and has uniformly controlled $L^p$ moments for any $p \geq 1$ (Cor. \ref{cor:list_of_MP_quantities_lambda_positive}). Next, we compute
$$\frac1D w_\star^T \left( \frac1n XX^T \right) w_\star = \what^T (I_D + \zeta vv^T) \left( \frac1n ZZ^T \right) (I_D + \zeta vv^T) \what$$
$$ = \what^T \left( \frac1n ZZ^T \right) \what + 2 \zeta \rho v^T \left( \frac1n ZZ^T \right) \what + \zeta^2 \rho^2 v^T \left( \frac1n ZZ^T \right) v$$
$$\asconverge 1 + 2\zeta \rho (\rho) + \zeta^2 \rho^2 (1) = 1 + ((\zeta + 1)^2 - 1) \rho^2 = 1 + \gamma \rho^2.$$
Here, we have invoked Cor. \ref{cor:list_of_MP_quantities_lambda_positive}, and furthermore note that $\frac1D w_\star^T \left( \frac1n XX^T \right) w_\star$ is moment controlled because it is the sum of moment controlled quantities (Prop. \ref{prop:uniform_Lp_control}). Next,
$$\frac{2}{n} \epsilon^T \left( \frac{1}{\sqrt{D}} X^T w_\star \right) = \frac{2}{n} \epsilon^T Z^T (I_D + \zeta vv^T) \what = \frac{2}{n} \epsilon^T Z^T \what + \frac{2 \zeta \rho}{n} \epsilon^T Z^T v \asconverge 0.$$
Here, we have used the facts that $\frac1n \epsilon^T Z^T \what \asconverge 0$ and $\frac1n \epsilon^T Z^T v \asconverge 0$ and that these quantities are moment controlled (Cor. \ref{cor:list_of_MP_quantities_lambda_positive}).
Altogether, we find that 
$$\frac{y^T y}{n} \asconverge 1 + \gamma \rho^2 + \sigma^2$$
and that it is moment controlled because it is a sum of moment controlled quantities (Prop. \ref{prop:uniform_Lp_control}).

Next, we turn to $\frac{\lambda y^T A^{-1} y}{n}$. If $\lambda = 0$, this quantity is trivially $0$, so it suffices to focus on the case of $\lambda > 0$. Recall Eq. \ref{eq:A_inv_formula}: 
$$A^{-1} = A_0^{-1} - \frac{\frac{\gamma}{D} A_0^{-1} Z^T vv^T Z A_0^{-1}}{1 + \frac{\gamma}{D} v^T Z A_0^{-1} Z^T v}.$$
To proceed, we may expand this quantity as
$$\frac{\lambda y^T A^{-1} y}{n} = \frac{1}{n/D} \what^T \left( \frac{\lambda}{D} X A^{-1} X^T\right) \what + \frac{2\lambda}{n} \epsilon^T A^{-1} X^T \what + \frac{\lambda}{n} \epsilon^T A^{-1} \epsilon $$
$$ = \frac{\lambda}{n/D} \what^T M_1 \what + 2\lambda \frac{\epsilon^T A^{-1} X^T \what}{n} + \lambda \frac{\epsilon^T A^{-1} \epsilon}{n} $$
$$\asconverge \frac{\lambda}{\psi}(\alim + \blim \rho^2) + \lambda \sigma^2 g(-\lambda).$$
Here, we have noted that for any fixed $\lambda > 0$, $\what^T M_1 \what \asconverge \alim + \blim \rho^2$, $\frac{\epsilon^T A^{-1} X^T \what}{n} \asconverge 0$, and $\frac{\epsilon^T A^{-1} \epsilon}{n} \asconverge \sigma^2 g(-\lambda)$, and that each of these quantities is moment controlled (Cor. \ref{cor:A_inv_MP_quantities}, Cor. \ref{cor:M_1_M_2_expectation_terms}). Thus, $\frac{\lambda y^T A^{-1} y}{n}$ is moment controlled as well.

Finally, we turn to $\frac{D}{n^2}y^T K_x y$. Recalling that $X = (I_D + \zeta vv^T) Z$, $I_D + \gamma vv^T = (I_D + \zeta vv^T)^2$, and $K_x = K_0 + \frac{\gamma}{D} Z^T vv^T Z$ (where $K_0 = \frac{Z^T Z}{D}$) yields
$$\frac{D}{n^2}y^T K_x y = \frac{D}{n^2} \left( X^T \what + \epsilon \right)^T \frac{X^T X}{D} \left( X^T \what + \epsilon \right)$$
$$ = \what^T \left( \frac1n XX^T \right)^2 \what + \frac{2}{(n/D)^2} \frac{\epsilon^T \left( \frac{X^T X}{D} \right) X^T \what}{D} + \frac{1}{n/D} \frac{\epsilon^T \left( \frac{X^T X}{D} \right) \epsilon}{n}.$$
Now, we make the following convenient definitions:
$$S_1 := \what^T \left( \frac1n XX^T \right)^2 \what $$
$$S_2 := \frac{2}{(n/D)^2} \frac{\epsilon^T \left( \frac{X^T X}{D} \right) X^T \what}{D} $$
$$S_3 := \frac{1}{n/D} \frac{\epsilon^T \left( \frac{X^T X}{D} \right) \epsilon}{n}. $$
We simplify each of these terms, one by one, each time using Cor. \ref{cor:list_of_MP_quantities_lambda_positive} to simplify intermediate quantities and show that they are moment controlled. First,
\begin{align*}
    S_1 = \what^T \Sigma^{\frac12} \frac{ZZ^T}{n} \Sigma \frac{ZZ^T}{n} \Sigma^{\frac12} \what \\
    = \what^T (I_D + \zeta vv^T) \frac{ZZ^T}{n} (I_D + \gamma vv^T) \frac{ZZ^T}{n} (I_D + \zeta vv^T) \what \\
\end{align*}
\begin{align*}
    = (\what + \zeta \rho v)^T \left( \left(\frac1n ZZ^T \right)^2 + \gamma \left(\frac1n ZZ^T \right) vv^T \left( \frac1n ZZ^T \right) \right) (\what + \zeta \rho v)  \\
\end{align*}
\begin{align*}
    = \what^T \left( \frac1n ZZ^T \right)^2 \what + 2\zeta \rho \what^T \left( \frac1n ZZ^T \right)^2 v + \zeta^2 \rho^2 v^T \left( \frac1n ZZ^T \right)^2 v \\
    + \gamma \left( \what^T \left( \frac1n ZZ^T \right) v \right)^2 + 2\gamma \zeta \rho \left(  \what^T \left( \frac1n ZZ^T \right) v \right) \left( v^T \left( \frac1n ZZ^T \right) v \right) \\
    + \gamma \zeta^2 \rho^2 \left( v^T \left( \frac1n ZZ^T \right) v \right)^2
\end{align*}
\begin{align*}
    \asconverge \left(1 + \frac{1}{\psi}\right) + 2\zeta \rho \left(\rho \cdot \left(1 + \frac{1}{\psi}\right)\right) + \zeta^2 \rho^2 \left(1 + \frac{1}{\psi}\right) \\
    + \gamma \rho^2 + 2\zeta \gamma \rho (\rho) (1) + \zeta^2 \gamma \rho^2 (1)^2 
\end{align*}
\begin{align*}
    = \left(1 + \left((\zeta + 1)^2 - 1\right)\rho^2 \right) \left(1 + \frac{1}{\psi}\right) + \gamma \rho^2 \left(1+\zeta  \right)^2 \\
    = (1 + \gamma \rho^2) \left(1 + \frac{1}{\psi}\right) + \gamma (\gamma + 1)\rho^2
\end{align*}
\begin{equation}    \label{eq:y_Kx_y_term1}
    = 1 + \frac{1}{\psi} + \gamma\rho^2 \left(2 + \gamma + \frac{1}{\psi} \right).
\end{equation}
Here, we have noted that, for any $u_1, u_2 \in \{\what, v\}$, $u_1^T \left(\frac1n ZZ^T \right)^j u_2 \asconverge (u_1^T u_2) \int x^j \MP^{1/\psi}(dx)$ (Cor. \ref{cor:list_of_MP_quantities_lambda_positive}), where $\MP^{1/\psi}$ refers to the limiting MP law for the Wishart matrix $\frac1n ZZ^T$. When $j = 1$, we note that $\int x \MP^{1/\psi}(dx) = 1$; when $j = 2$, we note that $\int x^2 \MP^{1/\psi}(dx) = 1 + \frac{1}{\psi}$ (Prop. \ref{prop:properties_of_MP_and_Stieltjes}).

Next, we turn to $S_2$:
$$S_2 = \frac{2}{(n/D)^2} \frac{\epsilon^T \left( \frac{Z^T \Sigma Z}{D} \right) Z^T \Sigma^{\frac12} \what}{D} =  \frac{2}{(n/D)^2} \frac{\epsilon^T \left(K_0 + \frac{\gamma}{D} Z^T v v^T Z \right) Z^T (I_D + \zeta vv^T) \what}{D}$$
\begin{align}   \label{eq:y_Kx_y_term2}
    = \frac{2}{(n/D)^2} \frac{\epsilon^T K_0 Z^T \what}{D} + \frac{2 \zeta \rho}{(n/D)^2} \frac{\epsilon^T K_0 Z^T v}{D} + \frac{2 \gamma}{n/D} \frac{\epsilon^T Z^T v}{D} \cdot \frac{v^T ZZ^T \what}{n} + \frac{2 \gamma \zeta \rho}{n/D} \frac{\epsilon^T Z^T v}{D} \cdot \frac{v^T ZZ^T v}{n} \asconverge 0
\end{align}
where we have noted that $n/D \to \psi$, $\frac{\epsilon^T K_0 Z^T \what}{D} \asconverge 0$, $\frac{\epsilon^T K_0 Z^T v}{D} \asconverge 0$, $\frac{\epsilon^T Z^T v}{D} \asconverge 0$, $\frac{v^T ZZ^T v}{n} \asconverge 1$, and $\frac{v^T ZZ^T \what}{n} \asconverge \rho$, and that each of these quantities is moment controlled (Cor. \ref{cor:list_of_MP_quantities_lambda_positive}). Finally, we turn to $S_3$:
\begin{align}   \label{eq:y_Kx_y_term3}
    S_3 = \frac{1}{n/D} \frac{\epsilon^T \left( \frac{Z^T \Sigma Z}{D} \right) \epsilon}{n} = \frac{1}{n/D} \frac{\epsilon^T \left( \frac{Z^T (I_D + \gamma vv^T) Z}{D} \right) \epsilon}{n} = \frac{1}{n/D} \frac{\epsilon^T \left( \frac{Z^T Z}{D} \right) \epsilon}{n} + \frac{\gamma}{(n/D)^2} \left(\frac{\epsilon^T Z^T v}{D}\right)^2 \asconverge \frac{\sigma^2}{\psi}
\end{align}
where we have used the facts that $\frac{\epsilon^T \left( \frac{Z^T Z}{D} \right) \epsilon}{n} \asconverge \sigma^2$ and $\frac{\epsilon^T Z^T v}{D} \asconverge 0$, and that these quantities are moment controlled (Cor. \ref{cor:list_of_MP_quantities_lambda_positive}).
Altogether, combining Eqs. \ref{eq:y_Kx_y_term1}, \ref{eq:y_Kx_y_term2}, and \ref{eq:y_Kx_y_term3}, we find that
$$\frac{D}{n^2} y^T K_x y = S_1 + S_2 + S_3 \asconverge 1 + \frac{1+\sigma^2}{\psi} + \gamma\rho^2\left(2 + \gamma + \frac{1}{\psi} \right).$$

\end{proof}

\begin{prop}   \label{prop:bcl_properties}
    Let the constant $\bcl$ be as defined in Theorem \ref{thm:inflation_constant_extended}. Then, the numerator term
    \begin{equation}    \label{eq:bcl_numerator_positive}
        \frac{\lambda}{\psi} \bkl \left(1 + \sigma^2 + \gamma \rho^2 - \lambda \left( \frac{\alim + \blim \rho^2}{\psi} + \sigma^2 g(-\lambda) \right) \right) \geq 0
    \end{equation}
    with strict inequality if and only if both $\lambda > 0$ and $\bkl > 0$. In particular, $\bcl \geq 1$ with equality if and only if either $\lambda = 0$ or $\bkl = 0$.
    
    Next, the denominator term
    \begin{equation}    \label{eq:bcl_denominator_positive}
        1 + \bkl \left( 1 + \frac{1+\sigma^2}{\psi} + \gamma \rho^2 \left( 2 + \gamma + \frac{1}{\psi} \right) - \frac{\lambda}{\psi}\left( 1 + \sigma^2 + \gamma \rho^2 \right) + \frac{\lambda^2}{\psi} \left( \frac{\alim + \blim \rho^2}{\psi} + \sigma^2 g(-\lambda) \right) \right) \geq 1.
    \end{equation}

    As a consequence, 
    \begin{equation}
        \left. \frac{\partial}{\partial \bkl} \right. \bcl \geq 0
    \end{equation}
    for any $\bkl \geq 0$, with strict inequality if and only if $\lambda > 0$. Moreover,  
    \begin{equation}
        \left. \frac{\partial}{\partial \lambda} \right|_{\lambda = 0} \bcl \geq 0
    \end{equation}
    with strict inequality if and only if $\bkl > 0$.
    
\end{prop}

\begin{proof}
First, granted that Inequalities \ref{eq:bcl_numerator_positive} and \ref{eq:bcl_denominator_positive} hold, note that
   $$\left. \frac{\partial}{\partial \bkl} \right. \bcl = \frac{\frac{\lambda}{\psi} \left( 1 + \sigma^2 + \gamma \rho^2 - \lambda \left( \frac{\alim + \blim \rho^2}{\psi} + \sigma^2 g(-\lambda) \right)\right)}{\left( 1 + \bkl \left( 1 + \frac{1+\sigma^2}{\psi} + \gamma \rho^2 \left( 2 + \gamma + \frac{1}{\psi} \right) - \frac{\lambda}{\psi}\left( 1 + \sigma^2 + \gamma \rho^2 \right) + \frac{\lambda^2}{\psi} \left( \frac{\alim + \blim \rho^2}{\psi} + \sigma^2 g(-\lambda) \right) \right)  \right)^2} \geq 0$$ for any $\bkl \geq 0$ with strict inequality if and only if $\lambda > 0$. Similarly,
    $$\left. \frac{\partial}{\partial \lambda} \right|_{\lambda = 0} \bcl = \frac{\frac{1}{\psi} \bkl \left( 1 + \sigma^2 + \gamma \rho^2  \right)}{1 + \bkl \left( 1 + \frac{1+\sigma^2}{\psi} + \gamma \rho^2 \left( 2 + \gamma + \frac{1}{\psi} \right) \right)} \geq 0$$
    with strict inequality if and only if $\bkl > 0$.
    Thus, we turn to proving Inequalities \ref{eq:bcl_numerator_positive} and \ref{eq:bcl_denominator_positive}.

    To see why Inequality \ref{eq:bcl_numerator_positive} holds,
    note that when $\lambda > 0$,  $$\frac{\lambda \alim}{\psi} = \lambda (1 - \lambda g(-\lambda)) = \lambda \left(1 - \lambda \int \frac{1}{x+\lambda} \MP^{\psi}(dx) \right) = \int x \cdot \frac{\lambda}{x+\lambda} \MP^{\psi}(dx)$$
$$ < \int x \MP^{\psi}(dx) = 1,$$
$$\frac{\lambda \blim \rho^2}{\psi} = \frac{\lambda \left( \frac{\gamma \tau_1 (1 - \tau_1)}{1 + \gamma \tau_1} \right) \rho^2}{\psi} \leq \frac{\gamma \lambda \tau_1 \rho^2}{\psi}$$ $$\leq \frac{\gamma \lambda  \psi \left(1 - \lambda \int \frac{1}{x+\lambda} \MP^{\psi}(dx) \right) \rho^2}{\psi} = \left( \int x \cdot \frac{\lambda}{x+\lambda} \MP^{\psi}(dx) \right) \gamma \rho^2$$ 
$$ \leq \left( \int x \MP^{\psi}(dx) \right) \gamma \rho^2 = \gamma \rho^2, \text{ and }$$
$$\sigma^2 \lambda g(-\lambda) = \sigma^2 \int 
\frac{\lambda}{x+\lambda} \MP^{\psi}(dx) \leq \sigma^2.$$
Here, we have used the fact that $$\tau_1 = \psi\left( 1 - \lambda \int \frac{1}{x+\lambda} \MP^{\psi}(dx) \right) = \psi \int \frac{x}{x+\lambda} \MP^{\psi}(dx) \geq 0$$ and (by extension) that $1 - \tau_1 \leq 1$. Combining the above inequalities yields 
$$\frac{\lambda}{\psi} \bkl \left(1 + \sigma^2 + \gamma \rho^2 - \lambda \left( \frac{\alim + \blim \rho^2}{\psi} + \sigma^2 g(-\lambda) \right) \right) \geq 0$$
with strict inequality precisely when both $\lambda > 0$ and $\bkl > 0$.

To see why Inequality \ref{eq:bcl_denominator_positive} holds, observe that, by Eq. \ref{eq:cl_expanded_expression} and Lemma \ref{lemma:bcl_intermediate_quantities_converge}, the denominator of $\bcl$ is the almost sure limit of 
$$1 + k_\ell \cdot \frac{D}{n} \cdot \frac1n y^T K_x A^{-1} K_x y = 1 + k_\ell \left( \frac{D}{n^2} y^T K_x y - \lambda \frac{D}{n} \frac{y^T y}{n} + \lambda^2 \frac{D}{n} \frac{y^T A^{-1} y}{n} \right)$$
as $n, D \to \infty$ in the proportional limit. Since $\frac{D}{n} \cdot \frac1n y^T K_x A^{-1} K_x y \geq 0$ (deterministically, as $K_x A^{-1} K_x$ is symmetric positive semi-definite), it follows that its almost sure limit is also nonnegative almost surely. Furthermore, since the limit is a deterministic constant, it follows that this limiting constant is nonnegative. As a result, 
$$1 + \frac{1+\sigma^2}{\psi} + \gamma \rho^2 \left( 2 + \gamma + \frac{1}{\psi} \right) - \frac{\lambda}{\psi}\left( 1 + \sigma^2 + \gamma \rho^2 \right) + \frac{\lambda^2}{\psi} \left( \frac{\alim + \blim \rho^2}{\psi} + \sigma^2 g(-\lambda) \right) \geq 0$$
and we conclude that
$$1 + \bkl \left( 1 + \frac{1+\sigma^2}{\psi} + \gamma \rho^2 \left( 2 + \gamma + \frac{1}{\psi} \right) - \frac{\lambda}{\psi}\left( 1 + \sigma^2 + \gamma \rho^2 \right) + \frac{\lambda^2}{\psi} \left( \frac{\alim + \blim \rho^2}{\psi} + \sigma^2 g(-\lambda) \right) \right) \geq 1.$$

\end{proof}

\begin{proof}[Proof of Theorem \ref{thm:inflation_constant_extended}]

To compute the almost sure limit of $\cl$ and to show that it is moment controlled, we invoke Lemma \ref{lemma:bcl_intermediate_quantities_converge}.
Indeed, using our results from Eqs. \ref{eq:pure_y_term}, \ref{eq:y_Ainv_y_term}, and \ref{eq:y_Kx_y_term}, we find that  
$$\cl = 1 + \frac{\lambda \frac{D}{n} k_\ell \left( \frac1n y^T y - \lambda \frac1n y^T A^{-1} y \right)}{1 + k_\ell \left( \frac{D}{n^2} y^T K_x y - \lambda \frac{D}{n} \frac{y^T y}{n} + \lambda^2 \frac{D}{n} \frac{y^T A^{-1} y}{n} \right)}$$
has an almost sure limit and is itself moment controlled. To see why, first note that by Lemma \ref{lemma:bcl_intermediate_quantities_converge} and because $ \lim_{n, D \uparrow \infty} k_\ell = \bkl$ is a fixed constant (independent of $D$ and $n$), 
the terms in the numerator and denominator of the above expression are moment controlled, and the denominator is
$$1 + k_\ell \left( \frac{D}{n^2}y^T K_x y - \frac{\lambda}{n/D} \frac1n y^T y + \frac{\lambda^2}{n/D} \frac1n y^T A^{-1} y \right) = 1 + k_\ell \frac{D}{n^2} y^T K_x A^{-1} K_x y \geq 1$$
meaning that it is quotient controlled. Since the numerator is moment controlled, it follows that $\cl$ is moment controlled (Prop. \ref{prop:uniform_Lp_control}). Moreover, because the denominator term is bounded below by $1$, $\cl$ has a well-defined almost sure limit that can be found by substituting in the almost sure limits for $\frac1n y^T y$, $\frac{\lambda}{n} y^T A^{-1} y$, and $\frac{D}{n^2} y^T K_x y$, and the limits for $k_\ell$ and $\frac{n}{D}$ (e.g. by Prop. \ref{prop:as_p_convergence}):

\begin{equation}        \label{eq:cl_convergence}
    \cl \asconverge 1 + \frac{\frac{\lambda}{\psi} \bkl\left( 1 + \sigma^2 + \gamma \rho^2 - \lambda \left( \frac{\alim + \blim \rho^2}{\psi} + \sigma^2 g(-\lambda) \right) \right)}{1 + \bkl \left( 1 + \frac{1+\sigma^2}{\psi} + \gamma \rho^2 \left( 2 + \gamma + \frac{1}{\psi} \right) - \frac{\lambda}{\psi}\left( 1 + \sigma^2 + \gamma \rho^2 \right) + \frac{\lambda^2}{\psi} \left( \frac{\alim + \blim \rho^2}{\psi} + \sigma^2 g(-\lambda) \right) \right)}.
\end{equation}

In particular, if $k_\ell = o_D(1)$, then $k_\ell \ndlim \bkl = 0$, so substituting $\bkl = 0$ into Eq. \ref{eq:cl_convergence} yields $\cl \asconverge 1$.

Finally, Prop. \ref{prop:bcl_properties} implies that $\bcl \geq 1$, with strict inequality if and only if both $\bkl > 0$ and $\lambda > 0$.

\end{proof}

Across either scaling regime of $k_\ell$, defining $\bcl$ as the (deterministic) almost sure limit of $\cl$ and noting that $\cl$ is moment controlled, Cor. \ref{cor:Lp_convergence_Vitali} implies that $\EE[\cl] \to \bcl$. More generally, Cor. \ref{cor:expectation_factorizes_Lp} implies that if $Y$ is a random variable that is moment controlled as $n, D \to \infty$ and with $Y \asconverge y$, where $y$ is a nonrandom constant, then $\EE[\cl Y] \ndlim \bcl y$. Furthermore, Theorem \ref{thm:inflation_constant_extended} implies that when $\eta^2 \ell^3 \ll D$, the one-step feature learning estimator $\hat{f}_{\feat}$ coincides with that of the baseline estimator $\hat{f}_{\init}$ in the asymptotic limit. Finally, we establish the following useful corollary:

\begin{cor} \label{cor:bcl_upper_bound}
    Under the same hypotheses as in Theorem \ref{thm:inflation_constant}, $$\bcl \leq 1 + \bkl \left(1 + \frac{1+\sigma^2}{\psi} + \gamma \rho^2 \left(2 + \gamma + \frac{1}{\psi} \right) \right)$$
    across all $\lambda \geq 0$. In particular, $\bcl$ is $O_\lambda(1)$.
\end{cor}

\begin{proof}
    First, at finite $n$ and $D$, 
    $$\cl = 1 + \frac{\lambda \kl \frac{D}{n} \frac1n y^T K_x A^{-1} y}{1 + \kl \frac{D}{n} \frac1n y^T K_x A^{-1} K_x y}$$
    where $\kl = \frac{\eta^2 d_\ell}{D}$.
    Observe that $\lambda K_x A^{-1} = K_x \cdot \lambda (K_x + \lambda I_n)^{-1} \preceq K_x$. As a result, 
    $$\lambda \cdot \frac1n y^T K_x A^{-1} y \leq \frac1n y^T K_x y.$$
    Since $1 + \kl \frac{D}{n} \frac1n y^T K_x A^{-1} K_x y \geq 1$, it follows that
    $$\cl \leq 1 + \kl \cdot \frac{D}{n} \cdot \frac1n y^T K_x y = 1 + \kl \cdot \frac{D}{n^2} y^T K_x y.$$
    In the overparameterized proportional limit, $\cl \asconverge \bcl$ (Thm. \ref{thm:inflation_constant}) and 
    $$\frac{D}{n^2} y^T K_x y \asconverge 1 + \frac{1+\sigma^2}{\psi} + \gamma\rho^2\left(2 + \gamma + \frac{1}{\psi} \right) $$
    using Eq. \ref{eq:y_Kx_y_term}.
    Finally, since $\kl \to \bkl$, it follows that
    $$\bcl \leq 1 + \bkl \left( 1 + \frac{1+\sigma^2}{\psi} + \gamma\rho^2\left(2 + \gamma + \frac{1}{\psi} \right) \right).$$
\end{proof}

\subsection{Calculating bias, variance, and generalization error}   
\label{app:gen_error}

\subsubsection{Proof of Theorem \ref{thm:bias_variance_gen_error}}
\label{app:bias_var_generror_formula_proof}

In this section, we prove Theorem \ref{thm:bias_variance_gen_error}, which establishes closed-form asymptotic formulas for the bias, variance, and generalization error of any estimator $\hat{f}_{\lambda; \kl}$.

Having established results regarding the convergence of each of $a$, $b$, $\tilde{a}$, $\tilde{b}$, as well as the almost sure convergence of $\cl$, we are now ready to derive an asymptotic formula for the generalization error of the one-step feature learning estimator (arising from kernel ridge regression). We may also calculate the generalization error for baseline ridge estimators, which arise from setting $\eta = 0$ --- or equivalently, setting $\cl = 1$.

\begin{lemma}  \label{lemma:bias_variance_decomp}
    Recall the definitions of bias, variance, and generalization error:
    \begin{equation}    \label{eq:bias}
        \mathcal{B}(\hat{f}) = \mathbb{E}_{x \sim \mathcal{N}(0, \Sigma)}\left[ \left( \mathbb{E}_{(X, y) \sim \mathcal{D}}[\hat{f}(x)] - \frac{1}{\sqrt{D}} w_\star^T x \right)^2\right]
    \end{equation}
    \begin{equation}    \label{eq:variance}
        \mathcal{V}(\hat{f}) = \mathbb{E}_{x \sim \mathcal{N}(0, \Sigma)}\left[ \mathbb{E}_{(X, y) \sim \mathcal{D}}\left[ \left( \mathbb{E}_{(X, y) \sim \mathcal{D}}\left[ \hat{f}(x)\right] - 
    \hat{f}(x) \right)^2 \right]\right] 
    \end{equation}
    \begin{equation}    \label{eq:gen_error_SUPP}
        \mathcal{G}(\hat{f}) := \mathbb{E}_{x \sim \mathcal{N}(0, \Sigma)} \left[ \mathbb{E}_{(X, y) \sim \mathcal{D}} \left[ \left( \hat{f}(x) - \frac{1}{\sqrt{D}} w_\star^T x \right)^2 \right] \right]
    \end{equation}

    Then, the identity $\gen(\hat{f}) = \bias(
    \hat{f}) + \variance(\hat{f})$ holds.
\end{lemma}

\begin{proof}
    Observe that 
    $$\left( \hat{f}(x) - \frac{1}{\sqrt{D}} w_\star^T x \right)^2 = \left( \left( \hat{f}(x) - \EE_{(X, y) \sim \mathcal{D}}\left[ \hat{f}(x) \right] \right) + \left( \EE_{(X, y) \sim \mathcal{D}}\left[ \hat{f}(x) \right] - \frac{1}{\sqrt{D}} w_\star^T x \right) \right)^2 $$
    \begin{multline*}
        = \left( \hat{f}(x) - \EE_{(X, y) \sim \mathcal{D}}\left[ \hat{f}(x) \right] \right)^2 + 2 \left( \hat{f}(x) - \EE_{(X, y) \sim \mathcal{D}}\left[ \hat{f}(x) \right] \right) \left( \EE_{(X, y) \sim \mathcal{D}}\left[ \hat{f}(x) \right] - \frac{1}{\sqrt{D}} w_\star^T x \right) \\ + \left( \EE_{(X, y) \sim \mathcal{D}}\left[ \hat{f}(x) \right] - \frac{1}{\sqrt{D}} w_\star^T x \right)^2
    \end{multline*}
    For any fixed $x \in \RR^D$, observe that $ \EE_{(X, y) \sim \mathcal{D}}\left[ \hat{f}(x) \right] - \frac{1}{\sqrt{D}} w_\star^T x$ is a constant. As a result, for any random variable $Y$, the following simplification holds:  
    $$\EE_{(X, y) \sim \mathcal{D}}\left[ \left(  \EE_{(X, y) \sim \mathcal{D}}\left[ \hat{f}(x) \right] - \frac{1}{\sqrt{D}} w_\star^T x \right) Y  \right] = \left(  \EE_{(X, y) \sim \mathcal{D}}\left[ \hat{f}(x) \right] - \frac{1}{\sqrt{D}} w_\star^T x \right) \EE_{(X, y) \sim \mathcal{D}}[Y].$$ 
    This means that
    \begin{align*}
        \EE_{(X, y) \sim \mathcal{D}}\left[ \left( \hat{f}(x) - \frac{1}{\sqrt{D}} w_\star^T x \right)^2  \right] = \EE_{(X, y) \sim \mathcal{D}}\left[ \left( \hat{f}(x) - \EE_{(X, y) \sim \mathcal{D}}\left[ \hat{f}(x) \right] \right)^2 \right] \\ 
        + 2 \left( \EE_{(X, y) \sim \mathcal{D}}\left[ \hat{f}(x) \right] - \frac{1}{\sqrt{D}} w_\star^T x \right) \EE_{(X, y) \sim \mathcal{D}}\left[ \left( \hat{f}(x) - \EE_{(X, y) \sim \mathcal{D}}\left[ \hat{f}(x) \right] \right) \right] \\
        + \left( \EE_{(X, y) \sim \mathcal{D}}\left[ \hat{f}(x) \right] - \frac{1}{\sqrt{D}} w_\star^T x \right)^2 
    \end{align*}
    \begin{align}   \label{eq:gen_error_fixed_test_input}
        = \left( \EE_{(X, y) \sim \mathcal{D}}\left[ \hat{f}(x) \right] - \frac{1}{\sqrt{D}} w_\star^T x \right)^2  + \EE_{(X, y) \sim \mathcal{D}}\left[ \left( \hat{f}(x) - \EE_{(X, y) \sim \mathcal{D}}\left[ \hat{f}(x) \right] \right)^2 \right]
    \end{align}
    where we have used the fact that
    $$\EE_{(X, y) \sim \mathcal{D}}\left[ \left( \hat{f}(x) - \EE_{(X, y) \sim \mathcal{D}}\left[ \hat{f}(x) \right] \right) \right] = 0.$$
    Finally, starting with Eq. \ref{eq:gen_error_fixed_test_input} and taking the expectation over test samples $x \sim \mathcal{N}(0, \Sigma)$ yields the identity $\gen(\hat{f}) = \bias(
    \hat{f}) + \variance(\hat{f})$.
\end{proof}

We now state and prove theorems that establish the asymptotic formulas for the bias and variance. As a convenient shorthand, we simply refer to the estimator $\hat{f}_{\lambda;\kl}$ as $\hat{f}$, making it clear that any mention of $\hat{f}$ will implicitly depend on $\lambda$ and $\kl$. Note that setting $\kl = 0$ means that $\hat{f}$ is a baseline ``no learning'' estimator, and choosing $\kl > 0$ results in $\hat{f}$ being a one-step feature learning estimator.

\begin{thm} \label{thm:asymptotic_bias}
    Recall (from Eq. \ref{eq:f_lambda_kl_defn}) that the one-step feature learning estimator is given by $$\hat{f}(x) = \frac{\cl}{\sqrt{D}} w_\star^T M_1^T x + \frac{\cl}{D} \epsilon^T A^{-1} X^T x.$$ 
Under Assumption \ref{assump:data_gen_process}, we asymptotically obtain the following limiting formula for the bias of $\hat{f}$:
\begin{equation}    \label{eq:bias_limit_SUPP}
 \bias(\hat{f}) \ndlim (1 - \alim \bcl)^2 + \rho^2 \blim \bcl \left(\blim \bcl - 2 (1 - \alim \bcl) \right) + \gamma \rho^2 (1 - (\alim + \blim)\bcl)^2.    
\end{equation}

\end{thm}

\begin{proof}
    Starting with the formula for the estimator bias, given in Eq. \ref{eq:bias}, we have that 
$$\mathcal{B}(\hat{f}) = \mathbb{E}_{x \sim \mathcal{N}(0, \Sigma)}\left[ \left( \mathbb{E}_{(X, y) \sim \mathcal{D}}[\hat{f}(x)] - \frac{1}{\sqrt{D}} w_\star^T x \right)^2\right]$$
$$ = \mathbb{E}_{x}\left[ \frac{1}{D} \left( w_\star^T \EE_{X, \epsilon}\left[ \cl M_1 - I_D \right] + \EE_{X, \epsilon}\left[ \frac{\cl}{\sqrt{D}}\epsilon^T A^{-1}X^T \right] \right) xx^T \left( \EE_{X, \epsilon}\left[ \cl M_1 - I_D \right]  w_\star + \EE_{X, \epsilon}\left[ \frac{\cl}{\sqrt{D}}XA^{-1} \epsilon \right] \right) \right]$$
$$ = \frac{1}{D} \left( w_\star^T \EE_{X, \epsilon}\left[ \cl M_1 - I_D \right] + \EE_{X, \epsilon}\left[ \frac{\cl}{\sqrt{D}}\epsilon^T A^{-1}X^T \right] \right) \Sigma \left( \EE_{X, \epsilon}\left[ \cl M_1 - I_D \right]  w_\star + \EE_{X, \epsilon}\left[ \frac{\cl}{\sqrt{D}}XA^{-1} \epsilon \right] \right) $$
\begin{align*}
    = \frac{1}{D} w_\star^T \EE_{X, \epsilon}\left[ \cl M_1 - I_D \right] \left( I_D + \gamma vv^T \right) \EE_{X, \epsilon}\left[ \cl M_1 - I_D \right]  w_\star \\
    + \frac2D \EE_{X, \epsilon}\left[ \frac{\cl}{\sqrt{D}}\epsilon^T A^{-1}X^T \right] \left( I_D + \gamma vv^T \right) \EE_{X, \epsilon}\left[ \cl M_1 - I_D \right] w_\star \\
    + \frac1D \EE_{X, \epsilon}\left[ \frac{\cl}{\sqrt{D}}\epsilon^T A^{-1}X^T \right] \left( I_D + \gamma vv^T \right) \EE_{X, \epsilon}\left[ \frac{\cl}{\sqrt{D}}XA^{-1} \epsilon \right].
\end{align*}

We now make the following definitions: $$\delta_1 := \frac2D \EE_{X, \epsilon}\left[ \frac{\cl}{\sqrt{D}}\epsilon^T A^{-1}X^T \right] \left( I_D + \gamma vv^T \right) \EE_{X, \epsilon}\left[ \cl M_1 - I_D \right] w_\star$$
$$\delta_2 := \frac1D \EE_{X, \epsilon}\left[ \frac{\cl}{\sqrt{D}}\epsilon^T A^{-1}X^T \right] \left( I_D + \gamma vv^T \right) \EE_{X, \epsilon}\left[ \frac{\cl}{\sqrt{D}}XA^{-1} \epsilon \right].$$
Note that if $\lambda = 0$, then $\cl = 1$ and so $\EE_{X, \epsilon}\left[ \frac{\cl}{\sqrt{D}}\epsilon^T A^{-1}X^T \right] = \EE_{X, \epsilon}\left[ \frac{1}{\sqrt{D}}\epsilon^T A^{-1}X^T \right] = 0$ (Cor. \ref{cor:integrability_of_inverses_lambda_zero}), noting that $\epsilon$ and $X$ are independent. In this case, $\delta_1 = \delta_2 = 0$. To address the case of $\lambda > 0$, let $v_\epsilon := \EE_{X, \epsilon}\left[ \frac{\cl}{D}X A^{-1} \epsilon \right]$. By Jensen's inequality, 
$$\norm{v_\epsilon}_2^2 \leq \EE_{X, \epsilon}\left[ \norm{\frac{\cl}{D}X A^{-1} \epsilon}_2^2 \right] = \EE_{X, \epsilon}\left[ \frac{\cl^2}{D}\epsilon^T A^{-1} \frac{X^T X}{D} A^{-1} \epsilon \right]$$
$$ = \EE_{X, \epsilon}\left[ \frac{\cl^2}{D}\epsilon^T A^{-1} (A - \lambda I_n) A^{-1} \epsilon \right] = \EE_{X, \epsilon}\left[ {\cl^2} \frac{\epsilon^T A^{-1} \epsilon}{D} \right] - \lambda\EE_{X, \epsilon}\left[ {\cl^2} \frac{\epsilon^T A^{-2} \epsilon}{D} \right]$$
\begin{equation}    \label{eq:v_epsilon_bounded}
    \ndlim \psi \bcl^2 \sigma^2 g(-\lambda) - \lambda \psi \bcl^2 \sigma^2 g'(-\lambda) = O_D(1)
\end{equation}
where we have used the fact that $\cl \asconverge \bcl$, $\cl$ is moment controlled (Thm. \ref{thm:inflation_constant_extended}) and that $\frac{\epsilon^T A^{-1} \epsilon}{D} = \frac{\epsilon^T A^{-1} \epsilon}{n} \cdot \frac{n}{D} \asconverge \psi \sigma^2 g(-\lambda)$, $\frac{\epsilon^T A^{-2} \epsilon}{D} = \frac{\epsilon^T A^{-2} \epsilon}{n} \cdot \frac{n}{D} \asconverge \psi \sigma^2 g'(-\lambda)$, and that $\frac{\epsilon^T A^{-1} \epsilon}{D}$ and $\frac{\epsilon^T A^{-2} \epsilon}{D}$ are moment controlled, when $\lambda > 0$ (Cor. \ref{cor:A_inv_MP_quantities}). Let $\hat{v}_\epsilon := \frac{v_\epsilon}{\norm{v_\epsilon}_2}$; by definition, $\hat{v}_\epsilon$ is a fixed unit vector on $\SS^{D-1}$.
Returning to $\delta_1$ and $\delta_2$, expanding terms gives
\begin{align*}
    |\delta_1| = \left| 2\EE_{X, \epsilon}\left[ \frac{\cl}{D}\epsilon^T A^{-1}X^T \right] \EE_{X, \epsilon}\left[(\cl - \bcl)M_1 + \bcl M_1 - I_D \right] \frac{w_\star}{\sqrt{D}} \right. \\
    \left. + 2\gamma \EE_{X, \epsilon}\left[ \frac{\cl}{D}\epsilon^T A^{-1}X^T v \right]  v^T \left(\EE_{X, \epsilon}\left[ \cl M_1 \right] - I_D \right) \frac{w_\star}{\sqrt{D}} \right| \\
\end{align*}
\begin{align*}
    \leq 2 \norm{v_\epsilon}_2 \left| \EE_{X, \epsilon}\left[(\cl - \bcl) \hat{v}_\epsilon^T M_1 \what \right] \right| \\
    + 2 \left| \EE_{X, \epsilon}\left[ \frac{\cl}{D}\epsilon^T A^{-1}X^T \right] \EE_{X, \epsilon}\left[ \bcl M_1 - I_D \right] \what \right| \\
    + 2\gamma \left| \EE_{X, \epsilon}\left[ \frac{\cl}{D}\epsilon^T A^{-1}X^T v \right]  v^T \left(\EE_{X, \epsilon}\left[ \cl M_1 \right] - I_D \right) \what \right| \\
\end{align*}
\begin{align*}
    \leq 2 \norm{v_\epsilon}_2 \EE_{X, \epsilon}\left[\left|\cl - \bcl\right| \cdot \left| \hat{v}_\epsilon^T M_1 \what \right| \right] \\
    + 2 \left| \EE_{X, \epsilon}\left[ \frac{\cl}{D}\epsilon^T A^{-1}X^T \right] \left( \bcl \left(aI_D + b vv^T \right) - I_D \right) \what \right| \\
    + 2\gamma \left| \EE_{X, \epsilon}\left[ \frac{\cl}{D}\epsilon^T A^{-1}X^T v \right]  \left(\EE_{X, \epsilon}\left[ \cl v^T M_1 \what \right] - \rho \right) \right|  \\
\end{align*}
\begin{align*}
    \leq 2 \norm{v_\epsilon}_2 \EE_{X, \epsilon}\left[|\cl - \bcl| \opnorm{M_1} \right] \\
    + 2 \left|a\bcl - 1\right| \cdot \left|\EE_{X, \epsilon}\left[ \cl \frac{\epsilon^T A^{-1}X^T \what}{D} \right] \right|  \\ 
    + 2 |\rho| |b| \bcl \left| \EE_{X, \epsilon}\left[ \cl \frac{\epsilon^T A^{-1}X^T v}{D} \right] \right| \\
    + 2\gamma \left| \EE_{X, \epsilon}\left[ \cl \frac{\epsilon^T A^{-1}X^T v}{D} \right] \right| \cdot \left(\EE_{X, \epsilon}\left[ \cl \opnorm{M_1} \right] + |\rho| \right)  \\
\end{align*}
\begin{align*}
    \lesssim \EE_{X, \epsilon}\left[|\cl - \bcl| \right] 
    + \EE_{X, \epsilon}\left[ \cl \left| \frac{\epsilon^T A^{-1}X^T \what}{D} \right| \right]  \\ 
    + \EE_{X, \epsilon}\left[  \cl \left| \frac{\epsilon^T A^{-1}X^T v}{D} \right| \right]
    + \EE_{X, \epsilon}\left[ \cl \left| \frac{\epsilon^T A^{-1}X^T v}{D} \right| \right]  \left(\EE_{X, \epsilon}\left[ |\cl| \right] + |\rho| \right)  \\
    \overset{n, D \nearrow \infty}{\longrightarrow} 0.
\end{align*}
Here, we have used Eq. \ref{eq:v_epsilon_bounded} along with the facts that $\opnorm{M_1} \leq 1$, $\cl \asconverge \bcl$ (Thm. \ref{thm:inflation_constant_extended}), $\frac{\epsilon^T A^{-1}X^T \what}{D} \asconverge 0$, and $\frac{\epsilon^T A^{-1}X^T v}{D} \asconverge 0$, and that each of $\cl$, $\frac{\epsilon^T A^{-1}X^T \what}{D}$, and $\frac{\epsilon^T A^{-1}X^T v}{D}$ is moment controlled when $\lambda > 0$ (Cor. \ref{cor:A_inv_MP_quantities}). Next,
\begin{align*}
    0 \leq \delta_2 = \EE_{X, \epsilon}\left[ \frac{\cl}{D}\epsilon^T A^{-1}X^T \right] \left( I_D + \gamma vv^T \right) \EE_{X, \epsilon}\left[ \frac{\cl}{D}XA^{-1} \epsilon \right] \\
    = \norm{ \EE_{X, \epsilon}\left[ \frac{\cl}{D}XA^{-1} \epsilon \right] }_2^2 + \gamma \EE_{X, \epsilon}\left[ \cl \frac{\epsilon^T A^{-1}X^T v}{D} \right]^2 \\
    = \norm{\EE_{X, \epsilon}\left[ \frac{\bcl}{D} XA^{-1} \epsilon \right] + \EE_{X, \epsilon}\left[ \frac{\cl - \bcl}{D} XA^{-1} \epsilon \right]}_2^2 + \gamma \EE_{X, \epsilon}\left[ \cl  \frac{\epsilon^T A^{-1}X^T v}{D}  \right]^2 \\
    = \norm{\EE_{X, \epsilon}\left[ \frac{\cl - \bcl}{D} XA^{-1} \epsilon \right]}_2^2 + \gamma \EE_{X, \epsilon}\left[ \cl  \frac{\epsilon^T A^{-1}X^T v}{D}  \right]^2 \\
\end{align*}
\begin{align*}
    \leq \EE_{X, \epsilon}\left[ \frac{(\cl - \bcl)^2}{D^2} \norm{XA^{-1} \epsilon}_2^2 \right] + \gamma \EE_{X, \epsilon}\left[ \cl \left| \frac{\epsilon^T A^{-1}X^T v}{D} \right| \right]^2 \\
    = \EE_{X, \epsilon}\left[ \frac{(\cl - \bcl)^2}{D} \epsilon^T A^{-1} \frac{X^T X}{D} A^{-1} \epsilon \right] + \gamma \EE_{X, \epsilon}\left[ \cl \left| \frac{\epsilon^T A^{-1}X^T v}{D} \right| \right]^2 \\
    = \EE_{X, \epsilon}\left[ (\cl - \bcl)^2 \frac{\epsilon^T A^{-1} (A - \lambda I_n) A^{-1} \epsilon}{D} \right] + \gamma \EE_{X, \epsilon}\left[ \cl \left| \frac{\epsilon^T A^{-1}X^T v}{D} \right| \right]^2 \\
    = \EE_{X, \epsilon}\left[ (\cl - \bcl)^2 \frac{\epsilon^T A^{-1} \epsilon}{D} \right] - \lambda \EE_{X, \epsilon}\left[ (\cl - \bcl)^2 \frac{\epsilon^T A^{-2} \epsilon}{D} \right]  + \gamma \EE_{X, \epsilon}\left[ \cl \left| \frac{\epsilon^T A^{-1}X^T v}{D} \right| \right]^2 \\
    \overset{n, D \nearrow \infty}{\longrightarrow} 0.
\end{align*}
Here, we have used the facts that $\cl \asconverge \bcl$ (Thm. \ref{thm:inflation_constant_extended}), $\frac{\epsilon^T A^{-1} \epsilon}{n} \asconverge \sigma^2 g(-\lambda)$, $\frac{\epsilon^T A^{-2} \epsilon}{n} \asconverge \sigma^2 g'(-\lambda)$, $\frac{\epsilon^T A^{-1}X^T v}{D} \asconverge 0$, and that for $\lambda > 0$ it is the case that each of these quantities is moment controlled (Cor. \ref{cor:A_inv_MP_quantities}). We have also used the fact that $\EE_{X, \epsilon}\left[ \frac{\bcl}{D}XA^{-1} \epsilon \right] = \bcl \EE_{X}\left[ \frac{1}{D}XA^{-1} \right] \EE_{\epsilon}\left[ \epsilon \right]  = 0$, which relies (in part) on the fact that $\opnorm{\frac{1}{\sqrt{D}} A^{-1} X^T} = \opnorm{M_2}^{1/2} \leq \lambda^{-1/2}$ when $\lambda > 0$. Thus,
$$\mathcal{B}(\hat{f}) = \frac{1}{D} w_\star^T \EE_{X, \epsilon}\left[ \cl M_1 - I_D \right] \left( I_D + \gamma vv^T \right) \EE_{X, \epsilon}\left[ \cl M_1 - I_D \right]  w_\star + o_D(1)$$
$$ = \frac1D w_\star^T \EE_{X, \epsilon}\left[ I_D - \cl M_1 \right]^2 w_\star + \gamma \left( \frac{1}{\sqrt{D}}w_\star^T \left( \EE_{X, \epsilon}\left[ I_D - \cl M_1 \right] \right) v \right)^2 + o_D(1) $$
$$ = \frac1D w_\star^T \EE_{X, \epsilon}\left[ I_D - \cl M_1 \right]^2 w_\star + \gamma \left( \frac{1}{\sqrt{D}}w_\star^T \left( I_D - \EE_{X, \epsilon}\left[ \cl M_1 \right] \right) v \right)^2 + o_D(1) $$
$$ = \frac1D w_\star^T \left( \EE_{X}\left[ I_D - \bcl M_1 \right] - \EE_{X, \epsilon}\left[ (\cl - \bcl) M_1 \right] \right)^2 w_\star + \gamma \left( \frac{1}{\sqrt{D}} \left( w_\star^T v - w_\star^T\EE_{X, \epsilon}\left[ \cl M_1 \right] v \right) \right)^2 + o_D(1)$$
\begin{equation}
\label{eq:bias_with_zeroing_terms}
\begin{aligned}
     = \frac1D w_\star^T \left( I_D - \bcl \EE_{X}[M_1] \right)^2 w_\star
     - \frac{2}{D} w_\star^T \left( I_D - \bcl \EE_{X}[M_1] \right)
     \left( \EE_{X, \epsilon}[(\cl - \bcl) M_1] \right) w_\star \\ 
     + \frac{1}{D}w_\star^T \EE_{X, \epsilon}[(\cl - \bcl) M_1]^2 w_\star
     + \gamma \left( \frac{1}{\sqrt{D}}
     \left( w_\star^T v - \bcl w_\star^T\EE_{X}\left[ M_1 \right] v
     - w_\star^T \EE_{X, \epsilon}[(\cl - \bcl) M_1] v \right) \right)^2
     + o_D(1).
\end{aligned}
\end{equation}

We now demonstrate that terms in Eq. \ref{eq:bias_with_zeroing_terms} involving $\cl - \bcl$ go to zero as $n, D \to \infty$ in the proportional regime. This certainly holds when $\lambda = 0$ because $\cl = 1 = \bcl$ in this case. When $\lambda > 0$, we first have that 
$$\left| \frac{1}{D} w_\star^T \left( I_D - \bcl \EE_{X}[M_1] \right) \left( \EE_{X, \epsilon}[(\cl - \bcl) M_1] \right) w_\star\right| = \left| \what^T \left( (1 - \bcl a) I_D - b \bcl vv^T \right) \EE_{X, \epsilon}[(\cl - \bcl) M_1] \what \right| $$
$$ = \left| (1 - \bcl a) \EE_{X, \epsilon}\left[(\cl - \bcl) \what^T M_1 \what\right] - \rho b \bcl \EE_{X, \epsilon}\left[(\cl - \bcl) \what^T M_1 v\right] \right| $$
$$\leq \left|1 - \bcl a\right| \EE_{X, \epsilon}\left[|\cl - \bcl| \opnorm{M_1}\right] + |\rho| |b| \bcl \EE_{X, \epsilon}\left[|\cl - \bcl| \opnorm{M_1}\right]$$
$$ = \left|1 - \bcl a\right| \EE_{X, \epsilon}\left[|\cl - \bcl| \right] + |\rho| |b| \bcl \EE_{X, \epsilon}\left[|\cl - \bcl| \right] \overset{n, D \nearrow \infty}{\longrightarrow} 0.$$
Here, we have used the facts that $\opnorm{M_1} \leq 1$ (for $n < D$), $\cl \asconverge \bcl$, and that $\cl$ is moment controlled (Thm. \ref{thm:inflation_constant_extended}). Consequently, Cor. \ref{cor:Lp_convergence_Vitali} implies that $\EE_{X, \epsilon}[|\cl - \bcl|] \to 0$. Next,
$$\left| \frac1D \ws^T \EE_{X, \epsilon}[(\cl - \bcl)M_1]^2 \ws \right| = \norm{\EE_{X, \epsilon}[(\cl - \bcl)M_1 \what]}_2^2$$
$$\leq \EE_{X, \epsilon}[\norm{(\cl - \bcl)M_1 \what}_2^2] = \EE_{X, \epsilon}[(\cl - \bcl)^2 \what^T M_1^2 \what ] $$
$$ \leq \EE_{X, \epsilon}\left[|\cl - \bcl|^2 \opnorm{M_1^2} \right] \leq \EE_{X, \epsilon}\left[|\cl - \bcl|^2 \opnorm{M_1}^2 \right]$$
$$\leq \EE_{X, \epsilon}\left[|\cl - \bcl|^2 \right] \overset{n, D \nearrow \infty}{\longrightarrow} 0.$$
This latter convergence statement again follows from Cor. \ref{cor:Lp_convergence_Vitali}, noting that $|\cl - \bcl|^2 \asconverge 0$ and that $|\cl - \bcl|^2$ is moment controlled. Finally,
$$\left| \frac{1}{\sqrt{D}}\ws^T \EE_{X, \epsilon}[(\cl - \bcl) M_1] v \right| = \left| \EE_{X, \epsilon}[(\cl - \bcl) \what^T M_1 v] \right| \leq \EE_{X, \epsilon}\left[|\cl - \bcl| \opnorm{M_1}\right]$$
$$\leq \EE\left[|\cl - \bcl| \right] \overset{n, D \nearrow \infty}{\longrightarrow} 0$$
where we have again used Cor. \ref{cor:Lp_convergence_Vitali}. Thus, in the proportional asymptotic limit of $n, D \to \infty$, we find that the bias converges to a nonrandom constant (using Cor. \ref{cor:a_b_atilde_btilde}):
$$\mathcal{B}(\hat{f}) = \frac1D w_\star^T \left( I_D - \bcl \EE_{X}[M_1] \right)^2 w_\star + \frac{\gamma}{D}\left( \ws^T v - \bcl \ws^T \EE_{X}[M_1] v \right)^2 + o_D(1)$$
$$ \ndlim \what^T \left( I_D - \bcl (\alim I_D + \blim vv^T) \right)^2 \what + \gamma \left( \what^T v - \bcl \what^T (\alim I_D + \blim vv^T) v \right)^2$$
$$ = \what^T \left( (1 - \alim \bcl)^2 I_D + (\blim^2 \bcl^2 - 2\blim \bcl (1 - \alim \bcl)) vv^T \right) \what + \gamma \left( \rho - \alim \bcl \rho - \blim \bcl \rho \right)^2$$
\begin{equation}    \label{eq:final_bias_formula}
= (1 - \alim \bcl)^2 + \rho^2 \blim \bcl \left(\blim \bcl - 2 (1 - \alim \bcl) \right) + \gamma \rho^2 (1 - (\alim + \blim)\bcl)^2.
\end{equation}
\end{proof}

\begin{thm} \label{thm:asymptotic_variance}
    Recall that the one-step feature learning estimator is given by $$\hat{f}(x) = \frac{\cl}{\sqrt{D}} w_\star^T M_1^T x + \frac{\cl}{D} \epsilon^T A^{-1} X^T x.$$ 
Under Assumption \ref{assump:data_gen_process}, we asymptotically obtain the following limiting formulas for the variance of $\hat{f}$:
\begin{equation}    \label{eq:variance_limit_SUPP}
    \variance(\hat{f}) \ndlim \bcl^2 \left( \alim + \blim \rho^2 - \lambda \left(\atlim + \btlim \rho^2 \right) - \alim^2 - \rho^2 \left( 2\alim \blim + \blim^2 \right) + \sigma^2 \atlim \right).
\end{equation}

\end{thm}

\begin{proof}
    Starting with the formula for the estimator variance, as given by Eq. \ref{eq:variance}, we find that
$$\mathcal{V}(\hat{f}) = \mathbb{E}_{x \sim \mathcal{N}(0, \Sigma)}\left[ \mathbb{E}_{(X, y) \sim \mathcal{D}}\left[ \left( \mathbb{E}_{(X, y) \sim \mathcal{D}}\left[ \hat{f}(x)\right] - \hat{f}(x) \right)^2 \right]\right]$$
$$ = \mathbb{E}_{x \sim \mathcal{N}(0, \Sigma)}\left[ \mathbb{E}_{(X, y) \sim \mathcal{D}}\left[\hat{f}(x)^2 \right] - \mathbb{E}_{(X, y) \sim \mathcal{D}}\left[ \hat{f}(x)\right]^2  \right]$$
\begin{equation}    \label{eq:variance_formula}
    = \mathbb{E}_{X, \epsilon}\left[ \mathbb{E}_{x \sim \mathcal{N}(0, \Sigma)}\left[ \hat{f}(x)^2 \right]  \right] - \mathbb{E}_{x \sim \mathcal{N}(0, \Sigma)}\left[ \EE_{X, \epsilon}\left[ \hat{f}(x) \right]^2 \right].
\end{equation}
Observe that 
$$\EE_{X, \epsilon}\left[ \hat{f}(x) \right] = \EE_{X, \epsilon}\left[ \frac{\cl}{\sqrt{D}} w_\star^T M_1^T x + \frac{\cl}{D} \epsilon^T A^{-1} X^T x \right]$$
$$ = \EE_{X, \epsilon}\left[ \frac{\bcl}{\sqrt{D}} w_\star^T M_1^T x\right] + \EE_{X, \epsilon}\left[ \frac{\cl - \bcl}{\sqrt{D}} w_\star^T M_1^T x\right] + \EE_{X, \epsilon}\left[\frac{\bcl}{D} \epsilon^T A^{-1} X^T x \right] + \EE_{X, \epsilon}\left[\frac{\cl - \bcl}{D} \epsilon^T A^{-1} X^T x \right]$$
$$ = \frac{\bcl}{\sqrt{D}} w_\star^T \EE_{X, \epsilon}\left[M_1 \right]^T x + \EE_{X, \epsilon}\left[ \frac{\cl - \bcl}{\sqrt{D}} w_\star^T M_1^T x\right] + \EE_{X, \epsilon}\left[\frac{\cl - \bcl}{D} \epsilon^T A^{-1} X^T x \right]$$
$$ = \frac{\bcl}{\sqrt{D}} w_\star^T \left(a I_D + bvv^T \right) x + \EE_{X, \epsilon}\left[ \frac{\cl - \bcl}{\sqrt{D}} w_\star^T M_1^T x\right] + \EE_{X, \epsilon}\left[\frac{\cl - \bcl}{D} \epsilon^T A^{-1} X^T x \right].$$
Here, we have used the fact that $\EE_{X, \epsilon}\left[\frac{\bcl}{D} \epsilon^T A^{-1} X^T \right] = \bcl \EE_{\epsilon}[\epsilon]^T \EE_{X}\left[\frac{1}{D} A^{-1} X^T \right] = 0$, which follows from Cor. \ref{cor:integrability_of_inverses_lambda_zero} when $\lambda = 0$ and by the fact that $\opnorm{\frac{1}{\sqrt{D}} A^{-1} X^T} = \opnorm{M_2}^{1/2} \leq \lambda^{-1/2}$ when $\lambda > 0$ (Eq. \ref{eq:M2_opnorm_lambda_positive}). As a result,
\begin{multline*}
    \mathbb{E}_{x \sim \mathcal{N}(0, \Sigma)}\left[ \EE_{X, \epsilon}\left[ \hat{f}(x) \right]^2 \right] = \\ \mathbb{E}_{x \sim \mathcal{N}(0, \Sigma)}\left[ \left( \frac{\bcl}{\sqrt{D}} w_\star^T \left(a I_D + bvv^T \right) x + \EE_{X, \epsilon}\left[ \frac{\cl - \bcl}{\sqrt{D}} w_\star^T M_1^T x\right] + \EE_{X, \epsilon}\left[\frac{\cl - \bcl}{D} \epsilon^T A^{-1} X^T x \right] \right)^2 \right]
\end{multline*}
\begin{align*}
    = \bcl^2 \what^T \left(a I_D + bvv^T \right) \Sigma \left(a I_D + bvv^T \right) \what + \EE_{X, \epsilon}\left[ (\cl - \bcl) \what^T M_1^T\right] \Sigma \EE_{X, \epsilon}\left[ (\cl - \bcl) M_1 \what \right] \\  
    + \EE_{X, \epsilon}\left[\frac{\cl - \bcl}{D} \epsilon^T A^{-1} X^T \right] \Sigma \EE_{X, \epsilon}\left[\frac{\cl - \bcl}{D} X A^{-1} \epsilon \right] + 2  \bcl \what^T \left(a I_D + bvv^T \right) \Sigma \EE_{X, \epsilon}\left[ (\cl - \bcl) M_1 \what \right]  \\
    + 2 \bcl \what^T \left(a I_D + bvv^T \right) \Sigma \EE_{X, \epsilon}\left[\frac{\cl - \bcl}{D} X A^{-1} \epsilon \right] + 2 \EE_{X, \epsilon}\left[ (\cl - \bcl) \what^T M_1^T\right] \Sigma \EE_{X, \epsilon}\left[\frac{\cl - \bcl}{D} X A^{-1} \epsilon \right].
\end{align*}
\begin{align*}
    = \bcl^2 \what^T \left(a I_D + bvv^T \right) \left( I_D + \gamma vv^T \right) \left(a I_D + bvv^T \right) \what \\ + \EE_{X, \epsilon}\left[ (\cl - \bcl) \what^T M_1^T\right] \left( I_D + \gamma vv^T \right) \EE_{X, \epsilon}\left[ (\cl - \bcl) M_1 \what \right] \\  
    + u_\epsilon^T \left( I_D + \gamma vv^T \right) u_\epsilon + 2 \bcl \what^T \left(a I_D + bvv^T \right) \left( I_D + \gamma vv^T \right) \EE_{X, \epsilon}\left[ (\cl - \bcl) M_1 \what \right]  \\
    + 2 \bcl \what^T \left(a I_D + bvv^T \right) \left( I_D + \gamma vv^T \right) u_\epsilon + 2 \EE_{X, \epsilon}\left[ (\cl - \bcl) \what^T M_1^T\right] \left( I_D + \gamma vv^T \right) u_\epsilon
\end{align*}
where we have defined $u_\epsilon :=  \EE_{X, \epsilon}\left[\frac{\cl - \bcl}{D} X A^{-1} \epsilon \right]$.
Since $\cl = \bcl = 1$ when $\lambda = 0$, all of the terms above except for $\bcl^2 \what^T \left(a I_D + bvv^T \right) \left( I_D + \gamma vv^T \right) \left(a I_D + bvv^T \right) \what$ vanish. When $\lambda > 0$, we show that each of the other terms converges to $0$ as $n, D \to \infty$. In particular, define the following terms:
\begin{align*}
    \delta_3 = \EE_{X, \epsilon}\left[ (\cl - \bcl) \what^T M_1^T\right] \left( I_D + \gamma vv^T \right) \EE_{X, \epsilon}\left[ (\cl - \bcl) M_1 \what \right] \\ 
    \delta_4 = u_\epsilon^T \left( I_D + \gamma vv^T \right) u_\epsilon \\
    \delta_5 = 2 \bcl \what^T \left(a I_D + bvv^T \right) \left( I_D + \gamma vv^T \right) \EE_{X, \epsilon}\left[ (\cl - \bcl) M_1 \what \right] \\
    \delta_6 = 2 \bcl \what^T \left(a I_D + bvv^T \right) \left( I_D + \gamma vv^T \right) u_\epsilon \\
    \delta_7 = 2 \EE_{X, \epsilon}\left[ (\cl - \bcl) \what^T M_1^T\right] \left( I_D + \gamma vv^T \right) u_\epsilon
\end{align*}
Note first that, for fixed $\lambda > 0$,
\begin{align*}
    0 \leq \delta_3 = \norm{\EE_{X, \epsilon}\left[ (\cl - \bcl) M_1 \what \right]}_2^2 + \gamma \left(\EE_{X, \epsilon}\left[ (\cl - \bcl) v^T M_1 \what \right]\right)^2 \\
    \leq \EE_{X, \epsilon}\left[ (\cl - \bcl)^2 \norm{M_1 \what}_2^2 \right] + \gamma \EE_{X, \epsilon}\left[ (\cl - \bcl)^2 \left| v^T M_1 \what \right|^2 \right] \\ 
    \leq \EE_{X, \epsilon}\left[ (\cl - \bcl)^2 \what^T M_1^2 \what \right] + \gamma \EE_{X, \epsilon}\left[ (\cl - \bcl)^2 \opnorm{M_1}^2 \right] \\
    \leq \EE_{X, \epsilon}\left[ (\cl - \bcl)^2 \opnorm{M_1^2} \right] + \gamma \EE_{X, \epsilon}\left[ (\cl - \bcl)^2 \opnorm{M_1}^2 \right]
\end{align*}
\begin{align*}
    \leq \EE_{X, \epsilon}\left[ (\cl - \bcl)^2 \opnorm{M_1}^2 \right] + \gamma \EE_{X, \epsilon}\left[ (\cl - \bcl)^2 \opnorm{M_1}^2 \right] \\
    \leq (1+\gamma) \EE_{X, \epsilon}\left[ (\cl - \bcl)^2 \right]  \overset{n, D \nearrow \infty}{\longrightarrow} 0.
\end{align*}
Here we have used the fact that $\opnorm{M_1} \leq 1$ (for $n < D$), that $\cl \asconverge \bcl$, and that $\cl$ is moment controlled (Thm. \ref{thm:inflation_constant_extended}). Next,
\begin{align*}
    0 \leq \delta_4 = \norm{u_\epsilon}_2^2 + \gamma \left(v^T u_\epsilon\right)^2 = \norm{\EE_{X, \epsilon}\left[\frac{\cl - \bcl}{D} X A^{-1} \epsilon \right]}_2^2 + \gamma \left(\EE_{X, \epsilon}\left[\frac{\cl - \bcl}{D} v^T X A^{-1} \epsilon \right]\right)^2 \\
    \leq \EE_{X, \epsilon}\left[\frac{(\cl - \bcl)^2}{D^2} \norm{X A^{-1} \epsilon}_2^2 \right] + \gamma \EE_{X, \epsilon}\left[(\cl - \bcl)^2\right] \EE_{X, \epsilon}\left[\left( \frac{v^T X A^{-1} \epsilon}{D} \right)^2 \right]\\
    = \EE_{X, \epsilon}\left[(\cl - \bcl)^2 \frac{\epsilon^T A^{-1} \frac{X^T X}{D} A^{-1} \epsilon}{D} \right] + \gamma \EE_{X, \epsilon}\left[(\cl - \bcl)^2\right] \EE_{X, \epsilon}\left[\left( \frac{v^T X A^{-1} \epsilon}{D} \right)^2 \right]\\
\end{align*}
\begin{align*}
    = \EE_{X, \epsilon}\left[(\cl - \bcl)^2 \frac{\epsilon^T A^{-1} (A - \lambda I_n) A^{-1} \epsilon}{D} \right] + \gamma \EE_{X, \epsilon}\left[(\cl - \bcl)^2\right] \EE_{X, \epsilon}\left[\left( \frac{v^T X A^{-1} \epsilon}{D} \right)^2 \right]\\
    = \EE_{X, \epsilon}\left[(\cl - \bcl)^2 \frac{\epsilon^T A^{-1} \epsilon}{D} \right] - \lambda\EE_{X, \epsilon}\left[(\cl - \bcl)^2 \frac{\epsilon^T A^{-2} \epsilon}{D} \right] + \gamma \EE_{X, \epsilon}\left[(\cl - \bcl)^2\right] \EE_{X, \epsilon}\left[\left( \frac{v^T X A^{-1} \epsilon}{D} \right)^2 \right]\\
    \overset{n,D \nearrow \infty}{\longrightarrow} 0.
\end{align*}
Here, we have used the facts that $\cl \asconverge \bcl$ (Thm. \ref{thm:inflation_constant_extended}), $\frac{\epsilon^T A^{-1} \epsilon}{n} \asconverge \sigma^2 g(-\lambda)$, $\frac{\epsilon^T A^{-2} \epsilon}{n} \asconverge \sigma^2 g'(-\lambda)$, $\frac{\epsilon^T A^{-1}X^T v}{D} \asconverge 0$, and that for $\lambda > 0$ it is the case that each of these quantities is moment controlled (Cor. \ref{cor:A_inv_MP_quantities}). Next,
\begin{align*}
    |\delta_5| = \left| 2\bcl \left( a \EE_{X, \epsilon}\left[ (\cl - \bcl) \what^T M_1 \what \right] + (b + a\gamma + b\gamma) (\what^T v) \EE_{X, \epsilon}\left[ (\cl - \bcl) v^T M_1 \what \right] \right) \right| \\
    \leq 2 \bcl \left(|a| \EE_{X, \epsilon}\left[ |\cl - \bcl| \opnorm{M_1}\right] + |\rho| \cdot |b + a\gamma + b\gamma| \EE_{X, \epsilon}\left[ |\cl - \bcl| \opnorm{M_1} \right] \right) \\
    \leq 2 \bcl \left(|a| \EE_{X, \epsilon}\left[ |\cl - \bcl| \right] + |\rho| \cdot  |b + a\gamma + b\gamma| \EE_{X, \epsilon}\left[ |\cl - \bcl| \right] \right) \overset{n, D \nearrow \infty}{\longrightarrow} 0.
\end{align*}
because $ \EE_{X, \epsilon}\left[ |\cl - \bcl| \right] \to 0$ (Thm. \ref{thm:inflation_constant_extended}). Next,
\begin{align*}
    |\delta_6| = \left|2 \bcl \left( a \what^T u_\epsilon + (b+a\gamma+b\gamma)(\what^T v) v^T u_\epsilon \right) \right| \\
    = \left| 2 \bcl \left( a \EE_{X, \epsilon}\left[\frac{\cl - \bcl}{D} \what^T X A^{-1} \epsilon \right] + \rho (b+a\gamma+b\gamma) \EE_{X, \epsilon}\left[\frac{\cl - \bcl}{D} v^T X A^{-1} \epsilon \right] \right) \right| \\ 
    \leq 2 \bcl \left( |a| \EE_{X, \epsilon}\left[|\cl - \bcl| \left|\frac{\what^T X A^{-1} \epsilon}{D}\right| \right] + |\rho| \cdot |b+a\gamma+b\gamma| \EE_{X, \epsilon}\left[|\cl - \bcl| \left|\frac{v^T X A^{-1} \epsilon}{D}\right| \right] \right) \\
    \overset{n, D \nearrow \infty}{\longrightarrow } 0.
\end{align*}
Finally, letting $\hat{u}_\epsilon := \frac{u_\epsilon}{\norm{u_\epsilon}_2}$,
\begin{align*}
    |\delta_7| = \left|2 \EE_{X, \epsilon}\left[ (\cl - \bcl) \what^T M_1^T \hat{u}_\epsilon\right] \norm{u_\epsilon}_2 + 2 \gamma \EE_{X, \epsilon}\left[ (\cl - \bcl) \what^T M_1^T v \right] v^T u_\epsilon \right|
\end{align*}
\begin{multline*}
    \leq  2 \EE_{X, \epsilon}\left[ |\cl - \bcl| \opnorm{M_1} \right] \norm{ \EE_{X, \epsilon}\left[\frac{\cl - \bcl}{D} X A^{-1} \epsilon \right] }_2 \\+ 2 \gamma \EE_{X, \epsilon}\left[ |\cl - \bcl| \opnorm{M_1} \right] \left| \EE_{X, \epsilon}\left[\frac{\cl - \bcl}{D} v^T X A^{-1} \epsilon \right] \right| 
\end{multline*}
\begin{align*}
    \leq 2 \EE_{X, \epsilon}\left[ |\cl - \bcl| \right] \EE_{X, \epsilon}\left[\frac{(\cl - \bcl)^2}{D^2} \norm{X A^{-1} \epsilon}_2^2 \right]^{1/2} + 2 \gamma \EE_{X, \epsilon}\left[ |\cl - \bcl| \right] \EE_{X, \epsilon}\left[|\cl - \bcl| \cdot \left|\frac{v^T X A^{-1} \epsilon }{D} \right| \right] \\
\end{align*}
\begin{multline*}
    = 2 \EE_{X, \epsilon}\left[ |\cl - \bcl| \right] \EE_{X, \epsilon}\left[(\cl - \bcl)^2 \frac{\epsilon^T A^{-1} \frac{X^T X}{D} A^{-1} \epsilon}{D} \right]^{1/2} \\+ 2 \gamma \EE_{X, \epsilon}\left[ |\cl - \bcl| \right] \EE_{X, \epsilon}\left[|\cl - \bcl| \left|\frac{v^T X A^{-1} \epsilon }{D} \right| \right]
\end{multline*}
\begin{multline*}
    = 2 \EE_{X, \epsilon}\left[ |\cl - \bcl| \right] \EE_{X, \epsilon}\left[(\cl - \bcl)^2 \frac{\epsilon^T A^{-1} (A - \lambda I_n) A^{-1} \epsilon}{D} \right]^{1/2} \\+ 2 \gamma \EE_{X, \epsilon}\left[ |\cl - \bcl| \right] \EE_{X, \epsilon}\left[|\cl - \bcl| \left|\frac{v^T X A^{-1} \epsilon }{D} \right| \right] 
\end{multline*}
\begin{multline*}
    = 2 \EE_{X, \epsilon}\left[ |\cl - \bcl| \right] \left( \EE_{X, \epsilon}\left[(\cl - \bcl)^2 \frac{\epsilon^T A^{-1} \epsilon}{D} \right] - \lambda \EE_{X, \epsilon}\left[(\cl - \bcl)^2 \frac{\epsilon^T A^{-2} \epsilon}{D} \right] \right)^{1/2} \\ + 2 \gamma \EE_{X, \epsilon}\left[ |\cl - \bcl| \right] \EE_{X, \epsilon}\left[|\cl - \bcl| \left|\frac{v^T X A^{-1} \epsilon }{D} \right| \right]  \\
    \overset{n, D \nearrow \infty}{\longrightarrow} 0.
\end{multline*}
Altogether, using Cor. \ref{cor:a_b_atilde_btilde}, we conclude that 
$$\mathbb{E}_{x \sim \mathcal{N}(0, \Sigma)}\left[ \EE_{X, \epsilon}\left[ \hat{f}(x) \right]^2 \right] = \bcl^2 \what^T \left(a I_D + bvv^T \right) \left( I_D + \gamma vv^T \right) \left(a I_D + bvv^T \right) \what + o_D(1)$$
$$ = \bcl^2 \left( a^2 \what^T \what + (\what^T v)^2 (2ab + b^2 + 2ab \gamma + a^2 \gamma + b^2 \gamma) \right) + o_D(1) $$
\begin{equation}    \label{eq:variance_second_term}
    \overset{n, D \nearrow \infty}{\longrightarrow} \bcl^2 \left(\alim^2 + \rho^2 \left( 2\alim \blim + \blim^2 + \gamma (\alim + \blim)^2 \right) \right).
\end{equation}

Next, recalling that $\Sigma = I_D + \gamma vv^T$ and that $K_x = \frac{X^T X}{D} = A - \lambda I_n$, we calculate $\mathbb{E}_{X, \epsilon}\left[ \mathbb{E}_{x \sim \mathcal{N}(0, \Sigma)}\left[ \hat{f}(x)^2 \right]  \right]$:
$$\mathbb{E}_{X, \epsilon}\left[ \mathbb{E}_{x \sim \mathcal{N}(0, \Sigma)}\left[ \hat{f}(x)^2 \right]  \right] = \mathbb{E}_{X, \epsilon}\left[ \mathbb{E}_{x \sim \mathcal{N}(0, \Sigma)}\left[ \left( \frac{\cl}{\sqrt{D}} w_\star^T M_1^T x + \frac{\cl}{D} \epsilon^T A^{-1} X^T x \right)^2 \right]  \right]$$
\begin{align*}
    = \mathbb{E}_{X, \epsilon}\left[ \mathbb{E}_{x \sim \mathcal{N}(0, \Sigma)}\left[ \left(\cl M_1 \what + \frac{\cl}{D} X A^{-1} \epsilon  \right)^T xx^T \left(\cl M_1 \what + \frac{\cl}{D} X A^{-1} \epsilon  \right)  \right] \right] \\
    = \mathbb{E}_{X, \epsilon}\left[ \left(\cl M_1 \what + \frac{\cl}{D} X A^{-1} \epsilon  \right)^T \left(I_D + \gamma vv^T \right) \left(\cl M_1 \what + \frac{\cl}{D} X A^{-1} \epsilon  \right)  \right]  
\end{align*}
\begin{align*}
    = \EE_{X, \epsilon}\left[\cl^2 \what^T M_1 \left(I_D + \gamma vv^T \right) M_1 \what  + \frac{\cl^2}{D^2} {\epsilon^T A^{-1} X^T \left(I_D + \gamma vv^T \right) X A^{-1} \epsilon} \right. \\ \left. + 2 \frac{\cl^2}{D} \what^T M_1 \left(I_D + \gamma vv^T \right)  X A^{-1} \epsilon \right]
\end{align*}
\begin{align*}
    = \EE_{X, \epsilon}\left[\cl^2 \what^T M_1^2 \what \right] + \gamma \EE_{X, \epsilon}\left[ \cl^2 \left(\what^T M_1 v \right)^2\right] + \EE_{X, \epsilon}\left[\cl^2 \frac{\epsilon^T A^{-1} \frac{X^T X}{D} A^{-1} \epsilon}{D} \right] \\  
    + \gamma \EE_{X, \epsilon}\left[\cl^2 \left( \frac{v^T X A^{-1} \epsilon}{D} \right)^2 \right] + 2 \EE_{X, \epsilon} \left[ \frac{\cl^2}{D} \what^T M_1 X A^{-1} \epsilon \right] + 2 \gamma \EE_{X, \epsilon}\left[ \cl^2 \left(\what^T M_1 v \right) \left( \frac{v^T X A^{-1} \epsilon}{D} \right) \right]
\end{align*}
\begin{align*}
    = \EE_{X, \epsilon}\left[\cl^2 \what^T \frac{X A^{-1} \frac{X^T X}{D} A^{-1} X^T}{D} \what \right] + \gamma \EE_{X, \epsilon}\left[ \cl^2 \left(\what^T M_1 v \right)^2\right] + \EE_{X, \epsilon}\left[\cl^2 \frac{\epsilon^T A^{-1} (A - \lambda I_n) A^{-1} \epsilon}{D} \right] \\  
    + \gamma \EE_{X, \epsilon}\left[\cl^2 \left( \frac{v^T X A^{-1} \epsilon}{D} \right)^2 \right] + 2 \EE_{X, \epsilon} \left[ \frac{\cl^2}{D} \what^T X A^{-1} \frac{X^T X}{D} A^{-1} \epsilon \right] + 2 \gamma \EE_{X, \epsilon}\left[ \cl^2 \left(\what^T M_1 v \right) \left( \frac{v^T X A^{-1} \epsilon}{D} \right) \right]
\end{align*}
\begin{align*}
    = \EE_{X, \epsilon}\left[\cl^2 \what^T \frac{X A^{-1} (A - \lambda I_n) A^{-1} X^T}{D} \what \right] + \gamma \EE_{X, \epsilon}\left[ \cl^2 \left(\what^T M_1 v \right)^2\right] + \EE_{X, \epsilon}\left[\cl^2 \frac{\epsilon^T A^{-1} \epsilon}{D} \right] \\ - \EE_{X, \epsilon}\left[\lambda \cl^2 \frac{\epsilon^T A^{-2} \epsilon}{D} \right]   
    + \gamma \EE_{X, \epsilon}\left[\cl^2 \left( \frac{v^T X A^{-1} \epsilon}{D} \right)^2 \right] \\+ 2 \EE_{X, \epsilon} \left[ \frac{\cl^2}{D} \what^T X A^{-1} (A - \lambda I_n) A^{-1} \epsilon \right] + 2 \gamma \EE_{X, \epsilon}\left[ \cl^2 \left(\what^T M_1 v \right) \left( \frac{v^T X A^{-1} \epsilon}{D} \right) \right]
\end{align*}
\begin{align*}
    = \EE_{X, \epsilon}\left[ \cl^2 \what^T M_1 \what \right] -  \EE_{X, \epsilon}\left[\lambda \cl^2 \what^T M_2 \what \right] + \gamma \EE_{X, \epsilon}\left[ \cl^2 \left(\what^T M_1 v\right)^2 \right] \\
    + \EE_{X, \epsilon}\left[\cl^2 \frac{\epsilon^T A^{-1} \epsilon}{D} \right] - \EE_{X, \epsilon}\left[\lambda \cl^2 \frac{\epsilon^T A^{-2} \epsilon}{D} \right] \\
    + \gamma \EE_{X, \epsilon}\left[\cl^2 \left( \frac{v^T X A^{-1} \epsilon}{D} \right)^2 \right] + 2 \EE_{X, \epsilon} \left[ \cl^2 \frac{\what^T X A^{-1} \epsilon}{D} \right] - 2 \EE_{X, \epsilon} \left[\lambda \frac{\cl^2}{D} \what^T X A^{-2} \epsilon \right] \\
    + 2 \gamma \EE_{X, \epsilon}\left[ \cl^2 \left(\what^T M_1 v \right) \left( \frac{v^T X A^{-1} \epsilon}{D} \right) \right]
\end{align*}
\begin{multline}   \label{eq:important_variance_terms}
    = \EE_{X, \epsilon}\left[ \cl^2 \what^T M_1 \what \right] - \EE_{X, \epsilon}\left[ \lambda \cl^2 \what^T M_2 \what \right] + \gamma \EE_{X, \epsilon}\left[ \cl^2 \left(\what^T M_1 v\right)^2 \right] \\
    + \EE_{X, \epsilon}\left[\cl^2 \frac{\epsilon^T A^{-1} \epsilon}{D} \right] - \EE_{X, \epsilon}\left[\lambda \cl^2 \frac{\epsilon^T A^{-2} \epsilon}{D} \right] + \delta_8. 
\end{multline}
Here, we have defined 
\begin{multline*}
    \delta_8 := \gamma \EE_{X, \epsilon}\left[\cl^2 \left( \frac{v^T X A^{-1} \epsilon}{D} \right)^2 \right] + 2 \EE_{X, \epsilon} \left[ \cl^2 \frac{\what^T X A^{-1} \epsilon}{D} \right] - 2 \EE_{X, \epsilon} \left[\lambda \cl^2 \frac{\what^T X A^{-2} \epsilon}{D} \right] \\ + 2 \gamma \EE_{X, \epsilon}\left[ \cl^2 \left(\what^T M_1 v \right) \left( \frac{v^T X A^{-1} \epsilon}{D} \right) \right].
\end{multline*}
We claim that $\delta_8 \overset{n, D \nearrow \infty}{\longrightarrow} 0$. Indeed, when $\lambda = 0$, $c_\lambda$ is $1$; thus, noting that $\EE[\epsilon] = 0$, we find that 
\begin{multline*}
    \delta_8 = \gamma \EE_{X, \epsilon}\left[\left( \frac{v^T X A^{-1} \epsilon}{D} \right)^2 \right] + 2 \EE_{X, \epsilon} \left[\frac{\what^T X A^{-1} \epsilon}{D} \right] - 2 \EE_{X, \epsilon} \left[\lambda \frac{\what^T X A^{-2} \epsilon}{D} \right] \\ + 2 \gamma \EE_{X, \epsilon}\left[ \left(\what^T M_1 v \right) \left( \frac{v^T X A^{-1} \epsilon}{D} \right) \right] 
\end{multline*}
$$ = \gamma \EE_{X, \epsilon}\left[\left( \frac{v^T X A^{-1} \epsilon}{D} \right)^2 \right]$$
where we have noted that $\EE_{X, \epsilon} \left[\frac{\what^T X A^{-1} \epsilon}{D} \right] = 0$ (Cor. \ref{cor:integrability_of_inverses_lambda_zero}), that $\EE_{X, \epsilon} \left[\lambda \frac{\what^T X A^{-2} \epsilon}{D} \right] = 0$ when $\lambda = 0$, and that 
$$\EE_{X, \epsilon}\left[ \left(\what^T M_1 v \right) \left( \frac{v^T X A^{-1} \epsilon}{D} \right) \right] = \EE_{X}\left[ \left(\what^T M_1 v \right) \left( \frac{v^T X A^{-1}}{D} \right) \right] \EE_\epsilon[\epsilon] = 0.$$ 
This is valid because $|\what^T M_1 v| \leq \norm{\what}_2 \opnorm{M_1} \norm{v}_2 \leq 1$ (note that $\opnorm{M_1} \leq 1$) and
$$\EE_{X}\left[ \norm{\left(\what^T M_1 v \right) \left( \frac{v^T X A^{-1}}{D}  \right)}_2 \right] \leq \frac{1}{\sqrt{D}} \EE_{X}\left[ \opnorm{ \frac{X A^{-1}}{\sqrt{D}} } \right] \leq \frac{1}{\sqrt{D}} \EE_{X}\left[ \eigmin(K_x)^{-1/2} \right] < \infty $$
at any fixed, finite $D$ and $n$ with $D > n+1$ (Cor. \ref{cor:integrability_of_inverses_lambda_zero}). Proceeding, we obtain
$$\delta_8 = \gamma \EE_{X, \epsilon}\left[\left( \frac{v^T X A^{-1} \epsilon}{D} \right)^2 \right] = \gamma \EE_{X, \epsilon}\left[  \frac{v^T X A^{-1} \epsilon \epsilon^T A^{-1} X^T v}{D^2}  \right] = \gamma \EE_{X}\left[ \EE\left[ \left. \frac{v^T X A^{-1} \epsilon \epsilon^T A^{-1} X^T v}{D^2} \text{ } \right| \text{ } X \right]  \right] $$
$$ = \gamma \EE_{X}\left[  \frac{v^T X A^{-1} (\sigma^2 I_n) A^{-1} X^T v}{D^2}  \right] = \frac{\sigma^2 \gamma}{D} v^T \EE_X\left[ \frac{1}{D} XA^{-2} X^T  \right] v$$
$$ = \frac{\sigma^2 \gamma}{D} \EE_X\left[v^T M_2 v\right] \overset{n, D \to \infty}{\longrightarrow} 0$$
where we have noted that $\EE_X\left[v^T M_2 v\right] = O_D(1)$ when $\lambda = 0$ (Prop. \ref{prop:M_v_contraction}). Furthermore, if $\lambda > 0$, then
\begin{multline*}
    \delta_8 = \gamma \EE_{X, \epsilon}\left[\cl^2 \left( \frac{v^T X A^{-1} \epsilon}{D} \right)^2 \right] + 2 \EE_{X, \epsilon} \left[ \cl^2 \frac{\what^T X A^{-1} \epsilon}{D} \right] - 2\lambda \EE_{X, \epsilon} \left[ \cl^2 \frac{\what^T X A^{-2} \epsilon}{D} \right] \\ + 2 \gamma \EE_{X, \epsilon}\left[ \cl^2 \left(\what^T M_1 v \right) \left( \frac{v^T X A^{-1} \epsilon}{D} \right) \right] 
\end{multline*}
$$\overset{n, D \nearrow \infty}{\longrightarrow} 0. $$
Here, we have applied the facts that, for $\lambda > 0$, $\cl \asconverge \bcl$, $\frac{v^T X A^{-1} \epsilon}{D} \asconverge 0$, $\frac{\what^T X A^{-1} \epsilon}{D} \asconverge 0$, $\frac{\what^T X A^{-2} \epsilon}{D} \asconverge 0$, $\what^T M_1 v$ has an almost sure limit, and that each of these quantities is moment controlled (Thm. \ref{thm:inflation_constant_extended}, Cor. \ref{cor:A_inv_MP_quantities}, Cor. \ref{cor:M_1_M_2_expectation_terms}).
Thus, to proceed, it suffices to simplify the terms in Eq. \ref{eq:important_variance_terms}.
When $\lambda = 0$, $\cl = 1$ and Eq. \ref{eq:important_variance_terms} simplifies to 
\begin{equation}    \label{eq:variance_2nd_moment_term_lambda_zero}
\begin{aligned}
    \mathbb{E}_{X, \epsilon}\left[ \mathbb{E}_{x \sim \mathcal{N}(0, \Sigma)}\left[ \hat{f}(x)^2 \right]  \right] = \EE_{X, \epsilon}\left[ \what^T M_1 \what \right] + \gamma \EE_{X, \epsilon}\left[ \left(\what^T M_1 v\right)^2 \right] + \EE_{X, \epsilon}\left[\frac{\epsilon^T K_x^{-1} \epsilon}{D} \right] + o_D(1) \\
    =  \what^T \EE_{X}\left[ M_1 \right] \what + \gamma \EE_{X}\left[ \left(\what^T M_1 v\right)^2 \right] + \frac{n}{D} \Tr\left( \EE_{X}\left[\frac{K_x^{-1}}{n} \right] \EE_{\epsilon}[\epsilon \epsilon^T] \right) + o_D(1) \\
    = \what^T \left( aI_D + b vv^T \right) \what + \gamma \left(\what^T \EE_{X}\left[ M_1 \right] v\right)^2  + \frac{n}{D} \sigma^2 \EE\left[ \Tr\left( \frac1n K_x^{-1} \right)  \right] + o_D(1)
    \end{aligned}
\end{equation}
\begin{align*}
    = a + b \rho^2 + \gamma \left(\what^T \left(a I_D + b vv^T \right) v\right)^2  + \frac{n}{D} \sigma^2 \EE\left[ \Tr\left( \frac1n K_x^{-1} \right) \right] + o_D(1) \\
    = a + b \rho^2 + \gamma \rho^2 (a+b)^2 + \frac{n}{D} \sigma^2 \EE_X[g_{K_x}(0)] + o_D(1) \\
\end{align*}
\begin{align}   \label{eq:lambda_zero_main_variance_term}
    \overset{n, D \nearrow \infty}{\longrightarrow} \alim + \blim \rho^2 + \gamma \rho^2 (\alim+\blim)^2 + \psi \sigma^2 g(0).
\end{align}
Here, we have used Cor. \ref{cor:a_b_atilde_btilde} and the facts that $\frac{n}{D} \to \psi$, $\EE\left[ \Tr\left( \frac1n K_x^{-1} \right)  \right] \to g(0)$ (Prop. \ref{prop:expected_stieltjes_transform_at_zero}) and that $\what^T M_1 v$ converges almost surely to a nonrandom constant (Cor. \ref{cor:A_inv_MP_quantities}, part 3) as $n, D \to \infty$. Moreover, we specifically claim that $\what^T M_1 v \asconverge \rho(\alim+\blim)$ and hence that $\left( \what^T M_1 v \right)^2 \asconverge \rho^2 (\alim+\blim)^2$. Indeed, since $|\what^T M_1 v| \leq \opnorm{M_1} \leq 1$, the dominated convergence theorem implies that 
\begin{equation}
    \what^T M_1 v \aseq \EE[\what^T M_1 v] = \what^T (a I_D + bvv^T) v \ndlim \rho(\alim + \blim).
\end{equation}

Moreover, $|\what^T M_1 v| \leq \opnorm{M_1} \leq 1$ further implies that $\left( \what^T M_1 v \right)^2 \leq 1$; consequently, the dominated convergence theorem implies that both $\EE_X\left[ \left( \what^T M_1 v \right)^2 \right]$ and $\left( \what^T \EE_X[M_1] v \right)^2 = \left( \EE_X[\what^T M_1 v] \right)^2$ converge to $\rho^2 (\alim + \blim)^2$, the almost sure limit of $\left(\what^T M_1 v \right)^2$. This justifies our implicit assertion that $\EE_X\left[ \left( \what^T M_1 v \right)^2 \right] = \left( \EE_X[\what^T M_1 v] \right)^2 + o_D(1)$ in Eq. \ref{eq:variance_2nd_moment_term_lambda_zero}.

When $\lambda > 0$, $\cl \asconverge \bcl$, $\what^T M_1 \what \asconverge \alim + \blim \rho^2$, $\what^T M_2 \what \asconverge \atlim + \btlim \rho^2$, $\what^T M_1 v \asconverge \rho (\alim+\blim)$, $\frac{\epsilon^T A^{-1} \epsilon}{n} \asconverge \sigma^2 g(-\lambda)$, and $\frac{\epsilon^T A^{-2} \epsilon}{n} \asconverge \sigma^2 g'(-\lambda)$; moreover, each of these random variables is moment controlled (Thm. \ref{thm:inflation_constant_extended}, Cor. \ref{cor:A_inv_MP_quantities}, Cor. \ref{cor:M_1_M_2_expectation_terms}). Consequently, using Cor. \ref{cor:a_b_atilde_btilde} and Cor. \ref{cor:expectation_factorizes_Lp}, we find that
\begin{align*}
    \mathbb{E}_{X, \epsilon}\left[ \mathbb{E}_{x \sim \mathcal{N}(0, \Sigma)}\left[ \hat{f}(x)^2 \right]  \right] =
    \bcl^2 \what^T \EE_X[M_1] \what - \lambda \bcl^2 \what^T \EE_X[M_2] \what + \gamma \bcl^2 \left(\what^T \EE_X[M_1] v\right)^2  \\
    + \frac{n}{D} \EE_{X, \epsilon}\left[\cl^2 \frac{\epsilon^T A^{-1} \epsilon}{n} \right] - \frac{n}{D} \lambda \EE_{X, \epsilon}\left[ \cl^2 \frac{\epsilon^T A^{-2} \epsilon}{n} \right] + o_D(1) \\
    \overset{n, D \nearrow \infty}{\longrightarrow} \bcl^2 \left( \what^T \left( \alim I_D + \blim vv^T\right) \what - \lambda \what^T \left( \atlim I_D + \btlim vv^T \right) \what \right. \\
    \left. + \gamma \left( \what^T \left( \alim I_D + \blim vv^T \right) v \right)^2 + \sigma^2 \psi \left(g(-\lambda) - \lambda g'(-\lambda) \right) \right)
\end{align*}
\begin{align}   \label{eq:main_variance_term}
    = \bcl^2 \left( \alim + \blim \rho^2 - \lambda \left(\atlim + \btlim \rho^2 \right) + \gamma \rho^2 \left( \alim + \blim \right)^2 + \sigma^2 \psi \left(g(-\lambda) - \lambda g'(-\lambda) \right) \right).
\end{align}
By inspection, the expression in Eq. \ref{eq:main_variance_term} coincides with that of Eq. \ref{eq:lambda_zero_main_variance_term} when $\lambda = 0$ (in which case $\bcl = 1$), so we proceed with the expression given by Eq. \ref{eq:main_variance_term} for all $\lambda \geq 0$. In particular, substituting our expressions from Eq. \ref{eq:main_variance_term} and Eq. \ref{eq:variance_second_term} into Eq. \ref{eq:variance_formula}, we obtain
\begin{align*}
    \mathcal{V}(\hat{f}) = \mathbb{E}_{X, \epsilon}\left[ \mathbb{E}_{x \sim \mathcal{N}(0, \Sigma)}\left[ \hat{f}(x)^2 \right]  \right] - \mathbb{E}_{x \sim \mathcal{N}(0, \Sigma)}\left[ \EE_{X, \epsilon}\left[ \hat{f}(x) \right]^2 \right] 
\end{align*}
\begin{align*}
    \overset{n, D \nearrow \infty}{\longrightarrow} \bcl^2 \left( \alim + \blim \rho^2 - \lambda \left(\atlim + \btlim \rho^2 \right) + \gamma \rho^2 \left( \alim + \blim \right)^2 + \sigma^2 \psi \left(g(-\lambda) - \lambda g'(-\lambda) \right) \right) \\
    - \bcl^2 \left(\alim^2 + \rho^2 \left( 2\alim \blim + \blim^2 + \gamma (\alim+\blim)^2 \right) \right)
\end{align*}
\begin{align}   \label{eq:final_variance_formula}
    = \bcl^2 \left( \alim + \blim \rho^2 - \lambda \left(\atlim + \btlim \rho^2 \right) - \alim^2 - \rho^2 \left( 2\alim \blim + \blim^2 \right) + \sigma^2 \atlim \right).
\end{align}

\end{proof}

\begin{cor} \label{cor:asymptotic_gen_error}
    Under Assumption \ref{assump:data_gen_process}, we asymptotically obtain the following formula for the generalization error of the feature learning estimator $\hat{f}$:
    \begin{multline}    \label{eq:gen_error_limit_SUPP}
    \mathcal{G}(\hat{f}) \ndlim 1 - 2\bcl \alim - 2 \rho^2 \blim \bcl + \gamma \rho^2 (1 - \bcl( \alim + \blim ))^2 + \alim \bcl^2 \\+ \rho^2 \blim \bcl^2 - \lambda \bcl^2 (\atlim + \btlim \rho^2) + \bcl^2 \sigma^2 \atlim.
\end{multline}
\end{cor}

\begin{proof}
    This result follows from the identity $\gen(\hat{f}) = \bias(\hat{f}) + \variance(\hat{f})$ (Lemma \ref{lemma:bias_variance_decomp}) and the limiting formulas for the bias and variance established in Theorems \ref{thm:asymptotic_bias} and \ref{thm:asymptotic_variance}.
\end{proof}

As stated in the main text, we refer to the asymptotic formulas for bias, variance, and generalization error as $\Bt$, $\Vt$, and $\Gt$, respectively.

\subsubsection{Analysis of intermediate quantities} \label{app:intermediate_quantities}

Before further analyzing the above asymptotic formulas, we state key properties of the intermediate quantities $\alim$, $\blim$, $\atlim$, and $\btlim$.

\begin{lemma}   \label{lemma:alim_blim_properties}
    With the setup given in Section \ref{supp:setup} (in particular, noting that $\psi \in (0, 1)$), each of $\alim$, $\blim$, $\atlim$, and $\btlim$ is a differentiable function in $\lambda$ for all $\lambda \geq 0$. Moreover, the following statements hold:
    \begin{enumerate}
        \item $$\alim = \tau_1 = \psi \int \frac{x}{x+\lambda} \MP^{\psi}(dx)$$ is a decreasing function over $\lambda \geq 0$, with $\alim \in (0, \psi]$, such that $\lim_{\lambda \to \infty} \alim = 0$.

        \item $$\atlim = \tau_2 = \psi \int \frac{x}{(x + \lambda)^2} \MP^{\psi}(dx)$$ is a decreasing function over $\lambda \geq 0$, with $\atlim \in \left(0, \frac{\psi}{1 - \psi}\right]$, such that $\lim_{\lambda \to \infty} \lambda \atlim = 0$. 

        \item $\lim_{\lambda \to \infty} \blim = 0$ and $\lim_{\lambda \to \infty} \lambda \btlim = 0$.

        \item The following identities hold across all $\lambda \geq 0$:
        \begin{equation}
            \frac{\partial \alim}{\partial \lambda} = -\atlim 
        \end{equation}
        \begin{equation}
            \frac{\partial \blim}{\partial \lambda} = -\btlim
        \end{equation}
        \begin{equation}
            \frac{\partial \atlim}{\partial \lambda} = - \psi(2 g'(-\lambda) - \lambda g''(-\lambda)) = -2 \psi \int \frac{x}{(x+\lambda)^3} \MP^{\psi}(dx)
        \end{equation}            
    \end{enumerate}
\end{lemma}

\begin{proof}
Note that the Stieltjes transform $g$ and its derivative $g'$ are both differentiable, as $g$ is infinitely differentiable for all $\lambda \geq 0$ (Prop. \ref{prop:properties_of_MP_and_Stieltjes}). As a result, $\alim = \psi(1 - \lambda g(-\lambda)) = \tau_1$ and $\atlim = \psi(g(-\lambda) - \lambda g'(-\lambda)) = \tau_2$ are differentiable for all $\lambda \geq 0$. Since $\tau_1$ and $\tau_2$ are differentiable, we also find that $\blim = \frac{\gamma \tau_1 (1 - \tau_1)}{1 + \gamma \tau_1}$ and $\btlim = \frac{(\gamma+1)\tau_2}{(1 + \gamma \tau_1)^2} - \tau_2$ are differentiable for all $\lambda \geq 0$ as well. (Note that $1 + \gamma \tau_1 \geq 1$, so the denominator terms never vanish). We now prove each of the remaining results.

\paragraph{1.} We have that
$$\alim = \psi (1 - \lambda g(-\lambda)) = \psi \left(1 - \lambda \int \frac{1}{x+\lambda} \MP^{\psi}(dx) \right)$$
$$ = \psi \int \left(1 - \frac{\lambda}{x + \lambda} \right) \MP^{\psi}(dx) = \psi \int \frac{x}{x + \lambda} \MP^{\psi}(dx).$$
Thus, we see that $\alim \geq 0$ (as the MP law is supported above zero) and that $\alim \leq \psi$ because $\frac{x}{x + \lambda} \leq 1$ over the support of the MP law. Moreover, for any fixed $x$, $\lambda \mapsto \frac{x}{x + \lambda}$ is a decreasing function in $\lambda$ for $\lambda \geq 0$, so we conclude that $\alim$ is also decreasing over all $\lambda \geq 0$. In turn, $$\alim \leq \alim(\lambda = 0) = \psi \int \MP^{\psi}(dx) = \psi$$
meaning that $\alim \leq \psi$ with equality when $\lambda = 0$. Finally, we show that $\lim_{\lambda \to \infty} \alim = 0$. Indeed, since $\MP^\psi$ is compactly supported, $\alim = \psi \int \frac{x}{x+\lambda} \MP^\psi(dx) \to 0$ as $\lambda \to \infty$.

\paragraph{2.} We have that
$$\atlim = \psi (g(-\lambda) - \lambda g'(-\lambda)) = \psi \left(\int \frac{1}{x + \lambda} \MP^{\psi}(dx) - \lambda \int \frac{1}{(x+\lambda)^2} \MP^{\psi}(dx) \right)$$
$$ = \psi \int \left(\frac{1}{x + \lambda} - \frac{\lambda}{(x + \lambda)^2} \right) \MP^{\psi}(dx) = \psi \int \frac{x}{(x + \lambda)^2} \MP^{\psi}(dx).$$
Thus, we see that $\atlim \geq 0$ (as the MP law is supported above zero). Moreover, for any fixed $x$, $\lambda \mapsto \frac{x}{(x + \lambda)^2}$ is a decreasing function in $\lambda$ for $\lambda \geq 0$, so we conclude that $\atlim$ is also decreasing over all $\lambda \geq 0$. As a result, we also find that $$\atlim \leq \atlim(\lambda = 0) = \psi \int \frac{1}{x} \MP^{\psi}(dx) = \frac{\psi}{1 - \psi}$$
where we have noted that $\int \frac{1}{x} \MP^{\psi}(dx) = \frac{1}{1 - \psi}$ by Prop. \ref{prop:properties_of_MP_and_Stieltjes}. Finally, $\lim_{\lambda \to \infty} \lambda \atlim = 0$ because 
$$\lambda \int \frac{x}{(x+\lambda)^2} \MP^\psi(dx) = \int \frac{x}{x+ \lambda} \cdot \frac{\lambda}{x+\lambda} \MP^\psi(dx) \leq \int \frac{x}{x+\lambda} \MP^\psi(dx) \overset{\lambda \to \infty}{\longrightarrow} 0.$$

\paragraph{3.} Recall that $\blim = \frac{\gamma \tau_1 (1 - \tau_1)}{1 + \gamma \tau_1}$ and that $\lim_{\lambda \to \infty} \tau_1 = 0$ from part 1 of this lemma. As a result, $\lim_{\lambda \to \infty} \blim = 0$. Next, recall that $\btlim = \frac{(\gamma + 1)\tau_2}{(1+ \gamma \tau_1)^2} - \tau_2$ and that $\lim_{\lambda \to \infty} \lambda \tau_2 = 0$ (by part 2 of this Lemma). Combining this with the fact that $\lim_{\lambda \to \infty} \tau_1 = 0$, it follows that 
$$\lim_{\lambda \to \infty} \lambda \btlim = \lim_{\lambda \to \infty} \left( \frac{(\gamma + 1) \lambda \tau_2}{(1 + \gamma \tau_1)^2} - \lambda \tau_2 \right) = 0.$$

\paragraph{4.} First, recall that $g(0) = \frac{1}{1 - \psi}$ and $g'(0) = \frac{1}{(1 - \psi)^3}$ (Prop. \ref{prop:properties_of_MP_and_Stieltjes}). Next, observe that
$$\left. \frac{\partial \alim}{\partial \lambda} \right. = \left. \frac{\partial \tau_1}{\partial \lambda} \right. = \frac{\partial}{\partial \lambda} \left( \psi(1 - \lambda g(-\lambda)) \right)$$ 
$$= \psi(-g(-\lambda) + \lambda g'(-\lambda)) = -\psi(g(-\lambda) - \lambda g'(-\lambda)) = -\atlim.$$

$$\left. \frac{\partial \blim}{\partial \lambda} \right. = \frac{\partial}{\partial \lambda} \left( \frac{(\gamma + 1) \tau_1}{1 + \gamma \tau_1} - \tau_1 \right) = -\frac{\partial \tau_1}{\partial \lambda} + \frac{(\gamma+1) \left( \frac{\partial \tau_1}{\partial \lambda} (1 + \gamma \tau_1) - \tau_1 \left(\gamma \frac{\partial \tau_1}{\partial \lambda}\right) \right)}{(1 + \gamma \tau_1)^2}$$ $$ = \tau_2 - \frac{(\gamma+1) \tau_2}{(1 + \gamma \tau_1)^2} = -\btlim.$$
For the last equality, we have used the fact that $$\frac{\partial \tau_1}{\partial \lambda} = \frac{\partial \alim}{\partial \lambda} = -\atlim = -\tau_2.$$
Finally,
$$\left. \frac{\partial \atlim}{\partial \lambda} \right. = \left. \frac{\partial \tau_2}{\partial \lambda} \right. = \frac{\partial}{\partial \lambda} \left( \psi(g(-\lambda) - \lambda g'(-\lambda)) \right)$$
$$ = \psi(-g'(-\lambda) - g'(-\lambda) + \lambda g''(-\lambda)) = -\psi(2g'(-\lambda) - \lambda g''(-\lambda)).$$
Moreover, by differentiating $g$ under the integral, we find that 
$$g'(-\lambda) = \int \frac{1}{(x+\lambda)^2} \MP^{\psi}(dx) \text{ and } g''(-\lambda) = 2\int \frac{1}{(x+\lambda)^3} \MP^{\psi}(dx).$$
As a result, 
    $$2g'(-\lambda) - \lambda g''(-\lambda) = 2 \int \left( \frac{1}{(x + \lambda)^2} - \frac{\lambda}{(x + \lambda)^3} \right) \MP^{\psi}(dx)$$
    $$ = 2 \int \frac{x}{(x + \lambda)^3}  \MP^{\psi}(dx) > 0$$
    for all $\lambda \geq 0$.
It follows that 
$$\left. \frac{\partial \atlim}{\partial \lambda} \right. = -2 \psi \int \frac{x}{(x + \lambda)^3}  \MP^{\psi}(dx).$$
\end{proof}

Using Lemma \ref{lemma:alim_blim_properties}, we may now enumerate the following quantities:

\begin{enumerate}
    \item $\tau_1(\lambda = 0) = \psi$.

    \item $\tau_2(\lambda = 0) = \psi g(0) = \frac{\psi}{1 - \psi}$.

    \item $\alim(\lambda = 0) = \tau_1 = \psi$.
    
    \item $\blim(\lambda = 0) = \frac{\gamma \tau_1 (1 - \tau_1)}{1 + \gamma \tau_1} = \frac{\gamma \psi (1 - \psi)}{1 + \gamma \psi}$.

    \item $\atlim(\lambda = 0) = \tau_2 = \frac{\psi}{1 - \psi}$.

    \item $\btlim(\lambda = 0) = \frac{(\gamma+1) \tau_2}{\left(1 + \gamma \tau_1 \right)^2} - \tau_2 = \frac{(\gamma+1) \frac{\psi}{1-\psi}}{\left(1 + \gamma \psi \right)^2} - \frac{\psi}{1-\psi} = \frac{\psi}{1 - \psi}\left( \frac{1+\gamma}{(1 + \gamma \psi)^2} - 1 \right)$.

    \item $ \left. \frac{\partial \alim}{\partial \lambda} \right|_{\lambda = 0} = -\atlim(\lambda = 0) =  -\frac{\psi}{1 - \psi}$.

    \item $\left. \frac{\partial \blim}{\partial \lambda} \right|_{\lambda = 0} = -\btlim(\lambda = 0) = \frac{\psi}{1 - \psi} \left( 1 - \frac{1 + \gamma}{(1 + \gamma \psi)^2} \right)$.

    \item $\alim(\lambda = 0) + \blim(\lambda = 0) = \frac{\psi (1 + \gamma)}{1 + \gamma \psi}$ and $1 - (\alim(\lambda = 0) + \blim(\lambda = 0)) = \frac{1 - \psi}{1 + \gamma \psi}$.

    \item $\atlim(\lambda = 0) + \btlim(\lambda = 0) = \frac{\psi (1 + \gamma)}{(1 - \psi)(1 + \gamma \psi)^2}$.

    \item $\left. \frac{\partial \atlim}{\partial \lambda} \right|_{\lambda = 0} = -2 \psi g'(0) = \frac{-2 \psi}{(1 - \psi)^3}$.

    \item $\bcl(\lambda = 0) = 1$.

    \item $\left. \frac{\partial \bcl}{\partial \lambda} \right|_{\lambda = 0} = \frac{ \frac{1}{\psi} \bkl (1 + \sigma^2 + \gamma \rho^2)}{1 + \bkl \left( 1 + \frac{1+\sigma^2 + \gamma \rho^2}{\psi} + (2 + \gamma)\gamma \rho^2 \right)} \geq 0$, with strict inequality if $\bkl > 0$.
\end{enumerate}

We will repeatedly make use of these facts in the subsequent sections.

\subsubsection{Sensitivity analysis of bias and variance around zero ridge --- Proof of Proposition \ref{prop:bias_variance_lambda_derivatives}}
\label{app:bias_variance_lambda_sensitivity}

In this section, we prove Prop. \ref{prop:bias_variance_lambda_derivatives}. As in Sec. \ref{subsec:bias_variance_gen_error_expressions} of the main text, we assume that $\kl$ is held to be a fixed constant independent of $D$ and $n$, so that we may write $\kl$ in place of $\bkl$.

First, recall that 
$$\left. \frac{\partial \bcl}{\partial \lambda} \right|_{\lambda = 0} = \frac{\frac{1}{\psi} k_\ell (1 + \sigma^2 + \gamma \rho^2)}{1 + k_\ell \left( 1 + \frac{1+\sigma^2 + \gamma \rho^2}{\psi} + (2 + \gamma)\gamma \rho^2 \right)} \geq 0$$
with equality precisely when $\kl = 0$. Moreover, by inspection of the above formula, we also have that $\left. \frac{\partial \bcl}{\partial \lambda} \right|_{\lambda = 0} \leq 1$.

Note that $\cl = 1$ when $\lambda = 0$. To calculate the derivatives of bias and variance at $\lambda = 0$, we use the following fact: if $H(\lambda, \bcl)$ is a differentiable scalar function, then 
$$\left. \frac{\partial H}{\partial \lambda} \right|_{\lambda = 0} = \left. \frac{\partial}{\partial \lambda} \right|_{\lambda = 0} H(\lambda, 1) + \left( \left. \frac{\partial}{\partial \bcl} \right|_{\bcl = 1} H(0, \bcl) \right) \cdot \left. \frac{\partial \bcl}{\partial \lambda} \right|_{\lambda = 0}.$$
That is, we may nicely split the derivative up into a baseline (``no learning'') term and a feature learning-dependent term. This has the advantage of making clear the explicit contributions arising from baseline ridge regression and feature learning, respectively. We now proceed to take derivatives. For convenience, we will refer to the bias and variance as $\Bt(\lambda, \bcl)$ and $\Vt(\lambda, \bcl)$, since both formulas contain various instances of $\bcl$ in addition to other terms that depend on $\lambda$.

We start with the bias, recalling that
$$\Bt(\hat{f}) = (1 - \alim \bcl)^2 + \rho^2 \blim \bcl \left(\blim \bcl - 2 (1 - \alim \bcl) \right) + \gamma \rho^2 (1 - (\alim + \blim)\bcl)^2.$$
First, we compute the baseline (``no learning'') term of the derivative:
$$\left. \frac{\partial}{\partial \lambda} \right|_{\lambda = 0} \Bt(\lambda, 1) = \left. \frac{\partial}{\partial \lambda} \right|_{\lambda = 0} \left( (1 - \alim)^2 + \rho^2 \blim \left(\blim - 2 (1 - \alim) \right) + \gamma \rho^2 (1 - (\alim + \blim))^2 \right)$$
$$ = \left.\left[ -2 \frac{\partial \alim}{\partial \lambda} (1 - \alim) + 2 \rho^2 \left( \frac{\partial \blim}{\partial \lambda} (\alim + \blim - 1) + \frac{\partial \alim}{\partial \lambda} \blim \right) - 2 \gamma \rho^2 (1 - (\alim + \blim)) \left(\frac{\partial \alim}{\partial \lambda} + \frac{\partial \blim}{\partial \lambda} \right)\right] \right|_{\lambda = 0}$$
$$ = \left.\left[ 2 \atlim (1 - \alim) - 2 \rho^2 \left( \btlim (\alim + \blim - 1) + \atlim \blim \right) + 2 \gamma \rho^2 (1 - (\alim + \blim)) \left(\atlim + \btlim \right)\right] \right|_{\lambda = 0}.$$
We now calculate the above quantity, term by term. First,
$$\left. 2 \atlim (1 - \alim) \right|_{\lambda = 0} = 2 \left( \frac{\psi}{1 - \psi} \right) (1 - \psi) = 2 \psi.$$
Next, 
$$\left. \btlim (\alim + \blim - 1) \right|_{\lambda = 0} = \frac{\psi}{1 - \psi} \left( 1 - \frac{1 + \gamma}{(1 + \gamma \psi)^2} \right) \left( \frac{1 - \psi}{1 + \gamma \psi} \right)$$
$$ = \psi \left( \frac{(1 + \gamma \psi)^2 - (1 + \gamma)}{(1 + \gamma \psi)^3}\right) = \frac{\psi (\psi^2 \gamma^2 + 2 \psi \gamma - \gamma)}{(1 + \gamma \psi)^3}.$$
Next,
$$\left. \atlim \blim \right|_{\lambda = 0} = \frac{\psi}{1 - \psi} \cdot \frac{\gamma \psi (1 - \psi)}{1 + \gamma \psi} = \frac{\gamma \psi^2}{1 + \gamma \psi} = \frac{ \gamma \psi^2 (1 + \gamma \psi)^2}{(1 + \gamma \psi)^3}$$
$$ = \frac{\psi (\psi^3 \gamma^3 + 2 \psi^2 \gamma^2  + \psi \gamma )}{(1 + \gamma \psi)^3}.$$
As a result,
$$\left. \left[ \btlim (\alim + \blim - 1) + \atlim \blim \right] \right|_{\lambda = 0} = \frac{\psi(\psi^3 \gamma^3 + 3 \psi^2 \gamma^2 + 3 \psi \gamma - \gamma)}{(1 + \gamma \psi)^3}$$
$$ = \frac{\psi\left( (1 + \gamma \psi)^3 - (1 + \gamma)\right)}{(1 + \gamma \psi)^3} = \psi - \frac{\psi (1 + \gamma)}{(1 + \gamma \psi)^3}.$$
Finally,
$$\left. 2 \gamma \rho^2 (1 - (\alim + \blim)) \left(\atlim + \btlim \right) \right|_{\lambda = 0} = 2 \gamma \rho^2 \left( \frac{1 - \psi}{1 + \gamma \psi} \right) \frac{\psi(1 + \gamma)}{(1 - \psi) (1 + \gamma \psi)^2} $$
$$ = 2 \psi \rho^2 \frac{ \gamma (1 + \gamma)}{(1 + \gamma \psi)^3}.$$
Thus, combining each of these terms, we obtain
$$\left. \frac{\partial}{\partial \lambda} \right|_{\lambda = 0} \Bt(\lambda, 1) = 2\psi - 2 \rho^2 \left( \psi - \frac{\psi (1 + \gamma)}{(1 + \gamma \psi)^3} \right) + 2 \psi \rho^2 \frac{\gamma (1 + \gamma)}{(1 + \gamma \psi)^3}$$
$$ = 2 \psi \left( 1 - \rho^2 + \frac{\rho^2 (1 + \gamma)^2}{(1 + \gamma \psi)^3} \right).$$

Next, we turn to calculating the $\bcl$-dependent term:
\begin{multline*}
    \left. \frac{\partial}{\partial \bcl} \right|_{\bcl = 1} \Bt(0, \bcl) = \left[ -2 \alim (1 - \alim \bcl) + \rho^2 \blim \left(2 \blim \bcl + 4 \alim \bcl - 2 \right) \right. \\
    \left. - 2 \gamma \rho^2 (\alim + \blim)(1 - (\alim + \blim)) \right]_{\lambda = 0, \bcl = 1}
\end{multline*}
$$ = - 2\psi(1 - \psi) + 2 \rho^2 \left[ \blim (2\alim + \blim - 1) \right]|_{\lambda = 0} - 2 \gamma \rho^2 \frac{\psi (1 + \gamma)}{1 + \gamma \psi} \cdot \frac{1 - \psi}{1 + \gamma \psi}.$$
To simplify this expression, note that
$$\left. \left[ \blim (2\alim + \blim - 1) \right]\right|_{\lambda = 0} = \left.\frac{\gamma \psi (1 - \psi)}{1 + \gamma \psi} \cdot \left[ \alim + (\alim + \blim - 1) \right]\right|_{\lambda = 0}$$
$$ = \frac{\gamma \psi (1 - \psi)}{1 + \gamma \psi} \cdot \left( \psi - \frac{1 - \psi}{1 + \gamma \psi} \right) = \frac{\gamma \psi (1 - \psi)}{1 + \gamma \psi} \cdot \frac{\gamma \psi^2 + 2 \psi - 1}{1 + \gamma \psi}$$
$$ = \frac{\psi (1 - \psi) (\gamma^2 \psi^2 + 2\psi \gamma + 1 - (1 + \gamma))}{(1 + \gamma \psi)^2}$$
\begin{equation}    \label{eq:b_times_2aplusb_term}
    = \psi (1 - \psi) \left(1 - \frac{1 + \gamma}{(1 + \gamma \psi)^2} \right).
\end{equation}
As a result, we find that
$$\left. \frac{\partial}{\partial \bcl} \right|_{\bcl = 1} \Bt(0, \bcl) = - 2 \psi (1 - \psi) + 2 \psi (1 - \psi) \left(\rho^2 - \rho^2 \frac{1 + \gamma}{(1 + \gamma \psi)^2} \right) - 2 \psi (1 - \psi) \rho^2 \frac{\gamma (1 + \gamma)}{(1 + \gamma \psi)^2}$$
$$ = - 2 \psi (1 - \psi) \left(1 - \rho^2 + \rho^2 \frac{(1 + \gamma)^2}{(1 + \gamma \psi)^2} \right).$$
At last, combining all of our results, we obtain that
\begin{equation}
    \left. \frac{\partial \Bt}{\partial \lambda} \right|_{\lambda = 0} = 2 \psi \left( 1 - \rho^2 + \frac{\rho^2 (1 + \gamma)^2}{(1 + \gamma \psi)^3} \right) - 2 \psi (1 - \psi) \left(1 - \rho^2 +  \frac{\rho^2 (1 + \gamma)^2}{(1 + \gamma \psi)^2} \right) \left. \frac{\partial \bcl}{\partial \lambda}\right|_{\lambda = 0}. 
\end{equation}
Now, since $\left. \frac{\partial \bcl}{\partial \lambda} \right|_{\lambda = 0} = 0$ when $\kl = 0$, 
$$\left. \frac{\partial \Bt(\hat{f}_{\lambda;0})}{\partial \lambda} \right|_{\lambda = 0} = 2 \psi \left( 1 - \rho^2 + \frac{\rho^2 (1 + \gamma)^2}{(1 + \gamma \psi)^3} \right) > 0.$$
Furthermore, for any $\kl > 0$,
$$\left. \frac{\partial \Bt(\hat{f}_{\lambda; k_\ell})}{\partial \lambda} \right|_{\lambda = 0} - \left. \frac{\partial \Bt(\hat{f}_{\lambda; 0})}{\partial \lambda} \right|_{\lambda = 0} = - 2 \psi (1 - \psi) \left(1 - \rho^2 +  \frac{\rho^2 (1 + \gamma)^2}{(1 + \gamma \psi)^2} \right) \left. \frac{\partial \bcl}{\partial \lambda}\right|_{\lambda = 0; \kl} < 0.$$

Next, we turn to studying the derivative of the variance at $\lambda = 0$. We start with the asymptotic variance formula
$$\Vt(\hat{f}) = \bcl^2 \left( \alim + \blim \rho^2 - \lambda \left(\atlim + \btlim \rho^2 \right) - \alim^2 - \rho^2 \left( 2\alim \blim + \blim^2 \right) + \sigma^2 \atlim \right).$$
First, we compute the baseline (``no learning'') term of the derivative:
$$\left. \frac{\partial}{\partial \lambda} \right|_{\lambda = 0} \Vt(\lambda, 1) = \left. \frac{\partial}{\partial \lambda} \right|_{\lambda = 0} \left( \alim + \blim \rho^2 - \lambda \left(\atlim + \btlim \rho^2 \right) - \alim^2 - \rho^2 \left( 2\alim \blim + \blim^2 \right) + \sigma^2 \atlim \right)$$
$$ = \left.\left[ - 2 (\atlim + \rho^2 \btlim) + 2 \alim \atlim - \rho^2 \left( -2 \atlim \blim - 2 \alim \btlim - 2 \blim \btlim \right) + \sigma^2 \frac{\partial \atlim}{\partial \lambda} \right]\right|_{\lambda = 0} $$
$$ = \left.\left[ - 2 \atlim (1 - \alim) + 2\rho^2 \left( \atlim \blim + \btlim (\alim + \blim - 1) \right) \right] \right|_{\lambda = 0} - \frac{2 \psi \sigma^2}{(1 - \psi)^3}$$
$$ = -2 \frac{\psi}{1 - \psi} (1 - \psi) + 2 \rho^2 \left( \frac{\psi}{1 - \psi} \cdot \frac{\gamma \psi (1 - \psi)}{1 + \gamma \psi} + \frac{\psi}{1 - \psi}\left(1 - \frac{1 + \gamma}{(1 + \gamma \psi)^2} \right) \frac{1 - \psi}{1 + \gamma \psi} \right) - \frac{2 \psi \sigma^2}{(1 - \psi)^3}$$
$$ = -2 \psi + 2 \psi \rho^2 \left( \frac{\gamma \psi (1 + \gamma \psi)^2}{(1 + \gamma \psi)^3} + \frac{(1 + \gamma \psi)^2 - (1 +\gamma)}{(1 + \gamma \psi)^3} \right) - 2 \psi \frac{\sigma^2}{(1 - \psi)^3}$$
$$ = - 2 \psi \left( 1 - \rho^2 \left( \frac{(1 + \gamma \psi)^3 - (1 +\gamma)}{(1 + \gamma \psi)^3} \right) + \frac{\sigma^2}{(1 - \psi)^3} \right)$$
$$ = - 2 \psi \left(1 - \rho^2 + \frac{\rho^2 (1 + \gamma)}{(1 + \gamma \psi)^3} + \frac{\sigma^2}{(1 - \psi)^3} \right).$$

Next, we turn to calculating the $\bcl$-dependent term:
$$\left. \frac{\partial}{\partial \bcl} \right|_{\bcl = 1} \Vt(0, \bcl) = 2 \left. \left[ \alim + \blim \rho^2 - \lambda \left(\atlim + \btlim \rho^2 \right) - \alim^2 - \rho^2 \left( 2\alim \blim + \blim^2 \right) + \sigma^2 \atlim \right] \right|_{\lambda = 0} $$
$$ = 2 \left. \left[ \alim ( 1 - \alim ) - \rho^2 \blim \left( 2\alim + \blim - 1 \right) + \sigma^2 \atlim \right] \right|_{\lambda = 0} $$
$$ = 2 \left( \psi (1 - \psi) - \rho^2 \left( \psi (1 - \psi) \left(1 - \frac{1 + \gamma}{(1 + \gamma \psi)^2} \right) \right) + \frac{\psi \sigma^2}{1 - \psi} \right)$$
$$ = 2 \psi (1 - \psi) \left(1 - \rho^2 + \frac{\rho^2 (1 + \gamma)}{(1 + \gamma \psi)^2} \right) + \frac{2 \psi \sigma^2}{1 - \psi}.$$
Here, we have again used our earlier intermediate result (Eq. \ref{eq:b_times_2aplusb_term}) that 
$$\left[ \blim (2\alim + \blim - 1) \right]|_{\lambda = 0} = \psi (1 - \psi) \left(1 - \frac{1 + \gamma}{(1 + \gamma \psi)^2} \right).$$
Altogether, we find that
\begin{multline}
    \left. \frac{\partial \Vt}{\partial \lambda} \right|_{\lambda = 0} = - 2 \psi \left(1 - \rho^2 + \frac{\rho^2 (1 + \gamma)}{(1 + \gamma \psi)^3} + \frac{\sigma^2}{(1 - \psi)^3} \right) \\
    + \left( 2 \psi (1 - \psi) \left(1 - \rho^2 + \frac{\rho^2 (1 + \gamma)}{(1 + \gamma \psi)^2} \right) + \frac{2 \psi \sigma^2}{1 - \psi} \right) \left. \frac{\partial \bcl}{\partial \lambda}\right|_{\lambda = 0}. 
\end{multline}
Now, when $\kl = 0$, 
$$\left. \frac{\partial \Vt(\hat{f}_{\lambda;0})}{\partial \lambda} \right|_{\lambda = 0} = - 2 \psi \left(1 - \rho^2 + \frac{\rho^2 (1 + \gamma)}{(1 + \gamma \psi)^3} + \frac{\sigma^2}{(1 - \psi)^3} \right) < 0.$$
Furthermore, for any $\kl > 0$,
$$\left. \frac{\partial \Vt(\hat{f}_{\lambda; k_\ell})}{\partial \lambda} \right|_{\lambda = 0} - \left. \frac{\partial \Vt(\hat{f}_{\lambda; 0})}{\partial \lambda} \right|_{\lambda = 0} = \left( 2 \psi (1 - \psi) \left(1 - \rho^2 + \frac{\rho^2 (1 + \gamma)}{(1 + \gamma \psi)^2} \right) + \frac{2 \psi \sigma^2}{1 - \psi} \right) \left. \frac{\partial \bcl}{\partial \lambda}\right|_{\lambda = 0; \kl} > 0.$$
This concludes our proof of Prop. \ref{prop:bias_variance_lambda_derivatives}.

\subsubsection{Sufficient conditions for a feature learning advantage}  \label{app:FLA_proof}

In this section, we establish and prove sufficient conditions for when one step of feature learning results in a lower asymptotic generalization error than no learning. (Again, we assume that $\kl$ is a fixed constant independent of $D$ and $n$.) We do this by proving Theorem \ref{thm:global_FLA}, as stated in the main text. First, we establish the following preliminary result:

\begin{lemma}   \label{lemma:gen_error_infinite_lambda_limit}
Under Assumption \ref{assump:data_gen_process}, $\lim_{\lambda \to \infty} \Gt(\flk) = 1 + \gamma \rho^2$.
\end{lemma}

\begin{proof}
    From Lemma \ref{lemma:alim_blim_properties}, note that $\alim \to 0$, $\blim \to 0$, $\lambda \atlim \to 0$, and $\lambda \btlim \to 0$ as $\lambda \to \infty$. Moreover, by Cor. \ref{cor:bcl_upper_bound}, $\bcl = O_\lambda(1)$, meaning that it remains uniformly bounded above as $\lambda \to \infty$. Applying these facts to the asymptotic generalization error formula (Eq. \ref{eq:gen_error_limit}), all terms involving $\alim$, $\blim$, $\atlim$, $\btlim$, $\lambda \atlim$, and $\lambda \btlim$ vanish as $\lambda \to \infty$: 
    \begin{multline*}
        \Gt(\flk) =  1 - 2\bcl \alim - 2 \rho^2 \blim \bcl + \gamma \rho^2 (1 - \bcl( \alim + \blim ))^2 + \alim \bcl^2 \\+ \rho^2 \blim \bcl^2 - \lambda \bcl^2 (\atlim + \btlim \rho^2) + \bcl^2 \sigma^2 \atlim
    \end{multline*}
    $$\overset{\lambda \nearrow \infty}{\longrightarrow} 1 + \gamma \rho^2.$$
    
\end{proof}

We now prove Theorem \ref{thm:global_FLA}.

\begin{proof}[Proof of Theorem \ref{thm:global_FLA}]
We begin by assuming that the prescribed signal-to-noise inequality holds:

\begin{equation}    \label{eq:appendix_global_FLA_inequality}
        \frac{(1 + \gamma \psi)^2}{\gamma (1 + \gamma) (1-\psi)^2} < \frac{\rho^2}{\sigma^2} < \frac{(1 + \gamma \psi)^3}{\gamma (1 + \gamma) (1 - \psi)^3}. 
    \end{equation}

From Prop. \ref{prop:bias_variance_lambda_derivatives}, we have the following two results:
\begin{equation*}
    \left. \frac{\partial \Bt}{\partial \lambda} \right|_{\lambda = 0} = 2 \psi \left( 1 - \rho^2 + \frac{\rho^2 (1 + \gamma)^2}{(1 + \gamma \psi)^3} \right) - 2 \psi (1 - \psi) \left(1 - \rho^2 +  \frac{\rho^2 (1 + \gamma)^2}{(1 + \gamma \psi)^2} \right) \left. \frac{\partial \bcl}{\partial \lambda}\right|_{\lambda = 0} 
\end{equation*}
\begin{multline*}
    \left. \frac{\partial \Vt}{\partial \lambda} \right|_{\lambda = 0} = - 2 \psi \left(1 - \rho^2 + \frac{\rho^2 (1 + \gamma)}{(1 + \gamma \psi)^3} + \frac{\sigma^2}{(1 - \psi)^3} \right) \\
    + \left( 2 \psi (1 - \psi) \left(1 - \rho^2 + \frac{\rho^2 (1 + \gamma)}{(1 + \gamma \psi)^2} \right) + \frac{2 \psi \sigma^2}{1 - \psi} \right) \left. \frac{\partial \bcl}{\partial \lambda}\right|_{\lambda = 0} 
\end{multline*}
Adding them, we obtain that
\begin{equation}    \label{eq:appendix_gen_error_lambda_derivative}
    \left. \frac{\partial}{\partial \lambda} \right|_{\lambda = 0}\Gt\left(\hat{f}_{\lambda; k_\ell} \right) = 2\psi \left( \frac{ \rho^2 \gamma (1 + \gamma)}{(1 + \gamma \psi)^3} - \frac{\sigma^2}{(1-\psi)^3} \right) - 2 \psi \left( \frac{ \rho^2 (1 - \psi) \gamma (1 + \gamma)}{(1 + \gamma \psi)^2} - \frac{\sigma^2 }{1 - \psi} \right) \left. \frac{\partial \bcl}{\partial \lambda} \right|_{\lambda = 0}.
\end{equation}

From rearranging the inequality $\frac{\rho^2}{\sigma^2} < \frac{(1 + \gamma \psi)^3}{\gamma (1 + \gamma) (1 - \psi)^3}$, we precisely obtain that
$$\frac{ \rho^2 \gamma (1 + \gamma)}{(1 + \gamma \psi)^3} < \frac{\sigma^2}{(1-\psi)^3} $$
which implies that the $\bcl$-independent term of $\left. \frac{\partial}{\partial \lambda} \right|_{\lambda = 0}\Gt\left(\hat{f}_{\lambda; k_\ell} \right)$ (in Eq. \ref{eq:appendix_gen_error_lambda_derivative}) is negative. In particular, recalling that $\left. \frac{\partial \bcl}{\partial \lambda} \right|_{\lambda = 0} = 0$ when $\kl = 0$, it follows that 
\begin{equation}    \label{eq:gen_error_local_decrease}
    \left. \frac{\partial}{\partial \lambda} \right|_{\lambda = 0}\Gt\left(\hat{f}_{\lambda; 0} \right)  < 0.
\end{equation}
Thus, Result 1 of the theorem holds.

Next, observe that
$$\frac{ \rho^2 (1 - \psi) \gamma (1 + \gamma)}{(1 + \gamma \psi)^2} - \frac{\sigma^2 }{1 - \psi} > 0$$
precisely when
$$\frac{\rho^2}{\sigma^2} > \frac{(1 + \gamma \psi)^2}{\gamma (1 + \gamma) (1 - \psi)^2}$$
which is the lower bound of the SNR window. 
Consequently, using Prop. \ref{prop:bcl_properties}, the $\bcl$-dependent term in $\left. \frac{\partial}{\partial \lambda} \right|_{\lambda = 0}\Gt\left(\hat{f}_{\lambda; k_\ell} \right)$ (in Eq. \ref{eq:appendix_gen_error_lambda_derivative}) is negative for all $\kl > 0$:
$$- 2 \psi \left( \frac{ \rho^2 (1 - \psi) \gamma (1 + \gamma)}{(1 + \gamma \psi)^2} - \frac{\sigma^2 }{1 - \psi} \right) \left. \frac{\partial \bcl}{\partial \lambda} \right|_{\lambda = 0} < 0 \qquad \forall \text{ } \kl > 0.$$
As a result, the inequality
\begin{equation}    \label{eq:local_gen_error_inequality}
    \left. \frac{\partial}{\partial \lambda} \right|_{\lambda = 0}\Gt\left(\hat{f}_{\lambda; k_\ell} \right) < \left. \frac{\partial}{\partial \lambda} \right|_{\lambda = 0}\Gt\left(\hat{f}_{\lambda; 0} \right)
\end{equation}
holds for any $\kl > 0$, thereby establishing Result 2 of the theorem. In particular, Results 1 and 2 collectively imply that a local FLA holds.

Next, we set out to prove Result 3, namely that
\begin{equation}    \label{eq:local_featlearning_derivative}
\left. \frac{\partial \Gt(\flk)}{\partial \kl} \right|_{\kl = 0} < 0 \qquad \forall \text{ } \lambda > 0.
\end{equation}

First, observe that by the chain rule
$$\left. \frac{\partial}{\partial k_\ell} \right|_{k_\ell = 0} \Gt\left(\hat{f}_{\lambda; k_\ell} \right) = \left. \frac{\partial}{\partial \bcl} \right|_{\bcl = 1} \Gt\left(\hat{f}_{\lambda; k_\ell} \right) \cdot  \left. \frac{\partial}{\partial k_\ell} \right|_{k_\ell = 0} \bcl$$
where we have noted that $\Gt\left( \hat{f}_{\lambda; k_\ell} \right)$ only depends on $k_\ell$ through $\bcl$ and that $\bcl = 1$ when $k_\ell = 0$. By Prop. \ref{prop:bcl_properties}, note that
$$\left. \frac{\partial}{\partial k_\ell} \right|_{k_\ell = 0} \bcl = \frac{\lambda}{\psi} \left( 1 + \sigma^2 + \gamma \rho^2 - \lambda \left( \frac{\alim + \blim \rho^2}{\psi} + \sigma^2 g(-\lambda) \right)\right) \geq 0$$ with strict inequality if $\lambda > 0$. Thus, proving Result 3 reduces to showing that 
$$\left. \frac{\partial}{\partial \bcl} \right|_{\bcl = 1} \Gt\left(\hat{f}_{\lambda; k_\ell} \right) < 0$$
for all $\lambda > 0$. Using our formula for the asymptotic generalization error (Cor. \ref{cor:asymptotic_gen_error}), recall that
$$\Gt(\flk) = 1 - 2\bcl \alim - 2 \rho^2 \blim \bcl + \gamma \rho^2 (1 - \bcl( \alim + \blim ))^2 + \alim \bcl^2 + \rho^2 \blim \bcl^2 - \lambda \bcl^2 (\atlim + \btlim \rho^2) + \bcl^2 \sigma^2 \atlim.$$
As a result, we find that
\begin{multline*}
    \left. \frac{\partial}{\partial \bcl} \right|_{\bcl = 1} \Gt\left(\hat{f}_{\lambda; k_\ell} \right) = -2\alim - 2 \blim \rho^2 - 2 \gamma \rho^2 (\alim+\blim)(1 - (\alim+\blim)) \\ 
    + 2 \alim + 2 \blim \rho^2 - 2 \lambda (\atlim + \btlim \rho^2) + 2 \sigma^2 \atlim 
\end{multline*}
\begin{equation}    \label{eq:dG_dc_c=1}
     = - 2 \gamma \rho^2 (\alim+\blim)(1 - (\alim+\blim)) - 2 \lambda (\atlim + \btlim \rho^2) + 2 \sigma^2 \atlim.
\end{equation}
For any $\lambda > 0$, $$\atlim + \btlim \rho^2 = \tau_2 + \rho^2 \left( \frac{(1 + \gamma) \tau_2}{(1 + \gamma \tau_1)^2} - \tau_2 \right) = (1 - \rho^2) \tau_2 + \rho^2 \frac{(1 + \gamma) \tau_2}{(1 + \gamma \tau_1)^2} > 0$$
since $\tau_2 = \atlim > 0$ (by Lemma \ref{lemma:alim_blim_properties}) and $\rho^2 \in [0, 1]$.
Thus, $-2 \lambda (\atlim + \btlim \rho^2) < 0$ for all $\lambda > 0$. Consequently, in light of Eq. \ref{eq:dG_dc_c=1}, one sufficient condition for the result that $\left. \frac{\partial}{\partial \bcl} \right|_{\bcl = 1} \Gt\left(\hat{f}_{\lambda; k_\ell} \right) < 0$ for all $\lambda > 0$ is that 
\begin{equation} \label{eqn:dG_dc_suff_cond}
    2 \gamma \rho^2 (\alim+\blim)(1 - (\alim+\blim)) >  2 \sigma^2 \atlim \qquad \forall \text{ } \lambda > 0.
\end{equation}

To analyze Eq. \ref{eqn:dG_dc_suff_cond}, we present a few important facts.
First, since $\psi \in (0, 1)$, recall that both $\alim > 0$ and $\atlim > 0$ for all $\lambda > 0$ and that $\alim$ and $\atlim$ both strictly decrease over all $\lambda > 0$ (Lemma \ref{lemma:alim_blim_properties}). 
Since $\tau_1 = \alim$, this means that $1 - \tau_1$ is increasing in $\lambda$ and that $\frac{1}{1 + \gamma \tau_1}$ is also increasing in $\lambda$, and so $\frac{1 - \tau_1}{1 + \gamma \tau_1}$ is increasing over all $\lambda > 0$.
Moreover,
$$\alim + \blim = \tau_1 + \left( \frac{(\gamma + 1) \tau_1}{1 + \gamma \tau_1} - \tau_1\right) = \frac{(\gamma + 1) \tau_1}{1 + \gamma \tau_1} = 1 - \frac{1 - \tau_1}{1 + \gamma \tau_1}.$$
Putting these facts together, this implies that $\alim+\blim = 1 - \frac{1 - \tau_1}{1 + \gamma \tau_1} \in (0, 1)$ is decreasing as a function of $\lambda$, and that $1 - (\alim+\blim) \in (0, 1)$ is increasing as a function of $\lambda$, over all positive $\lambda$. 

By inspection, we may fulfill the sufficient condition listed in Eq. \ref{eqn:dG_dc_suff_cond} by choosing $\rho$ and $\sigma$ so that 
\begin{equation}    \label{eq:SNR_suff_cond_rephrasing}
    \begin{aligned}
        \frac{\rho^2}{\sigma^2} > \frac{\atlim}{\gamma(\alim+\blim)(1 - (\alim+\blim))} = \frac{\tau_2}{\gamma \left( \frac{(\gamma + 1) \tau_1}{1 + \gamma \tau_1} \right) \left( \frac{1 - \tau_1}{1 + \gamma \tau_1} \right)} \\
        = \frac{1}{\gamma (1 + \gamma)} \cdot \frac{\tau_2}{\tau_1} \cdot \frac{(1 + \gamma \tau_1)^2}{1 - \tau_1} \text{ } \forall \lambda \geq 0.
    \end{aligned}
\end{equation}
Note that $\tau_1, \tau_2 > 0$ for all $\lambda \geq 0$, and that $\alim = \tau_1 \in (0, \psi]$ is decreasing over all $\lambda \in [0, \infty)$. This means that $\frac{1}{1 - \tau_1}$ and $(1 + \gamma \tau_1)^2$ are both decreasing as functions of $\lambda$. In turn, $\frac{(1 + \gamma \tau_1)^2}{1 - \tau_1}$ is decreasing over $\lambda \geq 0$. Next, we claim that $\frac{\tau_2}{\tau_1}$ is also decreasing in $\lambda$ over $\lambda \geq 0$. Indeed, define 
$$F(\lambda) := \frac{\tau_2}{\tau_1} = \frac{\atlim}{\alim} = \frac{\int \frac{x}{(x+\lambda)^2} \MP^{\psi}(dx)}{\int \frac{x}{x+\lambda} \MP^{\psi}(dx)}.$$
We claim that $F$ is a non-increasing function in $\lambda$ (for $\lambda \geq 0$). Indeed, using the quotient rule as well as differentiation through the integral (valid by Prop. \ref{prop:properties_of_MP_and_Stieltjes}, part 4), we find that
$$F'(\lambda) = \frac{-2 \int \frac{x}{(x+\lambda)^3} \MP^{\psi}(dx) \int \frac{x}{x+\lambda} \MP^{\psi}(dx) + \left( \int \frac{x}{(x+\lambda)^2} \MP^{\psi}(dx) \right)^2}{\left( \int \frac{x}{x+\lambda} \MP^{\psi}(dx) \right)^2}.$$
Letting $f_1(x) := \left(\frac{x}{x+\lambda}\right)^{\frac12}$ and $f_2(x) := \left( \frac{x}{(x+\lambda)^3} \right)^{\frac12}$, the Cauchy-Schwarz inequality implies that
$$\left( \int \frac{x}{(x+\lambda)^2} \MP^{\psi}(dx) \right)^2 = \left( \int f_1(x) f_2(x) \MP^{\psi}(dx) \right)^2 \leq \int f_1(x)^2 \MP^{\psi}(dx) \int f_2(x)^2 \MP^{\psi}(dx)$$
$$ = \int \frac{x}{(x+\lambda)^3} \MP^{\psi}(dx) \int \frac{x}{x+\lambda} \MP^{\psi}(dx) < 2 \int \frac{x}{(x+\lambda)^3} \MP^{\psi}(dx) \int \frac{x}{x+\lambda} \MP^{\psi}(dx)$$
implying that $F'(\lambda) < 0$ for any $\lambda \geq 0$. 
Thus, $\frac{\tau_2}{\tau_1}$ is decreasing in $\lambda$, as claimed. 
Consequently, Eq. \ref{eq:SNR_suff_cond_rephrasing} (and, thus, Eq. \ref{eqn:dG_dc_suff_cond}) is satisfied if 
$$\frac{\rho^2}{\sigma^2} > \frac{1}{\gamma (1 + \gamma)} \cdot \left.\left[\frac{\tau_2}{\tau_1} \cdot \frac{(1 + \gamma \tau_1)^2}{1 - \tau_1} \right] \right|_{\lambda = 0} = \frac{1}{\gamma (1 + \gamma)} \cdot \frac{\frac{\psi}{1-\psi}}{\psi} \cdot \frac{(1 + \gamma \psi)^2}{1 - \psi},$$
resulting in a final condition that
\begin{equation}    \label{eqn:SNR_c_lower_bound}
    \frac{\rho^2}{\sigma^2} > \frac{(1 + \gamma \psi)^2}{\gamma (1 + \gamma) (1 - \psi)^2}.
\end{equation}
Thus, the inequality ultimately arises from simply evaluating $\frac{\atlim}{\gamma (\alim+\blim)(1 - (\alim+\blim))}$ at $\lambda = 0$. This establishes Result 3 of the theorem.

Finally, we set out to prove Result 4, i.e. that $\inf_{\lambda \geq 0} \Gt(\hat{f}_{\lambda;0})$ is attained at some finite $\lambda_\star \geq 0$. To do so, it suffices to show that $$\Gt(\hat{f}_{0;0}) \leq \lim_{\lambda \to \infty} \Gt(\hat{f}_{\lambda;0}) = 1 + \gamma \rho^2$$
where we have applied Lemma \ref{lemma:gen_error_infinite_lambda_limit}. Consequently, since $\Gt(\hat{f}_{\lambda;0})$ is continuous in $\lambda$, it will follow that $\Gt(\hat{f}_{\lambda;0})$ must have a global minimum for some finite $\lambda_\star \geq 0$; in fact, by Result 1 (established above, given the SNR window inequality), it must be the case that $\lambda_\star > 0$. Thus, to conclude, we calculate and bound $\Gt(\hat{f}_{0;0})$, using the intermediate quantities listed in Appendix Section \ref{app:intermediate_quantities} as well as the fact that $\sigma^2 < \frac{\gamma (1 + \gamma) (1 - \psi)^2}{(1 + \gamma \psi)^2} \rho^2$ (using the assumed SNR lower bound):
\begin{multline*}
    \Gt(\hat{f}_{0;0}) = \left.\left[1 - 2\bcl \alim - 2 \rho^2 \blim \bcl + \gamma \rho^2 (1 - \bcl( \alim + \blim ))^2 + \alim \bcl^2 \right. \right. \\ \left. \left. + \rho^2 \blim \bcl^2 - \lambda \bcl^2 (\atlim + \btlim \rho^2) + \bcl^2 \sigma^2 \atlim \right] \right|_{(\lambda, \kl) = (0, 0)}
\end{multline*}
\begin{align*}
    = \left.\left[1 - 2 \alim - 2 \blim \rho^2 + (1 - (\alim + \blim))^2 \gamma \rho^2 + \alim + \blim \rho^2 + \sigma^2 \atlim \right]\right|_{\lambda = 0}
\end{align*}
\begin{align*}
    = \left.\left[1 - \alim - \blim \rho^2 + (1 - (\alim + \blim))^2 \gamma \rho^2 + \sigma^2 \atlim \right]\right|_{\lambda = 0}
\end{align*}
\begin{align*}
    = 1 - \psi - \frac{\psi(1 - \psi)}{1 + \gamma \psi} \gamma \rho^2 + \left( \frac{1 - \psi}{1 + \gamma \psi}\right)^2 \gamma \rho^2 + \sigma^2 \frac{\psi}{1 - \psi}
\end{align*}
\begin{align*}
    \leq 1 - \psi + \left( \left( \frac{1 - \psi}{1 + \gamma \psi}\right)^2 - \frac{\psi(1 - \psi)}{1 + \gamma \psi} + \frac{(1 + \gamma) \psi (1 - \psi)}{(1 + \gamma \psi)^2} \right) \gamma \rho^2
\end{align*}
\begin{align*}
    = (1 - \psi) + \frac{1 - \psi}{1 + \gamma \psi}\left( \frac{1 - \psi}{1 + \gamma \psi} - \psi + \frac{(1 + \gamma) \psi}{1 + \gamma \psi} \right) \gamma \rho^2
\end{align*}
\begin{align*}
    = (1 - \psi) + \frac{(1 - \psi)^2}{1 + \gamma \psi} \gamma \rho^2
\end{align*}
\begin{align*}
    \leq 1 + \gamma \rho^2
\end{align*}
where we have noted that $1 - \psi \leq 1$ and that $\frac{(1 - \psi)^2}{1 + \gamma \psi} \leq 1$. This concludes our proof of Result 4. 

Furthermore, Results 3 and 4 collectively imply a global FLA. Indeed, Result 4 implies that $\inf_{\lambda \geq 0} \Gt(\hat{f}_{\lambda;0}) = \Gt(\hat{f}_{\lambda_\star;0})$ for some finite constant $\lambda_\star > 0$. Then, Result 3 implies that there exists some $\kl^{(\lambda_\star)} > 0$ such that $$\Gt(\hat{f}_{\lambda_\star;\kl^{(\lambda_\star)}}) < \Gt(\hat{f}_{\lambda_\star;0}) = \inf_{\lambda \geq 0} \Gt(\hat{f}_{\lambda;0})$$
thus establishing that
$$\inf_{\lambda \geq 0, \kl > 0} \Gt(\hat{f}_{\lambda;\kl}) < \inf_{\lambda \geq 0} \Gt(\hat{f}_{\lambda;0}),$$
as desired.
\end{proof}

\subsubsection{No feature learning advantage in the isotropic setting --- Proof of Theorem \ref{thm:no_isotropic_FLA}}  \label{app:no_isotropic_FLA}

In this section, we prove Theorem \ref{thm:no_isotropic_FLA}, as stated in the main text. (Again, we assume that $\kl$ is a fixed constant independent of $D$ and $n$.)
First, when $\gamma = 0$, observe that $\blim = \btlim = 0$. Furthermore, when $\rho = 0$, every $\gamma$- and $\rho$-dependent term in the asymptotic generalization error (Eq. \ref{eq:gen_error_limit_SUPP}) vanishes since all such terms appear with a factor of $\rho^2$.
Altogether, from Eq. \ref{eq:gen_error_limit_SUPP} we find that when $\gamma = 0$ or $\rho = 0$, the asymptotic formula for the generalization error reduces to
\begin{equation}    \label{eq:isotropic_feat_gen_error}
    \Gt(\flk) = 1 - 2\alim \bcl + \left(\alim + (\sigma^2 - \lambda) \atlim \right) \bcl^2.
\end{equation}
Furthermore, when $\kl = 0$ so that $\bcl = 1$, we find that 
\begin{equation}    \label{eq:isotropic_baseline_gen_error}
    \Gt(\hat{f}_{\lambda; 0}) = 1 - \alim  + (\sigma^2 - \lambda) \atlim.
\end{equation}
Using these formulas, we now prove a key part of Theorem \ref{thm:no_isotropic_FLA} through the following lemma:

\begin{lemma}       \label{lemma:isotropic_optimal_baseline}
Suppose that $\gamma = 0$ or $\rho = 0$, and that $\psi \in (0, 1)$. Then, the following results hold:
\begin{enumerate}
    \item Suppose that $\sigma^2 > 0$. When viewed as a function in $\lambda$, $\Gt(\hat{f}_{\lambda; 0})$ is strictly decreasing for $\lambda \in [0, \sigma^2)$ and strictly increasing for $\lambda \in (\sigma^2, \infty)$. In particular, $$\inf_{\lambda \geq 0} \Gt(\hat{f}_{\lambda;0}) = \Gt(\hat{f}_{\sigma^2;0})$$ 
    and this minimum is uniquely attained when $\lambda = \sigma^2$.

    \item Suppose that $\sigma^2 = 0$. Then, $\Gt(\hat{f}_{\lambda; 0})$ strictly increases over $\lambda \geq 0$, thereby attaining a unique minimum when $\lambda = 0$.
\end{enumerate}
\end{lemma}

\begin{proof}
    First, suppose that $\sigma^2 > 0$. Then, using Eq. \ref{eq:isotropic_baseline_gen_error},
    \begin{equation}    \label{eq:dG_dlambda_isotropic}
        \begin{aligned}
            \frac{\partial \Gt(\hat{f}_{\lambda;0})}{\partial \lambda} = \frac{\partial}{\partial \lambda} \left[ 1-\alim + (\sigma^2 - \lambda) \atlim \right] \\
            = \atlim - \atlim + (\sigma^2 - \lambda) \frac{\partial \atlim}{\partial \lambda} \\
            = \psi (\lambda - \sigma^2) (2 g'(-\lambda) - \lambda g''(-\lambda)) \\
            = 2 \psi (\lambda - \sigma^2) \int \frac{x}{(x + \lambda)^3}  \MP^{\psi}(dx).
        \end{aligned}
    \end{equation}
    Here, we have used Lemma \ref{lemma:alim_blim_properties}.
    Since $\int \frac{x}{(x + \lambda)^3}  \MP^{\psi}(dx) > 0$, it follows that $\frac{\partial \Gt(\hat{f}_{\lambda;0})}{\partial \lambda} < 0$ for all $\lambda \in [0, \sigma^2)$, $\frac{\partial \Gt(\hat{f}_{\lambda;0})}{\partial \lambda} > 0$ for all $\lambda \in (\sigma^2, \infty)$, and that $\frac{\partial \Gt(\hat{f}_{\lambda;0})}{\partial \lambda} = 0$ when $\lambda = \sigma^2$. This precisely establishes Result 1.

    To prove Result 2, note that when $\sigma^2 = 0$, Eq. \ref{eq:dG_dlambda_isotropic} reduces to
    \begin{equation}    \label{eq:dG_dlambda_isotropic_noiseless}
        \frac{\partial \Gt(\hat{f}_{\lambda;0})}{\partial \lambda} = 2 \lambda \psi \int \frac{x}{(x + \lambda)^3}  \MP^{\psi}(dx)
    \end{equation}
    which is $0$ when $\lambda = 0$ and positive for all $\lambda > 0$. Thus, $\Gt(\hat{f}_{\lambda; 0})$ strictly increases over $\lambda \geq 0$, with its minimum being attained at $\lambda = 0$.
\end{proof}

Using Lemma \ref{lemma:isotropic_optimal_baseline}, we now prove Theorem \ref{thm:no_isotropic_FLA}.

\begin{proof}[Proof of Theorem \ref{thm:no_isotropic_FLA}]
    Note that Lemma \ref{lemma:isotropic_optimal_baseline} precisely implies one part of the theorem. In particular, it remains to show that for any $\lambda, \kl \geq 0$,
    \begin{equation}    \label{eq:baseline_lower_bound}
        \Gt(\hat{f}_{\lambda;\kl}) \geq \Gt(\hat{f}_{\sigma^2;0})
    \end{equation}
    with equality if and only if one of two conditions is satisfied:
    \begin{enumerate}
        \item $\sigma^2 > 0$ and $(\lambda, \kl) = (\sigma^2, 0)$.

        \item $\sigma^2 = 0$ and $\lambda = 0$ (with no restrictions on $\kl$).
    \end{enumerate}

    Starting with the asymptotic generalization error formula in Eq. \ref{eq:isotropic_feat_gen_error}, this equation may be viewed as a quadratic function in $\bcl$ with $\bcl$ depending on $\lambda$. In particular, define the function
    \begin{equation}    \label{eq:G_l_c}
        G(\lambda, c) := 1 - 2\alim c + \left(\alim + (\sigma^2 - \lambda) \atlim \right) c^2.
    \end{equation}
    Then, for any $(\lambda, \kl)$ (which then determines the value of $\bcl$), to prove Eq. \ref{eq:baseline_lower_bound}, it suffices to show that $G(\lambda, c) \geq G(\sigma^2, 1)$ with equality if and only if $\lambda = \sigma^2$ and $c = 1$. 
    From this it will also follow that the inequality will be strict whenever $\sigma^2 > 0$ and $(\lambda, \kl) \neq (\sigma^2, 0)$.
    This is because the condition that $\lambda = \sigma^2 > 0$ and $\bcl = 1$ is equivalent to $(\lambda, \kl) = (\sigma^2, 0)$, e.g. by Prop. \ref{prop:bcl_properties}. When $\sigma^2 = 0$, then the condition that $\lambda = \sigma^2 = 0$ and $\bcl = 1$ is equivalent to the condition that $\lambda = 0$ and $\kl$ is anything nonnegative. Thus, we have established the equality cases for Eq. \ref{eq:baseline_lower_bound}, and we set out to prove our reframed assertion.
    
    To prove this assertion, first note that $\alim + (\sigma^2 - \lambda) \atlim > 0$ for all $\lambda \geq 0$. Indeed, 
    $$\alim + (\sigma^2 - \lambda) \atlim = \psi \int \frac{x}{x+\lambda} \MP^{\psi}(dx) - \psi \int \frac{\lambda x}{(x+\lambda)^2} \MP^{\psi}(dx) + \psi \int \frac{\sigma^2 x}{(x+\lambda)^2} \MP^{\psi}(dx)$$
    $$ = \psi \int \frac{x^2 + \sigma^2 x}{(x+\lambda)^2} \MP^{\psi}(dx) > 0$$
    for all $\lambda \geq 0$ and $\sigma^2 \geq 0$. As a result, for any fixed $\lambda \geq 0$, $G(\lambda, c)$ is a concave-upwards quadratic function in $c$, with a minimum value of 
    $$1 - \frac{\alim^2}{\alim + (\sigma^2 - \lambda) \atlim}$$
    uniquely attained when $$c = c^\star(\lambda) := \frac{\alim}{\alim + (\sigma^2 - \lambda) \atlim}.$$
    Thus,
    \begin{equation}
        G(\lambda, c) \geq 1 - \frac{\alim^2}{\alim + (\sigma^2 - \lambda) \atlim}.
    \end{equation}
    Defining $H(\lambda) := 1 - \frac{\alim^2}{\alim + (\sigma^2 - \lambda) \atlim}$,
    we find that
    $$\frac{\partial H}{\partial \lambda} = - \frac{2 \alim \frac{\partial \alim}{\partial \lambda} \left( \alim + (\sigma^2 - \lambda) \atlim \right) - \alim^2 \left( \frac{\partial \alim}{\partial \lambda} - \atlim + (\sigma^2 - \lambda) \frac{\partial \atlim}{\partial \lambda} \right)}{\left( \alim + (\sigma^2 - \lambda) \atlim \right)^2}$$
    $$ = -\frac{-2\alim \atlim \left( \alim + (\sigma^2 - \lambda) \atlim  \right) - \alim^2 \left( -2 \atlim + (\sigma^2 - \lambda) \frac{\partial \atlim}{\partial \lambda} \right) }{\left( \alim + (\sigma^2 - \lambda) \atlim \right)^2}$$
    $$ = \frac{(\sigma^2 - \lambda) \alim \left( 2 \atlim^2 + \alim \frac{\partial \atlim}{\partial \lambda} \right)}{\left( \alim + (\sigma^2 - \lambda) \atlim \right)^2}.$$
    Recall that $\alim > 0$ for all $\lambda \geq 0$ (Lemma \ref{lemma:alim_blim_properties}). Moreover, again using Lemma \ref{lemma:alim_blim_properties}, we find that
    $$2 \atlim^2 + \alim \frac{\partial \atlim}{\partial \lambda} = 2\psi^2 \left( \int \frac{x}{(x+\lambda)^2} \MP^{\psi}(dx) \right)^2 - 2\psi^2 \int \frac{x}{x+\lambda} \MP^{\psi}(dx) \int \frac{x}{(x+\lambda)^3} \MP^{\psi}(dx) \leq 0$$
    where we have applied the Cauchy-Schwarz inequality in $L^2(\RR, \MP^{\psi})$ using the functions $f_1(x) := \left(\frac{x}{x+\lambda} \right)^{\frac12}$ and $f_2(x) := \left(\frac{x}{(x+\lambda)^3} \right)^{\frac12}$. In fact, we claim that the above inequality must be strict. Indeed, equality holds for the Cauchy-Schwarz inequality if and only if there exists a constant $r$ such that $f_2(x) = r f_1(x)$ holds $\MP^{\psi}$-almost everywhere. This cannot be true because it would entail that $x + \lambda$ is constant almost everywhere in the support of $\MP^{\psi}$. As a result, it follows that $2 \atlim^2 + \alim \frac{\partial \atlim}{\partial \lambda} < 0$ for all $\lambda \geq 0$.
    In turn, we find that the sign of the derivative $\frac{\partial H}{\partial \lambda}$ coincides precisely with the sign of $\lambda - \sigma^2$. Specifically, $\frac{\partial H}{\partial \lambda} < 0$ for $\lambda \in [0, \sigma^2)$, $\frac{\partial H}{\partial \lambda} = 0$ when $\lambda = \sigma^2$, and $\frac{\partial H}{\partial \lambda} > 0$ for $\lambda \in (\sigma^2, \infty)$. Thus, $H$ has a unique global minimizer at $\lambda = \sigma^2$, which results in $H(\sigma^2) = 1 - \alim(\lambda = \sigma^2)$. Altogether, we have established the chain of inequalities 
    \begin{equation}
        G(\lambda, c) \geq 1 - \frac{\alim^2}{\alim + (\sigma^2 - \lambda) \atlim} \geq 1 - \alim(\lambda = \sigma^2).
    \end{equation}

    Furthermore, from setting $\lambda = \sigma^2$ and $c = 1$ in Eq. \ref{eq:G_l_c}, we find that
    $$G(\sigma^2, 1) = 1 - \alim(\lambda = \sigma^2)$$
    which establishes that $\inf_{\lambda \geq 0, c \geq 1} G(\lambda, c)$ is attained and that $(\lambda, c) = (\sigma^2, 1)$ is a minimizer of $G$.
    To show that this is the unique minimizer of $G$, first suppose that $\lambda \neq \sigma^2$. For this choice of $\lambda$, recall that $G(\lambda, c)$ is minimized along the $c$-axis when $c = c^\star(\lambda) = \frac{\alim(\lambda)}{\alim(\lambda) + (\sigma^2 - \lambda) \atlim(\lambda)}$, attaining a minimum value of 
    $$H(\lambda) = 1 - \frac{\alim(\lambda)^2}{\alim(\lambda) + (\sigma^2 - \lambda) \atlim(\lambda)}.$$
    Moreover, we have shown that $H$ is uniquely minimized by $\lambda = \sigma^2$. As a result, for $\lambda \neq \sigma^2$, $$G(\lambda, c) \geq G(\lambda, c^\star) = H(\lambda) > H(\sigma^2) = G(\sigma^2, 1).$$
    Furthermore, if $\lambda = \sigma^2$, then $G(\lambda, c)$ is a nondegenerate concave-upward quadratic function in $c$, meaning that it is uniquely minimized along the $c$-axis by $$c = c^\star(\lambda) = \frac{\alim(\lambda = \sigma^2)}{\alim(\lambda = \sigma^2) + (\sigma^2 - \sigma^2)\atlim(\lambda = \sigma^2)} = 1.$$
    Therefore, we conclude that $G(\lambda, c) \geq G(\sigma^2, 1)$ with equality if and only if $(\lambda, c) = (\sigma^2, 1)$, which completes our proof.
    
\end{proof}

\end{document}